\documentclass[letterpaper]{article} 
\usepackage{aaai25}  
\usepackage{times}  
\usepackage{helvet}  
\usepackage{courier}  
\usepackage[hyphens]{url}  
\usepackage{graphicx} 
\usepackage{natbib}  
\usepackage{caption} 
\usepackage{algorithm}

\usepackage{algpseudocode}
\usepackage{amsmath}
\usepackage{amssymb}
\usepackage{soul}
\usepackage{multirow}
\usepackage[dvipsnames]{xcolor}
\usepackage{caption}
\usepackage{subcaption}
\usepackage{booktabs} 
\usepackage{enumitem}

\newcommand{\norm}[1]{\left\lVert#1\right\rVert}
\newcommand{\pr}[1]{\textrm{Pr}(#1)}
\newcommand{\cpr}[2]{\textrm{Pr}(#1|#2)}
\definecolor{mygreen}{RGB}{0, 128, 0}
\definecolor{darkgoldenrod}{RGB}{184, 134, 11}

\usepackage{newfloat}
\usepackage{listings}
\DeclareCaptionStyle{ruled}{labelfont=normalfont,labelsep=colon,strut=off} 
\floatstyle{ruled}
\newfloat{listing}{tb}{lst}{}
\floatname{listing}{Listing}
\title{PixelMan: Consistent Object Editing with Diffusion Models\\via Pixel Manipulation and Generation}
\author{
    Liyao Jiang\textsuperscript{\rm 1,\rm 2},
    Negar Hassanpour\textsuperscript{\rm 2},
    Mohammad Salameh\textsuperscript{\rm 2},\\
    Mohammadreza Samadi\textsuperscript{\rm 2},
    Jiao He\textsuperscript{\rm 3},
    Fengyu Sun\textsuperscript{\rm 3},
    Di Niu\textsuperscript{\rm 1}
}
\affiliations {
\textsuperscript{\rm 1}Department of Electrical and Computer Engineering, University of Alberta\\
    \textsuperscript{\rm 2}Huawei Technologies Canada\\
    \textsuperscript{\rm 3}Huawei Kirin Solution, China\\
    \{liyao1, dniu\}@ualberta.ca\\
    \{negar.hassanpour2, mohammad.salameh, mohammadreza.samadi1, hejiao4\}@huawei.com\\
    sunfengyu@hisilicon.com\\
}

\usepackage{bibentry}

\begin{document}

\maketitle

\begin{abstract}
Recent research explores the potential of Diffusion Models (DMs) for consistent object editing, which aims to modify object position, size, and composition, etc., while preserving the consistency of objects and background without changing their texture and attributes.
Current inference-time methods often rely on DDIM inversion, which inherently compromises efficiency and the achievable consistency of edited images.
Recent methods also utilize energy guidance which iteratively updates the predicted noise and can drive the latents away from the original image, resulting in distortions. In this paper, 
we propose \hbox{PixelMan}, an inversion-free and training-free method for achieving consistent object editing via Pixel Manipulation and generation, where 
we directly create a duplicate copy of the source object at target location in the pixel space, and introduce an efficient sampling approach to iteratively harmonize the manipulated object into the target location and inpaint its original location, while ensuring image consistency by anchoring the edited image to be generated to the pixel-manipulated image as well as by introducing various consistency-preserving optimization techniques during inference.
Experimental evaluations based on benchmark datasets as well as extensive visual comparisons show that in as few as 16 inference steps, PixelMan outperforms a range of state-of-the-art training-based and training-free methods (usually requiring 50 steps) on multiple consistent object editing tasks.
\end{abstract}
%
\begin{links}
    \link{Project}{liyaojiang1998.github.io/projects/PixelMan/}
    \link{Extended version}{https://arxiv.org/abs/2412.14283}
\end{links}


\section{Introduction}
\label{sec:intro}

Diffusion Models (DMs) excel at generating stunning visuals from text prompts~\cite{rombach2022high, saharia2022photorealistic, chang2023muse},
yet with potentials extending beyond text-to-image generation.
A highly popular application is image editing, 
as evidenced by widespread tools such as Google Photos MagicEditor~\cite{google_magiceditor} and AI Editor in Adobe Photoshop~\cite{adobe_photoshop}.
Many research efforts~\cite{hertz2022prompt, tumanyan2023plug, alaluf2023cross, parmar2023zero} achieve promising results on text-prompt-guided rigid image editing involving tasks such as changing the color, texture, attributes, and style of the image.
However, \textit{consistent object editing}~\cite{kawar2023imagic, cao2023masactrl, duan2024tuning} is a distinct type of image editing that aims to preserve the consistency of objects and background in the image without changing their texture and attributes, 
while modifying only certain non-rigid attributes of the objects (e.g., changing the position, size, and composition of objects).
Typical consistent object editing tasks include object repositioning~\cite{epstein2023diffusion, mou2024dragondiffusion, mou2024diffeditor, wang2024repositioning, winter2024objectdrop}, object resizing~\cite{epstein2023diffusion, mou2024dragondiffusion, mou2024diffeditor}, and object pasting~\cite{chen2024anydoor, mou2024dragondiffusion, mou2024diffeditor}.
Consistent object editing tasks are complex and usually involve multiple sub-tasks such as: 
(i)~generating a faithful reproduction of the source object at the target location, 
(ii)~maintaining the background scene details, 
(iii)~harmonizing the edited object into its surrounding target context,
and (iv)~inpainting the original vacated location with a cohesive background.

To solve this problem, training-based methods have been proposed~\cite{rombach2022high, chen2024anydoor, wang2024repositioning, winter2024objectdrop}, which however require a costly training process and usually also require collecting task-specific datasets.
Alternatively, recent training-free methods~\cite{epstein2023diffusion, mou2024dragondiffusion, mou2024diffeditor} 
rely on DDIM inversion~\cite{dhariwal2021diffusion} 
to estimate
the initial noise corresponding to the source image. 
However, this process is inefficient as it often requires many (usually at least 50) inference steps.
Reducing the number of steps to, e.g., 16, significantly compromises editing quality (see Fig.~\ref{fig:examples_comparisons_1}).
Moreover, DDIM inversion struggles to produce a precise and consistent final reconstruction of the source image, often yielding a coarse approximation due to accumulation of errors at each timestep~\cite{duan2024tuning}. 
As a result, training-free methods that rely on DDIM inversion
are inherently limited in their ability to perform consistent edits.

To facilitate object generation at target location and reproduction of background,
DragonDiffusion~\cite{mou2024dragondiffusion} and DiffEditor~\cite{mou2024diffeditor} utilize Energy Guidance~(EG) to minimize the feature similarity between the source and target objects (backgrounds).
While EG iteratively refines the predicted noise, this process can inadvertently drive the latent representation away from that of the original image during inference, causing distortions in object appearance and background. 
Additionally, seamlessly inpainting the vacated region (if any) with a coherent background remains a challenge, as existing methods often struggle to fully remove the original object or introduce unintended elements (see Fig.~\ref{fig:examples_comparisons_1}).

In this paper, we propose \hbox{PixelMan}, an inversion-free and training-free method to achieve consistent object editing with existing pretrained text-to-image diffusion models via Pixel Manipulation and generation in as few as 16 steps that outperform all competitive training-based and training-free methods (usually requiring 50 steps) on a range of consistent object editing tasks. 
Rather than performing DDIM inversion and edited denoising, we directly create a duplicate copy of the source object at target location in the pixel space, and introduce an efficient sampling approach to iteratively harmonize the manipulated object into the target location and inpaint its original location, while ensuring image consistency by anchoring output image to be generated to the pixel-manipulated image as well as by introducing various consistency-preserving optimization techniques during inference.
Our contributions are summarized as follows:

\begin{itemize}
    \item
    We propose to perform pixel manipulation for achieving consistent object editing, 
    by creating a pixel-manipulated image where we copy the source object to the target location in the pixel space. 
    At each step, we always anchor the target latents to the pixel-manipulated latents, 
    which reproduces the object and background with high image consistency, while only focusing on generating the missing ``delta'' between the pixel-manipulated image and the target image to be generated.

    \item
    We design an efficient three-branched inversion-free sampling approach, which finds the delta editing direction to be added on top of the anchor, i.e., the latents of the pixel-manipulated image, by computing the difference between the predicted latents of the target image and pixel-manipulated image in each step. 
    This process also facilitates faster editing by reducing the required number of inference steps and number of Network Function Evaluations~(NFEs).

    \item 
    To inpaint the manipulated object's source location, we identify a root cause of many incomplete or incoherent inpainting cases in practice, which is attributed to information leakage from similar objects through the Self-Attention (SA) mechanism. To address this issue, we propose a leak-proof self-attention technique to prevent attention to source, target, and similar objects in the image to mitigate leakage and enable cohesive inpainting.
    
    \item
    Our method harmonizes the edited object with the target context,
    by leveraging editing guidance with latents optimization, and by using a source branch to preserve uncontaminated source $K,V$ features as the context for generating appropriate harmonization effects (e.g. lighting, shadow, and edge blending) at the target location.
    
\end{itemize}

We provide extensive quantitative and/or qualitative visual comparisons to a range of state-of-the-art training-free and training-based approaches designed for object repositioning, object resizing and object pasting (some of which can be found in Appendix). 
Quantitative results on the COCOEE and ReS datasets as well as extensive visual comparisons suggest that \hbox{PixelMan} achieves superior performance in consistency metrics for object, background, and semantic consistency between the source and edited image, while achieving higher or comparable performance in IQA metrics.
As a training-free method, \hbox{PixelMan} only requires 16 inference steps with lower average latency and a lower number of NFEs than current popular methods.

\section{Related Works}
\label{sec:related}

\subsubsection{Image editing with DMs.}
While standard text-to-image DMs are not directly designed for image editing, 
recent research is actively exploring their potential
for this task. 
\textit{Training-based} approaches 
\cite{saharia2022palette, brooks2023instructpix2pix, zhang2023adding} 
optimize the UNet for certain editing scenarios. 
\citet{wang2024repositioning} fine-tuned an inpainting model specifically for object repositioning task 
(by introducing and utilizing an ad-hoc dataset, namely ReS). 
However, these
approaches may require high 
computational resources only to learn a specific task. 
As such, there is a high motivation to explore methods for augmenting pretrained UNets with different editing capabilities 
without additional training.
In \textit{\hbox{training-free}} methods 
\cite{hertz2022prompt, alaluf2023cross, hertz2023style, tumanyan2023plug}, 
users can perform editing either by a descriptive text prompt
\cite{hertz2022prompt, brooks2023instructpix2pix, tumanyan2023plug, epstein2023diffusion},
or by specifying editing points within an image, 
called \textit{point-based} editing 
\cite{endoPG2022, pan2023drag, shi2023dragdiffusion, mou2024dragondiffusion, mou2024diffeditor}. 
The main advantage of point-based editing is the granular control over the edit region. 
In this work, we propose a point-based training-free approach for consistent object editing using DM, which preserves the consistency between the source and edited image.

\subsubsection{Training-free consistent object editing.}
\citet{epstein2023diffusion} introduced Energy Guidance (EG) 
(see Appendix for details) 
and proposed \hbox{SelfGuidance}, a prompt-based editing method
that guides the sampling process based on specific energy functions 
defined on attentions and activations. 
\citet{mou2024dragondiffusion} proposed DragonDiffusion, 
a point-based editing approach that leverages EG to update the sampled noise. 
Building on this, DiffEditor~\cite{mou2024diffeditor} improved the content consistency 
by introducing regional SDE sampling and score-based gradient guidance \cite{song2020score}.
Despite their success, EG-based methods 
require computationally expensive tricks to propagate the guidance from $\epsilon$ to $z_t$.
Different from EG-based methods that update the estimated noise $\epsilon$, our method directly updates the latents $z_t$ for consistent object editing, which reduces the latency as well as NFEs, while maintaining consistency and image quality.

\subsubsection{Inverting real images.}
Preserving the consistency between the original and edited image is crucial for consistent image editing. 
Training-free methods often utilize the inversion techniques to convert the source image into a convertible initial noise ($z_T$). 
DDIM inversion~\cite{dhariwal2021diffusion} is a common but computationally expensive technique as it usually requires 50 inference steps.
ReNoise~\cite{garibi2024renoise} is a recent inversion technique that can utilize few-steps models 
\cite{luo2023latent, stabilityai_sdxl_turbo}, 
but its repeated UNet calls in its refinement phase still leads to high computation costs. 
An alternative approach is Denoising Diffusion Consistent Model (DDCM)~\cite{xu2023infedit},
which facilitates inversion-free prompt-guided rigid image editing for changing the texture and attribute of objects. 
In contrast to DDCM, our method does not use any prompt, and instead we propose an inversion-free approach for efficient consistent object editing which focuses on preserving the consistency of objects and background in the image without changing their texture and attributes while modifying only certain non-rigid attributes of the objects (e.g., changing the position, size, and composition of objects).

\subsubsection{Attention control for editing.}
Recent studies on training-free editing techniques \cite{cao2023masactrl, hertz2023style, tumanyan2023plug, hertz2022prompt, parmar2023zero} 
explore
either integrating or manipulating Cross-Attentions~(CAs) and Self-Attentions~(SAs) 
to exert precise control over the editing process.
Manipulating CAs has been demonstrated to offer control over object composition. 
\citet{hertz2022prompt} proposed an injection approach for swapping objects and changing the global style. 
Alternatively, since SAs incorporate information about pixel interactions
in the spatial domain, 
manipulating them affects overall style, texture, and object interaction
\cite{alaluf2023cross, hertz2023style, jeong2024visual, cao2023masactrl}. 
Building on this, 
\citet{patashnik2023localizing} 
presented a SA injection method to selectively preserve a set of objects while altering other regions.
Following these insights, we propose a leakproof self-attention technique to 
ensure a complete and cohesive inpainting of the vacated area with the background, by preventing a root cause of failed inpainting which is information leakage from the source or similar objects.

\section{Method}
\label{sec:method}

\begin{figure*}[ht]
    \centering
    \includegraphics[width=0.94\linewidth]{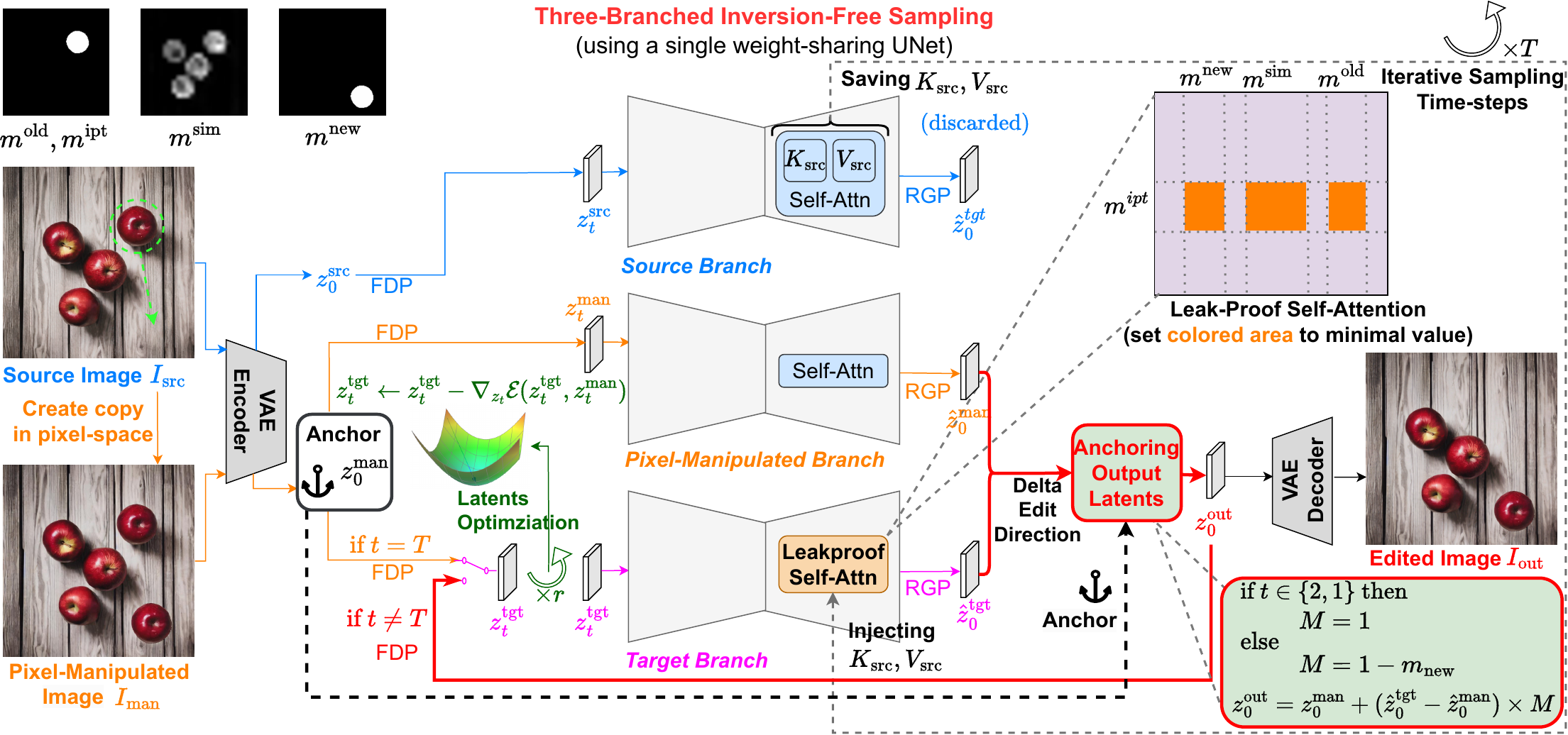}
    \caption{
        \textbf{Overview of PixelMan.} 
        An efficient inversion-free sampling approach for consistent image editing, which copies the object to target location in pixel-space, and ensure image consistency by anchoring to the latents of pixel-manipulated image. We design a leak-proof self-attention mechanism to achieve complete and cohesive inpainting by mitigating information leakage.
        }
    \label{fig:overview}
\end{figure*}

To enhance computational efficiency and preserve image consistency during object editing, we introduce \hbox{PixelMan}, an efficient \textit{inversion-free} and \textit{training-free} method that performs \textit{consistent object editing} with DMs via Pixel Manipulation and generation in \textit{few~inference~steps}.
The following subsections describe our proposed 
three-branched inversion-free sampling approach, 
leakproof self-attention technique,
and editing guidance with latents optimization method.

\subsection{Three-Branched Inversion-Free Sampling}

Our goal is to achieve the following three objectives with high efficiency:
(i)~consistent reproduction of the object and background; 
(ii)~object-background harmonization;
and (iii)~cohesive inpainting of the vacated location.
As the backbone of \hbox{PixelMan}, we propose a three-branched sampling paradigm that achieves these three objectives using a single pretrained DM, while also bypassing inversion to facilitate faster editing by reducing inference steps and NFEs.

Specifically, we utilize three separate branches: \textit{source branch}, \textit{pixel-manipulated branch}, and \textit{target branch}. 
Each branch maintains its own nosiy latents that is initialized, denoised (using the same UNet), and updated in different manners throughout the $T$ sampling time-steps $t\in [1,T]$.
We denote the noisy latents of the source branch, pixel-manipulated branch, and target branch at time-step $t$ respectively with $z_t^{\textrm{src}}$, $z_t^{\textrm{man}}$, $z_t^{\textrm{tgt}}$.

We create a \textit{pixel-manipulated image} $I_{\textrm{man}}$ by copying the source object to the target location in pixel-space.
For the object resizing task, we interpolate the object pixels based on the resizing scale before making a copy at the target location. For object pasting, the source object comes from a separate reference image $I_{\textrm{ref}}$, and is copied into the source image $I_{\textrm{src}}$ at the target location to create $I_{\textrm{man}}$.
Then, using the VAE encoder, we encode the pixel-manipulated image $I_{\textrm{man}}$ and the source image $I_{\textrm{src}}$ respectively into the pixel-manipulated latents $z_0^{\textrm{man}}$ and source latents $z_0^{\textrm{src}}$.

\subsubsection{Pixel-manipulated latents as anchor.}

At each time-step $t$ of our sampling process, our goal is to directly obtain a latent space estimation of the edited output image $I_{\textrm{out}}$ 
which we denote as output latents $z_0^{\textrm{out}}$. 
First, we ask the question of \textit{what would be a reasonable estimate of the output latents $z_0^{\textrm{out}}$}.
Intuitively, our estimation of output latents $z_0^{\textrm{out}}$ should be identical to the source latents $z_0^{\textrm{src}}$ so we can exactly reproduce the source image. 

However, we want to reproduce the source object at the new target location, so we set our estimation of the output latents $z_0^{\textrm{out}}$ to be identical to the pixel-manipulated latents $z_0^{\textrm{man}}$, which already have the source object reproduced at the target location through pixel manipulation. By using this naive estimation $z_0^{\textrm{out}}=z_0^{\textrm{man}}$, we can already effortlessly preserve the original background and consistently reproduce the object at the target location. Therefore, we refer to this pixel-manipulated latents $z_0^{\textrm{man}}$ as the anchor.

In addition to image consistency, we also want to achieve cohesive inpaitning of the vacated location, and harmonize the object and background with realistic effects. So, there should be a delta editing direction $\Delta z$ added on top of the anchor $z_0^{\textrm{man}}$ to achieve the inpainting and harmonization edits.
More concretely, at each time-step $t$, we set the output latent $z_0^{\textrm{out}}$ as:
\begin{equation}
    \label{eq:latent_out_1}
    z_0^{\textrm{out}} = z_0^{\textrm{man}} + \Delta z.
\end{equation}

With our simple estimation of output latents $z_0^{\textrm{out}}$ using the sum of the anchor $z_0^{\textrm{man}}$ and the delta edit direction $\Delta z$, we can preserve the object and background consistency without any inversion, which improves both the efficiency and image consistency by avoiding the computation bottleneck and accumulated reconstruction error of the DDIM inversion~\cite{dhariwal2021diffusion} process. 
Next, we introduce our method for obtaining the delta edit direction $\Delta z$.

\subsubsection{Obtaining delta edit direction.}
We aim to obtain the delta editing direction $\Delta z$ that can achieve cohesive inpainting of the vacated location, and harmonize the object and background with realistic effects (e.g., lighting, shadow, edge blending).
To achieve this, we propose to apply several editing guidance techniques (introduced in the later sections) for generating the inpainting and harmonization edits in the target branch, including leak-proof self-attention, editing guidance with latent optimization, and injection of source $K,V$ features into the target branch. Meanwhile, we keep the pixel-manipulated branch consistent with the anchor $z_0^{\textrm{man}}$, and obtain $\Delta z$ by finding the difference in the output of the two branches. 

Specifically, we calculate the difference between the
predicted target latents $\hat{z}_0^{\textrm{tgt}}$ from the target branch and predicted pixel-manipulated latents $\hat{z}_0^{\textrm{man}}$ from the pixel-manipulated branch:
\begin{equation}
    \label{eq:delta_z}
    \Delta z = \hat{z}_0^{\textrm{tgt}} - \hat{z}_0^{\textrm{man}}.
\end{equation}

To obtain $\hat{z}_0^{\textrm{man}}$ from the \textbf{pixel-manipulated branch}, 
we always analytically compute the noisy latents $z_t^{\textrm{man}}$ at each sampling time-step $t$ from the pixel-manipulated latents $z_0^{\textrm{man}}$ 
(i.e., the anchor that ensures consistency to $I_{\textrm{man}}$)
which has already reproduced object at the target location and the original background.

Specifically, at each time-step $t$, we first follow the FDP equation to obtain $z_t^{\textrm{man}}$ by adding random Gaussian noise $\epsilon\!\sim\!\mathcal{N}(0, I)$ to $z_0^{\textrm{src}}$:
\begin{equation}
    \label{eq:fdp_man}
    z^{\textrm{man}}_t = \sqrt{\bar{\alpha}_t} \times z^{\textrm{man}}_0 + \sqrt{1-\bar{\alpha}_t} \times \epsilon.
\end{equation}
Then, we pass the noisy source latents $z^{\textrm{man}}_t$ to the denoising UNet (parameterized by $\theta$) to get the predicted noise $\hat{\epsilon}^{\textrm{man}}_t$ at time-step t:
\begin{equation}
    \label{eq:unet_man}
\hat{\epsilon}^{\textrm{man}}_t = \textrm{UNet}(z^{\textrm{man}}_t, t).
\end{equation}
Finally, we obtain the predicted pixel-manipulated latents $\hat{z}_0^{\textrm{man}}$ 
based on the noisy pixel-manipulated latents $z^{\textrm{man}}_t$ and the UNet predicted noise $\hat{\epsilon}^{\textrm{man}}_t$ at time-step t, using the Reverse Generative Process (RGP) (more details in Eq.~(\ref{eq:rgp}) in the Appendix):
\begin{equation}
\label{eq:rgp_man}
\hat{z}_0^{\textrm{man}} = \textrm{RGP}(z_t^{\textrm{man}}, \hat{\epsilon}_t^{\textrm{man}}, t).
\end{equation}

Next, we obtain $\hat{z}_0^{\textrm{tgt}}$ from the \textbf{target branch}. Before the initial timestep $t=T$, we initialize the target latents $z_0^{\textrm{tgt}}$ to be the same as $z_0^{\textrm{man}}$ which corresponds to the anchor $I_{\textrm{man}}$. 
In contrast, at each sampling time-step $t$, 
we instead utilize the FDP similar to Eq.~(\ref{eq:fdp_man}) to analytically compute 
the noisy target latents $z_t^{\textrm{tgt}}$ from the 
estimated output latents $z_0^{\textrm{out}}$ of previous time-step.

Then, $z_t^{\textrm{tgt}}$ is updated with latents optimization (detailed in ``Editing Guidance with Latents Optimization''). Next, we pass $z_t^{\textrm{tgt}}$ along with the saved source branch $K_{\textrm{src}}$, $V_{\textrm{src}}$ (detailed in ``Feature-preserving source branch'') to the UNet to obtain the predicted noise $\hat{\epsilon}^{\textrm{src}}_t$, where 
$\hat{\epsilon}^{\textrm{src}}_t = \textrm{UNet}(z^{\textrm{src}}_t, t; \{K_{\textrm{src}}, V_{\textrm{src}}\}).$
Next, we obtain the predicted target latents $\hat{z}^{\textrm{tgt}}_0$ using the RGP similar to Eq.~(\ref{eq:rgp_man}).

After calculating both $\hat{z}_0^{\textrm{tgt}}$ and $\hat{z}_0^{\textrm{man}}$, we finally obtain the delta editing direction $\Delta z$. To estimate the output image, we combine the anchor $z_0^{\textrm{man}}$ and the delta editing direction in Eq.~(\ref{eq:delta_z}), while applying a masked-blending approach with mask $(1-m_{\textrm{new})}$ (i.e., the object target location):
\begin{equation}
\label{eq:mask_blend}
    z_0^{\textrm{out}} = z_0^{\textrm{man}} + (\hat{z}_0^{\textrm{tgt}} - \hat{z}_0^{\textrm{man}}) \times (1 - m_{\textrm{new}}).
\end{equation}
The masked-blending is applied throughout the sampling time-steps to remove the delta editing direction $\Delta z$ in the target location, and only use the anchor $z_0^{\textrm{man}}$ to achieve object consistency. While allowing $\Delta z$ to change the background for inpainting and harmonization.
For the last few time-steps no masking is applied, which encourages seamless object-background blending and allows the DM to refine the details of the output image.

\subsubsection{Feature-preserving source branch.}
At each time-step $t$, we always analytically compute the noisy source latents $z_t^{\textrm{src}}$ from $z_0^{\textrm{src}}$
(making $z_t^{\textrm{src}}$ consistent with $z_0^{\textrm{src}}$). 
Specifically, at each time-step $t$, we first follow the FDP equation similar to Eq.~(\ref{eq:fdp_man}) to obtain $z_t^{\textrm{src}}$ by adding random Gaussian noise $\epsilon\!\sim\!\mathcal{N}(0, I)$ to $z_0^{\textrm{src}}$.
Then, we pass the noisy source latents $z^{\textrm{src}}_t$ to the denoising UNet to get the predicted noise $\hat{\epsilon}^{\textrm{src}}_t$ at time-step t:
$\hat{\epsilon}^{\textrm{src}}_t, \{K_{\textrm{src}}, V_{\textrm{src}}\} = \textrm{UNet}(z^{\textrm{src}}_t, t).$
Note that the $\hat{\epsilon}^{\textrm{src}}_t$ is discarded here, and we save the self-attention $K_{\textrm{src}}$ and $V_{\textrm{src}}$ matrices from the source branch and inject%
\footnote{
    Injection refers to overwriting the respective attention K and V matrices with the previously saved ones.
}
them back during the UNet call on $z_t^{\textrm{tgt}}$ in the target branch, which is inspired by the 
mutual self-attention technique proposed in~\cite{cao2023masactrl}.
The saved $K_{\textrm{src}}$ and $V_{\textrm{src}}$ preserve the original visual details from $I_{\textrm{src}}$, and the injection into target branch serves as context for generating appropriate harmonization effects (e.g., lighting, shadow, and edge blending), and also for inpainting the vacated area. 

\subsection{Leak-Proof Self-Attention}
\label{sec:SAM}

Our objective is to achieve complete and cohesive inpainting of the vacated region after the edited object moves out.
However, current methods often either struggle to remove all traces of the object 
(e.g., object is not entirely removed in columns (d), (e), and (f) of Fig.~\ref{fig:examples_comparisons_1} by the SOTA method DiffEditor~\cite{mou2024diffeditor}),
or hallucinate new unwanted artifacts in the vacated region.
We attribute these issues to information leakage from similar objects through the SA mechanism~\cite{dahary2024yourself},
and propose a leak-proof self-attention technique that prevents the attention to source object, target object, and similar objects in the image. Leak-proof SA leverages and controls the inter-region dependencies captured by SA to alleviate information leakage.

Intuitively, areas $m_\textrm{old}$, $m_\textrm{new}$, and $m_\textrm{sim}$ all contain information about the to-be-edited object,
and this information can be leaked to area $m_\textrm{ipt}$ through the SA mechanism,
where 
$m_\textrm{old}$ is mask of to-be-edit object at the \textbf{source/old} location;
$m_\textrm{new}$ is $m_\textrm{old}$ shifted to the the \textbf{target/new} location;
$m_\textrm{sim}$ is mask of other \textbf{similar} objects to the to-be-edited object 
(e.g., other apples in the multi-apple image in Fig.~\ref{fig:overview};
see details on how to automatically obtain $m_\textrm{sim}$ in the Appendix);
and
$m_\textrm{ipt}$ equals the mask from $(m_\textrm{old}-m_\textrm{new})$ 
which represents the \textbf{to-be-inpainted} vacated region.
To minimize the information leakage 
of the to-be-edited object and similar objects on
the inpainted region, 
we strategically reset the corresponding elements 
(i.e., $m_\textrm{old} \cup m_\textrm{new} \cup m_\textrm{sim}$)
in $QK^T$ to a minimal value (i.e., $-\infty$).
This strategy is activated for the target branch UNet call in all SA layers and at all time-steps to mitigate leakage and enable cohesive inpainting.

\subsection{Editing Guidance with Latents Optimization}
\label{sec:our_gsn}

\citet{mou2024dragondiffusion} propose a set of energy functions, which enforce feature correspondence to provide editing guidance. We utilize the same energy functions from DragonDiffusion~\cite{mou2024dragondiffusion} to obtain additional editing guidance for object generation, harmonization, inpainting, and background consistency in our target branch.

Moreover, we aim to improve the efficiency of editing guidance. 
\citet{mou2024diffeditor} showed that having a refinement loop 
that applies the editing guidance multiple times at a single time-step 
significantly enhances the performance.
However, EG-based methods update the predicted noise $\epsilon$ 
while the loss function operates on the noisy latents $z_t$. 
To bridge this gap and propagate the guidance from $\epsilon$ to $z_t$, 
\cite{mou2024diffeditor} introduced ``time travel''
that requires a repetitive second round of DDIM inversion~\cite{dhariwal2021diffusion}.
Therefore, the EG-based editing guidance can be computationally expensive in terms of NFEs. 

Different from EG-based methods which updates the predicted noise $\epsilon$, 
we propose a more efficient refinement strategy by applying the editing guidance directly to the target noisy latents $z^{\textrm{tgt}}_t$ at each time-step $t$:
\begin{equation}
\label{eq:our_update}
    z_t^{\textrm{tgt}}  \leftarrow z_t^{\textrm{tgt}} - \nabla_{z_t}\mathcal{E}(z_t^{\textrm{tgt}}, z_t^{\textrm{man}}).
\end{equation}

The direct application of guidance at $z^{\textrm{tgt}}_t$ eliminates the need for expensive tricks such as time travel. 
Our strategy is grounded in a solid theoretical foundation,
as it leverages \hbox{inference-time} gradient descent optimization, which is also known as GSN~\cite{chefer2023attendandexcite} in the text-to-image generation DM literatures.
Furthermore, we demonstrate this efficiency through an ablation experiment in 
the Appendix.

In a refinement loop, 
we iteratively compute and apply the edit guidance to the target noisy latents$z_t^{\textrm{tgt}}$ as in Eq.~(\ref{eq:our_update}).
This iterative guidance process guarantees progressive deviation of $z_t^{\textrm{tgt}}$ from $z_t^{\textrm{man}}$,
resulting in the removal of the object from its old location and 
harmonization of the object with the context at its new location.

\algrenewcommand\algorithmicensure{\textbf{Output:}}

\begin{algorithm}[t]
    \caption{Algorithm Overview of PixelMan}
    \label{alg:ours}
    \small
    \begin{algorithmic}[1]
        \Require $\textrm{VAE Encoder: }z_t = \mathcal{E}(I)$;
        \quad $\textrm{VAE Decoder: }I = \mathcal{D}(z_t)$
        \Require $\hat{\epsilon}_t, \{K, V\} = \textrm{UNet}(z_t, t)$
        \Require $z_t = \textrm{FDP}(z_0, \epsilon)$; 
        \quad $\hat{z}_0 = \textrm{RGP}(z_t, \hat{\epsilon}_t, t)$
        \Require $z_0^{\textrm{src}}=\mathcal{E}(I_{\textrm{src}})$;
        \quad $z_0^{\textrm{man}}=\mathcal{E}(I_{\textrm{man}})$;
        \quad $z_0^{\textrm{out}} =\mathcal{E}(I_{\textrm{man}})$
        \Require $\textrm{source, target, and inpaint mask: }
        m_{\textrm{old}}$, $m_{\textrm{new}}$, $m_{\textrm{ipt}}$
        
        \For{time-step $t\in \{T,T-1,...,1\}$}
            \State $\epsilon \sim \mathcal{N}(0, I)$
            \State $z_t^{\textrm{src}} = \textrm{FDP}(z_0^{\textrm{src}}, \epsilon$)
            \State $z_t^{\textrm{man}} = \textrm{FDP}(z_0^{\textrm{man}}, \epsilon$)
            \State $z_t^{\textrm{tgt}} = \textrm{FDP}(z_0^{\textrm{out}}, \epsilon$)
        
            \For{repeat $r$} \Comment{latents optimization}
                \State $z_t^{\textrm{tgt}} \leftarrow z_t^{\textrm{tgt}} - \nabla_{z_t}\mathcal{E}(z_t^{\textrm{tgt}}, z_t^{\textrm{man}})$
            \EndFor
            
            \State $\hat{\epsilon}_t^{\textrm{man}} = \textrm{UNet}(z_t^{\textrm{man}}, t)$
            \State $\hat{\epsilon}_t^{\textrm{src}}, \{K_{\textrm{src}}, V_{\textrm{src}}\} = \textrm{UNet}(z_t^{\textrm{src}}, t)$ \Comment{save K,V}
            \State $\hat{\epsilon}_t^{\textrm{tgt}} = 
            \textrm{UNet}(z_t^{\textrm{tgt}}, t;\{K_{\textrm{src}}, V_{\textrm{src}}\})$ \Comment{apply leak-proof SA}

            \State $\hat{z}_0^{\textrm{man}} = \textrm{RGP}(z_t^{\textrm{man}}, \hat{\epsilon}_t^{\textrm{man}}, t)$
            \State $\hat{z}_0^{\textrm{tgt}} = \textrm{RGP}(z_t^\textrm{{tgt}}, \hat{\epsilon}_t^{\textrm{tgt}}, t)$
            
            \If{$t \in \{2,1\}$} \Comment{i.e., last few time-steps}
                \State $z_0^{\textrm{out}} = z_0^{\textrm{man}} + (\hat{z}_0^{\textrm{tgt}} - \hat{z}_0^{\textrm{man}})$ \Comment{no masked-blending}
            \Else
                \State $z_0^{\textrm{out}} = z_0^{\textrm{man}} + (\hat{z}_0^{\textrm{tgt}} - \hat{z}_0^{\textrm{man}}) \times (1-m_{\textrm{new}})$ \Comment{with mask}
            \EndIf
        \EndFor
        
        \Ensure $I_{\textrm{out}} = \mathcal{D}(z_0^{\textrm{out}})$ \Comment{the edited output image}
    \end{algorithmic}
\end{algorithm}

\section{Experiments}
\label{sec:exp}

\begin{figure*}[!ht]
    \vspace{-4mm}
    \centering
    \setlength{\tabcolsep}{0.4pt}
    \renewcommand{\arraystretch}{0.4}
    {\footnotesize
    \begin{tabular}{c c c c c c c c c c}
        &
        \multicolumn{1}{c}{(a)} &
        \multicolumn{1}{c}{(b)} &
        \multicolumn{1}{c}{(c)} &
        \multicolumn{1}{c}{(d)} &
        \multicolumn{1}{c}{(e)} &
        \multicolumn{1}{c}{(f)} &
        \multicolumn{1}{c}{(g)} &
        \multicolumn{1}{c}{(h)} \\

        {\raisebox{0.34in}{
        \multirow{1}{*}{\rotatebox{0}{Input}}}} &
        \includegraphics[width=0.09\textwidth]{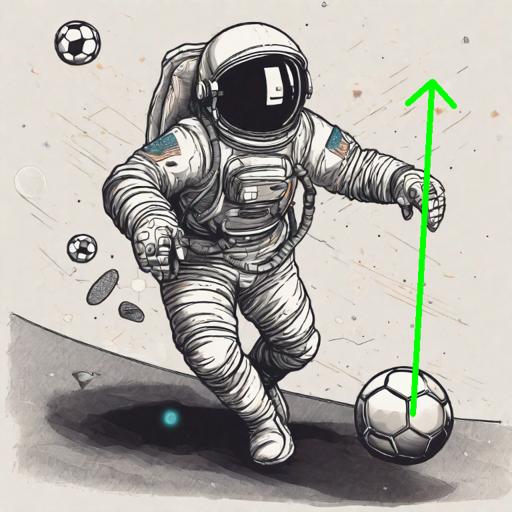} &
        \includegraphics[width=0.09\textwidth]{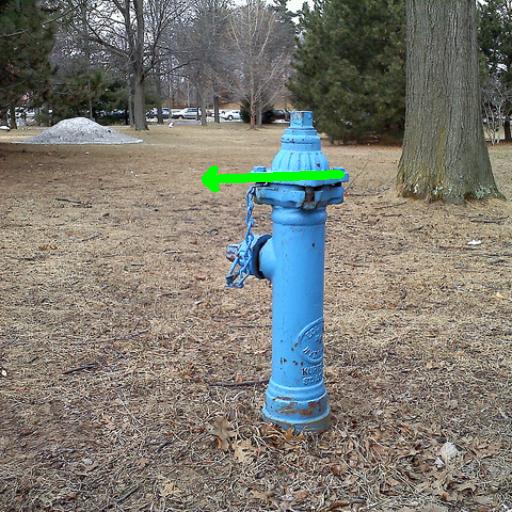} &
        \includegraphics[width=0.09\textwidth]{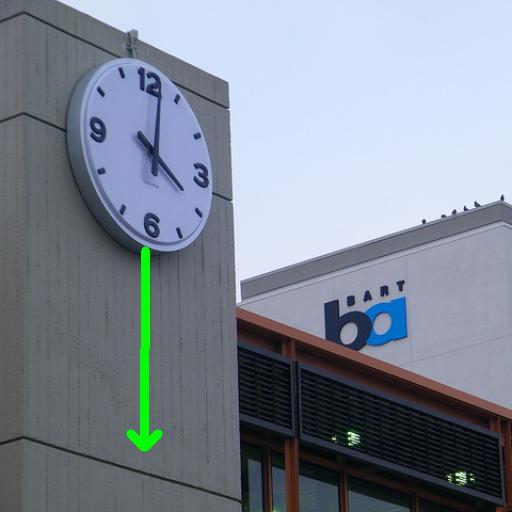} &
        \includegraphics[width=0.09\textwidth]{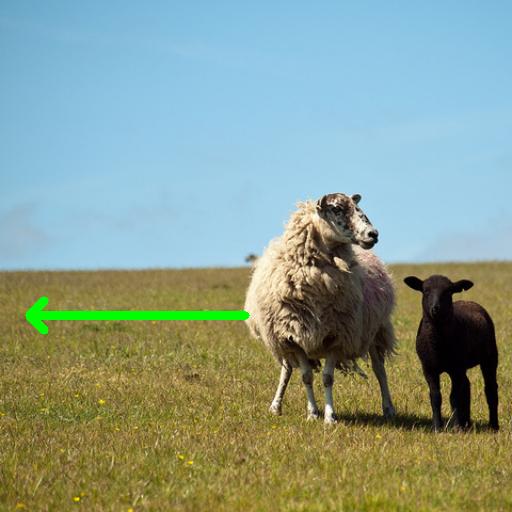} &
        \includegraphics[width=0.09\textwidth]{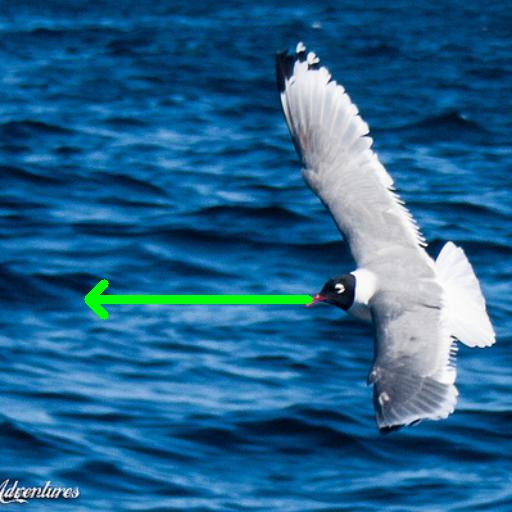} &
        \includegraphics[width=0.09\textwidth]{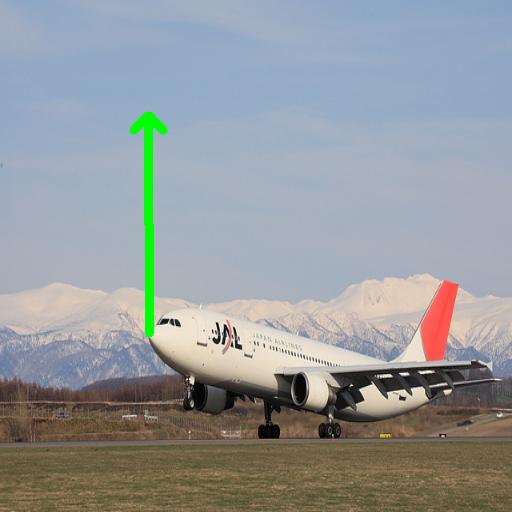} &
        \includegraphics[width=0.09\textwidth]{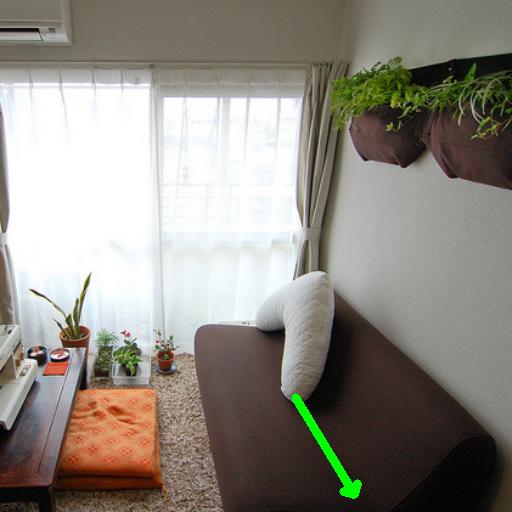} &
        \includegraphics[width=0.09\textwidth]{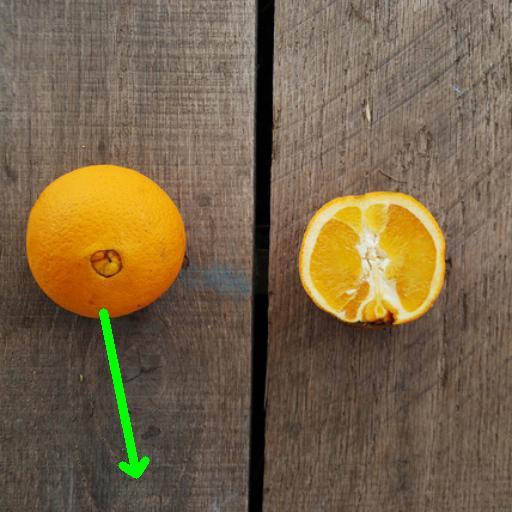} &\\

        {\raisebox{0.42in}{\multirow{1}{*}{\begin{tabular}{c}SDv2-Inpainting\\+AnyDoor \\ (50 steps, 15s)\end{tabular}}}} &
        \includegraphics[width=0.09\textwidth]{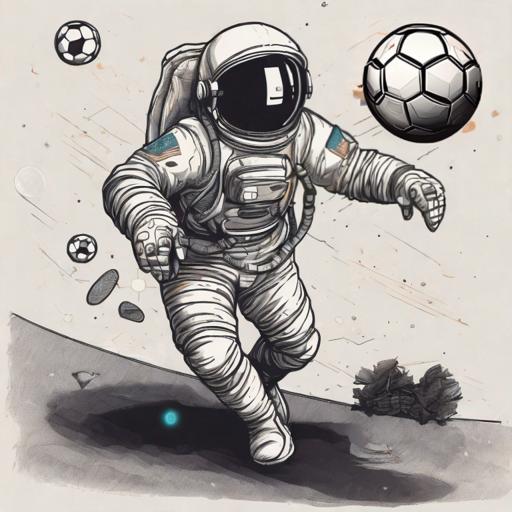} &
        \includegraphics[width=0.09\textwidth]{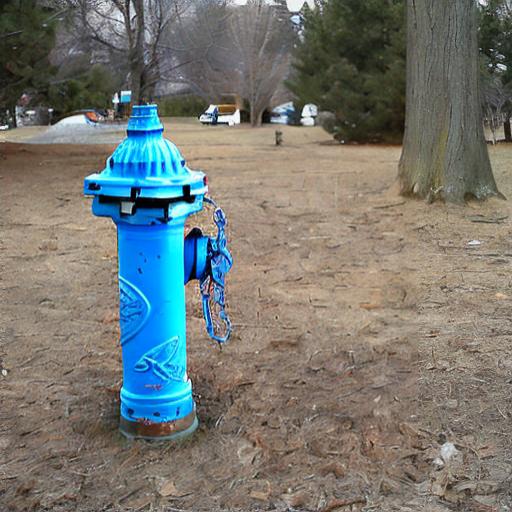} &
        \includegraphics[width=0.09\textwidth]{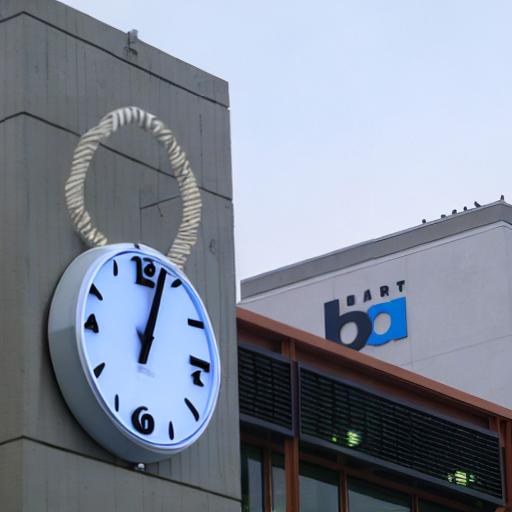} &
        \includegraphics[width=0.09\textwidth]{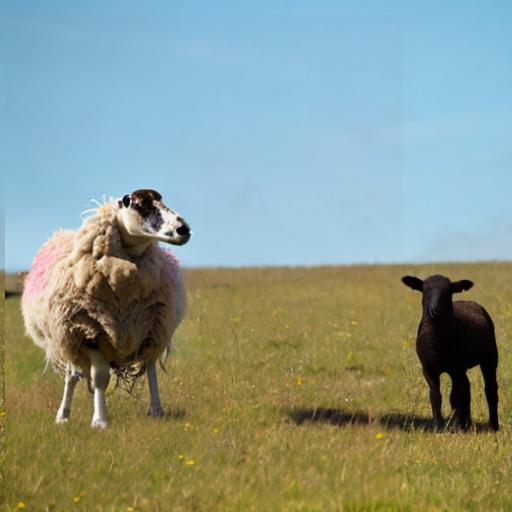} &
        \includegraphics[width=0.09\textwidth]{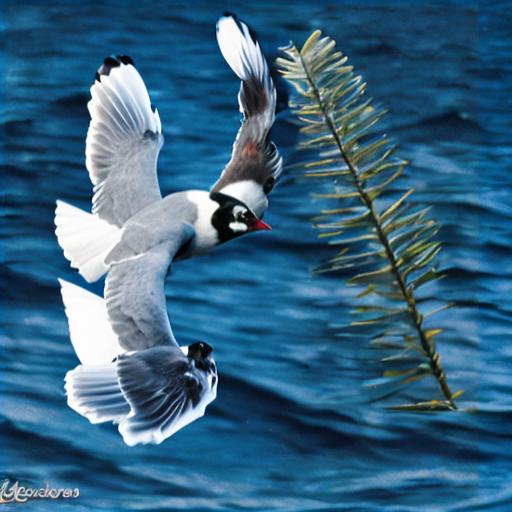} &
        \includegraphics[width=0.09\textwidth]{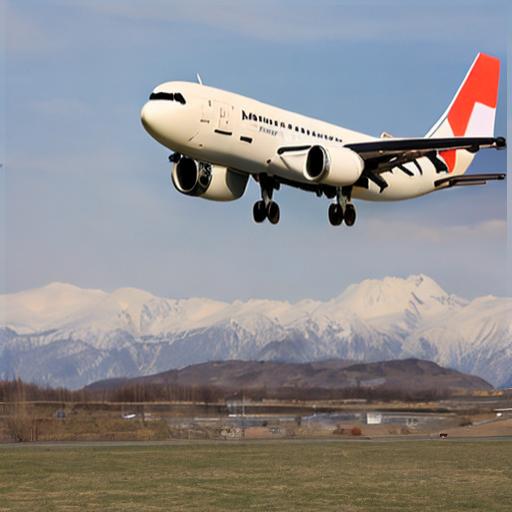} &
        \includegraphics[width=0.09\textwidth]{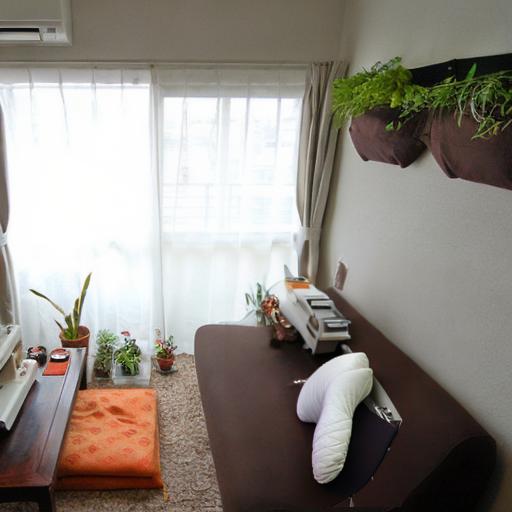} &
        \includegraphics[width=0.09\textwidth]{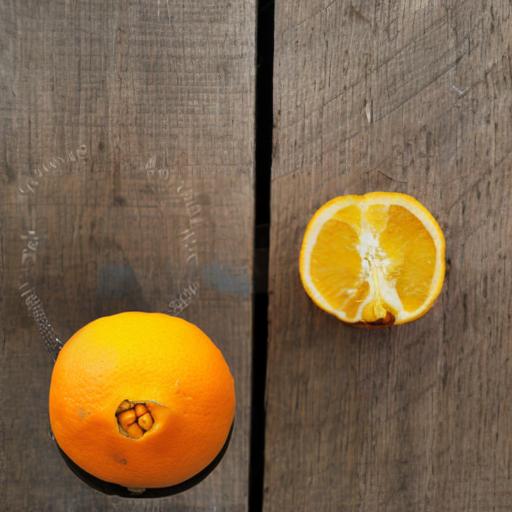} &\\

        {\raisebox{0.37in}{\multirow{1}{*}{\begin{tabular}{c}SelfGuidance \\ (50 steps, 11s)\end{tabular}}}} &
        \includegraphics[width=0.09\textwidth]{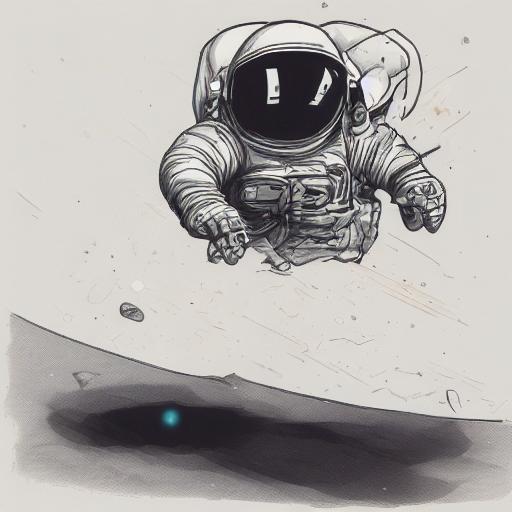} &
        \includegraphics[width=0.09\textwidth]{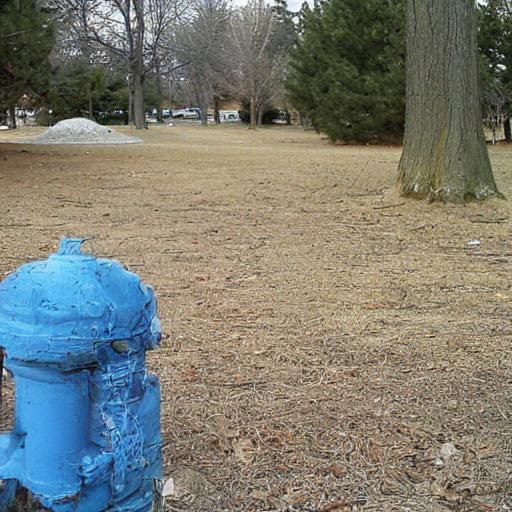} &
        \includegraphics[width=0.09\textwidth]{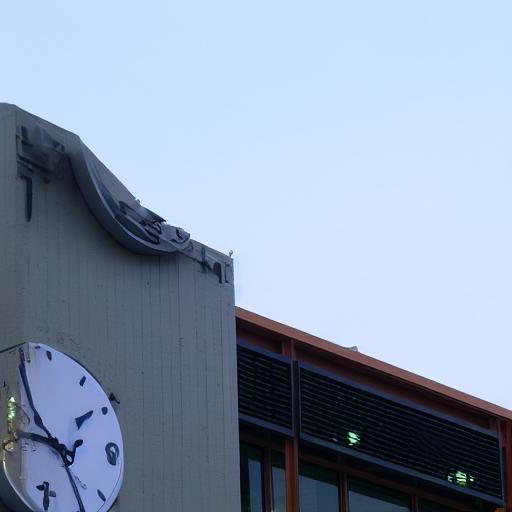} &
        \includegraphics[width=0.09\textwidth]{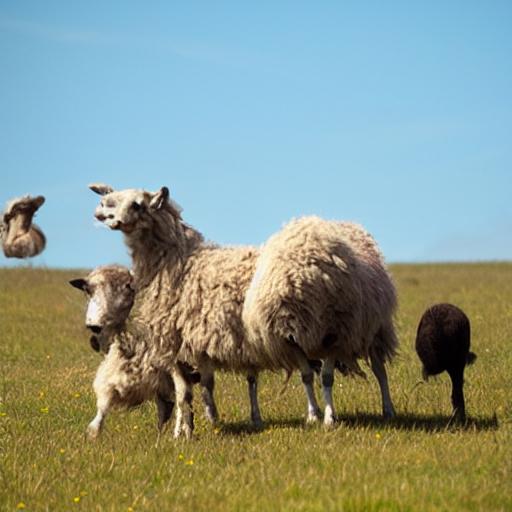} &
        \includegraphics[width=0.09\textwidth]{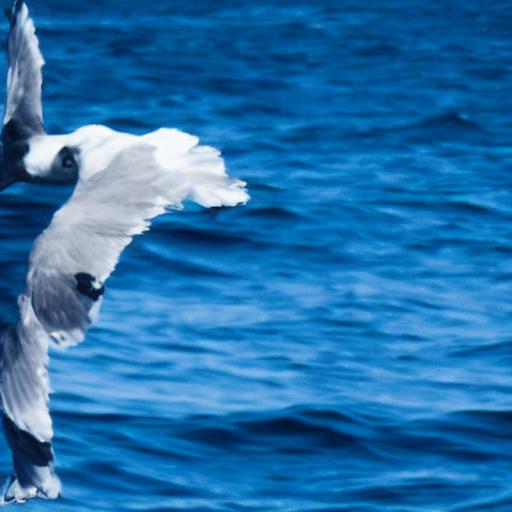} &
        \includegraphics[width=0.09\textwidth]{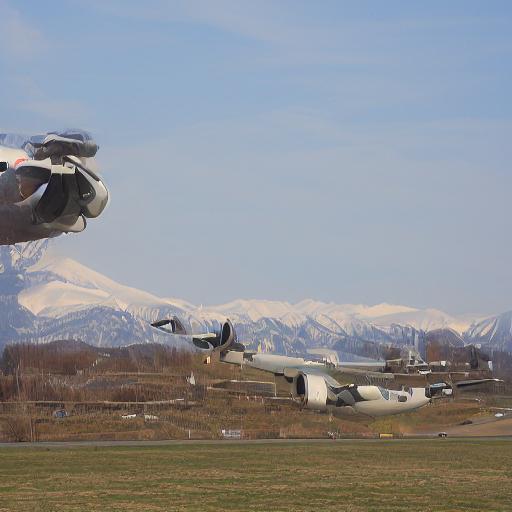} &
        \includegraphics[width=0.09\textwidth]{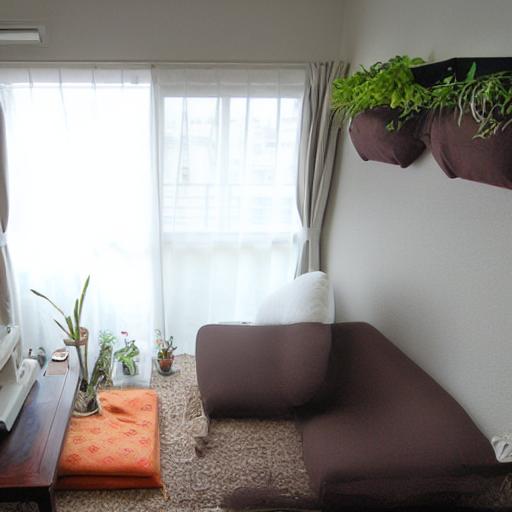} &
        \includegraphics[width=0.09\textwidth]{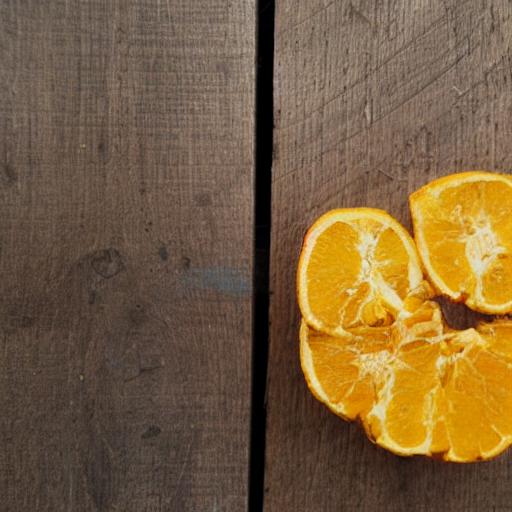} &\\
        
        {\raisebox{0.37in}{\multirow{1}{*}{\begin{tabular}{c}DragonDiffusion \\ (50 steps, 23s)\end{tabular}}}} &
        \includegraphics[width=0.09\textwidth]{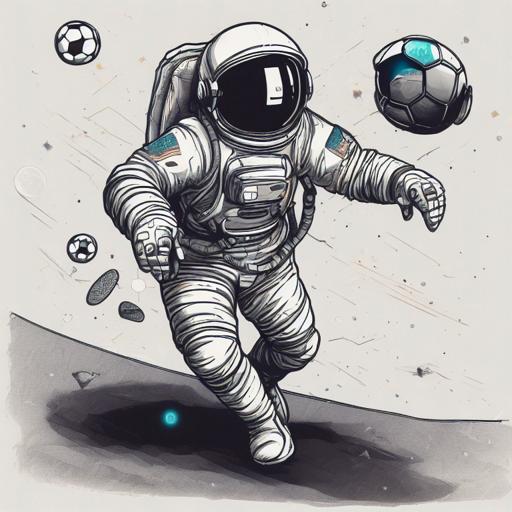} &
        \includegraphics[width=0.09\textwidth]{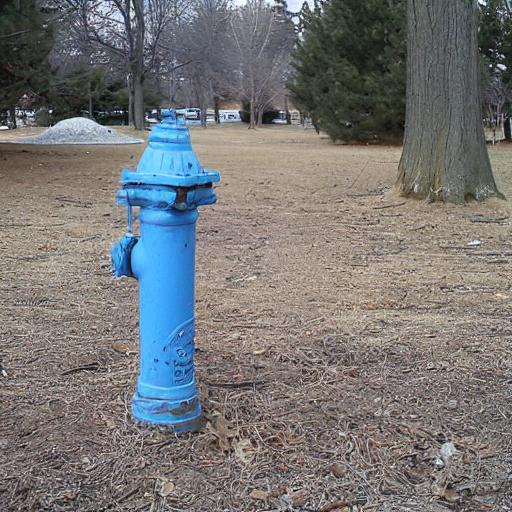} &
        \includegraphics[width=0.09\textwidth]{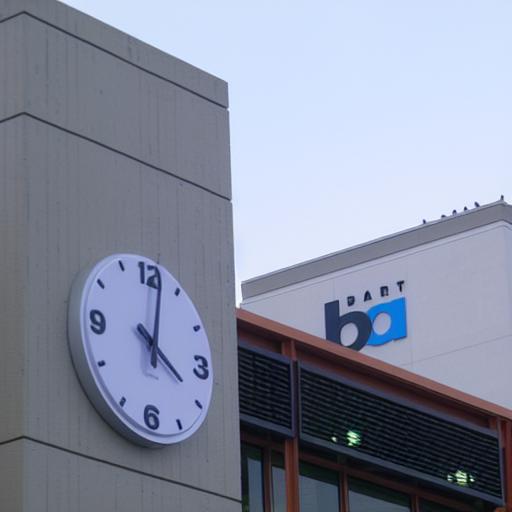} &
        \includegraphics[width=0.09\textwidth]{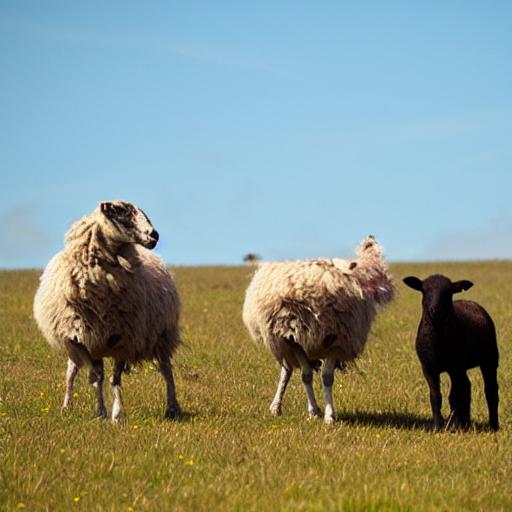} &
        \includegraphics[width=0.09\textwidth]{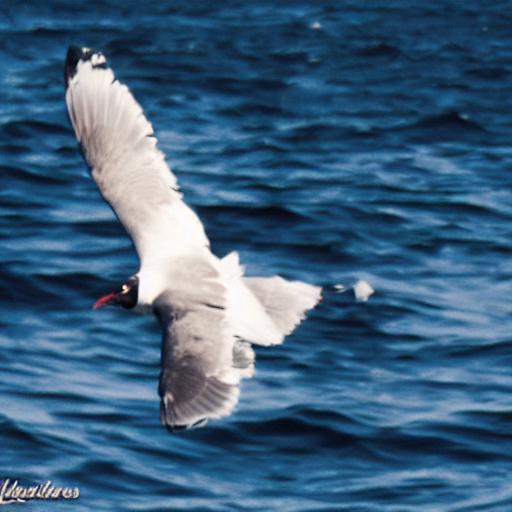} &
        \includegraphics[width=0.09\textwidth]{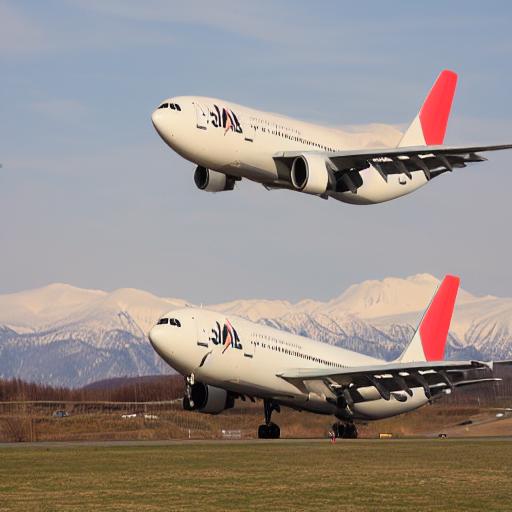} &
        \includegraphics[width=0.09\textwidth]{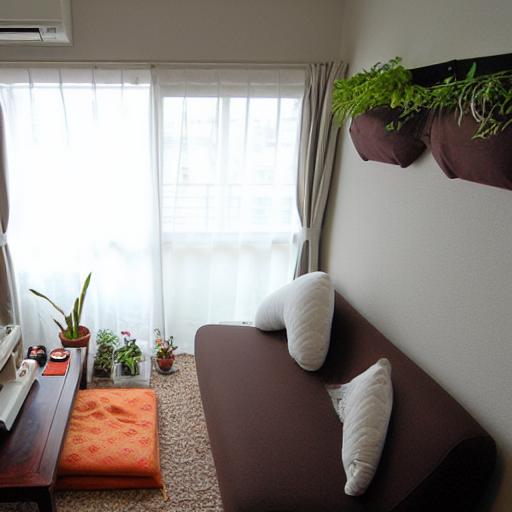} &
        \includegraphics[width=0.09\textwidth]{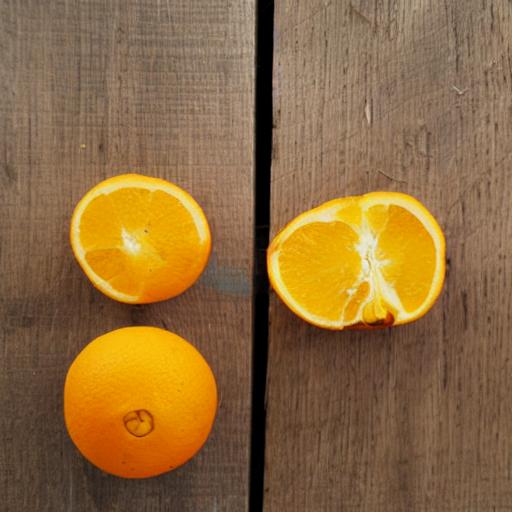} &\\
        
        {\raisebox{0.37in}{\multirow{1}{*}{\begin{tabular}{c}DiffEditor \\ (50 steps, 24s)\end{tabular}}}} &
        \includegraphics[width=0.09\textwidth]{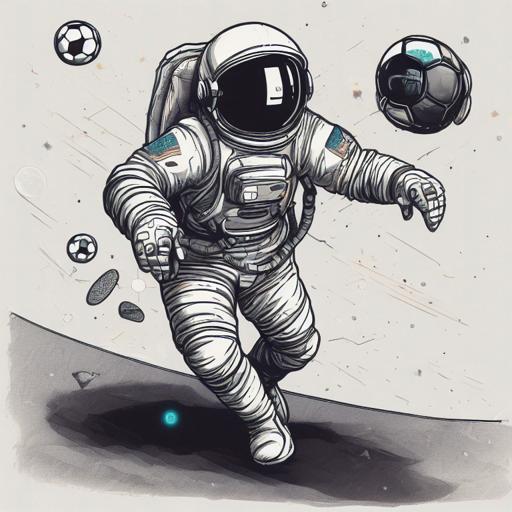} &
        \includegraphics[width=0.09\textwidth]{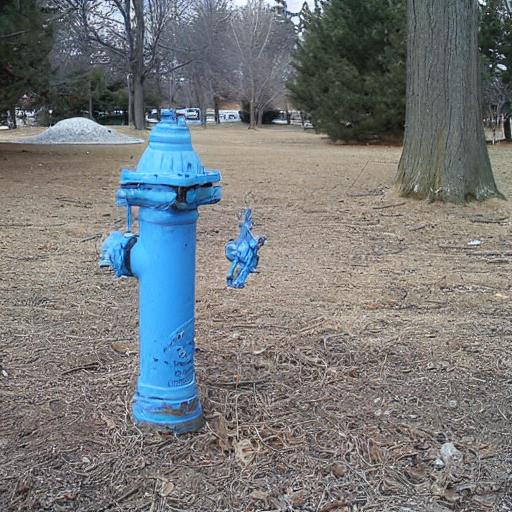} &
        \includegraphics[width=0.09\textwidth]{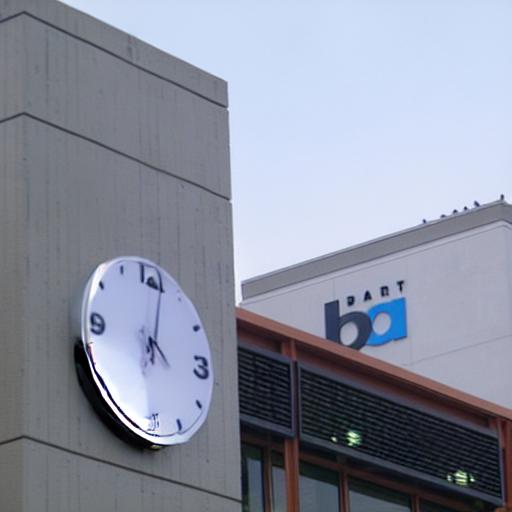} &
        \includegraphics[width=0.09\textwidth]{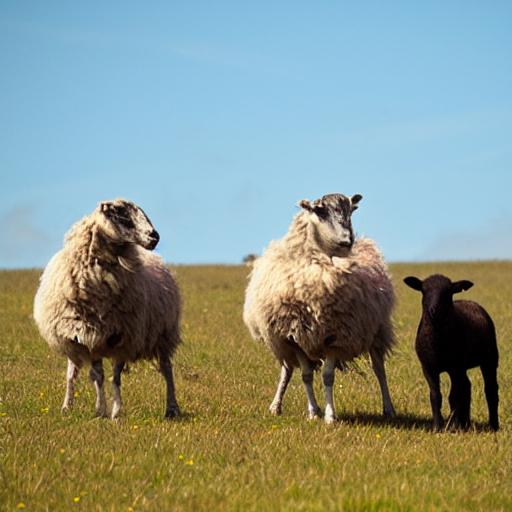} &
        \includegraphics[width=0.09\textwidth]{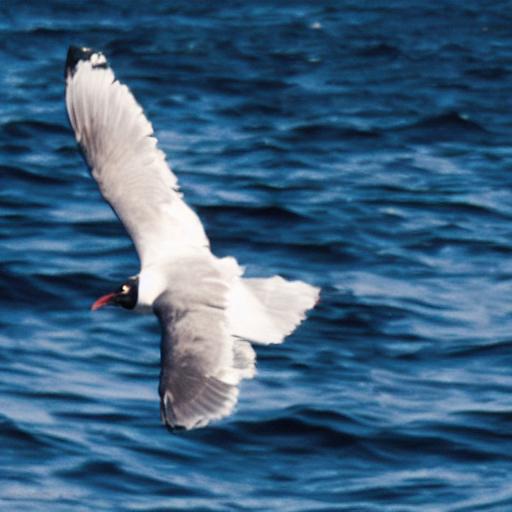} &
        \includegraphics[width=0.09\textwidth]{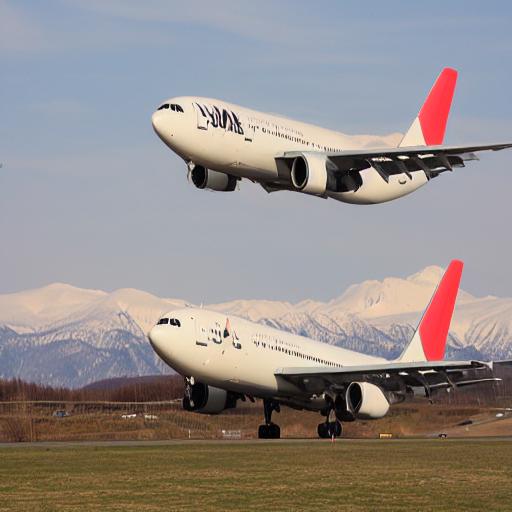} &
        \includegraphics[width=0.09\textwidth]{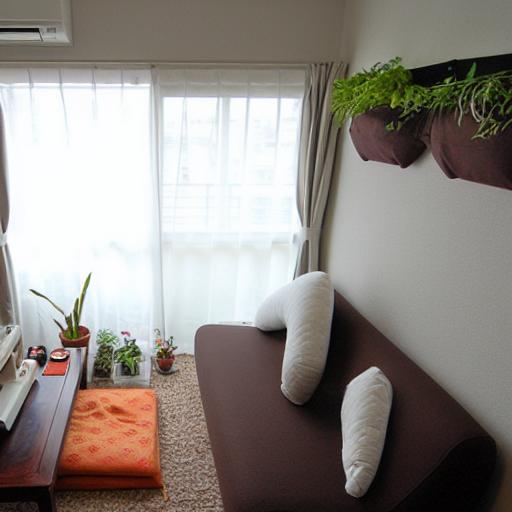} &
        \includegraphics[width=0.09\textwidth]{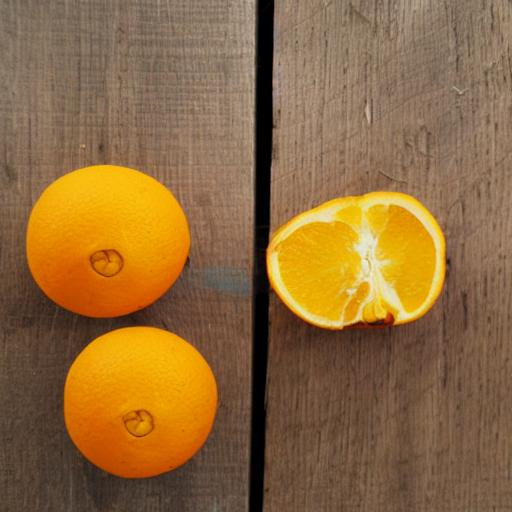} &\\

        {\raisebox{0.37in}{\multirow{1}{*}{\begin{tabular}{c}DiffEditor \\ (16 steps, 9s)\end{tabular}}}} &
        \includegraphics[width=0.09\textwidth]{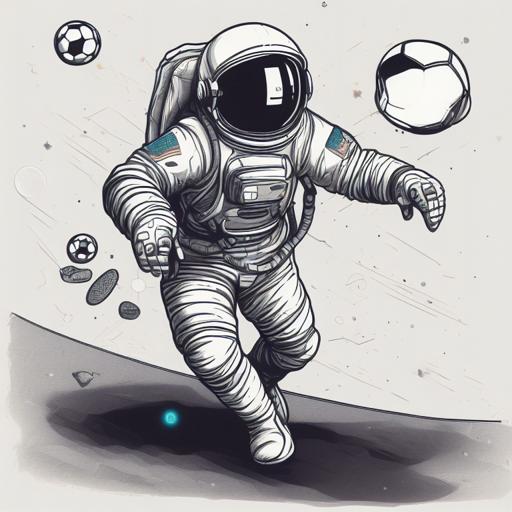} &
        \includegraphics[width=0.09\textwidth]{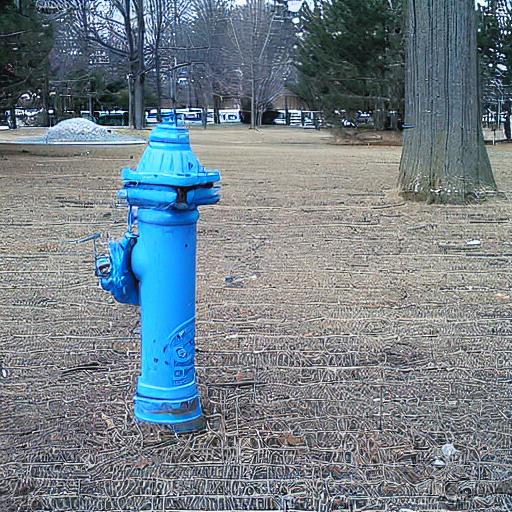} &
        \includegraphics[width=0.09\textwidth]{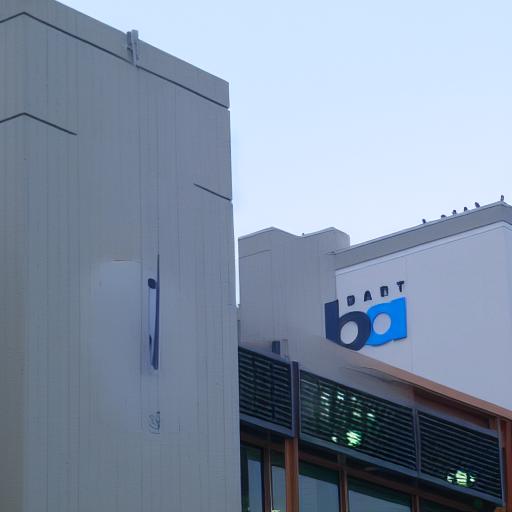} &
        \includegraphics[width=0.09\textwidth]{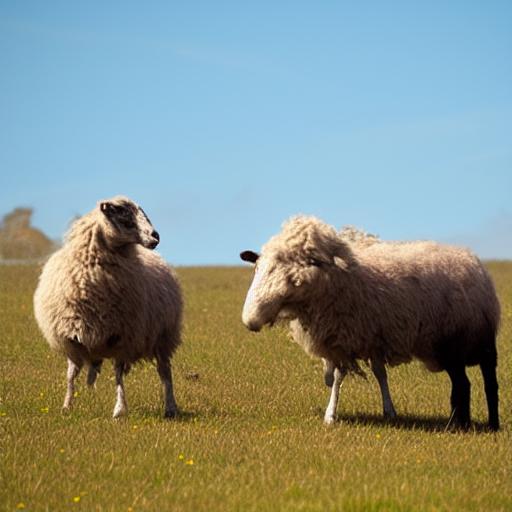} &
        \includegraphics[width=0.09\textwidth]{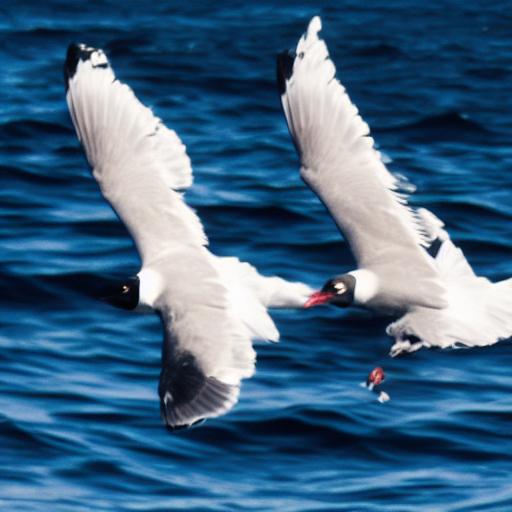} &
        \includegraphics[width=0.09\textwidth]{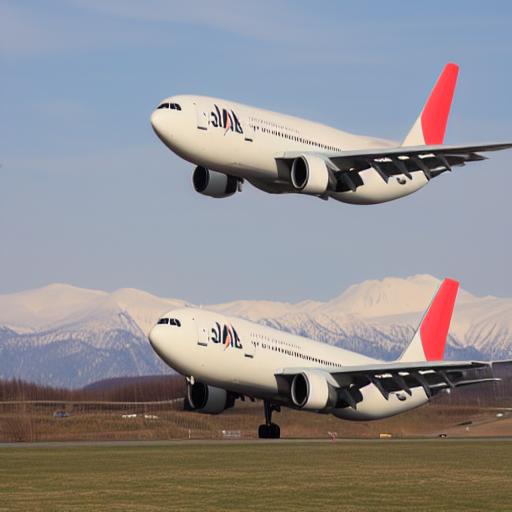} &
        \includegraphics[width=0.09\textwidth]{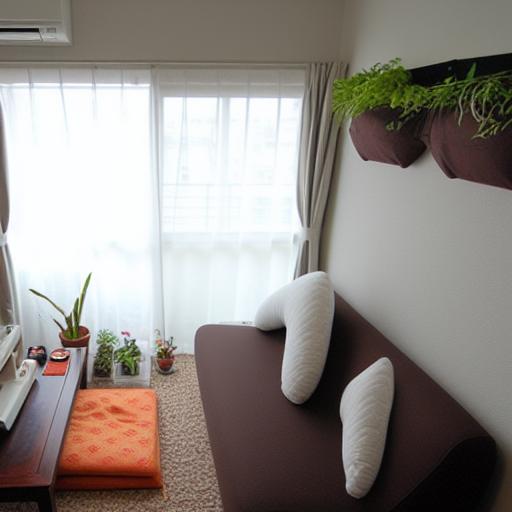} &
        \includegraphics[width=0.09\textwidth]{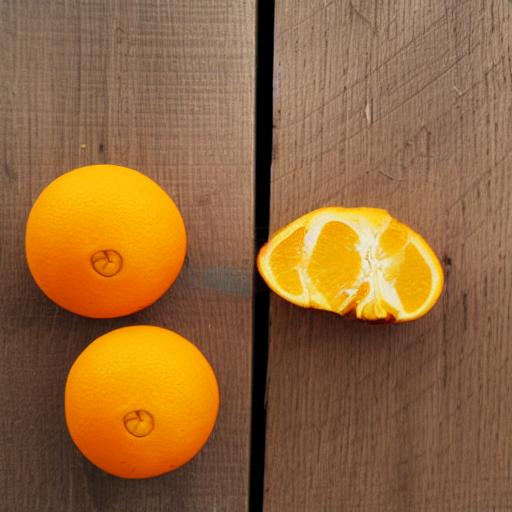} &\\

        {\raisebox{0.37in}{\multirow{1}{*}{\begin{tabular}{c}\textbf{PixelMan} \\ (16 steps, 9s)\end{tabular}}}} &
        \includegraphics[width=0.09\textwidth]{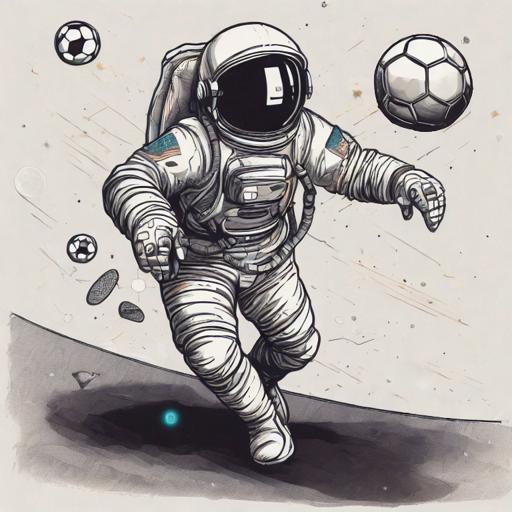} &
        \includegraphics[width=0.09\textwidth]{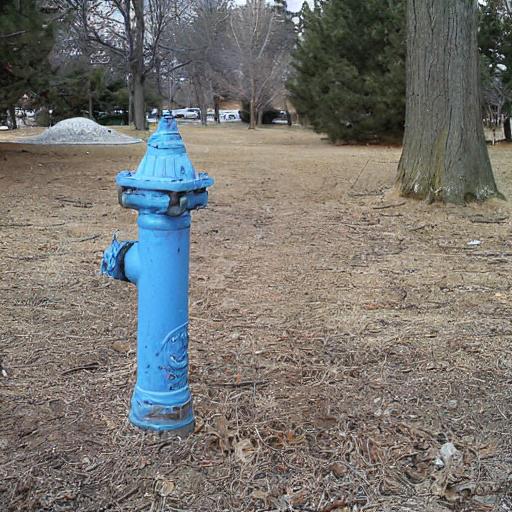} &
        \includegraphics[width=0.09\textwidth]{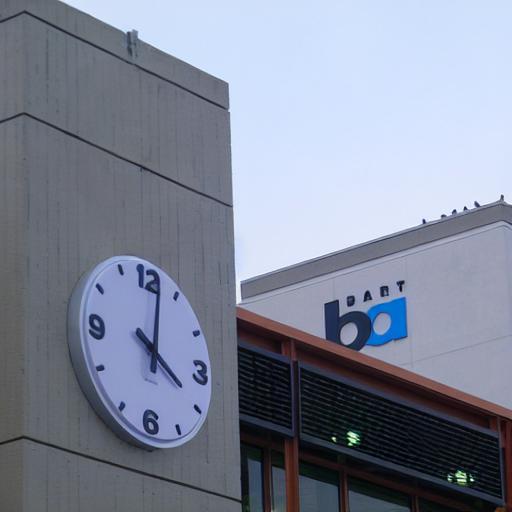} &
        \includegraphics[width=0.09\textwidth]{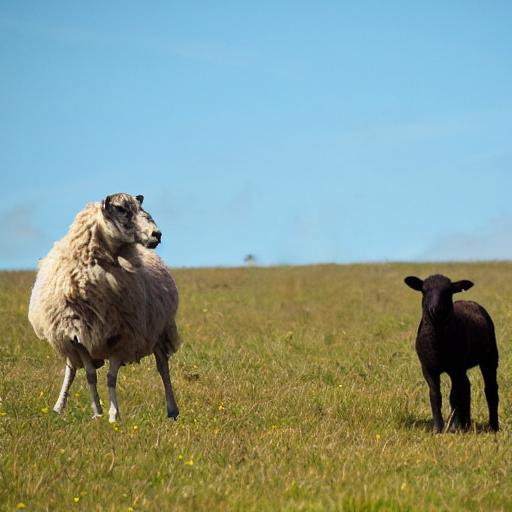} &
        \includegraphics[width=0.09\textwidth]{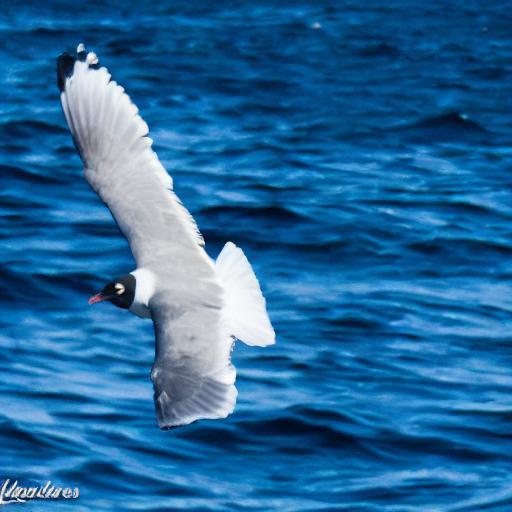} &
        \includegraphics[width=0.09\textwidth]{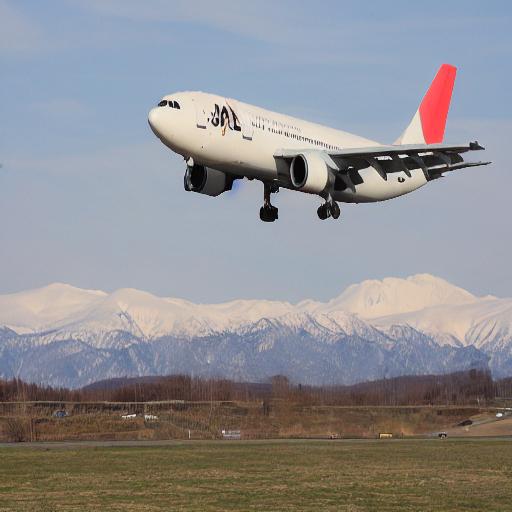} &
        \includegraphics[width=0.09\textwidth]{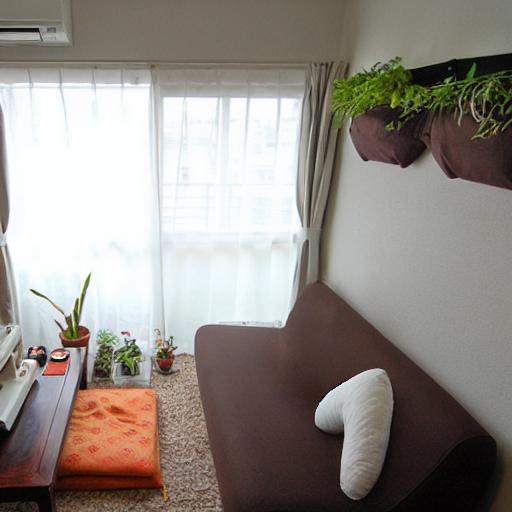} &
        \includegraphics[width=0.09\textwidth]{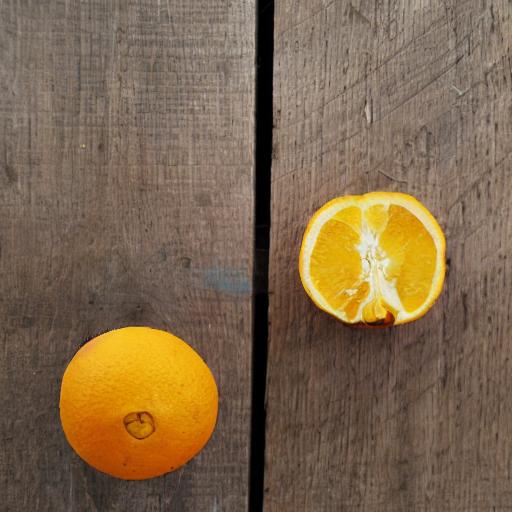} &\\
        
    \end{tabular}
    }
    \caption{
        \textbf{Visual comparisons on COCOEE dataset.} 
        PixelMan achieves consistent object editing for object repositioning with lower latency and fewer inference steps, while better preserving image consistency and achieving cohesive inpainting.
        }
    \label{fig:examples_comparisons_1}
\end{figure*}

\newlength{\imagewidth}
\setlength{\imagewidth}{0.19\textwidth}  

\newlength{\imagespace}
\setlength{\imagespace}{1cm}  

\begin{figure*}
\captionsetup[subfigure]{justification=centering}
\centering  
    \begin{subfigure}[t]{\textwidth}
        \centering
        \includegraphics[width=0.6\textwidth]{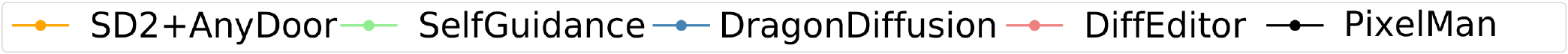}
    \end{subfigure}
    \begin{subfigure}[t]{\imagewidth}
        \includegraphics[width=\textwidth]{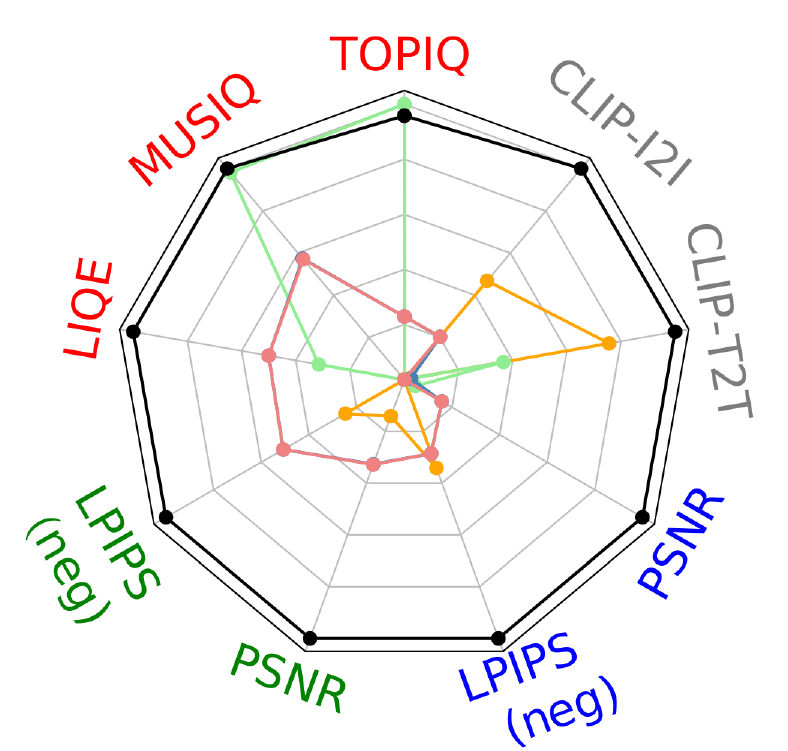}
        \caption{COCOEE dataset, all methods using 8 steps}
        \label{fig:radar_cocoee_8}
    \end{subfigure}
    \begin{subfigure}[t]{\imagewidth}
        \includegraphics[width=\textwidth]{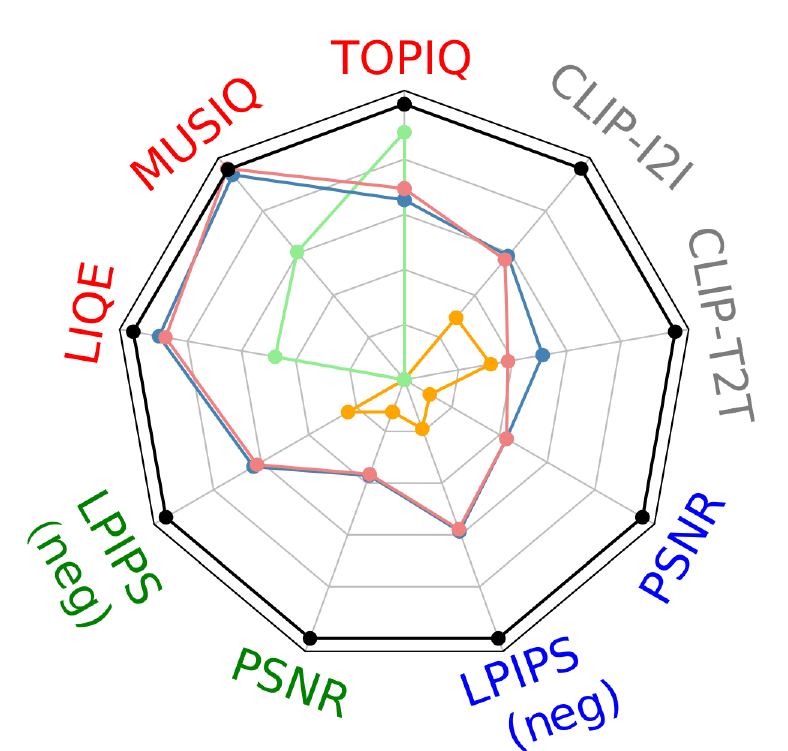}
        \caption{COCOEE dataset, all methods using 16 steps}
        \label{fig:radar_cocoee_16}
    \end{subfigure}
    \begin{subfigure}[t]{\imagewidth}
        \includegraphics[width=\textwidth]{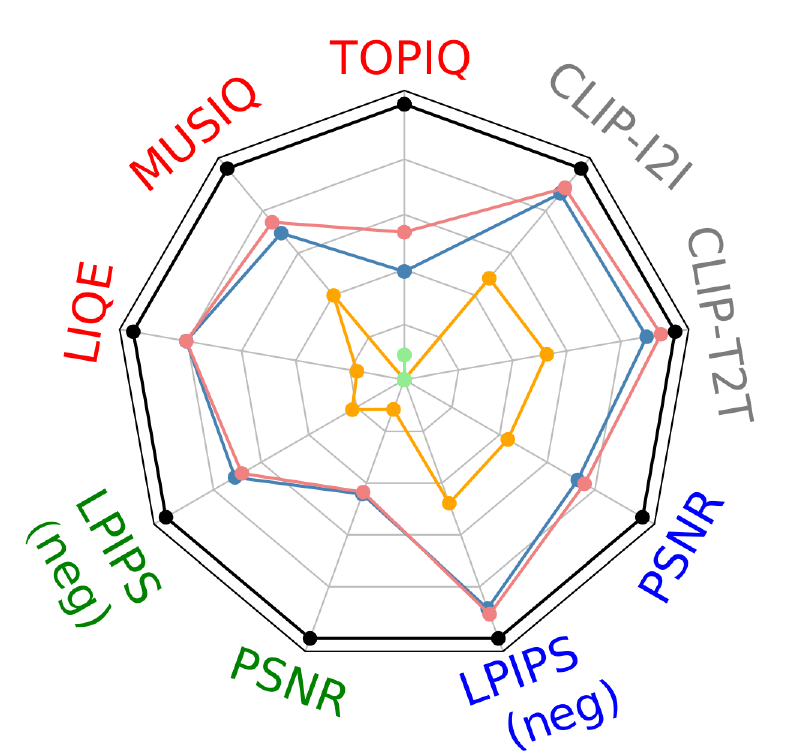}
        \caption{COCOEE dataset, all methods using 50 steps}
        \label{fig:radar_cocoee_50}
    \end{subfigure}
    \begin{subfigure}[t]{\imagewidth}
        \includegraphics[width=\textwidth]{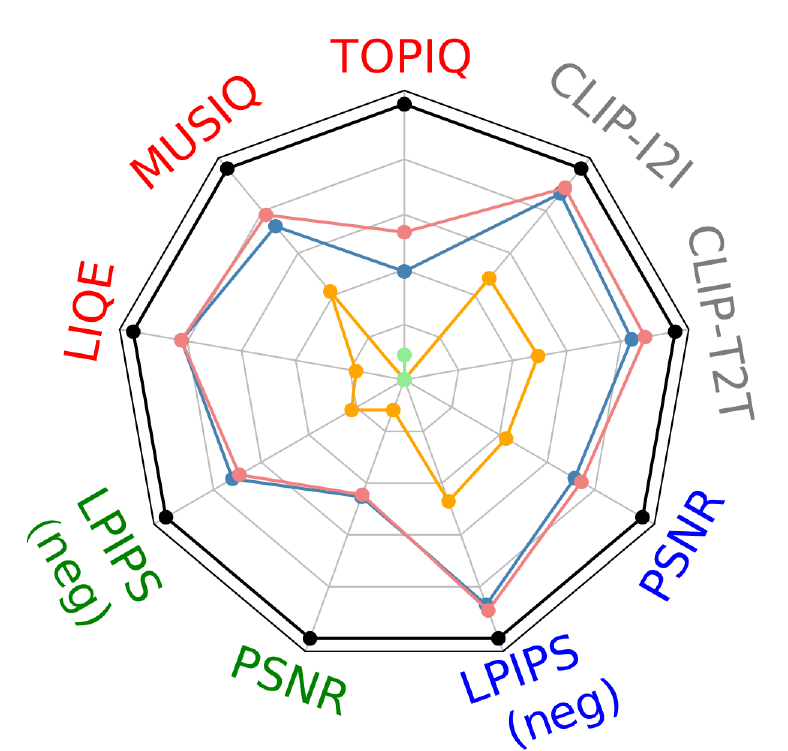}
        \caption{COCOEE dataset, PixelMan using 16 steps, others using 50 steps}
        \label{fig:radar_cocoee_16_50}
    \end{subfigure}
    \begin{subfigure}[t]{\imagewidth}
        \includegraphics[width=\textwidth]{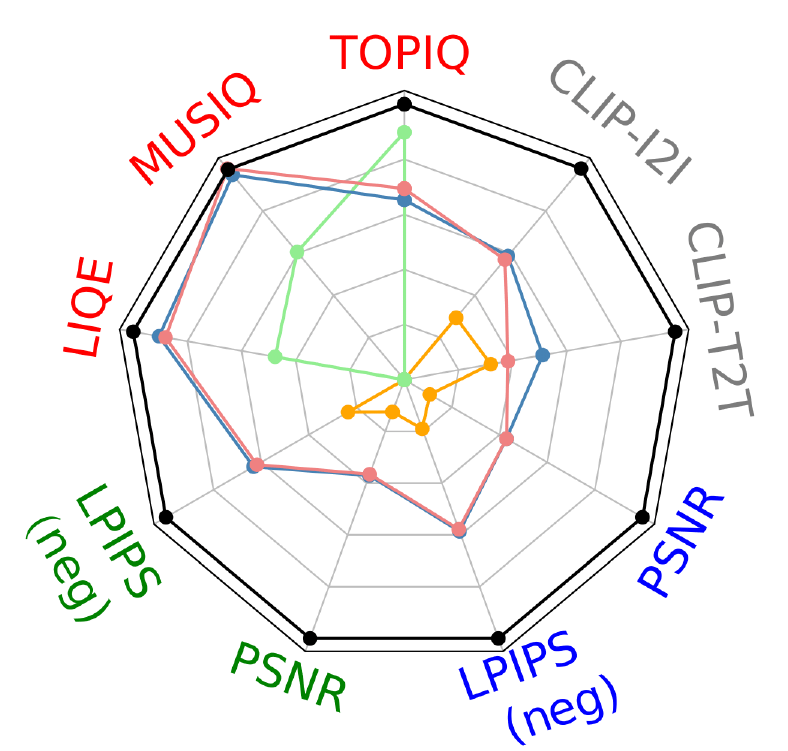}
        \caption{ReS dataset, PixelMan using 16 steps, others using 50 steps}
        \label{fig:radar_res_16_50}
    \end{subfigure}
    \caption{
            \textbf{Radar charts} that shows 
            \textit{normalized} evaluation metric values of different methods.
            \textcolor{red}{TOPIQ}, \textcolor{red}{MUSIQ}, \textcolor{red}{LIQE} belong to \textcolor{red}{IQA};
            \textcolor{mygreen}{LPIPS (neg)} and \textcolor{mygreen}{PSNR} belong to \textcolor{mygreen}{Object Consistency};
            \textcolor{blue}{LPIPS (neg)} and \textcolor{blue}{PSNR} belong to \textcolor{blue}{Background Consistency};
            and \textcolor{gray}{CLIP-T2T} and \textcolor{gray}{CLIP-I2I} belong to \textcolor{gray}{Semantic Consistency}. \textbf{Detailed results and additional comparisons in Appendix.}
            }
    \label{fig:radar}
\end{figure*}

First, we evaluate the effectiveness of \hbox{PixelMan} in the representative consistent object editing task, which is object repositioning. 
In addition, we also apply \hbox{PixelMan} on other consistent object editing tasks including object resizing, and object pasting to demonstrate the generalizability to different tasks (see Appendix). 
For the object repositioning task, we perform extensive quantitative evaluation and visual comparisons,
both against the existing methods, 
as well as ablation studies on the various components of \hbox{PixelMan}
(placed in the Appendix due to space constraints). 
In the Appendix, we also provide complete details of all the quantitative results and more visual comparisons.

For our experiments, we adopted subsets of two challenging datasets, 
namely COCOEE~\cite{lin2014microsoft, yang2022paint} and ReS~\cite{wang2024repositioning}
(detailed in the Appendix).
To comprehensively evaluate performance quantitatively, 
we adopted 9 metrics from 4 categories
(elaborated in the Appendix): 
\textit{Image Quality Assessment}~(3), 
\textit{Object Consistency}~(2), 
\textit{Background Consistency}~(2), 
and \textit{Semantic Consistency}~(2). 
For \textit{Efficiency}, we compare the number of inference steps, 
NFEs (i.e., number of UNet calls), and the average latency over 10 runs.

\subsection{Results on Object Repositioning}
\label{sec:results}

We compare \hbox{PixelMan} with three existing object repositioning methods including, 
SelfGuidance~\cite{epstein2023diffusion}, 
DragonDiffusion~\cite{mou2024dragondiffusion}, and 
DiffEditor~\cite{mou2024diffeditor}~(SOTA). 
All training-free methods are evaluated based on SDv1.5~\cite{rombach2022high, stabilityai_sd1p5} to align with \cite{mou2024dragondiffusion, mou2024diffeditor}. 
For thorough evaluations, 
we also consider a training-based baseline using SDv2-Inpainting Model~\cite{rombach2022high, stabilityai_sd2_ipt} to inpaint the original location and then use AnyDoor~\cite{chen2024anydoor} for inserting the object at the target location.

\textbf{Overall performance.}
In Figs.~\ref{fig:radar_cocoee_8},~\ref{fig:radar_cocoee_16},~and~\ref{fig:radar_cocoee_50},
we compare \hbox{PixelMan} against the four contenders on the COCOEE dataset at the same number of inference steps 
(8, 16, and 50 steps respectively). 
At 50 steps, \hbox{PixelMan} outperforms all other methods in 9 out of 9 metrics. At 16 steps, \hbox{PixelMan} outperforms other methods in 8 out of 9 metrics, while being second place on the MUSIQ IQA metric.
At 8 steps, \hbox{PixelMan} scores the best in 8 out of 9 metrics, 
while being second place on the TOPIQ IQA metric. 
Overall, \hbox{PixelMan} outperforms the other methods at the same number of steps. 
In the Appendix, we provide the full quantitative results and visual comparisons of all methods at both 16 and 50 steps.

\textbf{Efficiency.}
\begin{table}[ht]																
  \centering
  \small
    \begin{tabular}{lccrr}
    \toprule
	&	&   & \multicolumn{1}{c}{\textbf{COCOEE}} & \multicolumn{1}{c}{\textbf{ReS}} \\
        \cline{4-5}
	&	\textbf{\#Steps}	&	\textbf{NFEs}	&	\textbf{avg(lat.)}	&	\textbf{avg(lat.)}	\\
        \midrule
        SD2+AnyDoor	    &	50	&	100	&	15	&	16	\\
        SelfGuidance	&	50	&	100	&	11	&	14	\\
        DragonDiffusion	&	50	&	160	&	23	&	30	\\
        DiffEditor	    &	50	&	176	&	24	&	32	\\
        PixelMan (ours)	&	16	&	64	&   9	&	11	\\
    \bottomrule
    \end{tabular}
    \caption{\textbf{Efficiency comparisons.} PixelMan at 16 steps performs 112 fewer NFEs and is 15 seconds faster than DiffEditor~\cite{mou2024diffeditor} on the COCOEE dataset.}
    \label{tab:latency_16_50}
\end{table}		
More importantly, 
\hbox{PixelMan} achieves superior performance with fewer NFEs than existing methods. We attribute this to our three-branched inversion-free sampling approach that avoids quality degradation at 16 steps,
seen in methods~\cite{epstein2023diffusion,mou2024dragondiffusion,mou2024diffeditor}
that rely on DDIM inversion 
(e.g., row ``DiffEditor 16 steps'' of Fig.~\ref{fig:examples_comparisons_1}).
As shown in Table~\ref{tab:latency_16_50}, 
\hbox{PixelMan} at 16 steps requires 112 fewer computations and is 15 seconds faster than the SOTA DiffEditor on COCOEE. 
Despite being faster, \hbox{PixelMan}'s quality at 16 steps surpasses DiffEditor’s at 50 steps 
(additional examples in the Appendix). 
Therefore, hereafter, we directly compare \hbox{PixelMan} at 16 steps to other methods at 50 steps in the following evaluation categories.

\textbf{Image quality.}
In Fig.~\ref{fig:radar_cocoee_16_50}, \hbox{PixelMan} (16 steps) achieves significantly better image quality in all three IQA metrics than the other methods (50 steps) on COCOEE. 
In Fig.~\ref{fig:radar_res_16_50}, \hbox{PixelMan} has similar image quality to DragonDiffusion 
and DiffEditor
on ReS dataset even when using significantly fewer steps. 
In visual comparisons,
we observe \hbox{PixelMan} achieves overall better image quality than other methods while being more efficient. 
This includes less artifacts, more natural colors, well-blended objects and backgrounds, and natural lighting and shadow. 

\textbf{Object consistency.}
PixelMan (16 steps) excels in object consistency on both COCOEE~and~ReS datasets (Figs.~\ref{fig:radar_cocoee_16_50},~\ref{fig:radar_res_16_50}), as 
measured by LPIPS (neg) and PSNR. 
Our three-branched inversion-free sampling approach helps the faithful reproduction of the object at the new location 
since we always anchor the output latents to the pixel-manipulated latents which ensures
the moved object to be consistent with the original object.
This is evident in Fig.~\ref{fig:examples_comparisons_1} and Fig.~\ref{fig:examples_comparisons_2} in Appendix, 
where \hbox{PixelMan} consistently preserves details like shape, color, and texture 
(e.g., clock, bird, airplane, orange).

\textbf{Background consistency.}
On both COCOEE and ReS datasets 
(i.e., Fig.~\ref{fig:radar_cocoee_16_50} and Fig.~\ref{fig:radar_res_16_50}), 
\hbox{PixelMan} outperforms all other methods in both background consistency metrics LPIPS (neg) and PSNR. 
In the visual examples in Fig.~\ref{fig:examples_comparisons_1}, 
we observe the background in \hbox{PixelMan}'s edited images are more consistent with the source image 
(e.g., the grass texture and color in (b) and the water color in (e)).

\textbf{Inpainting.}
We provide abundant visual comparisons to assess the inpainting quality in Fig.~\ref{fig:examples_comparisons_1}
and in Figs.~\ref{fig:examples_comparisons_2},~\ref{fig:examples_full_1},~\ref{fig:examples_full_2},~\ref{fig:examples_res_1}, and~\ref{fig:examples_res_2} (in the Appendix).
Here, we see \hbox{PixelMan} excels at removing objects (e.g., plane, pillow, orange in Fig.~\ref{fig:examples_comparisons_1}) while preserving the surrounding scene. 
Conversely, other methods either leave traces of the original object or introduce new artifacts in the inpainted area.

\textbf{Semantic consistency.}
PixelMan outperforms all methods on COCOEE and is best in CLIP-I2I on ReS and remains competitive in CLIP-T2T
(see Fig.~\ref{fig:radar}). 
\hbox{PixelMan} preserves the original semantics of the source image, while maintaining consistency in object, background and better inpainting quality
(e.g., 2 instead of 3 oranges in Fig.~\ref{fig:examples_comparisons_1} (h)).

\section{Conclusion}
\label{sec:con}
We propose \hbox{PixelMan}, an inversion-free and training-free method for achieving consistent object editing via Pixel Manipulation and generation.
PixelMan maintains image consistency by directly creating a duplicate copy of the source object at target location in the pixel space, and we introduce an efficient sampling approach to iteratively harmonize the manipulated object into the target location and inpaint its original location. 
The key to image consistency is anchoring the output image to be generated to the pixel-manipulated image and introducing various consistency-preserving optimization techniques during inference.
Moreover, we propose a leak-proof SA technique to enable cohesive inpainting by addressing the attention leakage issue which is a root cause of failed inpainting.
Quantitative results on the COCOEE and ReS datasets and extensive visual comparisons show that \hbox{PixelMan} achieves superior performance in consistency metrics for object, background, and image semantics while achieving higher or comparable performance in IQA metrics.
As a training-free method, \hbox{PixelMan} only requires 16 inference steps with lower latency and a lower number of NFEs than current popular methods.

\bibliography{aaai25}


\bigskip
\appendix
\setcounter{page}{1}

\noindent\textbf{\Large{Technical Appendix}}
\bigskip

This technical appendix to our main paper has the following sections:
\begin{itemize}
    \item In Section ``Background and Related Works'', we provide more details of the background and related works.
    \item In Section ``Implementation and Evaluation Details'', we provide the implementation details and evaluation settings including the datasets and metrics.
    \item In Section ``Ablation Study'', we present an ablation study on the design of \hbox{PixelMan}.
    \item In Section ``Detailed Results'', we present more details of our quantitative results and additional qualitative (visual) comparison examples. Moreover, we provided comparisons to additional baseline methods.

\end{itemize}

\section{Background and Related Works}
\label{sec:app_related}

\subsection{Diffusion Models}
\label{sec:dm}

Diffusion Models~(DMs)~\cite{rombach2022high, saharia2022photorealistic, ramesh2022hierarchical, gafni2022make, chang2023muse, yu2022scaling, kang2023scaling}
are a class of generative models that learn to draw high-fidelity samples from the complex real-world data-distribution.
They employ a Forward Diffusion Process~(FDP) that progressively adds noise to a real data point $z_0$, 
transforming it into a sample $z_T$ from the unit Gaussian.
A model is then trained to iteratively denoise $z_T$ through the Reverse Generative Process~(RGP). 
This "denoising" capability allows the model to generate new data by iteratively removing noise from any sample from $\mathcal{N}(0, I)$.
To reduce computational costs, 
it is common to work in the latent space of a VAE~\cite{rezende2015variational}
instead of the high-dimensional pixel space.
These models are referred to as Latent Diffusion Models (LDMs)~\cite{rombach2022high}.

DMs learn to iteratively denoise a randomly sampled noise from a unit Gaussian and construct a meaningful data point (e.g., an image).
The training and inference happens in two phases.
The first phase is the Forward Diffusion Process~(FDP), 
in which we corrupt a data point sampled from the real data distribution with a known noise sampled from the unit Gaussian,
as follows:
\begin{equation}
    z_t = \sqrt{\bar{\alpha}_t} \times z_0 + \sqrt{1-\bar{\alpha}_t} \times \epsilon, \quad \epsilon \sim \mathcal{N}(0, I),
\end{equation}
where $z_0$ is the 
(latent~\cite{rombach2022high} representation of) ground truth data point,
$z_t$ is the corrupted sample,
and $\bar{\alpha}_t$ is a time-dependent coefficient.
We then train the model to estimate this added noise:
\begin{equation}
    \mathcal{L}(\theta) = \mathbb{E}_{t\sim\mathcal{U}(1, T), \epsilon\sim\mathcal{N}(0, I)}\big[\norm{\epsilon - \epsilon_\theta(z_t, t, y)}^2\big],
\end{equation}
where $y$ could be a conditioning signal such as a text prompt.

The second phase, 
which is done during inference,
is referred to as the Reverse Generative Process~(RGP), 
where, starting from a pure noise sampled from the unit Gaussian,
we can not only obtain a direct estimate of the initial latent $\hat{z}_0$, 
but also iteratively update our prediction of the latent of the previous time-step (i.e., $z_{t-1}$),
both using the noise estimations from the trained model:
\begin{equation}
    \label{eq:rgp}
    z_{t-1}, \hat{z}_0 = f\big(z_t, \epsilon_\theta(z_t, t, y), t\big),
\end{equation}
where $f$ represents a sampling strategy such as DDPM~\cite{ho2020denoising}, DDIM~\cite{song2020denoising}, PNDMS~\cite{liu2022pseudo}, etc.

\subsection{Self-Attention}
Self-Attention (SA) score matrix (i.e., $\textrm{Softmax}(QK^T/\sqrt{d})$) \cite{vaswani2017attention}
plays a crucial role in understanding the relationships between objects in an image~\cite{dosovitskiy2020image}.
It calculates a probability distribution for each element 
(here, a pixel in an intermediate feature map),  
that indicates the relative importance of that element with respect to all other elements.
Specifically, the value at any given row $i$ and column $j$ in the 
$QK^T$ matrix quantifies the impact of pixel $j$ on pixel $i$.
By modifying the SA matrix at certain indices corresponding to particular objects,
we can exert control over the editing process \cite{patashnik2023localizing}.

\subsection{Generative Semantic Nursing}
Generative Semantic Nursing (GSN)~\cite{chefer2023attendandexcite} 
refers to slightly updating the latents $z_t$ at every time-step during RGP, 
guided by a carefully designed loss function that encourages the success of the task at hand.
The update is done through gradient descent,
using the Jacobian of the loss i.e., 
\begin{equation}
\label{eq:GSN}
    z_t  \leftarrow z_t - \eta \nabla_{z_t}\mathcal{L}_{\textrm{GSN}}(\cdot)
\end{equation}
The loss in \cite{chefer2023attendandexcite} is designed to encourage better consideration of the semantic information in the text prompt while image generation.
Specifically, this loss leverages CA maps to enforce object presence,
enabling faithful \textit{text-to-image generation}. 
In this work, we optimize the latents for \textit{consistent object editing}.

\subsection{Editing with DMs}

\subsubsection{Energy guidance.}
\label{sec:eg}
Self-guidance~\cite{epstein2023diffusion} is the first work that introduced Energy Guidance (EG) to guide the edit process by updating the estimated noise.
EG is inspired by classifier-guidance~\cite{dhariwal2021diffusion, song2020score}, 
which was originally used to convert an unconditional DM, $\pr{x}$, into a conditional one, $\cpr{x}{y}$:
\begin{eqnarray}
    \cpr{z_t}{y} & \propto & \pr{z_t} \times \cpr{y}{z_t} \\
    \nabla_{z_{t}} \log{\cpr{z_t}{y}} & = & \nabla_{z_{t}} \log{\pr{z_t}} + \nabla_{z_{t}} \log{\cpr{y}{z_t}} \\
    \hat{\epsilon} & = & \epsilon_\theta - \nabla_{z_{t}} \log{\cpr{y}{z_t}},
\end{eqnarray}
where $\epsilon_\theta$ is the noise estimation of the unconditional model
and
$\cpr{y}{z_t}$ is a classifier that given the noisy latent,
yields the probability that this noisy image belongs to the desired class $y$ (i.e., condition),
and $\hat{\epsilon}$ is the guided noise directed towards generating an image that has a higher chance of belonging to class $y$.
It is common in the literature to multiply the classifier guidance term $\nabla_{z_{t}} \log{\cpr{y}{z_t}}$ with a [time-dependent] coefficient $\eta$ to control the guidance strength.

\citet{epstein2023diffusion} pointed out that, in general, 
any energy function (to be minimized) can be used to guide the estimated noise throughout the RGP, 
and not just a function of probabilities from a classifier 
(i.e., $-\nabla_{z_{t}} \log{\cpr{y}{z_t}}$).
The update is then as follows:
\begin{equation}
    \label{eq:EG}
    \hat{\epsilon} = \epsilon_\theta + \nabla_{z_{t}} \mathcal{E}(\cdot),
\end{equation}
where $\mathcal{E}(\cdot)$ denotes the energy function defined on some intermediate variables in the UNet.
Addressing the prompt-based editing task, \cite{epstein2023diffusion} define their energy function based on the Cross-Attention (CA) maps 
and require the prompt to provide directions for the desired edit.

\paragraph{DragonDiffsuon and DiffEditor.}
DragonDiffuson \cite{mou2024dragondiffusion} and DiffEditor \cite{mou2024diffeditor} consider the point-based editing task instead.
For object movement, they define an energy function with four components,
namely:
(i)~Edit, which makes sure object appears in the destination and looks the same;
(ii)~Content, which makes sure everything else stays the same;
(iii)~Contrast, which makes sure the patch whose object has moved no longer looks like before; and
(iv)~Inpaint, which makes sure the inpainted area blends well with the surroundings.
Each energy component has a coefficient.
(i.e., $\{k_1, k_2, k_3, k_4\}$)
The overall energy function $\mathcal{E}$ is defined as:
\begin{equation}
\label{eq:energy_comp}
    \mathcal{E} =   k_1 \times \mathcal{E}_{\textrm{edit}} + \
                    k_2 \times \mathcal{E}_{\textrm{content}} + \
                    k_3 \times \mathcal{E}_{\textrm{contrast}} + \
                    k_4 \times \mathcal{E}_{\textrm{inpaint}}
\end{equation}

After obtaining the initial noise from DDIM inversion~\cite{dhariwal2021diffusion}
---which requires 50 Network Function Evaluations (NFEs; i.e., UNet calls)---
the guidance is applied with respect to $z_t$ and $z_t^{\textrm{src}}$ within a guidance refinement loop.
Being an EG-based method, since it is the $\epsilon$ that is updated with each guidance step (see Eq.~(\ref{eq:EG})),
the algorithm requires to call on the DDIM inversion a second time (referred to as time travel in the paper),
to propagate the change from $\epsilon$ to the intermediate latents $z_t$ in order for the refinement loop to function properly.
This increases the NFEs even more.

\subsection{Inversion}
In order to maintain the consistency of the context of source image in the edited image,
the majority of training-free approaches require to obtain the initial noise that could have generated the source image given the pre-trained UNet.
They use the DDIM inversion technique \cite{dhariwal2021diffusion} to obtain this initial noise $z_T$,
and then,
apply their proposed guidance or manipulation on that initial noisy latent, either at the first step or throughout the entire RGP.

However, due to the asymmetry between RGP and DDIM inversion process, 
the skip time intervals must be very small, in order to obtain a valid initial noise.
Hence, the number of inference time-steps must be large (usually 50) which renders few-steps editing with it unachievable.
Renoise~\cite{garibi2024renoise} uses fixed-point iteration to refine its estimation of the noise at each time-step.
However, since at each refinement step, the UNet is called several times, the NFE is not reduced by much.

An alternative approach to inversion is DDCM~\cite{xu2023infedit},
which facilitates inversion-free prompt-based editing.
Specifically, starting from randomly sampled noise from the unit Gaussian ($z_T$), 
\cite{xu2023infedit} calculates the consistency noise ($\epsilon^{con}$) from the source image and $z_T$.
This $\epsilon^{con}$ is the golden noise that would take us back to the source image.
Then, they run the edit process in two parallel iterative branches:
one branch predicts the noise from $z_t^{src}$ and the original prompt $y^{src}$ (i.e., $\epsilon_t^{src}(z_t^{src}, y^{src})$)
and the other branch predicts the noise from $z_t^{tgt}$ and the edit prompt $y^{tgt}$ (i.e., $\epsilon_t^{tgt}(z_t^{tgt}, y^{tgt})$).
Next, the edit direction is determined by $\epsilon_t^{tgt} - \epsilon_t^{src}$, 
to which the golden consistency noise $\epsilon^{con}$ is added:
\begin{equation}
    \epsilon_t = \epsilon_t^{tgt} - \epsilon_t^{src} + \epsilon^{con}.
\end{equation}
Finally, this $\epsilon_t$ is used to estimate the denoised edited image $z_0^{tgt}$ and with that, the RGP continues.

\section{Implementation and Evaluation Details}
\label{sec:eval_details}
\subsection{Obtaining the $m_\textrm{sim}$ Mask}
\label{sec:sim}
To obtain mask $m_\textrm{sim}$, 
we leverage the average of $QK^T$ matrix rows that correspond to pixels in the $m_\textrm{new}$ region. 
This represents, on average, how much attention pixels in area $m_\textrm{new}$ are paying to other pixels. 
Since the edited object is in area $m_\textrm{new}$, 
pixels in this area are likely to pay more attention to the similar objects 
(e.g., the other similar apples in Fig.~\ref{fig:overview}). 
This averaged row is rearranged to a spatial self-attention map illustrated as $m_\textrm{sim}$ in Fig.~\ref{fig:overview}, 
which we convert into a binary mask with a threshold of $0.1$ (selected by comparing different values from 0.1 to 0.5).
This mask ($m_\textrm{sim}$) represents the area of objects similar to the to-be-edited object. 
We extract $m_\textrm{sim}$ from the $QK^T$ matrix at each time-step $t$, 
and use it in the next time-step $t-1$.%
\footnote{
    Note that $m_\textrm{sim}$ is empty at the first time-step.
}

\subsection{Implementation Details}
\label{sec:implementation_details}

We perform our experiments on Nvidia V100 (32G) GPU. For the reported latency, we measure the wall-clock time for editing one image, averaged over 10 runs on 1 x Nvidia V100 (32G) GPU.

All training-free methods are evaluated based on SDv1.5~\cite{rombach2022high, stabilityai_sd1p5} to align with \cite{mou2024dragondiffusion, mou2024diffeditor}. 
For thorough evaluations, 
we also consider a training-based baseline using SDv2-Inpainting Model~\cite{rombach2022high, stabilityai_sd2_ipt} to inpaint the original location and then use AnyDoor~\cite{chen2024anydoor} for inserting the object at the target location.
For the existing methods~\cite{chen2024anydoor,epstein2023diffusion,mou2024dragondiffusion,mou2024diffeditor} that require a prompt describing the scene, we adopt the BLIP~\cite{li2022blip} image captioning model to generate a prompt for the given source image.

In Eq.~(\ref{eq:mask_blend}), we apply mask $M=1-m_\textrm{new}$ in timesteps $t < T-2$ (value 2 is selected by testing out values from 1 to 5), where T is the number of inference timesteps (i.e., mask M is not applied in the last two inference steps).
We apply a Gaussian blurring filter to the mask M with a kernel size of 9 (selected from testing out values of 5 to 11). The masking strategy ensures seamless blending with the context and allows the model to refine the details of the edited image.

Following DragonDiffusion~\cite{mou2024dragondiffusion}, we apply latents optimization updates in every timestep $t$ for $t < 0.2 \times T$, and once every two timesteps for $ 0.2 \times T \leq t < 0.6 \times T$. 
For timesteps $t$ where $0.4 \times T < t < 0.6 \times T$, we perform three repeated GSN updates (i.e., $r=3$ in Algorithm~\ref{alg:ours}). 
We use the same energy function coefficients and update step size as in DragonDiffusion~\cite{mou2024dragondiffusion}. 

The $K$ and $V$ matrix injection happens at all timesteps for the Self-Attention layers of all upsampling blocks in the UNet. For Self-Attention maps used to obtain $m_{\textrm{sim}}$, we use the average of SA layers in the last three upsampling blocks in the UNet.

\subsection{Evaluation Datasets}
\label{sec:data}
\paragraph{COCOEE dataset.}  
This is an editing benchmark which \citet{yang2022paint} compiled by manually selecting 3,500 images 
from the MSCOCO (MicroSoft Common Objects in COntext)~\cite{lin2014microsoft} validation set.
In our work, a human-operator used the Segment Anything model~\cite{kirillov2023segment} on a random subset of COCOEE to retrieve multiple segments from each image.
Then, they identified a reasonable segmented object and suggested a diff vector that determines where the respective object should be moved to.
The result is a benchmark with 100 images along with a mask and diff vector for each.

\paragraph{ReS dataset.}
Recently, \citet{wang2024repositioning} open-sourced the ReS 
(Repositioning the Subject within image) 
dataset,
which is a real-world benchmark of 100 pairs of images.%
\footnote{
    For each image pair, we can define two movement tasks: from the first image to the second image and vice versa (so 200 movement tasks in total).
}
This is a challenging dataset due to changes in perspective / scale of the moved objects, lighting, shadows, etc. 
For each pair, a single object is moved while everything else in the scene is kept intact.
Note that it is possible for the move to take a portion of the object either behind another object or outside of the image frame 
(we refer to these as \textit{occlusion} cases).
In this work, we do not consider tasks that involve occlusion
and only compare performance of the considered methods on 162 object movement tasks.

\subsection{Evaluation Metrics}
\label{sec:metric}
Besides efficiency, which we evaluate based on 
(i)~number of inference steps, 
(ii)~NFEs (in terms of number of UNet calls), 
and
(iii)~Latency (in terms of wall-clock time for editing one image, averaged over 10 runs),
we distinguish ourselves by also conducting a comprehensive quality evaluation using quantitative metrics on all compared methods. 
Here, we list the specific metrics used in this work:
\begin{itemize}
    \item For \textit{Image Quality Assessment (IQA)}, we evaluate the overall perceptual image quality by adopting TOPIQ~\cite{chen2024topiq}, MUSIQ~\cite{ke2021musiq}, and LIQE~\cite{zhang2023liqe}.
    \item To evaluate \textit{Object Consistency}, we measure the similarity of the moved object to the original object in the source image through LPIPS (neg)~\cite{zhang2018unreasonable} and PSNR metrics.
    \item For \textit{Background Consistency}, we evaluate the similarity of the background in the edited image to the background in the original image with the same metrics as in Object Consistency.
    \item For \textit{Semantic Consistency}, we consider the similarity between the semantics of the source image and the edited image through CLIP-I2I and CLIP-T2T~\cite{radford2021learning, chefer2023attendandexcite, li2022blip}. 
    CLIP-I2I is the image-to-image similarity (i.e., similarity between CLIP embeddings of the source and edited image). For the CLIP-T2T text-to-text similarity, we measure the similarity between CLIP embeddings of text caption of the source image and the caption of the edited image. Both captions are obtained using the BLIP~\cite{li2022blip} image captioning model.

\end{itemize}

\section{Ablation Study}
\label{sec:ablate_app}

\begin{figure*}[hbt!]
    \centering
    \setlength{\tabcolsep}{0.4pt}
    \renewcommand{\arraystretch}{0.4}
    {\footnotesize
    \begin{tabular}{c c c c c c c c}
        &
        \multicolumn{1}{c}{(a)} &
        \multicolumn{1}{c}{(b)} &
        \multicolumn{1}{c}{(c)} &
        \multicolumn{1}{c}{(d)} &
        \multicolumn{1}{c}{(e)} &
        \multicolumn{1}{c}{(f)} \\

        {\raisebox{0.34in}{
        \multirow{1}{*}{\rotatebox{0}{Input}}}} &
        \includegraphics[width=0.12\textwidth]{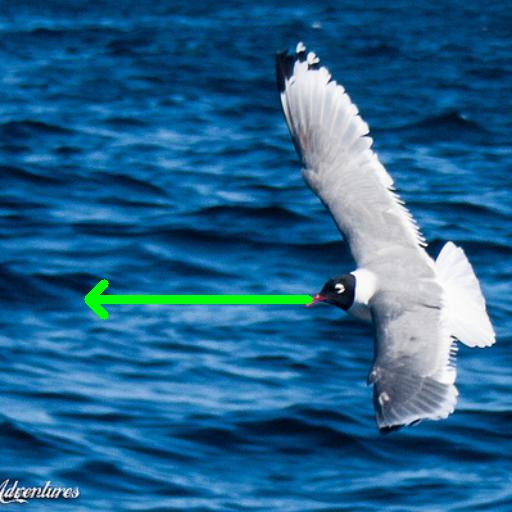} &
        \includegraphics[width=0.12\textwidth]{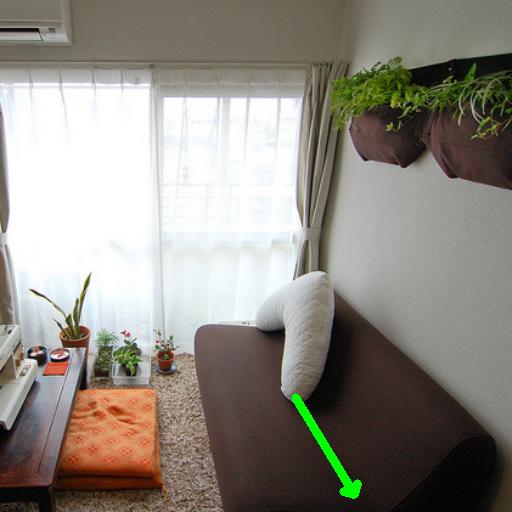} &
        \includegraphics[width=0.12\textwidth]{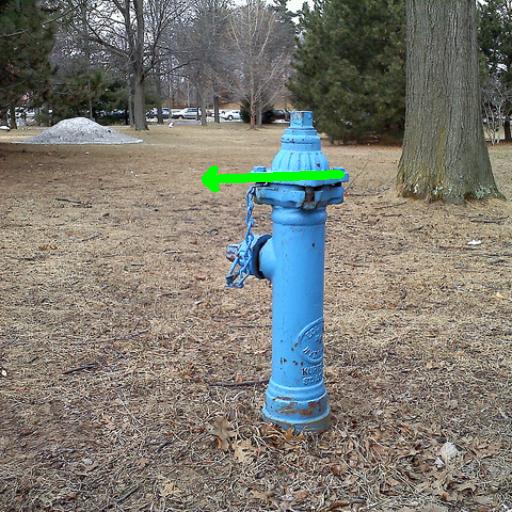} &
        \includegraphics[width=0.12\textwidth]{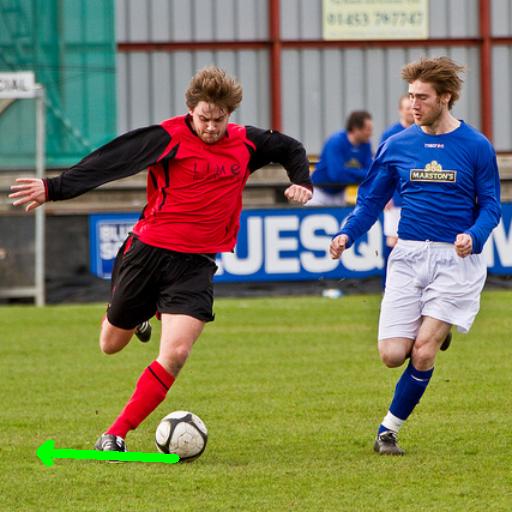} &
        \includegraphics[width=0.12\textwidth]{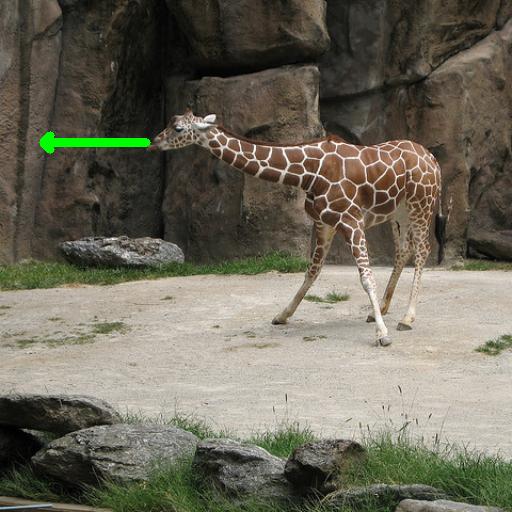} &
        \includegraphics[width=0.12\textwidth]{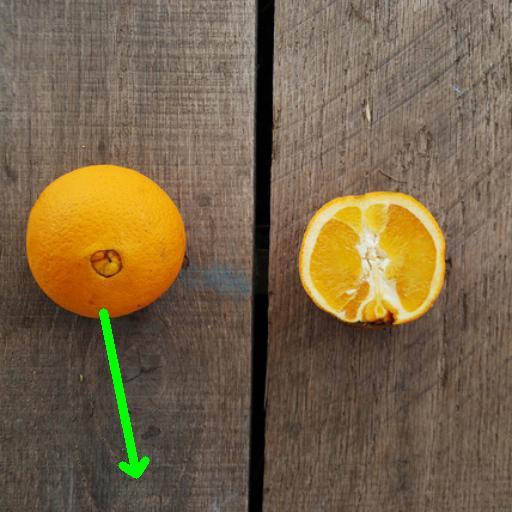} &\\

        {\raisebox{0.37in}{\multirow{1}{*}{\begin{tabular}{c}With EG\\(10s, 70 NFEs)\end{tabular}}}}
        &
        \includegraphics[width=0.12\textwidth]{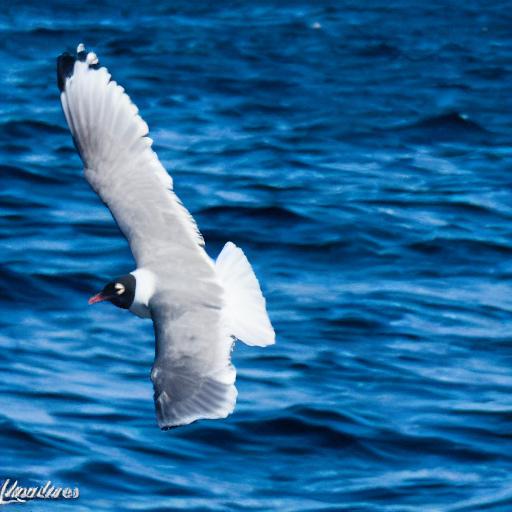} &
        \includegraphics[width=0.12\textwidth]{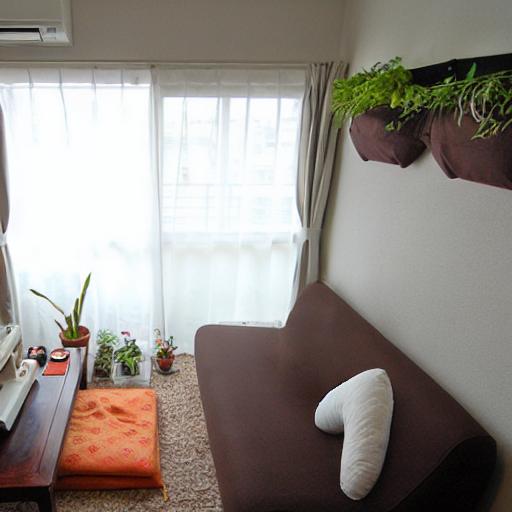} &
        \includegraphics[width=0.12\textwidth]{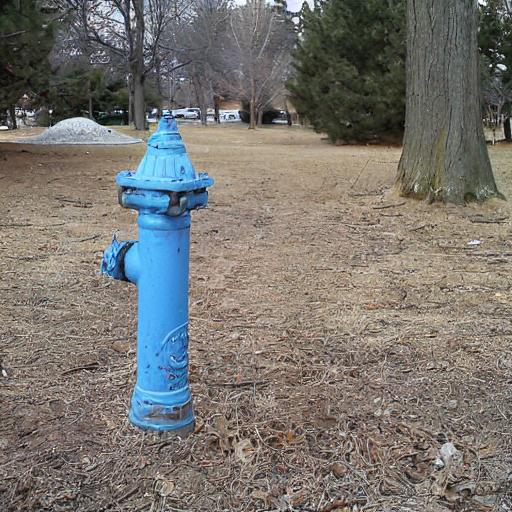} &
        \includegraphics[width=0.12\textwidth]{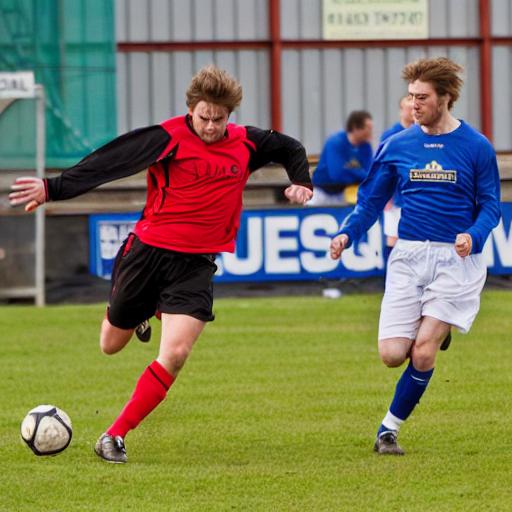} &
        \includegraphics[width=0.12\textwidth]{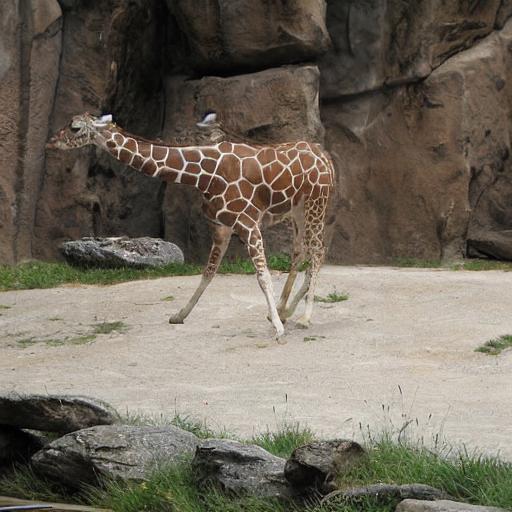} &
        \includegraphics[width=0.12\textwidth]{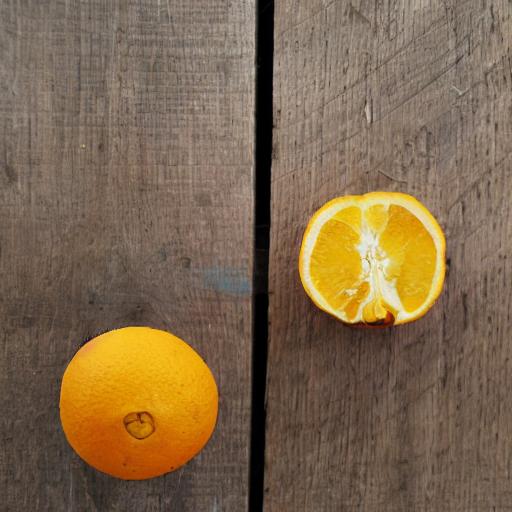} &\\
        
        {\raisebox{0.45in}{\multirow{1}{*}{\begin{tabular}{c}With DDIM\\Inversion\\(8s, 58 NFEs)\end{tabular}}}} &
        \includegraphics[width=0.12\textwidth]{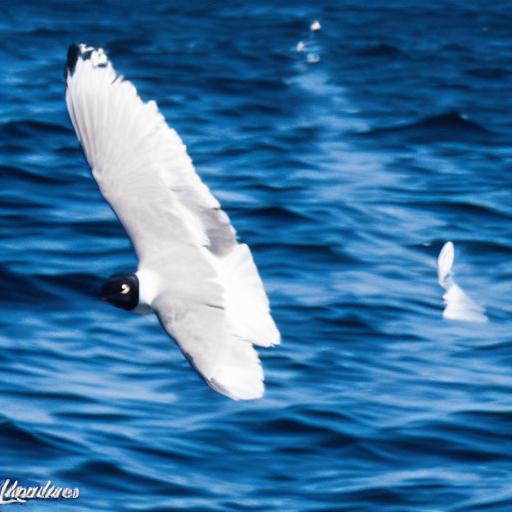} &
        \includegraphics[width=0.12\textwidth]{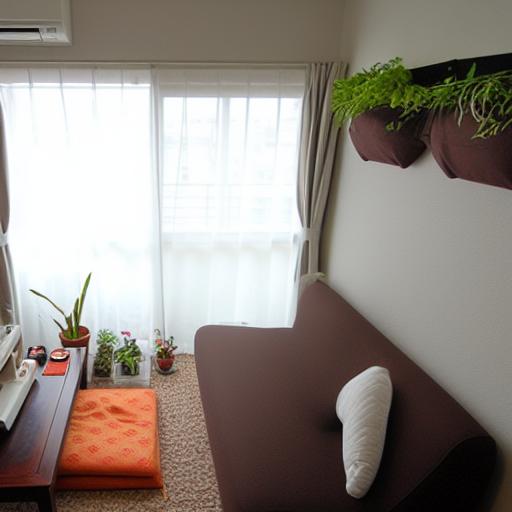} &
        \includegraphics[width=0.12\textwidth]{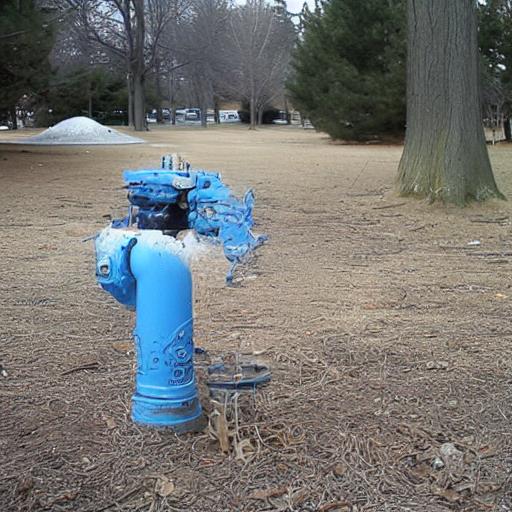} &
        \includegraphics[width=0.12\textwidth]{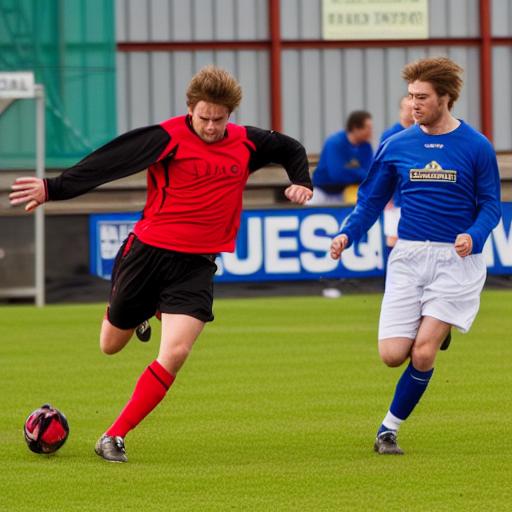} &
        \includegraphics[width=0.12\textwidth]{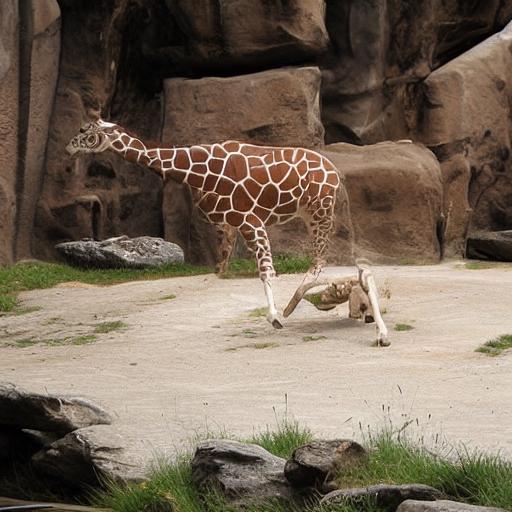} &
        \includegraphics[width=0.12\textwidth]{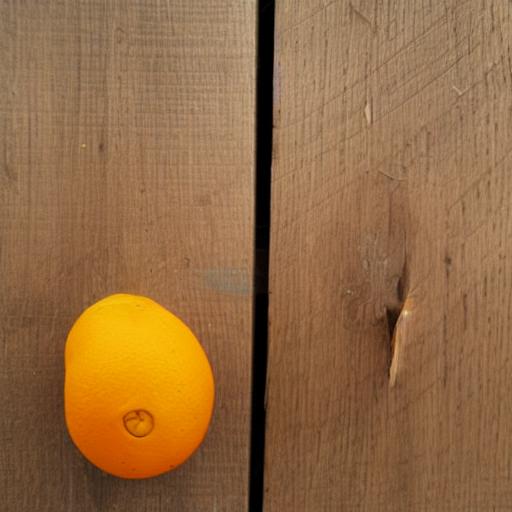} &\\
        
        {\raisebox{0.45in}{\multirow{1}{*}{\begin{tabular}{c}Without\\Leak-Proof SA\\(9s, 64 NFEs)\end{tabular}}}} &
        \includegraphics[width=0.12\textwidth]{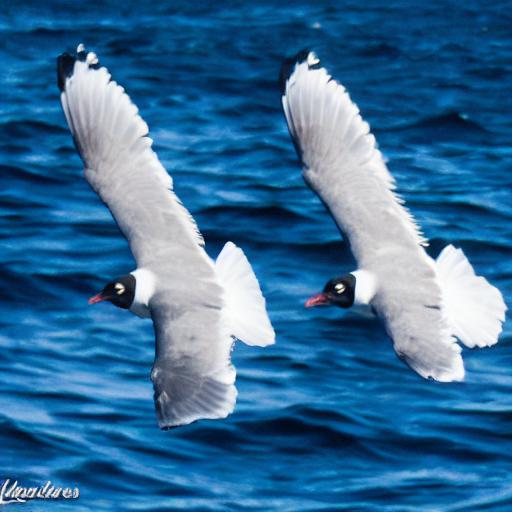} &
        \includegraphics[width=0.12\textwidth]{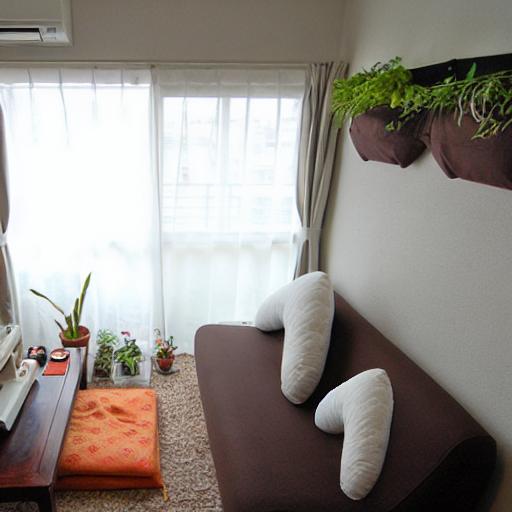} &
        \includegraphics[width=0.12\textwidth]{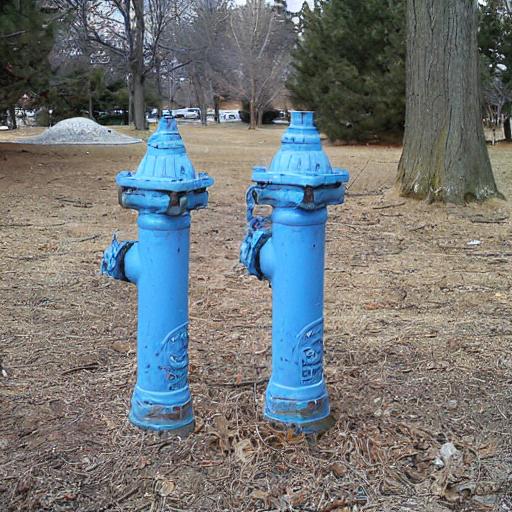} &
        \includegraphics[width=0.12\textwidth]{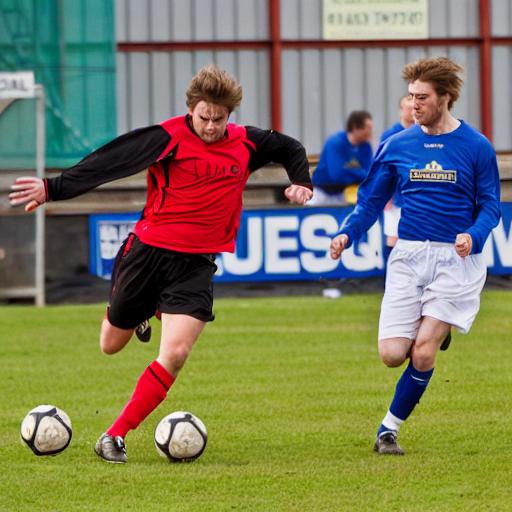} &
        \includegraphics[width=0.12\textwidth]{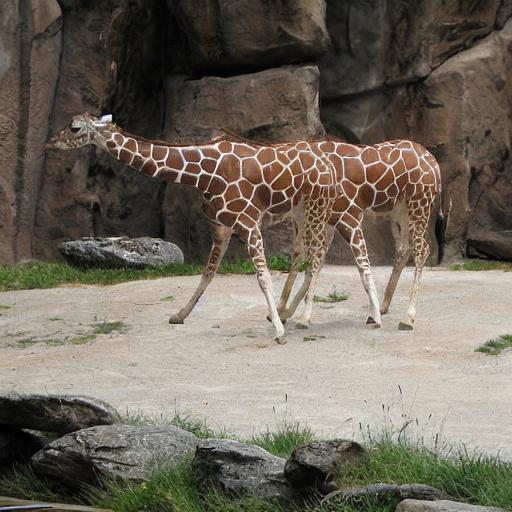} &
        \includegraphics[width=0.12\textwidth]{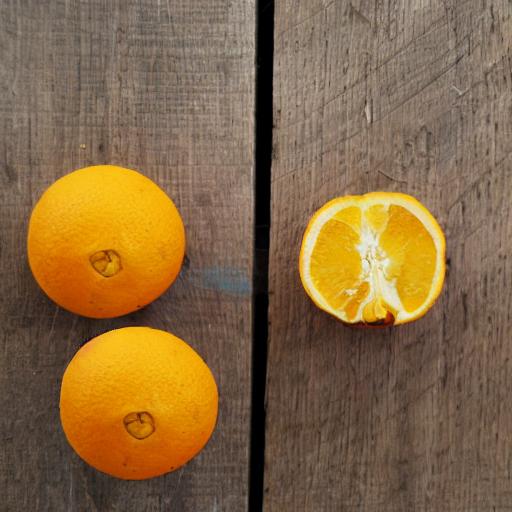} &\\

        {\raisebox{0.55in}{\multirow{1}{*}{\begin{tabular}{c}Without\\Pixel-Manipulated\\Branch\\(8s, 64 NFEs)\end{tabular}}}} &
        \includegraphics[width=0.12\textwidth]{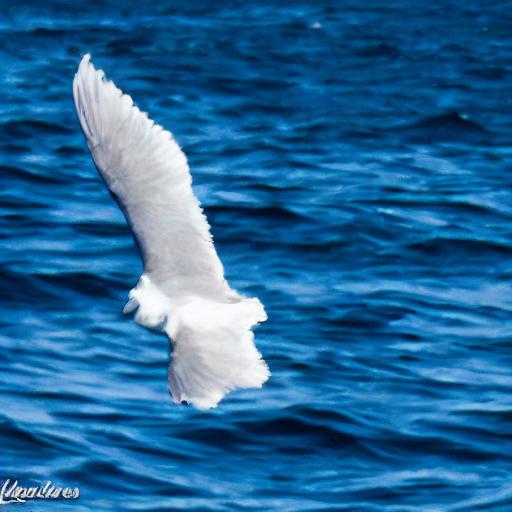} &
        \includegraphics[width=0.12\textwidth]{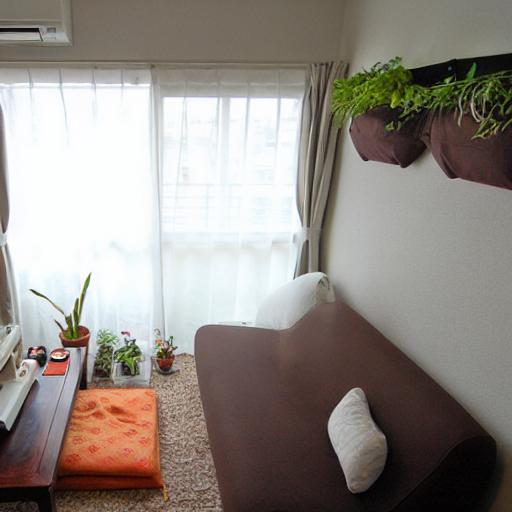} &
        \includegraphics[width=0.12\textwidth]{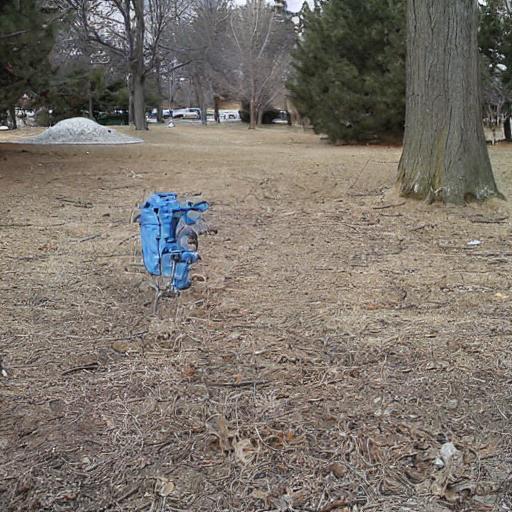} &
        \includegraphics[width=0.12\textwidth]{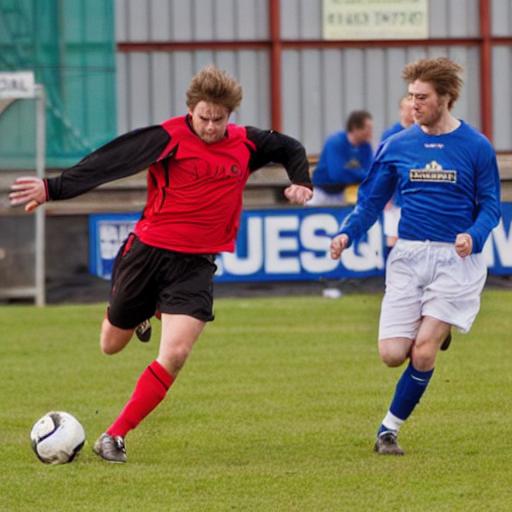} &
        \includegraphics[width=0.12\textwidth]{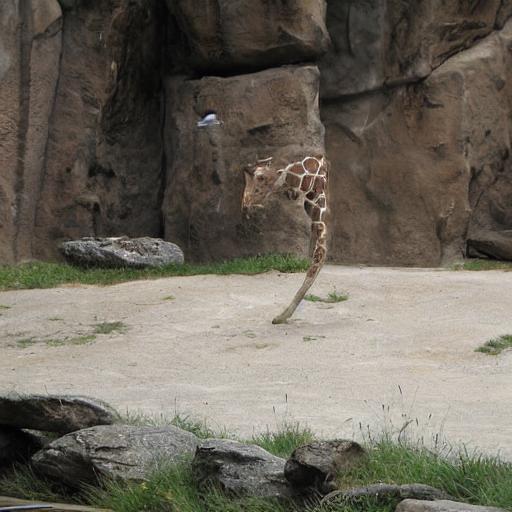} &
        \includegraphics[width=0.12\textwidth]{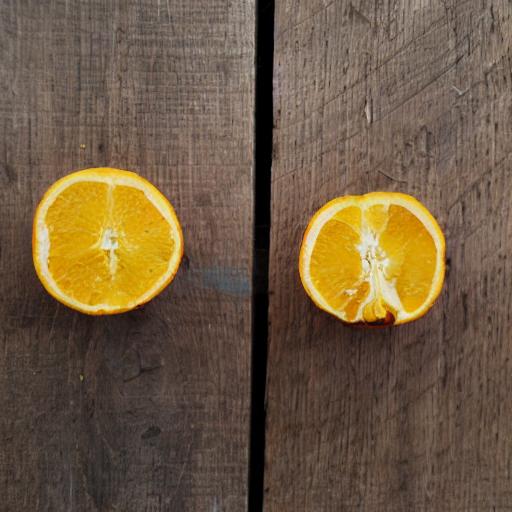} &\\
        
        {\raisebox{0.37in}{\multirow{1}{*}{\begin{tabular}{c}\textbf{PixelMan} \\ (9s, 64 NFEs)\end{tabular}}}} &
        \includegraphics[width=0.12\textwidth]{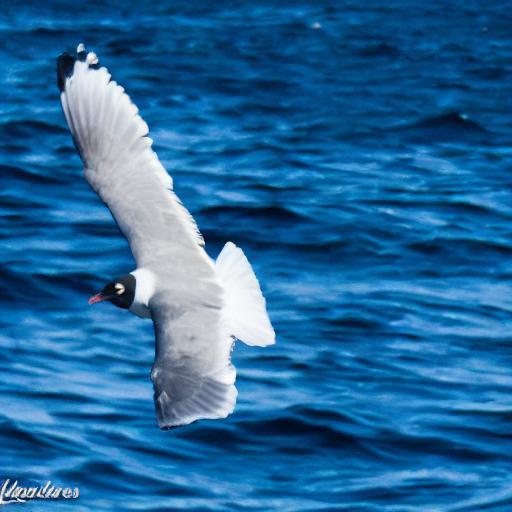} &
        \includegraphics[width=0.12\textwidth]{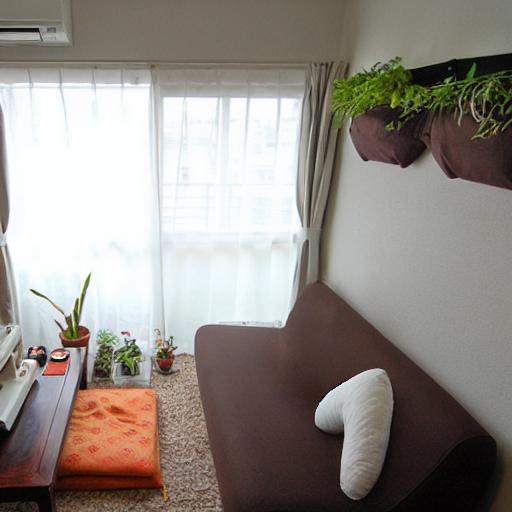} &
        \includegraphics[width=0.12\textwidth]{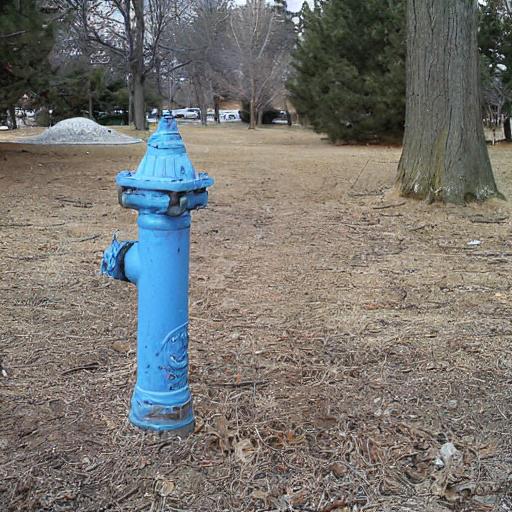} &
        \includegraphics[width=0.12\textwidth]{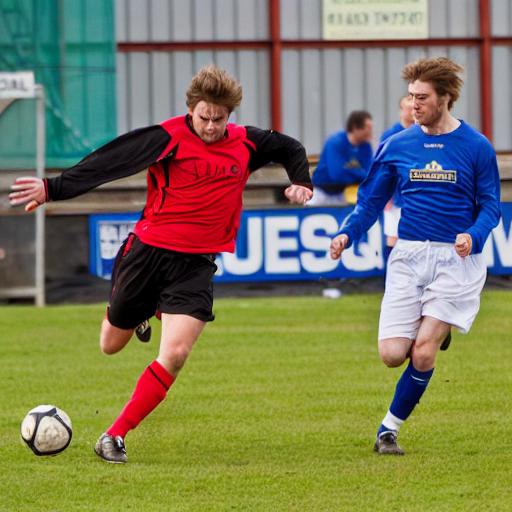} &
        \includegraphics[width=0.12\textwidth]{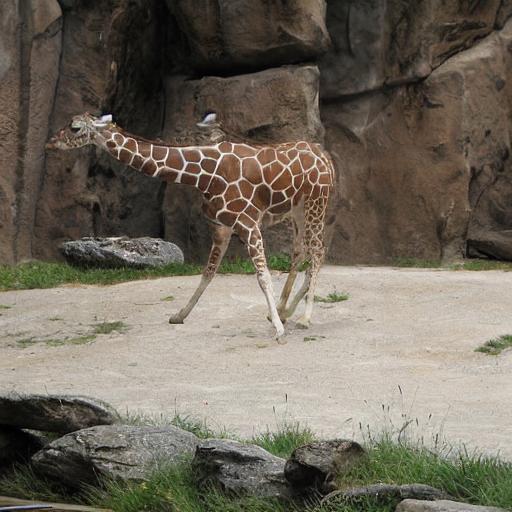} &
        \includegraphics[width=0.12\textwidth]{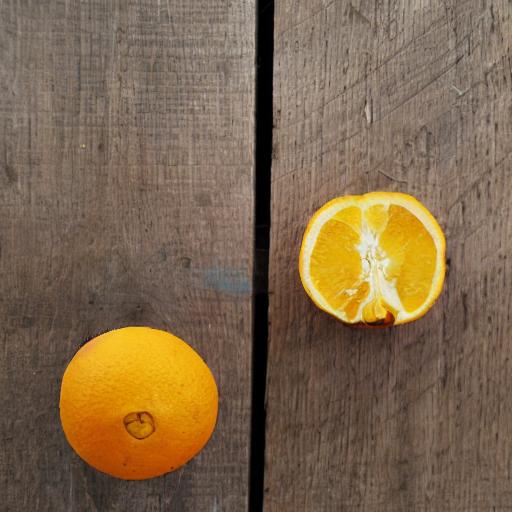} &\\

    \end{tabular}
    }
    \caption{
        \textbf{Ablation qualitative examples} on the COCOEE dataset at 16 steps. 
    }
    \label{fig:examples_ablation}
\end{figure*}
\begin{table*}[hbt!]
\centering
\small
\addtolength{\tabcolsep}{-2.9pt}

\begin{tabular}{l|cc|ccc|cc|cc|cc} 
\hline
\multirow{2}{*}{\textbf{Method}} &
\multicolumn{2}{c|}{\textbf{Efficiency}} &
\multicolumn{3}{c|}{\textbf{\begin{tabular}[c]{@{}c@{}}Image Quality\\ Assessment\end{tabular}}} &
\multicolumn{2}{c|}{\textbf{\begin{tabular}[c]{@{}c@{}}Object\\ Consistency\end{tabular}}} &
\multicolumn{2}{c|}{\textbf{\begin{tabular}[c]{@{}c@{}}Background\\ Consistency\end{tabular}}} & 
\multicolumn{2}{c}{\textbf{\begin{tabular}[c]{@{}c@{}}Semantic\\ Consistency\end{tabular}}} \\ 

\cline{2-12} & 
\multicolumn{1}{c}{\textbf{\begin{tabular}[c]{@{}c@{}}\# NFEs\\$\downarrow$\end{tabular}}} &
\multicolumn{1}{c|}{\textbf{\begin{tabular}[c]{@{}c@{}}Latency\\(secs)~$\downarrow$\end{tabular}}} &
\multicolumn{1}{c}{\textbf{\begin{tabular}[c]{@{}c@{}}TOPIQ\\$\uparrow$\end{tabular}}} &
\multicolumn{1}{c}{\textbf{\begin{tabular}[c]{@{}c@{}}MUSIQ\\$\uparrow$\end{tabular}}} &
\multicolumn{1}{c|}{\textbf{\begin{tabular}[c]{@{}c@{}}LIQE\\$\uparrow$\end{tabular}}} &
\multicolumn{1}{c}{\textbf{\begin{tabular}[c]{@{}c@{}}LPIPS\\$\downarrow$\end{tabular}}} &
\multicolumn{1}{c|}{\textbf{\begin{tabular}[c]{@{}c@{}}PSNR\\$\uparrow$\end{tabular}}} &
\multicolumn{1}{c}{\textbf{\begin{tabular}[c]{@{}c@{}}LPIPS\\$\downarrow$\end{tabular}}} &
\multicolumn{1}{c|}{\textbf{\begin{tabular}[c]{@{}c@{}}PSNR\\$\uparrow$\end{tabular}}} &
\multicolumn{1}{c}{\textbf{\begin{tabular}[c]{@{}c@{}}CLIP-\\T2T $\uparrow$\end{tabular}}} &
\multicolumn{1}{c}{\textbf{\begin{tabular}[c]{@{}c@{}}CLIP-\\I2I $\uparrow$\end{tabular}}} \\

\toprule
\multicolumn{12}{l}{\textbf{\begin{tabular}[l]{@{}l@{}}Optimization Target\end{tabular}}} \\
Energy Guidance & 70 & 10 & \textbf{0.607} & \textbf{70.01} & 4.34 & \textbf{0.015} & 35.56 & 0.076 & 26.30 & 0.945 & \textbf{0.974} \\
Latents Optimization & \textcolor{OliveGreen}{\textbf{64}} & \textcolor{OliveGreen}{\textbf{9}}  & 0.605 & 69.98 & \textbf{4.35} & \textbf{0.015} & \textbf{35.62} & \textbf{0.074} & \textbf{26.43} & \textbf{0.946} & \textbf{0.974} \\

\midrule
\multicolumn{12}{l}{\textbf{\begin{tabular}[l]{@{}l@{}}Sampling\end{tabular}}} \\
DDIM Inversion & \textbf{58} & \textbf{8}  & 0.565 & 68.41 & 4.15 & 0.041 & 27.44 & 0.134 & 22.78 & 0.924 & 0.942 \\
Three-Branched Inversion-Free & 64 & 9  & \textcolor{OliveGreen}{\textbf{0.605}} & \textcolor{OliveGreen}{\textbf{69.98}} & \textcolor{OliveGreen}{\textbf{4.35}} & \textcolor{OliveGreen}{\textbf{0.015}} & \textcolor{OliveGreen}{\textbf{35.62}} & \textcolor{OliveGreen}{\textbf{0.074}} & \textcolor{OliveGreen}{\textbf{26.43}} & \textcolor{OliveGreen}{\textbf{0.946}} & \textcolor{OliveGreen}{\textbf{0.974}} \\

\midrule
\multicolumn{12}{l}{\textbf{\begin{tabular}[l]{@{}l@{}}Leak-Proof SA\end{tabular}}} \\
No & \textbf{64} & \textbf{9}  & 0.602 & \textbf{70.77} & \textbf{4.42} & \textbf{0.015} & \textbf{35.62} & \textbf{0.064} & \textbf{27.42} & 0.891 & 0.969 \\
Yes & \textbf{64} & \textbf{9}  & \textbf{0.605} & 69.98 & 4.35 & \textbf{0.015} & \textbf{35.62} & 0.074 & 26.43 & \textcolor{OliveGreen}{\textbf{0.946}} & \textcolor{OliveGreen}{\textbf{0.974}} \\

\midrule
\multicolumn{12}{l}{\textbf{\begin{tabular}[l]{@{}l@{}}Pixel-Manipulated Branch\end{tabular}}} \\
No & \textbf{64} & \textbf{8}  & 0.570 & 67.52 & 4.19 & 0.077 & 23.59 & \textbf{0.066} & \textbf{26.68} & 0.896 & 0.945 \\
Yes & \textbf{64} & 9  & \textcolor{OliveGreen}{\textbf{0.605}} & \textcolor{OliveGreen}{\textbf{69.98}} & \textcolor{OliveGreen}{\textbf{4.35}} & \textcolor{OliveGreen}{\textbf{0.015}} & \textcolor{OliveGreen}{\textbf{35.62}} & 0.074 & 26.43 & \textcolor{OliveGreen}{\textbf{0.946}} & \textcolor{OliveGreen}{\textbf{0.974}} \\

\midrule
\multicolumn{12}{l}{\textbf{\begin{tabular}[l]{@{}l@{}}K, V Saving and Injection\end{tabular}}} \\
No Saving or Injection & 48 & 7  & 0.604 & 70.37 & 4.34 & \textbf{0.014} & 36.02 & 0.112 & 24.28 & 0.875 & 0.950 \\
From Manipulated Branch & 48 & 7  & \textbf{0.621} & \textbf{70.40} & \textbf{4.35} & \textbf{0.014} & \textbf{36.10} & \textbf{0.074} & \textbf{26.75} & 0.943 & 0.973 \\
From Source Branch & 64 & 9  & 0.605 & 69.98 & \textbf{4.35} & 0.015 & 35.62 & \textbf{0.074} & 26.43 & \textcolor{OliveGreen}{\textbf{0.946}} & \textcolor{OliveGreen}{\textbf{0.974}} \\

\bottomrule
\end{tabular}

\caption{\textbf{Ablation experiments on key techniques of PixelMan} on the COCOEE~\cite{yang2022paint} dataset at 16 steps.
The $\downarrow$ indicates lower is better, and the $\uparrow$ means the higher the better.
The best performance result is marked in \textbf{bold}. Our reported latency measures the average wall-clock time over 10 runs for generating 1 image on this dataset in seconds with a V100 GPU.}
\label{tab:ablation_results}
\end{table*}

To understand the contribution of each major technique in \hbox{PixelMan}, we perform ablation studies with both quantitative and qualitative (visual) comparisons. We show the ablation qualitative visual comparisons in Fig.~\ref{fig:examples_ablation} and we report the quantitative results in Table~\ref{tab:ablation_results}. 
For each of the techniques, we keep the rest of the method unchanged (i.e., all other components), and ablate for that specific technique.

\begin{figure}
  \centering
  \captionsetup[subfigure]{labelformat=empty}
  \begin{subfigure}[t]{.23\linewidth}
    \centering
    \includegraphics[width=\linewidth]{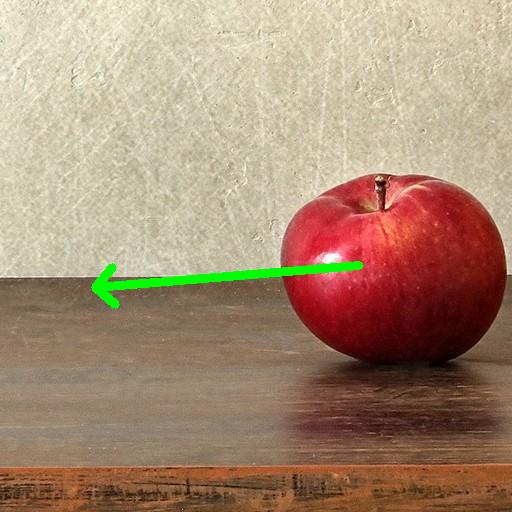}
    \caption{
    Source
    }
  \end{subfigure}
  \begin{subfigure}[t]{.23\linewidth}
    \centering
    \includegraphics[width=\linewidth]{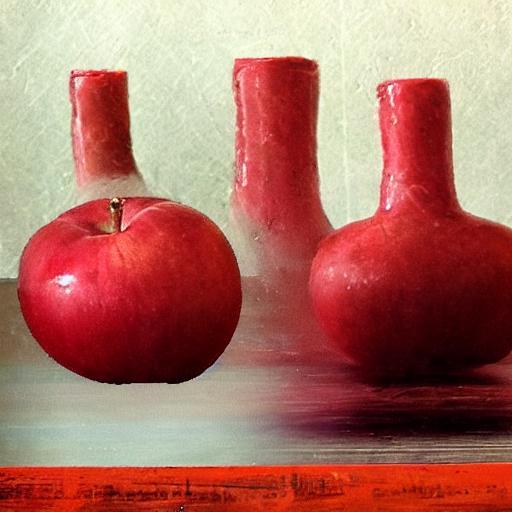}
    \caption{
    No Saving or Injection\\ \\(7s, 48 NFEs)
    }
  \end{subfigure}
  \begin{subfigure}[t]{.23\linewidth}
    \centering
    \includegraphics[width=\linewidth]{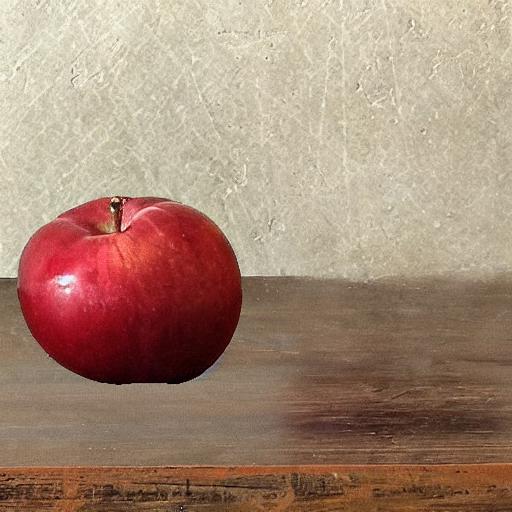}
    \caption{
    From Pixel-Manipulated Branch\\ (7s, 48 NFEs)
    }
  \end{subfigure}
  \begin{subfigure}[t]{.23\linewidth}
    \centering
    \includegraphics[width=\linewidth]{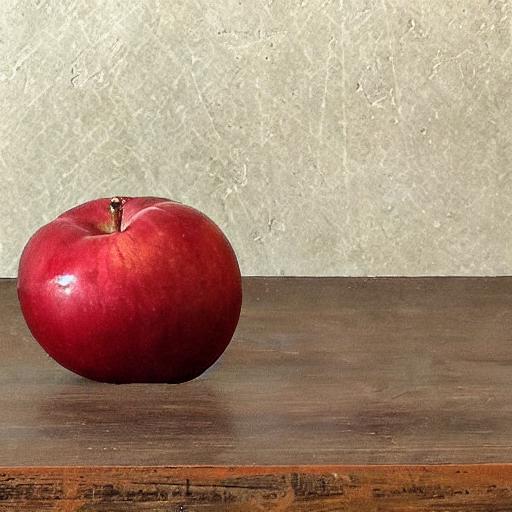}
    \caption{
    From Source Branch (\textbf{i.e., PixelMan})\\ (9s, 64 NFEs)
    }
  \end{subfigure}
  \caption{Ablation on K, V saving and injection.
  }
\label{fig:ablation_kv}
\end{figure}

\begin{itemize}
\item \textbf{Latents optimization~vs.~predicted~noise~update:}
we compare the latents ($z_t$) optimization (also known as GSN~\cite{chefer2023attendandexcite}) with the predicted noise ($\hat{\epsilon}_t$) update (i.e., Energy Guidance (EG)~\cite{mou2024dragondiffusion,mou2024diffeditor}).
In Table~\ref{tab:ablation_results}, our results show that latents optimization achieves comparable performance in most metrics with fewer NFEs than EG, demonstrating its efficiency. In Fig.~\ref{fig:examples_ablation}, the qualitative (visual) result of EG and latents update is also similar, while the latents optimization approach achieves this with better efficiency in fewer NFEs and lower latency. 

\item \textbf{Three-branched inversion-free sampling vs. DDIM inversion:}
we compare the three-branched inversion-free sampling approach with the DDIM inversion~\cite{dhariwal2021diffusion} technique. In Table~\ref{tab:ablation_results}, our inversion-free sampling approach significantly improves the performance in four metric categories (i.e., IQA, object consistency, background consistency, and semantic consistency). From the visual examples in Fig.~\ref{fig:examples_ablation}, using DDIM inversion results in inconsistent colors and artifacts, low object consistency and background consistency.

\item \textbf{Leak-proof~SA:}
we ablate the leak-proof SA technique by disabling it.
As shown in Table~\ref{tab:ablation_results}, it significantly improves semantic consistency while being comparable on other metrics. In Fig.~\ref{fig:examples_ablation}, without using leak-proof SA, we observe that the model often fails to remove the object at the original location. Our result reveals that leak-proof SA is the key to enhancing the model's ability to remove the object from its original location.

\item \textbf{Pixel-manipulated branch:}
we ablate the effect of the pixel-manipulated branch in our three-branched inversion-free sampling approach.
In Table~\ref{tab:ablation_results}, it significantly improves the performance on IQA, object consistency, and semantic consistency. In Fig.~\ref{fig:examples_ablation}, without using the pixel-manipulated branch, the moved object is often missing or inconsistent with the original object. Therefore, the pixel-manipulated branch with the pixel-manipulated image improves the model's ability to generate a faithful representation of the object at its new location.

\item \textbf{K,V saving \& injection:}
We ablate the effect of the $K,V$ saving and injection, where we save the $K, V$ from the UNet call in the feature-preserving source branch, and inject them in the target branch UNet call.
As shown in Table~\ref{tab:ablation_results}, using $K,V$ saved from either source or pixel-manipulated branch significantly improves the background consistency and semantic consistency, which is also demonstrated visually in Fig.~\ref{fig:ablation_kv} that the image background is significantly degraded without $K,V$ saving or injection.

Comparing saving the $K,V$ from the source branch or the pixel-manipulated branch, as shown in Fig.~\ref{fig:ablation_kv}, the edited apple is better harmonized with a more natural shadow and seamless blending when using the $K,V$ saved from the consistency-preserving source branch, although the introduction of the source branch will slightly increase the \#NFEs and latency. Also observed in Fig.~\ref{fig:ablation_kv}, saving the $K,V$ from the source branch results in a more complete and cohesive inpainting.

\end{itemize}

\section{Detailed Results}
\label{sec:detail_results}

\subsection{Detailed Quantitative Results}
For the experiments comparing to other methods in the main paper, we present the detailed results in Table~\ref{tab:full_results_cocoee} and Table~\ref{tab:full_results_res}, where we compare to existing methods on the COCOEE~\cite{yang2022paint} and ReS~\cite{wang2024repositioning} datasets at 8, 16, and 50 steps.

\subsection{Additional Qualitative Results}

In Figs.~\ref{fig:examples_comparisons_2},~\ref{fig:examples_full_1}~and~\ref{fig:examples_full_2}, we provide additional visual comparisons among \hbox{PixelMan} and other methods on the COCOEE~\cite{yang2022paint} datasets at both 16 and 50 steps. 
In Figs.~\ref{fig:examples_res_1}~and~\ref{fig:examples_res_2}, we present visual comparisons among \hbox{PixelMan} and other methods on the ReS~\cite{wang2024repositioning} datasets at both 16 and 50 steps. 

As shown in Figs.~\ref{fig:examples_comparisons_2},~\ref{fig:examples_full_1},~\ref{fig:examples_full_2},~\ref{fig:examples_res_1}, and~\ref{fig:examples_res_2}, \hbox{PixelMan} can better inpaint the original object while preserving the object consistency after the edit. In addition, the background is more consistent with less color shift and texture change. Most importantly, other methods have a significant quality drop when using 16 steps compared to 50 steps. Whereas \hbox{PixelMan} can efficiently edit images at 16 steps while having better quality than other methods at 50 steps.

In Fig.~\ref{fig:examples_8step_cocoee} and Fig.~\ref{fig:examples_8step_res}, we present the qualitative comparison of \hbox{PixelMan} and other methods at 8 inference steps on the COCOEE dataset and ReS dataset respectively. The other methods produce low-quality images at 8 steps, whereas \hbox{PixelMan} can still generate objects at the new location and inpaint the original location even at 8 steps.

\subsubsection{Other consistent object editing tasks.}
In Fig.~\ref{fig:examples_other_tasks}, we apply PixelMan to other consistent object editing tasks, including object resizing and object pasting.

\subsection{Comparison to Additional Baselines}
We present additional evaluation results in Tables~\ref{tab:additional_results_cocoee} and \ref{tab:additional_results_res}, where we compare to additional baseline methods PAIR Diffusion~\cite{goel2023pair} and InfEdit~\cite{xu2023infedit} on the COCOEE~\cite{yang2022paint} and ReS~\cite{wang2024repositioning} datasets at 8, 16, and 50 steps.

\subsubsection{Comparison to PAIR DIffusion.}
PixelMan is training-free, whereas PAIR-Diffusion requires costly DM model fine-tuning. Despite this distinction, for completeness, we conducted comparisons on object repositioning using the COCOEE and ReS datasets at 8, 16, and 50 inference steps (see Tables~\ref{tab:additional_results_cocoee} and \ref{tab:additional_results_res}). PixelMan achieves lower latency and also eliminates the need for costly model fine-tuning. Regarding image quality, PixelMan consistently outperforms PAIR-Diffusion in all evaluated metrics (except for an on par LIQE score).
In terms of visual quality, PixelMan delivers higher quality edits with more natural color, lighting, and shadow. It better preserves object identity and background details while fully removing the old object and seamlessly filling in the background. PAIR-Diffusion often struggles with these aspects. Visual comparisons are presented in Figs.~\ref{fig:examples_additional_cocoee} and \ref{fig:examples_additional_res}.

\subsubsection{Comparison to InfEdit.}
PixelMan differs from InfEdit~\cite{xu2023infedit} in both task focus and methodology. While InfEdit relies on prompt guidance to edit rigid attributes (e.g., color, texture) based on differences between original and edited prompts, PixelMan is prompt-free, focusing on non-rigid attributes (e.g., position, size). PixelMan leverages pixel-manipulated latents as anchors and employs a feature-preserving source branch to retain the original image details, allowing consistent object edits beyond the capability of prompt-based methods such as InfEdit.

PixelMan applies inference-time optimization of latents via energy functions tailored for object generation, harmonization, inpainting, and background consistency. To fill vacated regions, a leak-proof SA technique is introduced to prevent attention leakage to similar objects, ensuring cohesive inpainting.

To show PixelMan's advantages over InfEdit~\cite{xu2023infedit}, we extended InfEdit to non-rigid editing using DiffEditor~\cite{mou2024diffeditor}'s energy guidance for necessary editing guidance. In Tables~\ref{tab:additional_results_cocoee} and \ref{tab:additional_results_res}, we conducted object repositioning experiments comparing both methods on COCOEE and ReS datasets using 8, 16, and 50 steps.

PixelMan achieves lower latency than InfEdit~\cite{xu2023infedit}.
In terms of image quality, PixelMan delivers visibly higher-quality images with fewer artifacts and smoother object-background blending, whereas InfEdit struggles with partial inpainting and inconsistencies. Moreover, PixelMan excels in preserving object and background details (e.g., shape, color, texture), while fully removing old objects and filling in a coherent background, while InfEdit struggles to maintain object and semantic consistency. PixelMan outperforms InfEdit in object and semantic consistency while maintaining comparable IQA and background consistency scores. Visual comparisons are presented in Figs.~\ref{fig:examples_additional_cocoee} and \ref{fig:examples_additional_res}.

\begin{table*}[hbt!]
\centering
\small
\addtolength{\tabcolsep}{-2.9pt}

\begin{tabular}{c|ccc|ccc|cc|cc|cc} 
\hline
\multirow{2}{*}{\textbf{Method}} &
\multicolumn{3}{c|}{\textbf{Efficiency}} &
\multicolumn{3}{c|}{\textbf{\begin{tabular}[c]{@{}c@{}}Image Quality\\ Assessment\end{tabular}}} &
\multicolumn{2}{c|}{\textbf{\begin{tabular}[c]{@{}c@{}}Object\\ Consistency\end{tabular}}} &
\multicolumn{2}{c|}{\textbf{\begin{tabular}[c]{@{}c@{}}Background\\ Consistency\end{tabular}}} & 
\multicolumn{2}{c}{\textbf{\begin{tabular}[c]{@{}c@{}}Semantic\\ Consistency\end{tabular}}} \\ 

\cline{2-13} & 
\multicolumn{1}{c}{\textbf{\begin{tabular}[c]{@{}c@{}}\#\\Steps\end{tabular}}} &
\multicolumn{1}{c}{\textbf{\begin{tabular}[c]{@{}c@{}}\# NFEs\\$\downarrow$\end{tabular}}} &
\multicolumn{1}{c|}{\textbf{\begin{tabular}[c]{@{}c@{}}Latency\\(secs)~$\downarrow$\end{tabular}}} &
\multicolumn{1}{c}{\textbf{\begin{tabular}[c]{@{}c@{}}TOPIQ\\$\uparrow$\end{tabular}}} &
\multicolumn{1}{c}{\textbf{\begin{tabular}[c]{@{}c@{}}MUSIQ\\$\uparrow$\end{tabular}}} &
\multicolumn{1}{c|}{\textbf{\begin{tabular}[c]{@{}c@{}}LIQE\\$\uparrow$\end{tabular}}} &
\multicolumn{1}{c}{\textbf{\begin{tabular}[c]{@{}c@{}}LPIPS\\$\downarrow$\end{tabular}}} &
\multicolumn{1}{c|}{\textbf{\begin{tabular}[c]{@{}c@{}}PSNR\\$\uparrow$\end{tabular}}} &
\multicolumn{1}{c}{\textbf{\begin{tabular}[c]{@{}c@{}}LPIPS\\$\downarrow$\end{tabular}}} &
\multicolumn{1}{c|}{\textbf{\begin{tabular}[c]{@{}c@{}}PSNR\\$\uparrow$\end{tabular}}} &
\multicolumn{1}{c}{\textbf{\begin{tabular}[c]{@{}c@{}}CLIP-\\T2T $\uparrow$\end{tabular}}} &
\multicolumn{1}{c}{\textbf{\begin{tabular}[c]{@{}c@{}}CLIP-\\I2I $\uparrow$\end{tabular}}} \\

\hline
SDv2-Inpainting+AnyDoor & 50 & 100 & 15 & 0.549 & 67.61 & 3.98 & 0.068 & 24.28 & 0.172 & 21.52 & 0.905 & 0.934 \\ 
Self-Guidance   & 50 & 100 & 11 & 0.554 & 65.91 & 3.90 & 0.083 & 22.77 & 0.259 & 17.86 & 0.865 & 0.897 \\
DragonDiffusion   & 50 & 160 & 23 & 0.571 & 68.87 & \underline{4.27} & \underline{0.034} & \underline{28.59} & 0.098 & 23.99 & 0.933 & 0.965 \\
DiffEditor   & 50 & 176 & 24 & \underline{0.579} & \underline{69.09} & \underline{4.27} & 0.036 & 28.49 & \underline{0.094} & \underline{24.23} & \underline{0.937} & \underline{0.967} \\
PixelMan & 16 & 64  &  9 & \textbf{0.605} & \textbf{69.98} & \textbf{4.35} & \textbf{0.015} & \textbf{35.62} & \textbf{0.074} & \textbf{26.43} & \textbf{0.946} & \textbf{0.974} \\

\hline
SDv2-Inpainting+AnyDoor  & \multirow{5}{*}{\begin{tabular}[c]{@{}c@{}}50\end{tabular}}					
       & 100 & 15 & 0.549 & 67.61 & 3.98 & 0.068 & 24.28 & 0.172 & 21.52 & 0.905 & 0.934 \\ 
Self-Guidance   & & 100 & 11 & 0.554 & 65.91 & 3.90 & 0.083 & 22.77 & 0.259 & 17.86 & 0.865 & 0.897 \\
DragonDiffusion   & & 160 & 23 & 0.571 & 68.87 & \underline{4.27} & \underline{0.034} & \underline{28.59} & 0.098 & 23.99 & 0.933 & 0.965 \\
DiffEditor   & & 176 & 24 & \underline{0.579} & \underline{69.09} & \underline{4.27} & 0.036 & 28.49 & \underline{0.094} & \underline{24.23} & \underline{0.937} & \underline{0.967} \\
PixelMan & & 206 & 27 & \textbf{0.605} & \textbf{70.17} & \textbf{4.36} & \textbf{0.014} & \textbf{35.92} & \textbf{0.077} & \textbf{26.28} & \textbf{0.941} & \textbf{0.974} \\

\hline
SDv2-Inpainting+AnyDoor  & \multirow{5}{*}{\begin{tabular}[c]{@{}c@{}}16\end{tabular}}
       & 32 &  5 & 0.556 & 67.66 & 3.93 & 0.067 & 24.44 & 0.172 & 21.60 & 0.914 & 0.933 \\
Self-Guidance   & & 32 &  4 & \underline{0.600} & 69.07 & 4.13 & 0.083 & 22.85 & 0.195 & 21.02 & 0.899 & 0.916 \\ 
DragonDiffusion   & & 64 &  9 & 0.588 & 69.92 & \underline{4.31} & \underline{0.040} & \underline{27.58} & \underline{0.124} & \underline{23.34} & \underline{0.923} & \underline{0.950} \\ 
DiffEditor   & & 58 &  9 & 0.590 & \textbf{69.99} & 4.30 & 0.041 & 27.52 & 0.125 & \underline{23.34} & 0.917 & 0.949 \\ 
PixelMan & & 64 &  9 & \textbf{0.605} & \underline{69.98} & \textbf{4.35} & \textbf{0.015} & \textbf{35.62} & \textbf{0.074} & \textbf{26.43} & \textbf{0.946} & \textbf{0.974} \\

\hline
SDv2-Inpainting+AnyDoor  & \multirow{5}{*}{\begin{tabular}[c]{@{}c@{}}8\end{tabular}}					
       & 16 & 3 & 0.556 & 66.86 & 3.78 & 0.068 & 24.50 & \underline{0.177} & 21.49 & \underline{0.916} & \underline{0.929} \\
Self-Guidance   & & 16 & 2 & \textbf{0.604} & \underline{69.58} & 3.95 & 0.085 & 22.72 & 0.232 & 21.73 & 0.900 & 0.892 \\
DragonDiffusion   & & 32 & 5 & 0.567 & 68.45 & \underline{4.05} & \underline{0.050} & 26.84 & 0.186 & \underline{22.31} & 0.886 & 0.908 \\
DiffEditor   & & 32 & 5 & 0.567 & 68.44 & \underline{4.05} & \underline{0.050} & \underline{26.86} & 0.186 & \underline{22.31} & 0.885 & 0.908 \\
PixelMan & & 28 & 4 & \underline{0.602} & \textbf{69.63} & \textbf{4.32} & \textbf{0.016} & \textbf{35.33} & \textbf{0.071} & \textbf{26.70} & \textbf{0.926} & \textbf{0.971} \\

\hline
\end{tabular}

\caption{\textbf{Quantitative results on the COCOEE~\cite{yang2022paint} dataset.}
Comparing PixelMan with other methods including Self-Guidance~\cite{epstein2023diffusion}, DragonDiffusion~\cite{mou2024dragondiffusion}, DiffEditor~\cite{mou2024diffeditor}, and the training-based SDv2-Inpainting+AnyDoor~\cite{rombach2022high, stabilityai_sd2_ipt, chen2024anydoor} baseline.
The $\downarrow$ indicates lower is better, and the $\uparrow$ means the higher the better.
The best performance result is marked in \textbf{bold} and the second best result is annotated with \underline{underlines}.
Our reported latency measures the average wall-clock time over ten runs for generating one image on this dataset in seconds with a V100 GPU.}
\label{tab:full_results_cocoee}
\end{table*}

\begin{table*}[hbt!]
\small
\centering
\addtolength{\tabcolsep}{-2.9pt}
\begin{tabular}{c|ccc|ccc|cc|cc|cc} 
\hline
\multirow{2}{*}{\textbf{Method}} &
\multicolumn{3}{c|}{\textbf{Efficiency}} &
\multicolumn{3}{c|}{\textbf{\begin{tabular}[c]{@{}c@{}}Image Quality\\ Assessment\end{tabular}}} &
\multicolumn{2}{c|}{\textbf{\begin{tabular}[c]{@{}c@{}}Object\\ Consistency\end{tabular}}} &
\multicolumn{2}{c|}{\textbf{\begin{tabular}[c]{@{}c@{}}Background\\ Consistency\end{tabular}}}  &
\multicolumn{2}{c}{\textbf{\begin{tabular}[c]{@{}c@{}}Semantic\\ Consistency\end{tabular}}} \\ 

\cline{2-13} & 
\multicolumn{1}{c}{\textbf{\begin{tabular}[c]{@{}c@{}}\#\\Steps\end{tabular}}} &
\multicolumn{1}{c}{\textbf{\begin{tabular}[c]{@{}c@{}}\# NFEs\\$\downarrow$\end{tabular}}} &
\multicolumn{1}{c|}{\textbf{\begin{tabular}[c]{@{}c@{}}Latency\\(secs)~$\downarrow$\end{tabular}}} &
\multicolumn{1}{c}{\textbf{\begin{tabular}[c]{@{}c@{}}TOPIQ\\$\uparrow$\end{tabular}}} &
\multicolumn{1}{c}{\textbf{\begin{tabular}[c]{@{}c@{}}MUSIQ\\$\uparrow$\end{tabular}}} &
\multicolumn{1}{c|}{\textbf{\begin{tabular}[c]{@{}c@{}}LIQE\\$\uparrow$\end{tabular}}} &
\multicolumn{1}{c}{\textbf{\begin{tabular}[c]{@{}c@{}}LPIPS\\$\downarrow$\end{tabular}}} &
\multicolumn{1}{c|}{\textbf{\begin{tabular}[c]{@{}c@{}}PSNR\\$\uparrow$\end{tabular}}} &
\multicolumn{1}{c}{\textbf{\begin{tabular}[c]{@{}c@{}}LPIPS\\$\downarrow$\end{tabular}}} &
\multicolumn{1}{c|}{\textbf{\begin{tabular}[c]{@{}c@{}}PSNR\\$\uparrow$\end{tabular}}}   &
\multicolumn{1}{c}{\textbf{\begin{tabular}[c]{@{}c@{}}CLIP-\\T2T $\uparrow$\end{tabular}}} &
\multicolumn{1}{c}{\textbf{\begin{tabular}[c]{@{}c@{}}CLIP-\\I2I $\uparrow$\end{tabular}}} \\

\hline
SDv2-Inpainting+AnyDoor & 50 & 100  & 16  & 0.621 & 71.19 & 4.22 & 0.052 & 26.06 & 0.159 & 21.21 & 0.866 & 0.907 \\
Self-Guidance   & 50 & 100 & 14 & 0.586 & 69.41 & 3.61 & 0.064 & 24.21 & 0.273 & 17.92 & 0.817 & 0.869 \\
DragonDiffusion   & 50 & 160 & 30 & 0.690 & \textbf{74.95} & \underline{4.72} & \underline{0.030} & \underline{29.68} & \underline{0.083} & 25.38 & \textbf{0.902} & \underline{0.934} \\
DiffEditor   & 50 & 176 & 32 & \underline{0.691} & \underline{74.94} & \textbf{4.73} & 0.032 & 29.59 & \underline{0.083} & \underline{25.44} & \underline{0.899} & 0.933 \\
PixelMan & 16 & 64  & 11 & \textbf{0.696} & 74.66 & 4.70 & \textbf{0.015} & \textbf{35.90} & \textbf{0.070} & \textbf{27.18} & 0.898 & \textbf{0.939} \\

\hline
SDv2-Inpainting+AnyDoor  & \multirow{5}{*}{\begin{tabular}[c]{@{}c@{}}50\end{tabular}}					
       & 100 & 16 & 0.621 & 71.19 & 4.22 & 0.052 & 26.06 & 0.159 & 21.21 & 0.866 & 0.907 \\
Self-Guidance   & & 100 & 14 & 0.586 & 69.41 & 3.61 & 0.064 & 24.21 & 0.273 & 17.92 & 0.817 & 0.869 \\
DragonDiffusion   & & 160 & 30 & \underline{0.690} & \textbf{74.95} & 4.72 & \underline{0.030} & \underline{29.68} & \underline{0.083} & 25.38 & \textbf{0.902} & \underline{0.934} \\
DiffEditor   & & 176 & 32 & \textbf{0.691} & \underline{74.94} & \underline{4.73} & 0.032 & 29.59 & \underline{0.083} & \underline{25.44} & \underline{0.899} & 0.933 \\
PixelMan & & 206 & 34 & 0.688 & 74.72 & \textbf{4.75} & \textbf{0.015} & \textbf{36.26} & \textbf{0.073} & \textbf{26.74} & 0.896 & \textbf{0.940} \\

\hline
SDv2-Inpainting+AnyDoor  & \multirow{5}{*}{\begin{tabular}[c]{@{}c@{}}16\end{tabular}}					
       & 32 & 6  & 0.625 & 71.29 & 4.17 & 0.051 & 26.21 & 0.159 & 21.25 & 0.856 & 0.907 \\
Self-Guidance   & & 32 & 6  & 0.663 & 73.41 & 4.16 & 0.064 & 24.00 & 0.194 & 20.95 & 0.847 & 0.886 \\
DragonDiffusion   & & 64 & 12 & \textbf{0.697} & \textbf{75.21} & \textbf{4.72} & \underline{0.033} & \underline{29.19} & \underline{0.104} & 24.99 & \underline{0.894} & \underline{0.917} \\
DiffEditor   & & 58 & 11 & \textbf{0.697} & \underline{75.20} & \textbf{4.72} & \underline{0.033} & 29.15 & 0.105 & \underline{25.00} & 0.889 & \underline{0.917} \\
PixelMan & & 64 & 11 & \underline{0.696} & 74.66 & \underline{4.70} & \textbf{0.015} & \textbf{35.90} & \textbf{0.070} & \textbf{27.18} & \textbf{0.898} & \textbf{0.939} \\

\hline
SDv2-Inpainting+AnyDoor  & \multirow{5}{*}{\begin{tabular}[c]{@{}c@{}}8\end{tabular}}					
       & 16 & 3 & 0.627 & 70.92 & 4.04 & 0.051 & 26.31 & \underline{0.162} & 21.21 & 0.849 & \underline{0.902} \\
Self-Guidance   & & 16 & 3 & 0.678 & 73.07 & 4.01 & 0.065 & 23.97 & 0.255 & 20.76 & 0.851 & 0.845 \\
DragonDiffusion   & & 32 & 6 & \underline{0.692} & \textbf{74.62} & \underline{4.46} & \underline{0.038} & \underline{28.57} & 0.173 & \underline{22.68} & \underline{0.856} & 0.876 \\
DiffEditor   & & 32 & 6 & \underline{0.692} & \textbf{74.62} & \underline{4.46} & \underline{0.038} & \underline{28.57} & 0.173 & \underline{22.68} & 0.852 & 0.876 \\
PixelMan & & 28 & 5 & \textbf{0.695} & \underline{74.59} & \textbf{4.67} & \textbf{0.016} & \textbf{35.57} & \textbf{0.067} & \textbf{27.74} & \textbf{0.900} & \textbf{0.937} \\

\hline
\end{tabular}

\caption{\textbf{Quantitative results on the ReS~\cite{yang2022paint} dataset.}
Comparing PixelMan with other methods including Self-Guidance~\cite{epstein2023diffusion}, DragonDiffusion~\cite{mou2024dragondiffusion}, DiffEditor~\cite{mou2024diffeditor}, and the training-based SDv2-Inpainting+AnyDoor~\cite{rombach2022high, stabilityai_sd2_ipt, chen2024anydoor} baseline.
The $\downarrow$ indicates lower is better, and the $\uparrow$ means the higher the better.
The best performance result is marked in \textbf{bold} and the second best result is annotated with \underline{underlines}.
Our reported latency measures the average wall-clock time over ten runs for generating one image on this dataset in seconds with a V100 GPU.}
\label{tab:full_results_res}
\end{table*}
\begin{table*}[hbt!]
\centering
\small
\addtolength{\tabcolsep}{-2.9pt}

\begin{tabular}{c|ccc|ccc|cc|cc|cc} 
\hline
\multirow{2}{*}{\textbf{Method}} &
\multicolumn{3}{c|}{\textbf{Efficiency}} &
\multicolumn{3}{c|}{\textbf{\begin{tabular}[c]{@{}c@{}}Image Quality\\ Assessment\end{tabular}}} &
\multicolumn{2}{c|}{\textbf{\begin{tabular}[c]{@{}c@{}}Object\\ Consistency\end{tabular}}} &
\multicolumn{2}{c|}{\textbf{\begin{tabular}[c]{@{}c@{}}Background\\ Consistency\end{tabular}}} & 
\multicolumn{2}{c}{\textbf{\begin{tabular}[c]{@{}c@{}}Semantic\\ Consistency\end{tabular}}} \\ 

\cline{2-13} & 
\multicolumn{1}{c}{\textbf{\begin{tabular}[c]{@{}c@{}}\#\\Steps\end{tabular}}} &
\multicolumn{1}{c}{\textbf{\begin{tabular}[c]{@{}c@{}}\# NFEs\\$\downarrow$\end{tabular}}} &
\multicolumn{1}{c|}{\textbf{\begin{tabular}[c]{@{}c@{}}Latency\\(secs)~$\downarrow$\end{tabular}}} &
\multicolumn{1}{c}{\textbf{\begin{tabular}[c]{@{}c@{}}TOPIQ\\$\uparrow$\end{tabular}}} &
\multicolumn{1}{c}{\textbf{\begin{tabular}[c]{@{}c@{}}MUSIQ\\$\uparrow$\end{tabular}}} &
\multicolumn{1}{c|}{\textbf{\begin{tabular}[c]{@{}c@{}}LIQE\\$\uparrow$\end{tabular}}} &
\multicolumn{1}{c}{\textbf{\begin{tabular}[c]{@{}c@{}}LPIPS\\$\downarrow$\end{tabular}}} &
\multicolumn{1}{c|}{\textbf{\begin{tabular}[c]{@{}c@{}}PSNR\\$\uparrow$\end{tabular}}} &
\multicolumn{1}{c}{\textbf{\begin{tabular}[c]{@{}c@{}}LPIPS\\$\downarrow$\end{tabular}}} &
\multicolumn{1}{c|}{\textbf{\begin{tabular}[c]{@{}c@{}}PSNR\\$\uparrow$\end{tabular}}} &
\multicolumn{1}{c}{\textbf{\begin{tabular}[c]{@{}c@{}}CLIP-\\T2T $\uparrow$\end{tabular}}} &
\multicolumn{1}{c}{\textbf{\begin{tabular}[c]{@{}c@{}}CLIP-\\I2I $\uparrow$\end{tabular}}} \\

\hline
PAIR Diffusion   & 50 & 100 & 32 & 0.525 & 64.76 & 3.51 & 0.088 & 23.87 & 0.112 & 24.69 & 0.836 & 0.863 \\
InfEdit   & 50 & 230 & 35 & \underline{0.567} & \underline{69.07} & \underline{4.29} & \underline{0.034} & \underline{29.04} & \underline{0.077} & \underline{25.95} & \underline{0.908} & \underline{0.968} \\
PixelMan & 16 & 64  &  9 & \textbf{0.605} & \textbf{69.98} & \textbf{4.35} & \textbf{0.015} & \textbf{35.62} & \textbf{0.074} & \textbf{26.43} & \textbf{0.946} & \textbf{0.974} \\

\hline
PAIR Diffusion   & \multirow{3}{*}{\begin{tabular}[c]{@{}c@{}}50\end{tabular}} & 100 & 32 & 0.525 & 64.76 & 3.51 & 0.088 & 23.87 & 0.112 & 24.69 & 0.836 & 0.863 \\
InfEdit   & & 230 & 35 & \underline{0.567} & \underline{69.07} & \underline{4.29} & \underline{0.034} & \underline{29.04} & \textbf{0.077} & \underline{25.95} & \underline{0.908} & \underline{0.968} \\
PixelMan & & 206 & 27 & \textbf{0.605} & \textbf{70.17} & \textbf{4.36} & \textbf{0.014} & \textbf{35.92} & \textbf{0.077} & \textbf{26.28} & \textbf{0.941} & \textbf{0.974} \\

\hline
PAIR Diffusion   & \multirow{3}{*}{\begin{tabular}[c]{@{}c@{}}16\end{tabular}} & 32 &  12 & 0.538 & 65.08 & 3.54 & 0.088 & 23.84 & 0.113 & 24.53 & 0.838 & 0.868 \\ 
InfEdit   & & 70 &  11 & \underline{0.566} & \underline{69.00} & \underline{4.30} & \underline{0.034} & \underline{28.97} & \textbf{0.071} & \textbf{26.44} & \underline{0.893} & \underline{0.967} \\ 
PixelMan & & 64 &  9 & \textbf{0.605} & \textbf{69.98} & \textbf{4.35} & \textbf{0.015} & \textbf{35.62} & \underline{0.074} & \underline{26.43} & \textbf{0.946} & \textbf{0.974} \\

\hline
PAIR Diffusion   & \multirow{3}{*}{\begin{tabular}[c]{@{}c@{}}8\end{tabular}} & 16 & 7 & 0.545 & 65.12 & 3.45 & 0.090 & 23.63 & 0.119 & 24.10 & 0.832 & 0.861 \\
InfEdit   & & 28 & 5 & \underline{0.563} & \underline{68.78} & \underline{4.30} & \underline{0.035} & \underline{28.80} & \textbf{0.070} & \underline{26.64} & \underline{0.898} & \underline{0.967} \\
PixelMan & & 28 & 4 & \textbf{0.602} & \textbf{69.63} & \textbf{4.32} & \textbf{0.016} & \textbf{35.33} & \underline{0.071} & \textbf{26.70} & \textbf{0.926} & \textbf{0.971} \\

\hline
\end{tabular}

\caption{\textbf{Quantitative results on the COCOEE~\cite{yang2022paint} dataset.}
Comparing PixelMan with additional baselines including PAIR Diffusion~\cite{goel2023pair} and InfEdit~\cite{xu2023infedit}.
The $\downarrow$ indicates lower is better, and the $\uparrow$ means the higher the better.
The best performance result is marked in \textbf{bold} and the second best result is annotated with \underline{underlines}.
Latency measures the average wall-clock time over ten runs for generating one image on this dataset in seconds with a V100 GPU.}
\label{tab:additional_results_cocoee}
\end{table*}

\begin{table*}[hbt!]
\small
\centering
\addtolength{\tabcolsep}{-2.9pt}
\begin{tabular}{c|ccc|ccc|cc|cc|cc} 
\hline
\multirow{2}{*}{\textbf{Method}} &
\multicolumn{3}{c|}{\textbf{Efficiency}} &
\multicolumn{3}{c|}{\textbf{\begin{tabular}[c]{@{}c@{}}Image Quality\\ Assessment\end{tabular}}} &
\multicolumn{2}{c|}{\textbf{\begin{tabular}[c]{@{}c@{}}Object\\ Consistency\end{tabular}}} &
\multicolumn{2}{c|}{\textbf{\begin{tabular}[c]{@{}c@{}}Background\\ Consistency\end{tabular}}}  &
\multicolumn{2}{c}{\textbf{\begin{tabular}[c]{@{}c@{}}Semantic\\ Consistency\end{tabular}}} \\ 

\cline{2-13} & 
\multicolumn{1}{c}{\textbf{\begin{tabular}[c]{@{}c@{}}\#\\Steps\end{tabular}}} &
\multicolumn{1}{c}{\textbf{\begin{tabular}[c]{@{}c@{}}\# NFEs\\$\downarrow$\end{tabular}}} &
\multicolumn{1}{c|}{\textbf{\begin{tabular}[c]{@{}c@{}}Latency\\(secs)~$\downarrow$\end{tabular}}} &
\multicolumn{1}{c}{\textbf{\begin{tabular}[c]{@{}c@{}}TOPIQ\\$\uparrow$\end{tabular}}} &
\multicolumn{1}{c}{\textbf{\begin{tabular}[c]{@{}c@{}}MUSIQ\\$\uparrow$\end{tabular}}} &
\multicolumn{1}{c|}{\textbf{\begin{tabular}[c]{@{}c@{}}LIQE\\$\uparrow$\end{tabular}}} &
\multicolumn{1}{c}{\textbf{\begin{tabular}[c]{@{}c@{}}LPIPS\\$\downarrow$\end{tabular}}} &
\multicolumn{1}{c|}{\textbf{\begin{tabular}[c]{@{}c@{}}PSNR\\$\uparrow$\end{tabular}}} &
\multicolumn{1}{c}{\textbf{\begin{tabular}[c]{@{}c@{}}LPIPS\\$\downarrow$\end{tabular}}} &
\multicolumn{1}{c|}{\textbf{\begin{tabular}[c]{@{}c@{}}PSNR\\$\uparrow$\end{tabular}}}   &
\multicolumn{1}{c}{\textbf{\begin{tabular}[c]{@{}c@{}}CLIP-\\T2T $\uparrow$\end{tabular}}} &
\multicolumn{1}{c}{\textbf{\begin{tabular}[c]{@{}c@{}}CLIP-\\I2I $\uparrow$\end{tabular}}} \\

\hline
PAIR Diffusion   & 50 & 100 & 49 & 0.667 & 72.78 & 4.39 & 0.064 & 24.86 & 0.105 & 24.66 & 0.799 & 0.871 \\
InfEdit   & 50 & 230 & 47 & \underline{0.671} & \underline{74.56} & \textbf{4.72} & \underline{0.029} & \underline{29.85} & \textbf{0.068} & \underline{26.72} & \underline{0.875} & \underline{0.936} \\
PixelMan & 16 & 64  & 11 & \textbf{0.696} & \textbf{74.66} & \underline{4.70} & \textbf{0.015} & \textbf{35.90} & \underline{0.070} & \textbf{27.18} & \textbf{0.898} & \textbf{0.939} \\

\hline
PAIR Diffusion   & \multirow{3}{*}{\begin{tabular}[c]{@{}c@{}}50\end{tabular}} & 100 & 49 & 0.667 & 72.78 & 4.39 & 0.064 & 24.86 & 0.105 & 24.66 & 0.799 & 0.871 \\
InfEdit   & & 230 & 47 & \underline{0.671} & \underline{74.56} & \underline{4.72} & \underline{0.029} & \underline{29.85} & \textbf{0.068} & \underline{26.72} & \underline{0.875} & \underline{0.936} \\
PixelMan & & 206 & 34 & \textbf{0.688} & \textbf{74.72} & \textbf{4.75} & \textbf{0.015} & \textbf{36.26} & \underline{0.073} & \textbf{26.74} & \textbf{0.896} & \textbf{0.940} \\

\hline
PAIR Diffusion   & \multirow{3}{*}{\begin{tabular}[c]{@{}c@{}}16\end{tabular}} & 32 & 18 & 0.673 & 72.75 & 4.35 & 0.064 & 24.84 & 0.106 & 24.58 & 0.801 & 0.871 \\
InfEdit   & & 70 & 15 & \underline{0.676} & \textbf{74.67} & \textbf{4.74} & \underline{0.030} & \underline{29.74} & \textbf{0.067} & \underline{27.01} & \underline{0.870} & \underline{0.933} \\
PixelMan & & 64 & 11 & \textbf{0.696} & \underline{74.66} & \underline{4.70} & \textbf{0.015} & \textbf{35.90} & \underline{0.070} & \textbf{27.18} & \textbf{0.898} & \textbf{0.939} \\

\hline
PAIR Diffusion   & \multirow{3}{*}{\begin{tabular}[c]{@{}c@{}}8\end{tabular}} & 16 & 11 & \underline{0.679} & 72.74 & 4.27 & 0.064 & 24.76 & 0.110 & 24.33 & 0.795 & 0.867 \\
InfEdit   & & 28 & 7 & 0.673 & \underline{74.56} & \textbf{4.73} & \underline{0.030} & \underline{29.63} & \textbf{0.066} & \underline{27.32} & \underline{0.871} & \underline{0.931} \\
PixelMan & & 28 & 5 & \textbf{0.695} & \textbf{74.59} & \underline{4.67} & \textbf{0.016} & \textbf{35.57} & \underline{0.067} & \textbf{27.74} & \textbf{0.900} & \textbf{0.937} \\

\hline
\end{tabular}

\caption{\textbf{Quantitative on the ReS~\cite{yang2022paint} dataset.}
Comparing PixelMan with additional baselines including PAIR Diffusion~\cite{goel2023pair} and InfEdit~\cite{xu2023infedit}.
The $\downarrow$ indicates lower is better, and the $\uparrow$ means the higher the better.
The best performance result is marked in \textbf{bold} and the second best result is annotated with \underline{underlines}.
Latency measures the average wall-clock time over ten runs for generating one image on this dataset in seconds with a V100 GPU.}
\label{tab:additional_results_res}
\end{table*}

\begin{figure*}[hbt!]
    \centering
    \setlength{\tabcolsep}{0.4pt}
    \renewcommand{\arraystretch}{0.4}
    {\footnotesize
    \begin{tabular}{c c c c c c c c}
        &
        \multicolumn{1}{c}{(a)} &
        \multicolumn{1}{c}{(b)} &
        \multicolumn{1}{c}{(c)} &
        \multicolumn{1}{c}{(d)} &
        \multicolumn{1}{c}{(e)} &
        \multicolumn{1}{c}{(f)} \\

        {\raisebox{0.34in}{
        \multirow{1}{*}{\rotatebox{0}{Input}}}} &
        \includegraphics[width=0.105\textwidth]{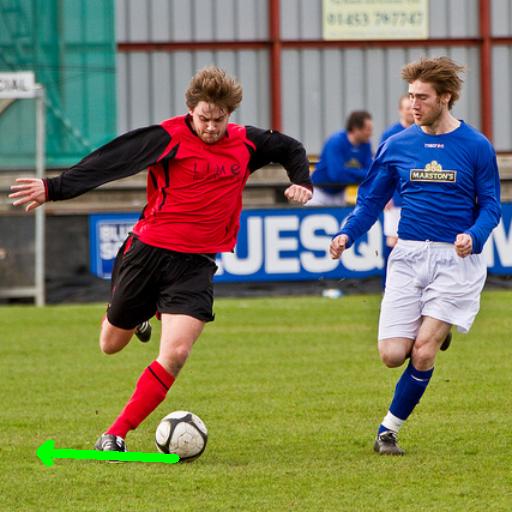} &
        \includegraphics[width=0.105\textwidth]{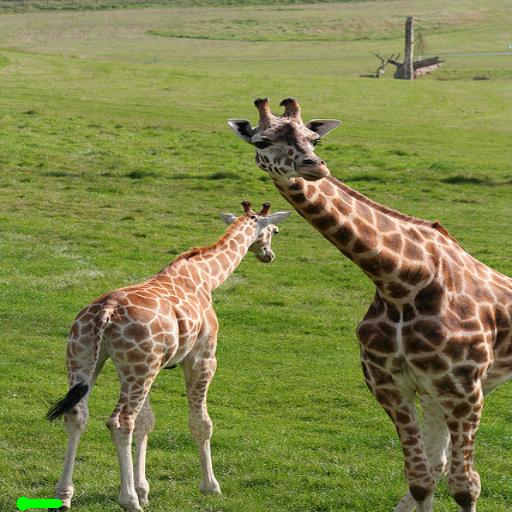} &
        \includegraphics[width=0.105\textwidth]{images/comparison/COCOEE/000000111930_GT_source.jpg} &
        \includegraphics[width=0.105\textwidth]{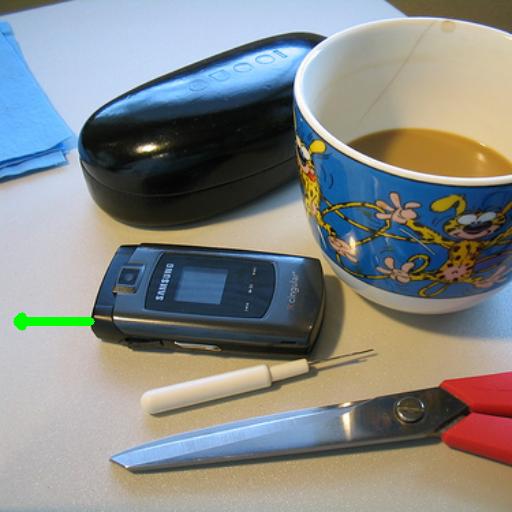} &
        \includegraphics[width=0.105\textwidth]{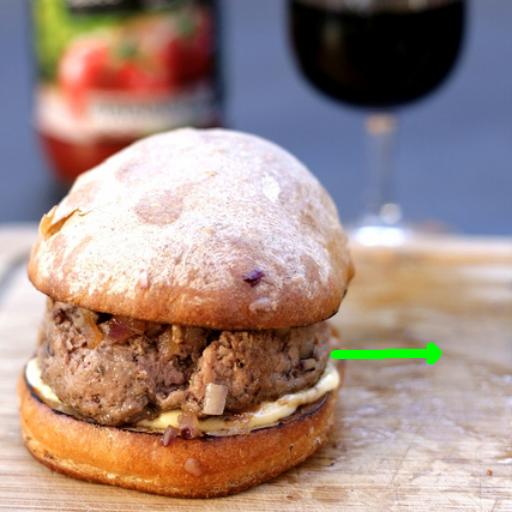} &
        \includegraphics[width=0.105\textwidth]{images/comparison/COCOEE/000001557820_GT_source.jpg} &\\

        {\raisebox{0.47in}{\multirow{1}{*}{\begin{tabular}{c}SDv2-Inpainting\\+AnyDoor \\ (50 steps, 15s)\end{tabular}}}}
        &
        \includegraphics[width=0.105\textwidth]{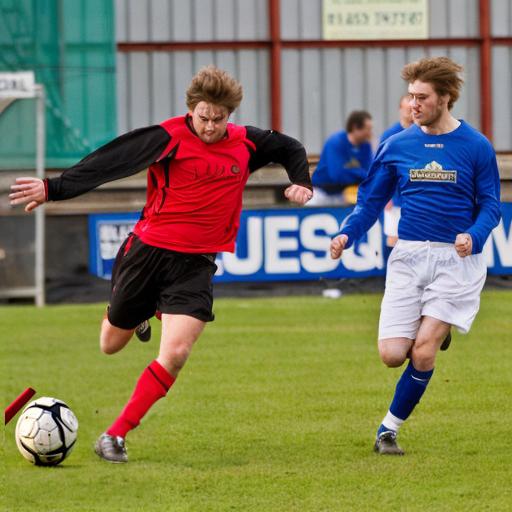} &
        \includegraphics[width=0.105\textwidth]{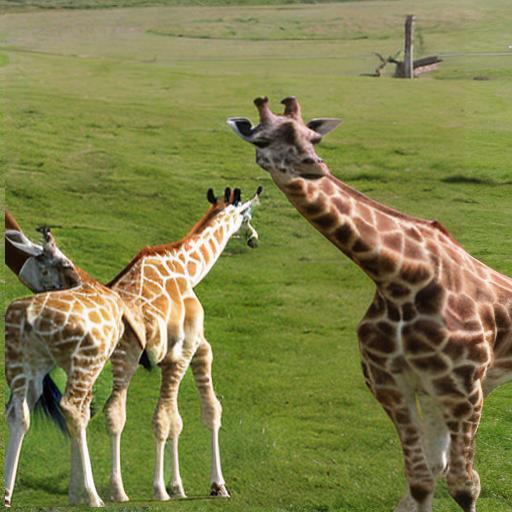} &
        \includegraphics[width=0.105\textwidth]{images/comparison/COCOEE/000000111930_GT_anydoor50.jpg} &
        \includegraphics[width=0.105\textwidth]{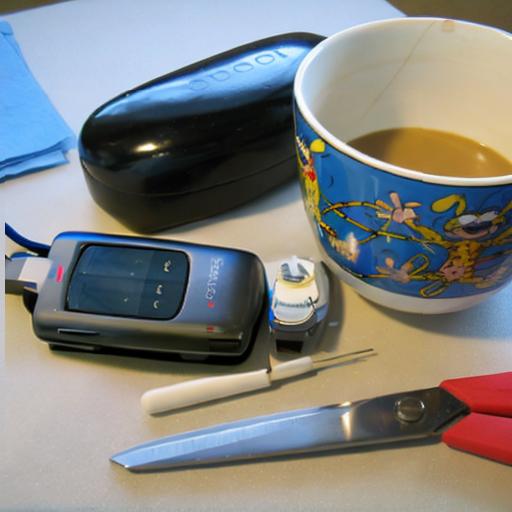} &
        \includegraphics[width=0.105\textwidth]{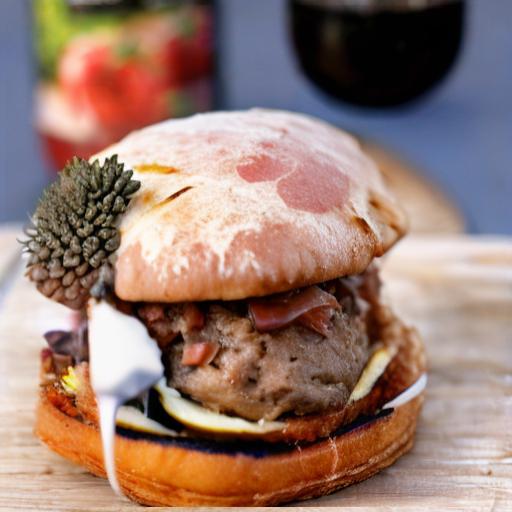} &
        \includegraphics[width=0.105\textwidth]{images/comparison/COCOEE/000001557820_GT_anydoor50.jpg} &\\

        {\raisebox{0.47in}{\multirow{1}{*}{\begin{tabular}{c}SDv2-Inpainting\\+AnyDoor \\ (16 steps, 5s)\end{tabular}}}}
        &
        \includegraphics[width=0.105\textwidth]{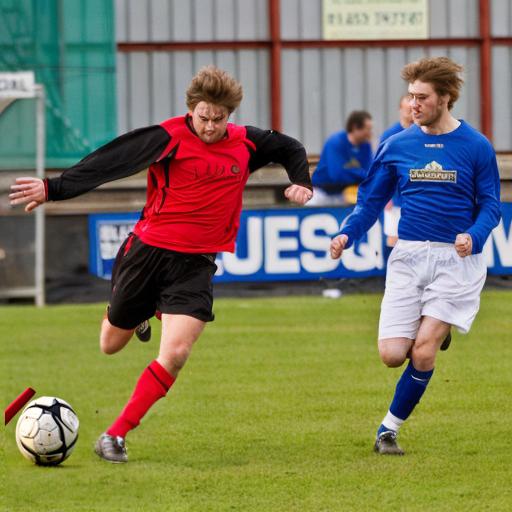} &
        \includegraphics[width=0.105\textwidth]{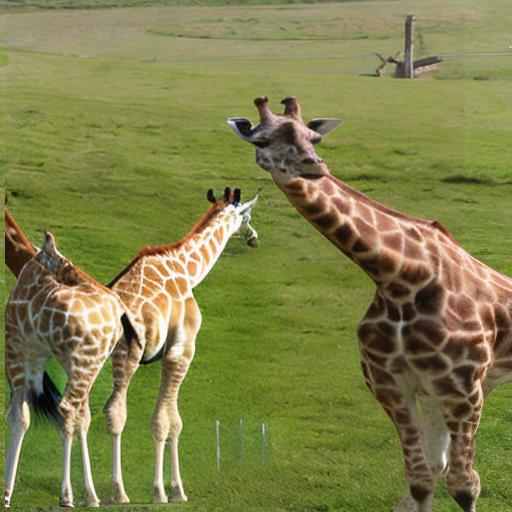} &
        \includegraphics[width=0.105\textwidth]{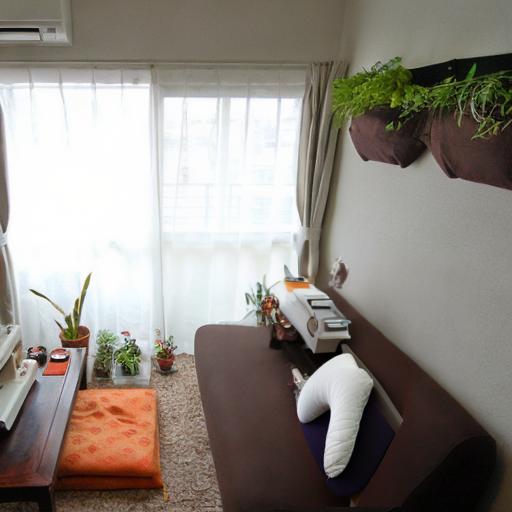} &
        \includegraphics[width=0.105\textwidth]{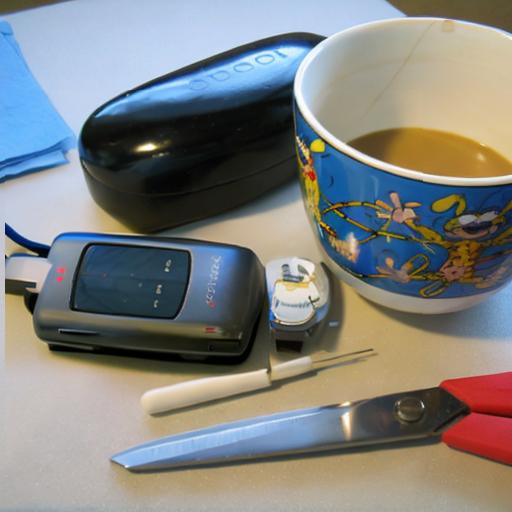} &
        \includegraphics[width=0.105\textwidth]{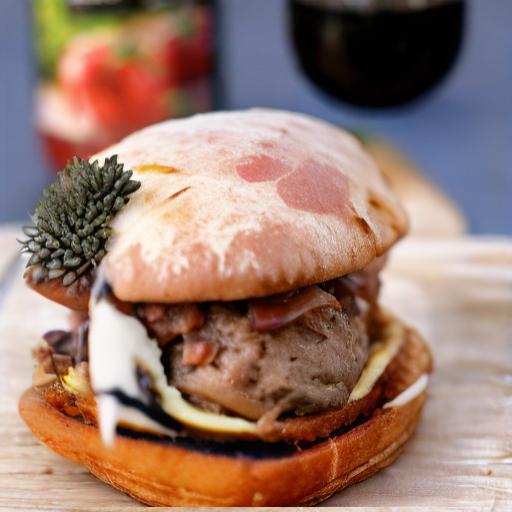} &
        \includegraphics[width=0.105\textwidth]{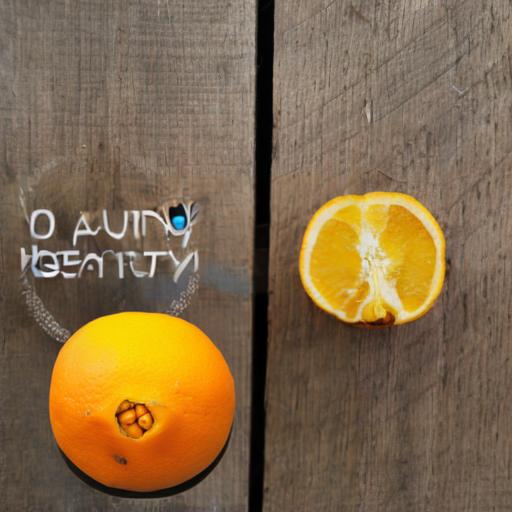} &\\
        
        {\raisebox{0.37in}{\multirow{1}{*}{\begin{tabular}{c}SelfGuidance \\ (50 steps, 11s)\end{tabular}}}} &
        \includegraphics[width=0.105\textwidth]{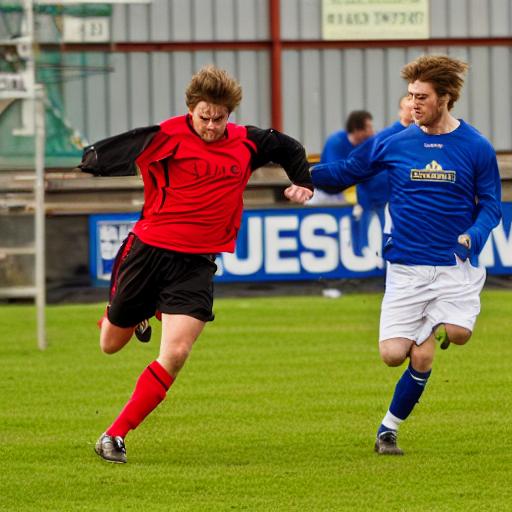} &
        \includegraphics[width=0.105\textwidth]{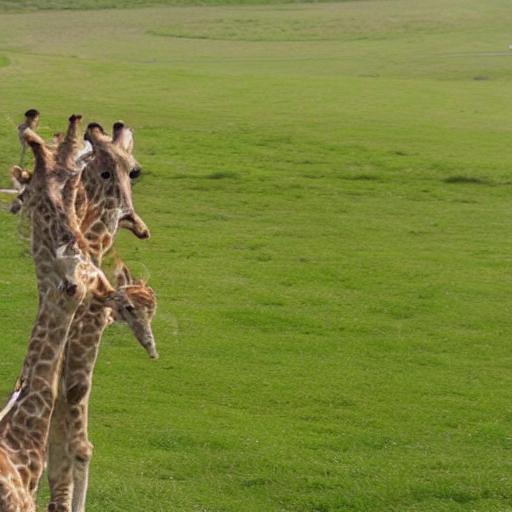} &
        \includegraphics[width=0.105\textwidth]{images/comparison/COCOEE/000000111930_GT_sg50.jpg} &
        \includegraphics[width=0.105\textwidth]{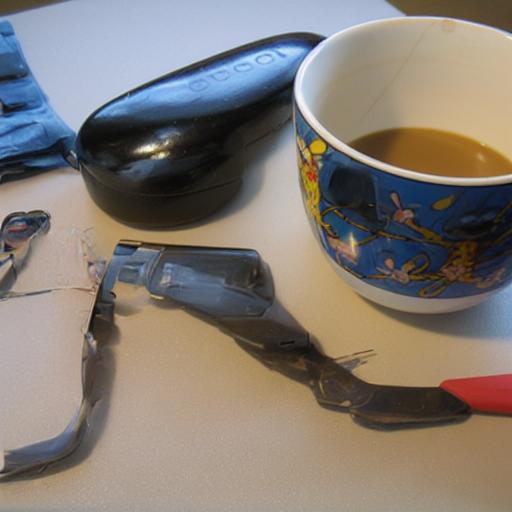} &
        \includegraphics[width=0.105\textwidth]{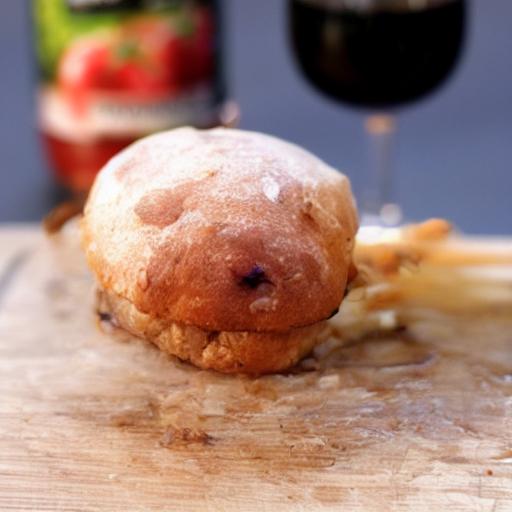} &
        \includegraphics[width=0.105\textwidth]{images/comparison/COCOEE/000001557820_GT_sg50.jpg} &\\
        
        {\raisebox{0.37in}{\multirow{1}{*}{\begin{tabular}{c}SelfGuidance \\ (16 steps, 4s)\end{tabular}}}} &
        \includegraphics[width=0.105\textwidth]{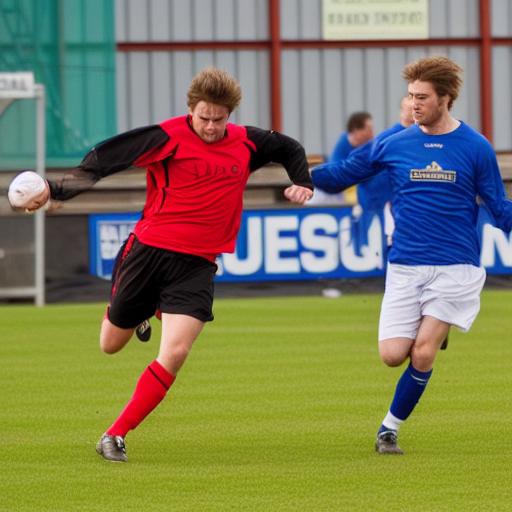} &
        \includegraphics[width=0.105\textwidth]{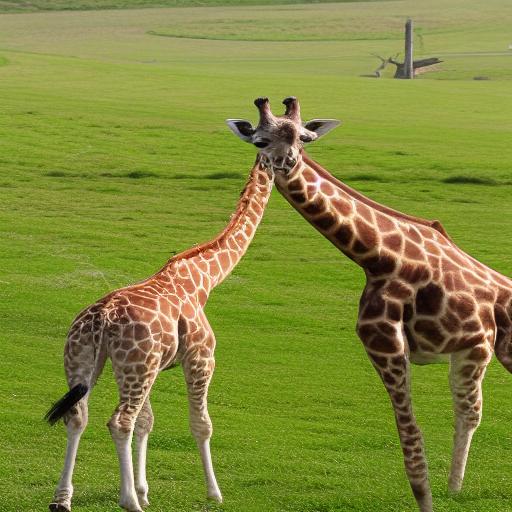} &
        \includegraphics[width=0.105\textwidth]{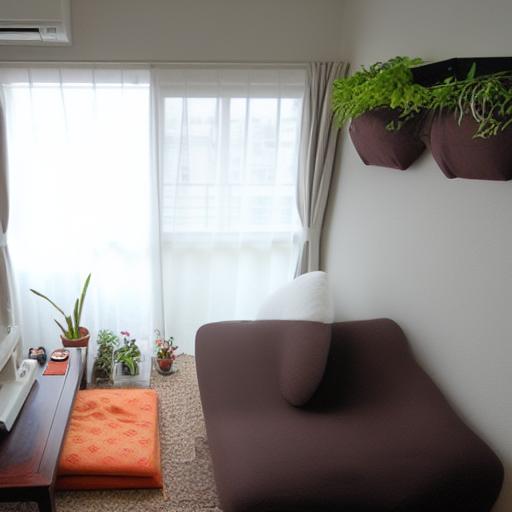} &
        \includegraphics[width=0.105\textwidth]{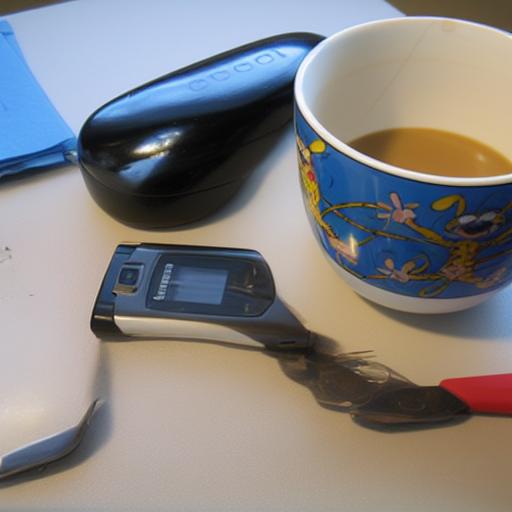} &
        \includegraphics[width=0.105\textwidth]{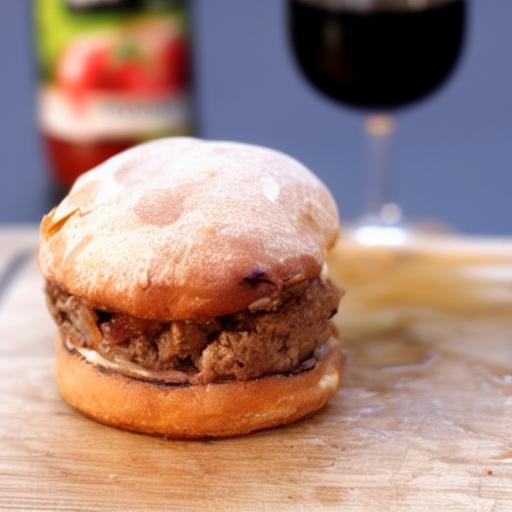} &
        \includegraphics[width=0.105\textwidth]{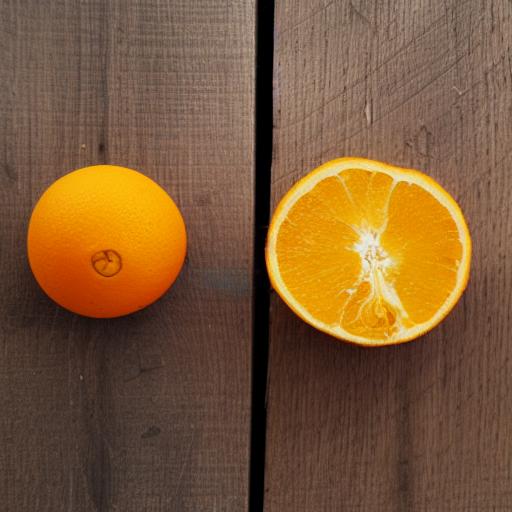} &\\
        
        {\raisebox{0.37in}{\multirow{1}{*}{\begin{tabular}{c}DragonDiffusion \\ (50 steps, 23s)\end{tabular}}}} &
        \includegraphics[width=0.105\textwidth]{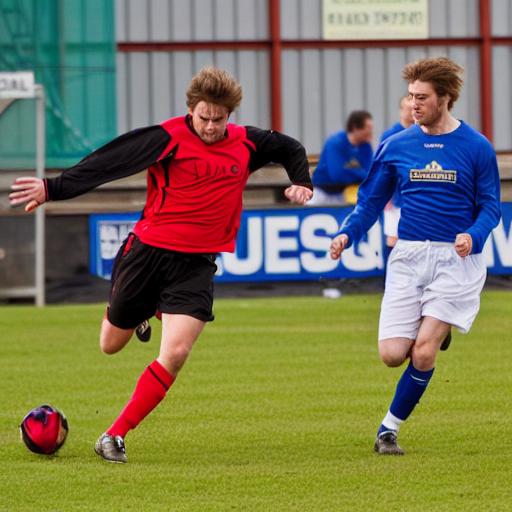} &
        \includegraphics[width=0.105\textwidth]{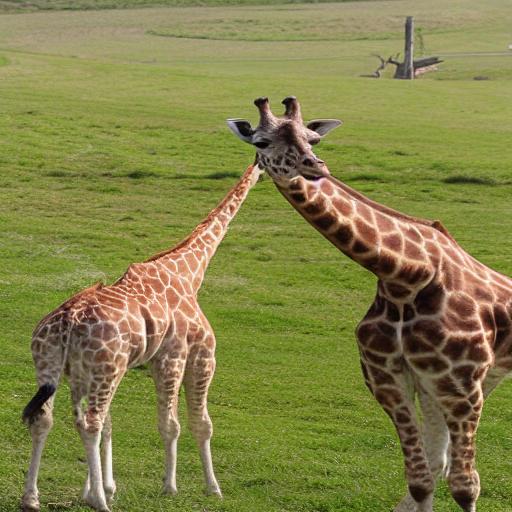} &
        \includegraphics[width=0.105\textwidth]{images/comparison/COCOEE/000000111930_GT_dd50.jpg} &
        \includegraphics[width=0.105\textwidth]{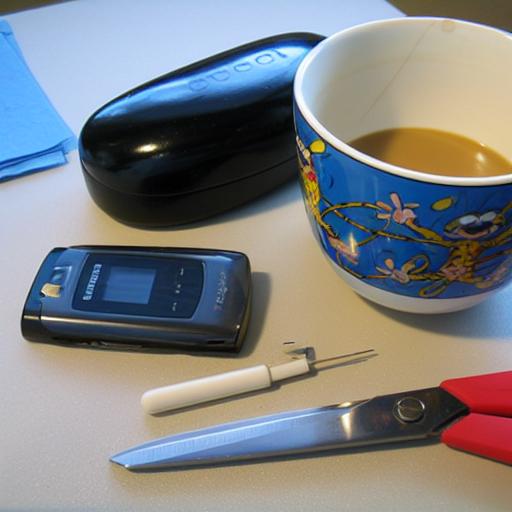} &
        \includegraphics[width=0.105\textwidth]{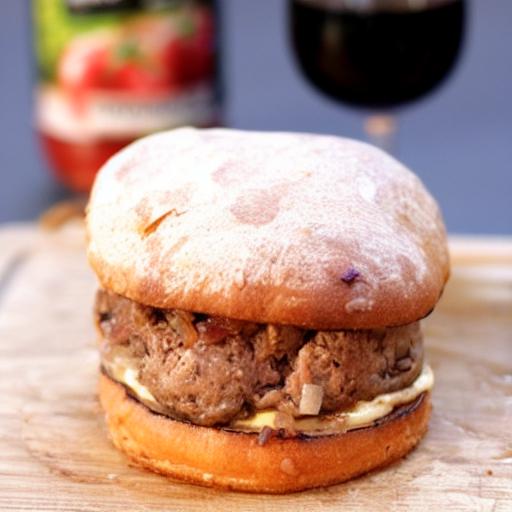} &
        \includegraphics[width=0.105\textwidth]{images/comparison/COCOEE/000001557820_GT_dd50.jpg} &\\

        {\raisebox{0.37in}{\multirow{1}{*}{\begin{tabular}{c}DragonDiffusion \\ (16 steps, 9s)\end{tabular}}}} &
        \includegraphics[width=0.105\textwidth]{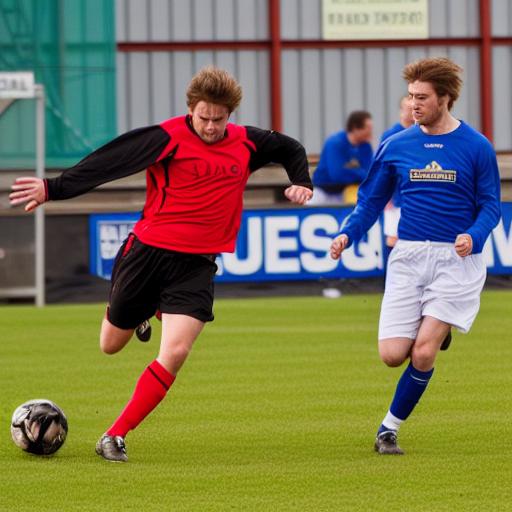} &
        \includegraphics[width=0.105\textwidth]{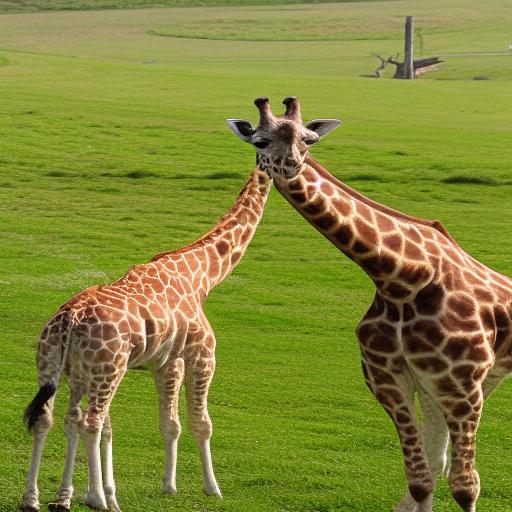} &
        \includegraphics[width=0.105\textwidth]{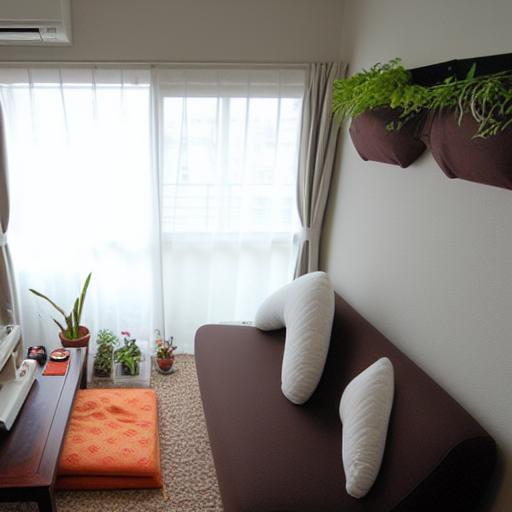} &
        \includegraphics[width=0.105\textwidth]{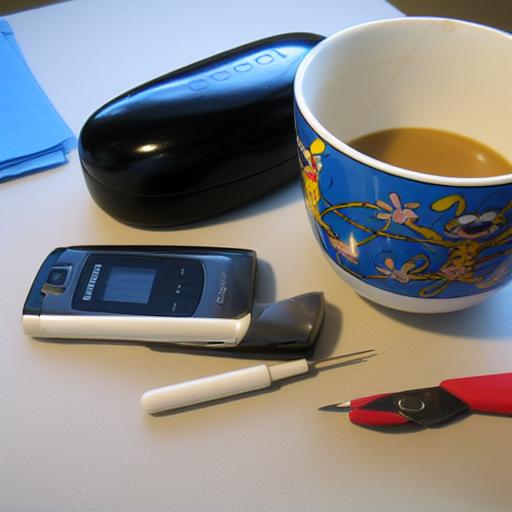} &
        \includegraphics[width=0.105\textwidth]{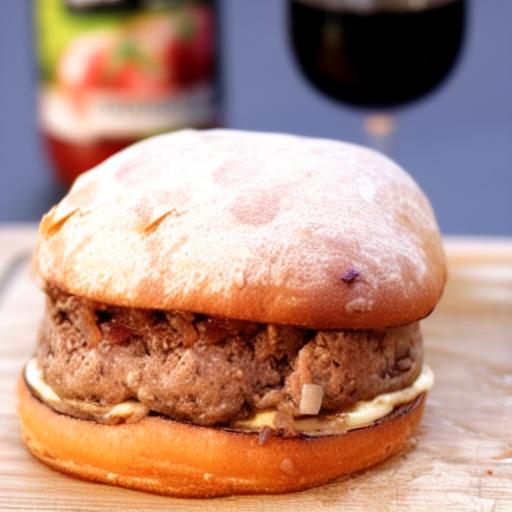} &
        \includegraphics[width=0.105\textwidth]{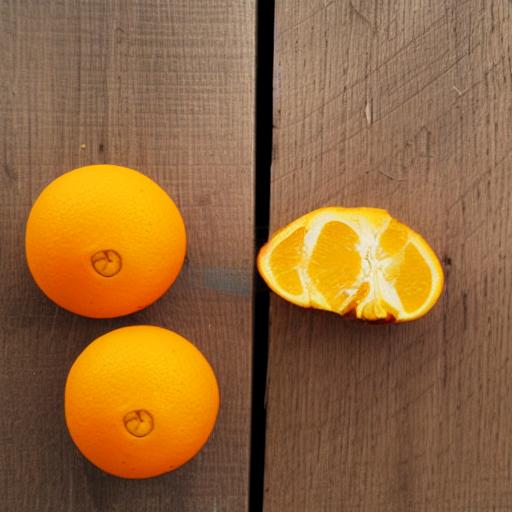} &\\

        {\raisebox{0.37in}{\multirow{1}{*}{\begin{tabular}{c}DiffEditor \\ (50 steps, 24s)\end{tabular}}}} &
        \includegraphics[width=0.105\textwidth]{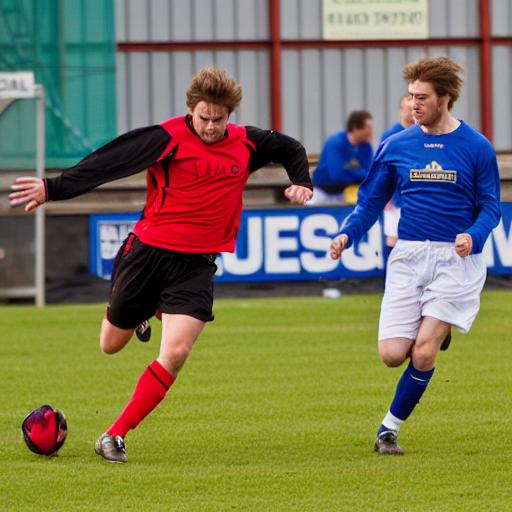} &
        \includegraphics[width=0.105\textwidth]{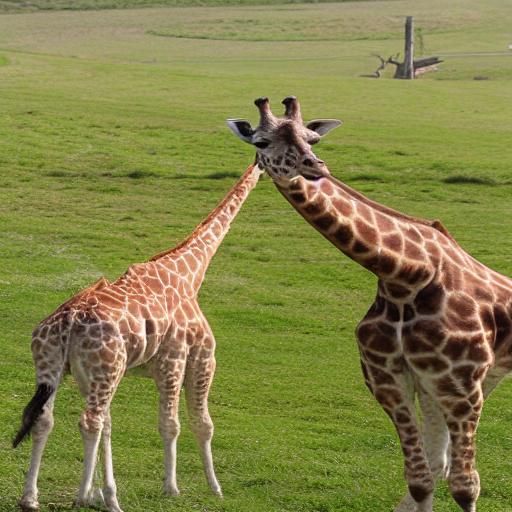} &
        \includegraphics[width=0.105\textwidth]{images/comparison/COCOEE/000000111930_GT_de50.jpg} &
        \includegraphics[width=0.105\textwidth]{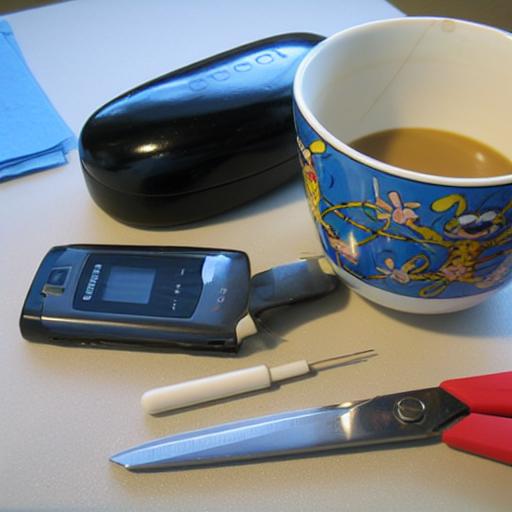} &
        \includegraphics[width=0.105\textwidth]{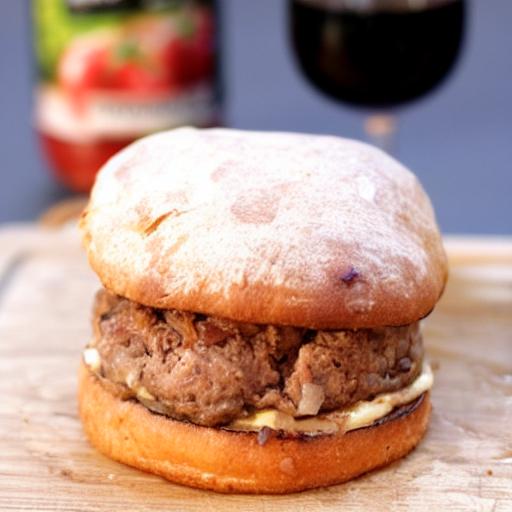} &
        \includegraphics[width=0.105\textwidth]{images/comparison/COCOEE/000001557820_GT_de50.jpg} &\\

        {\raisebox{0.37in}{\multirow{1}{*}{\begin{tabular}{c}DiffEditor \\ (16 steps, 9s)\end{tabular}}}} &
        \includegraphics[width=0.105\textwidth]{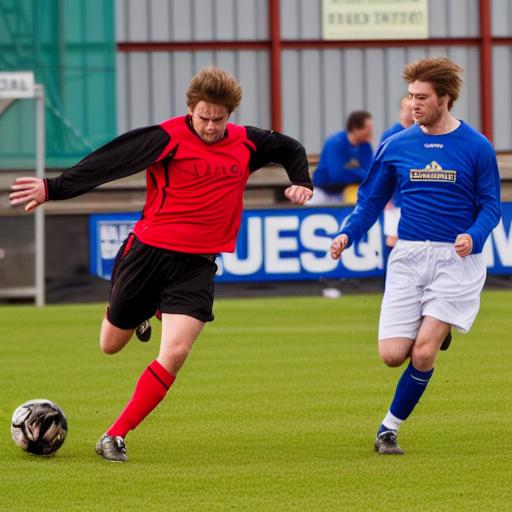} &
        \includegraphics[width=0.105\textwidth]{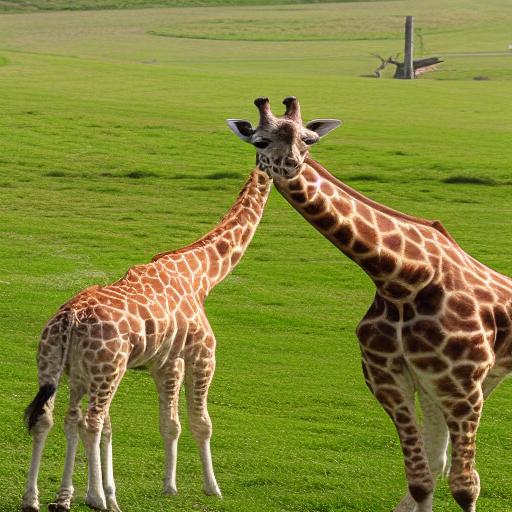} &
        \includegraphics[width=0.105\textwidth]{images/comparison/COCOEE/000000111930_GT_de16.jpg} &
        \includegraphics[width=0.105\textwidth]{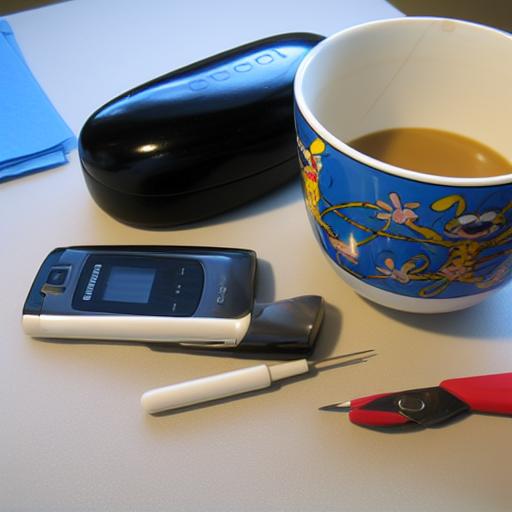} &
        \includegraphics[width=0.105\textwidth]{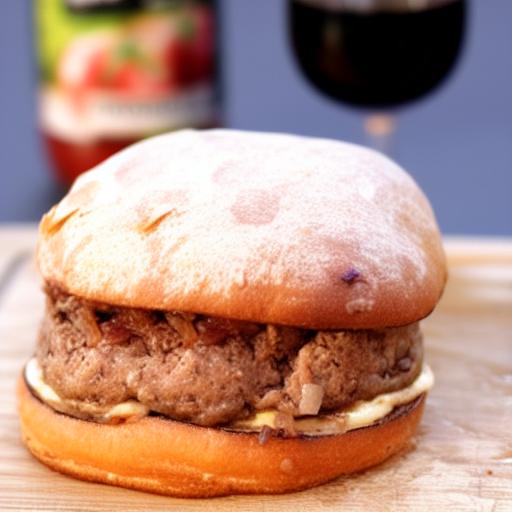} &
        \includegraphics[width=0.105\textwidth]{images/comparison/COCOEE/000001557820_GT_de16.jpg} &\\

        {\raisebox{0.37in}{\multirow{1}{*}{\begin{tabular}{c}\textbf{PixelMan} \\ (50 steps, 27s)\end{tabular}}}} &
        \includegraphics[width=0.105\textwidth]{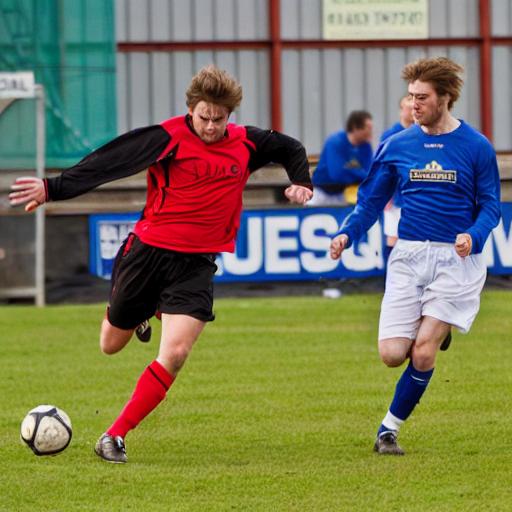} &
        \includegraphics[width=0.105\textwidth]{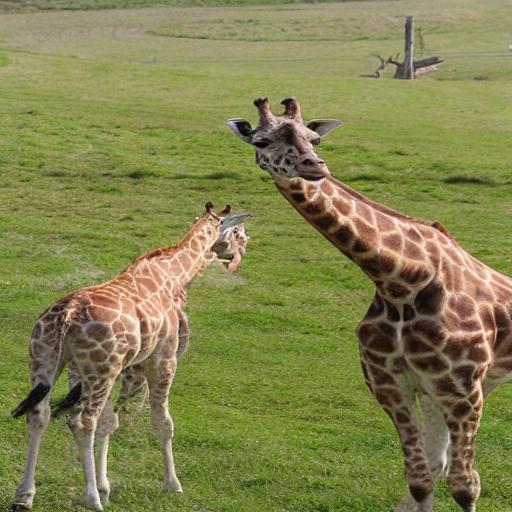} &
        \includegraphics[width=0.105\textwidth]{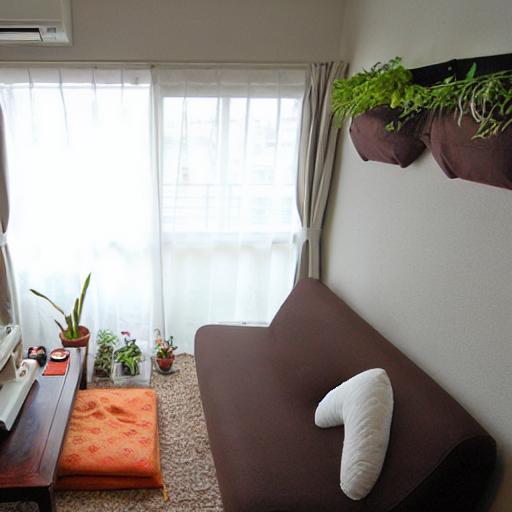} &
        \includegraphics[width=0.105\textwidth]{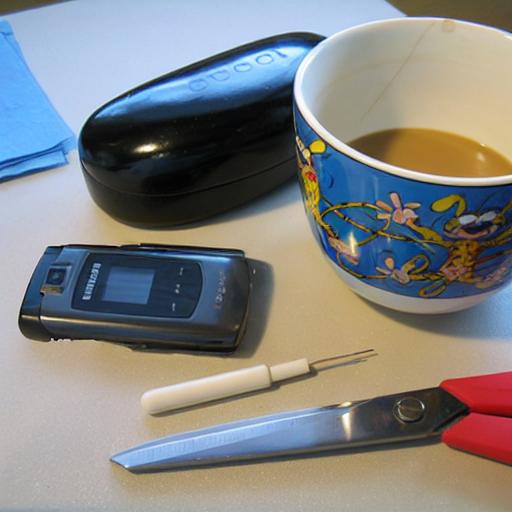} &
        \includegraphics[width=0.105\textwidth]{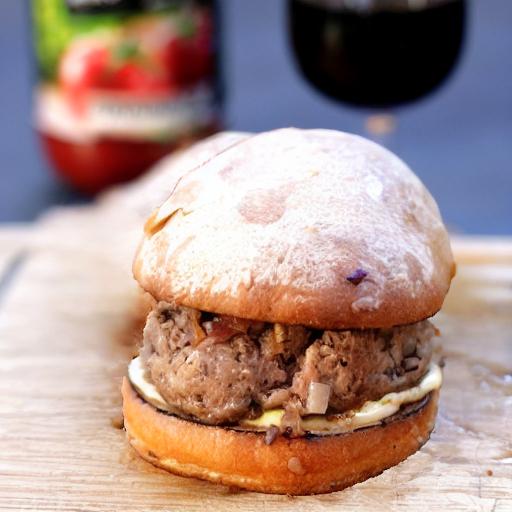} &
        \includegraphics[width=0.105\textwidth]{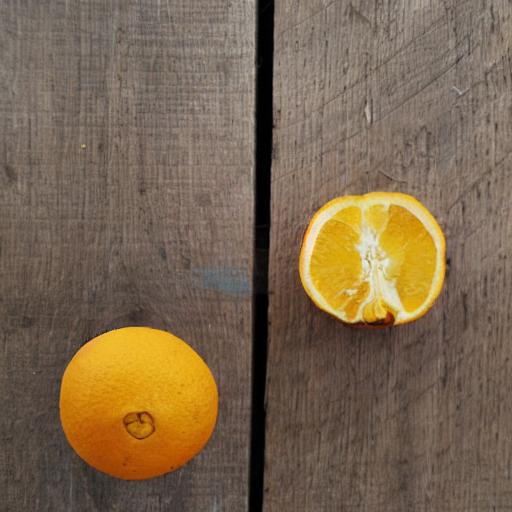} &\\
        
        {\raisebox{0.37in}{\multirow{1}{*}{\begin{tabular}{c}\textbf{PixelMan} \\ (16 steps, 9s)\end{tabular}}}} &
        \includegraphics[width=0.105\textwidth]{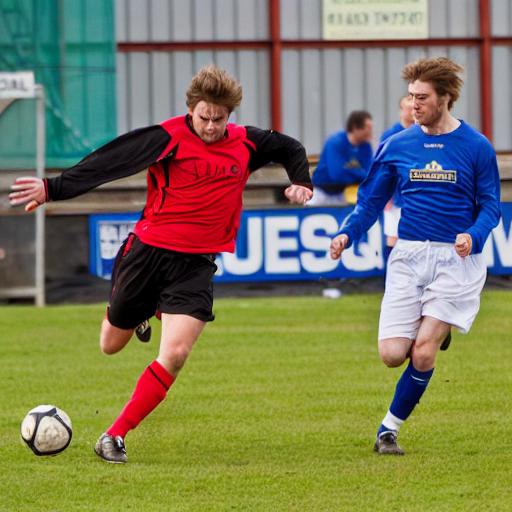} &
        \includegraphics[width=0.105\textwidth]{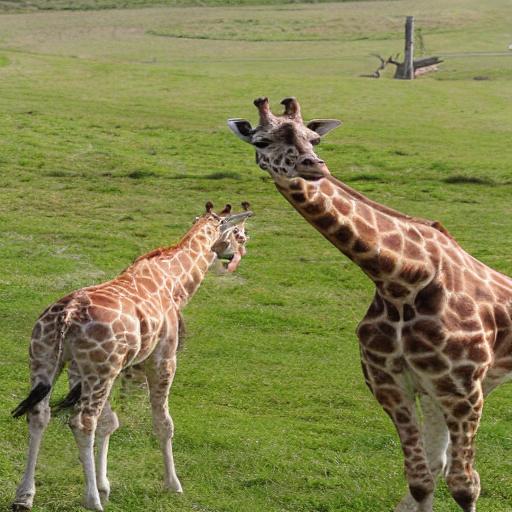} &
        \includegraphics[width=0.105\textwidth]{images/comparison/COCOEE/000000111930_GT_ours16.jpg} &
        \includegraphics[width=0.105\textwidth]{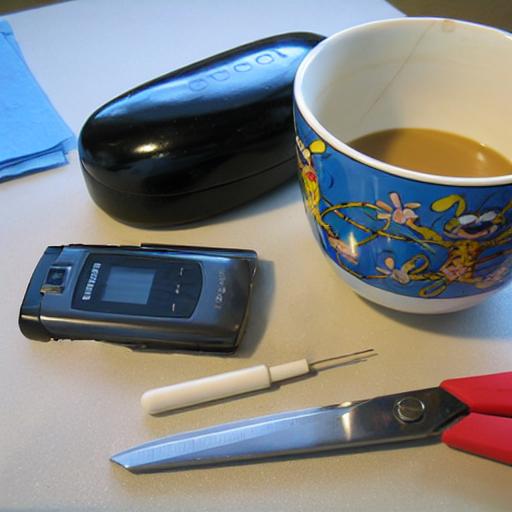} &
        \includegraphics[width=0.105\textwidth]{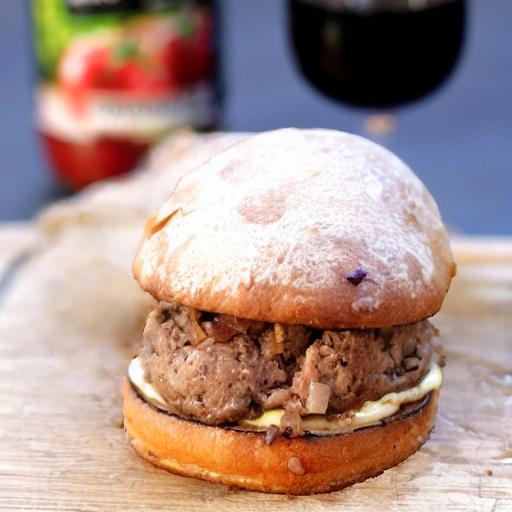} &
        \includegraphics[width=0.105\textwidth]{images/comparison/COCOEE/000001557820_GT_ours16.jpg} &\\

    \end{tabular}
    }
    \caption{
        \textbf{Additional qualitative comparison} on the COCOEE dataset at both 16 and 50 steps. 
    }
    \label{fig:examples_comparisons_2}
\end{figure*}
\begin{figure*}[hbt!]
    \centering
    \setlength{\tabcolsep}{0.4pt}
    \renewcommand{\arraystretch}{0.4}
    {\footnotesize
    \begin{tabular}{c c c c c c c c}
        &
        \multicolumn{1}{c}{(a)} &
        \multicolumn{1}{c}{(b)} &
        \multicolumn{1}{c}{(c)} &
        \multicolumn{1}{c}{(d)} &
        \multicolumn{1}{c}{(e)} &
        \multicolumn{1}{c}{(f)} \\

        {\raisebox{0.34in}{
        \multirow{1}{*}{\rotatebox{0}{Input}}}} &
        \includegraphics[width=0.105\textwidth]{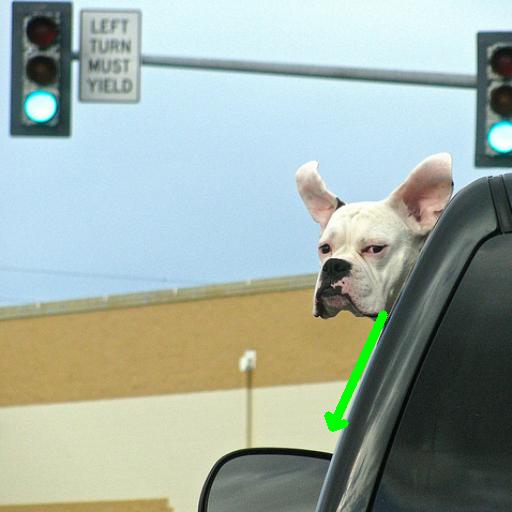} &
        \includegraphics[width=0.105\textwidth]{images/comparison/COCOEE/000000061097_GT_source.jpg} &
        \includegraphics[width=0.105\textwidth]{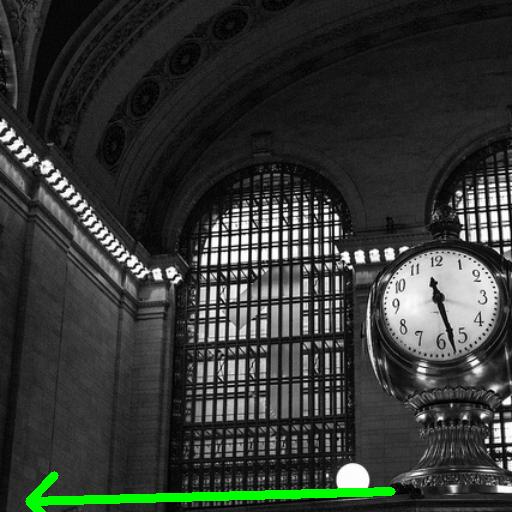} &
        \includegraphics[width=0.105\textwidth]{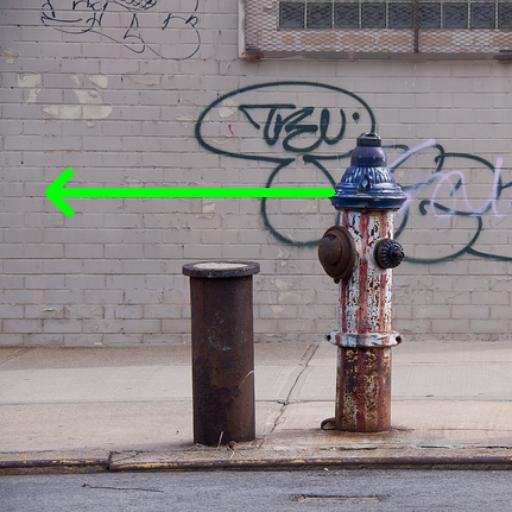} &
        \includegraphics[width=0.105\textwidth]{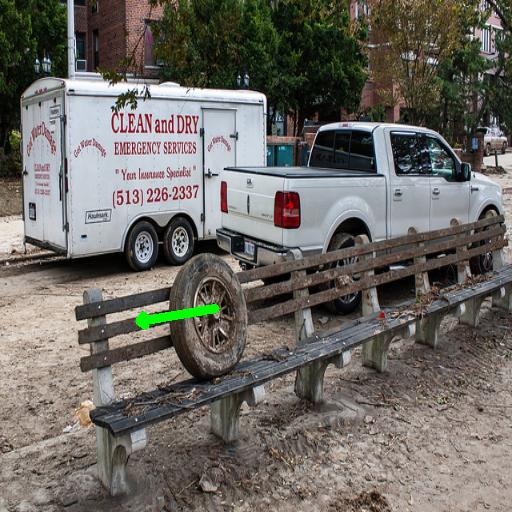} &
        \includegraphics[width=0.105\textwidth]{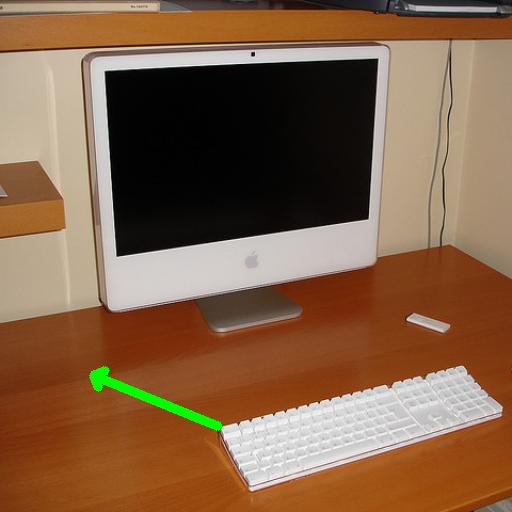} &\\

        {\raisebox{0.47in}{\multirow{1}{*}{\begin{tabular}{c}SDv2-Inpainting\\+AnyDoor\\ (50 steps, 15s)\end{tabular}}}}
        &
        \includegraphics[width=0.105\textwidth]{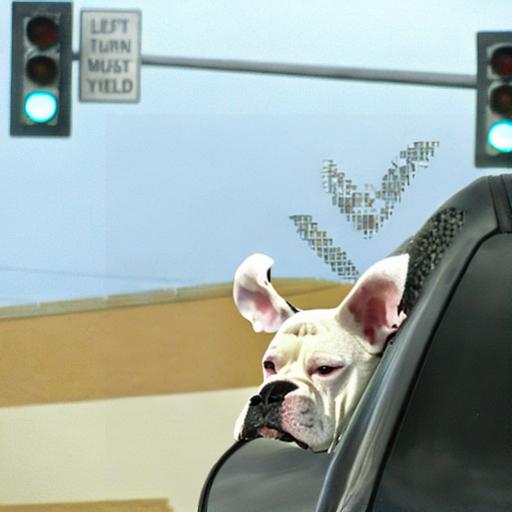} &
        \includegraphics[width=0.105\textwidth]{images/comparison/COCOEE/000000061097_GT_anydoor50.jpg} &
        \includegraphics[width=0.105\textwidth]{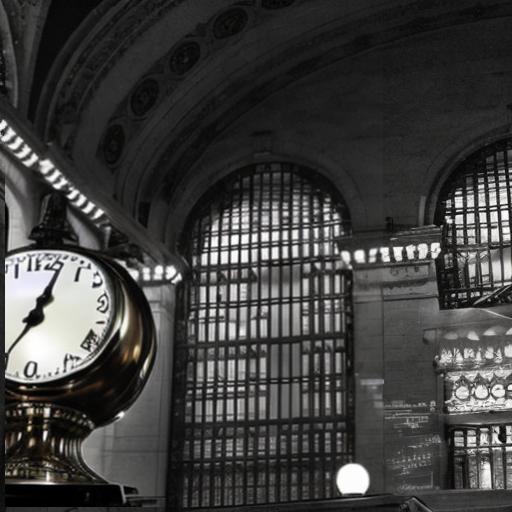} &
        \includegraphics[width=0.105\textwidth]{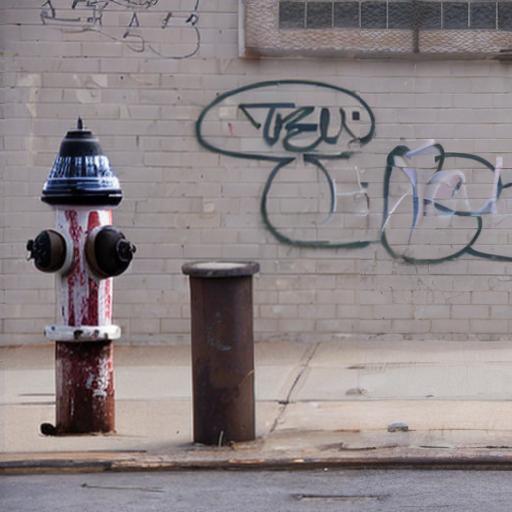} &
        \includegraphics[width=0.105\textwidth]{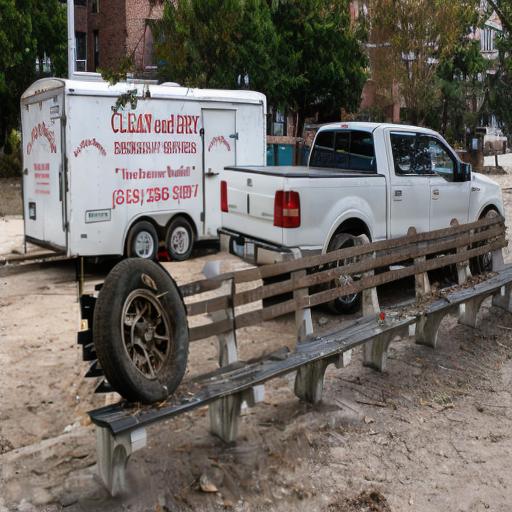} &
        \includegraphics[width=0.105\textwidth]{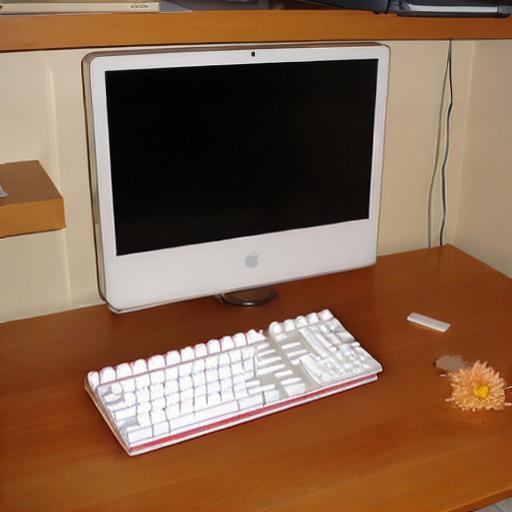} &\\

        {\raisebox{0.47in}{\multirow{1}{*}{\begin{tabular}{c}SDv2-Inpainting\\+AnyDoor\\ (16 steps, 5s)\end{tabular}}}}
        &
        \includegraphics[width=0.105\textwidth]{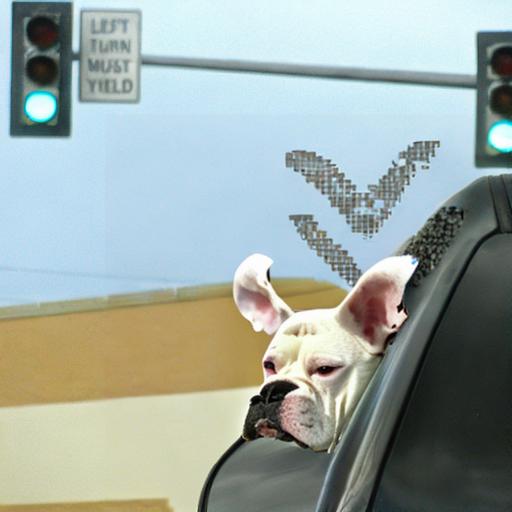} &
        \includegraphics[width=0.105\textwidth]{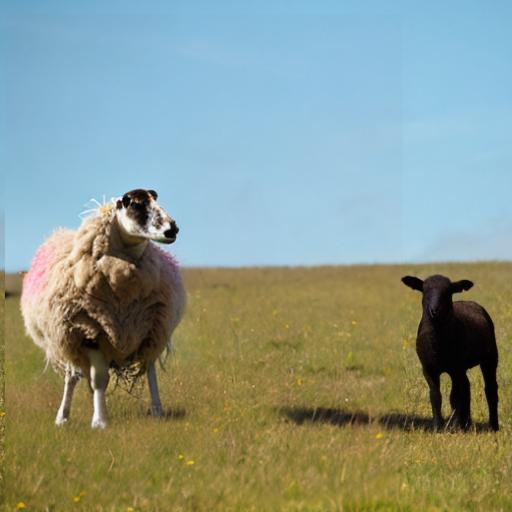} &
        \includegraphics[width=0.105\textwidth]{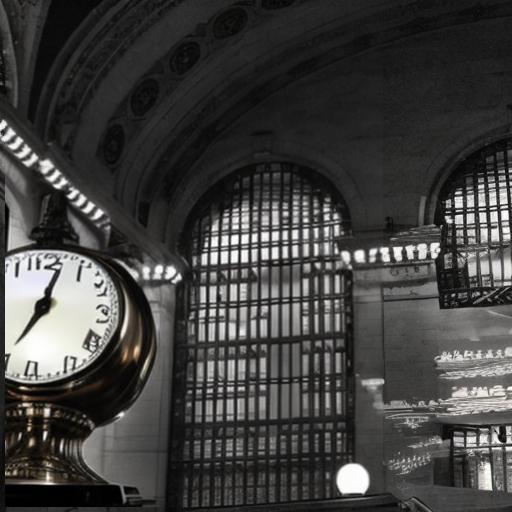} &
        \includegraphics[width=0.105\textwidth]{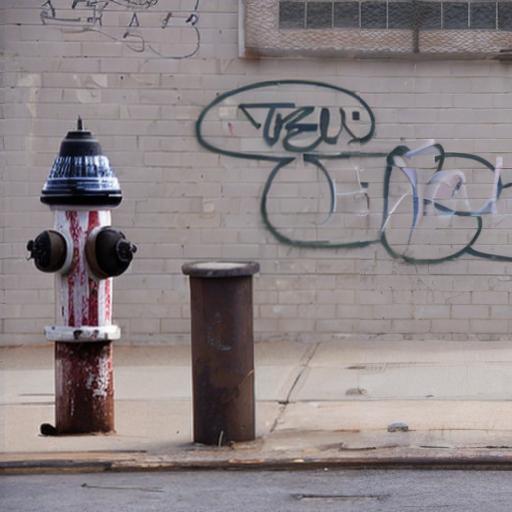} &
        \includegraphics[width=0.105\textwidth]{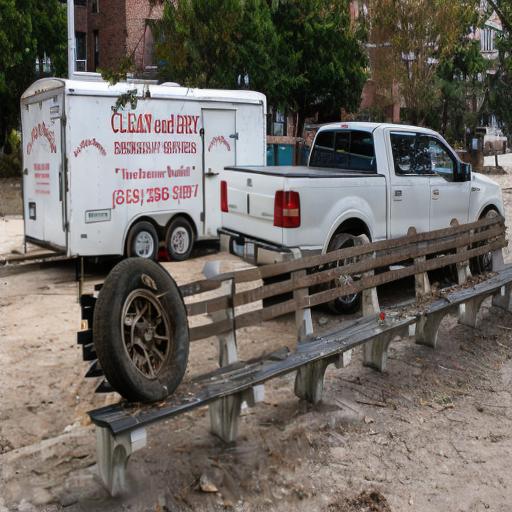} &
        \includegraphics[width=0.105\textwidth]{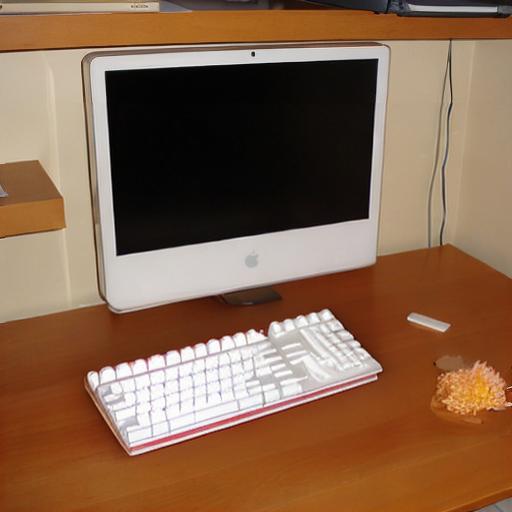} &\\
        
        {\raisebox{0.37in}{\multirow{1}{*}{\begin{tabular}{c}SelfGuidance \\ (50 steps, 11s)\end{tabular}}}} &
        \includegraphics[width=0.105\textwidth]{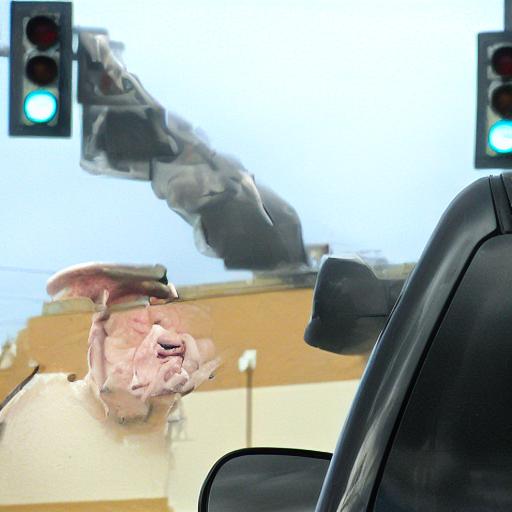} &
        \includegraphics[width=0.105\textwidth]{images/comparison/COCOEE/000000061097_GT_sg50.jpg} &
        \includegraphics[width=0.105\textwidth]{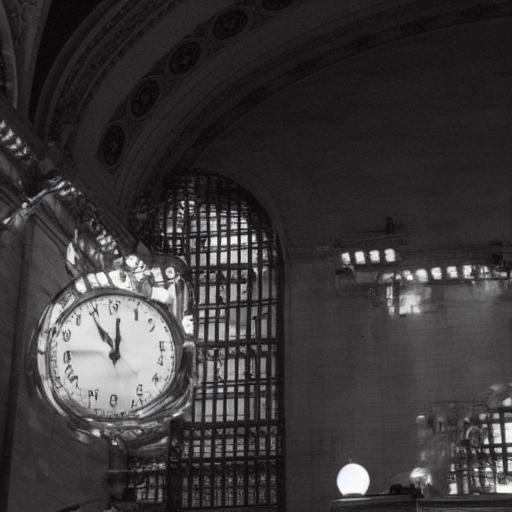} &
        \includegraphics[width=0.105\textwidth]{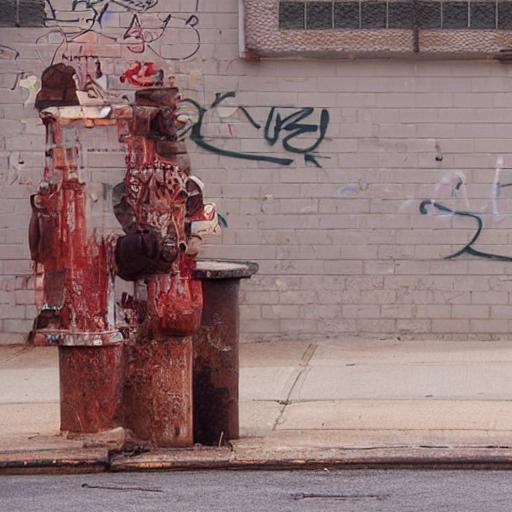} &
        \includegraphics[width=0.105\textwidth]{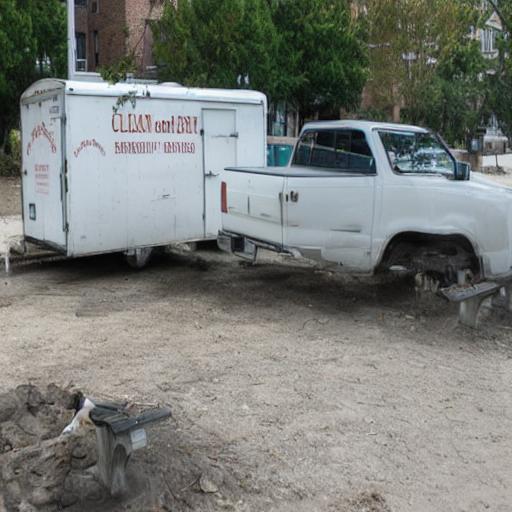} &
        \includegraphics[width=0.105\textwidth]{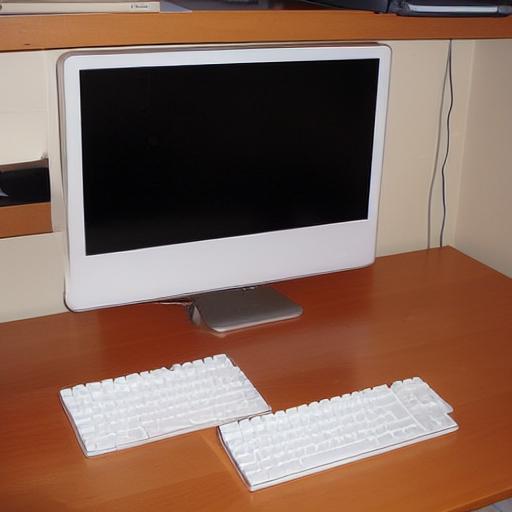} &\\
        
        {\raisebox{0.37in}{\multirow{1}{*}{\begin{tabular}{c}SelfGuidance \\ (16 steps, 4s)\end{tabular}}}} &
        \includegraphics[width=0.105\textwidth]{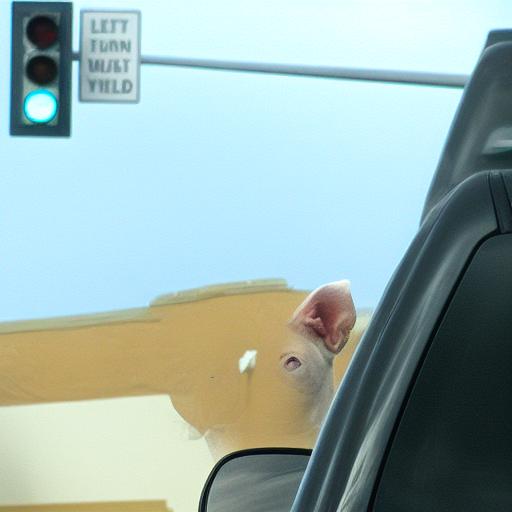} &
        \includegraphics[width=0.105\textwidth]{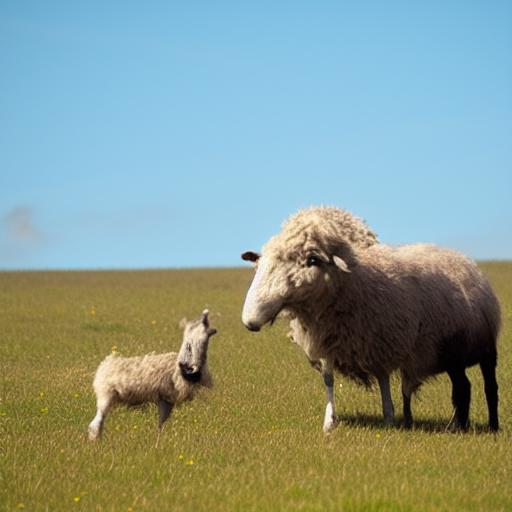} &
        \includegraphics[width=0.105\textwidth]{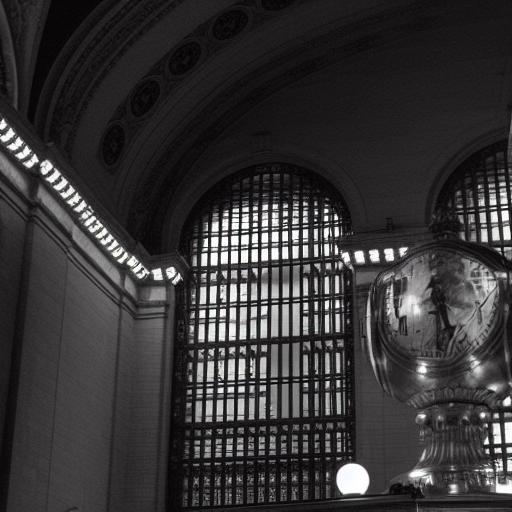} &
        \includegraphics[width=0.105\textwidth]{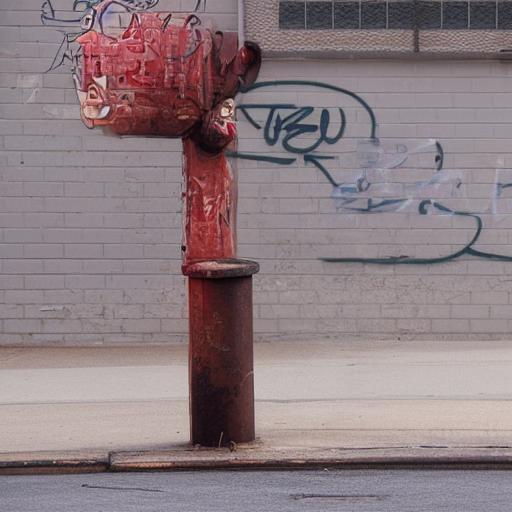} &
        \includegraphics[width=0.105\textwidth]{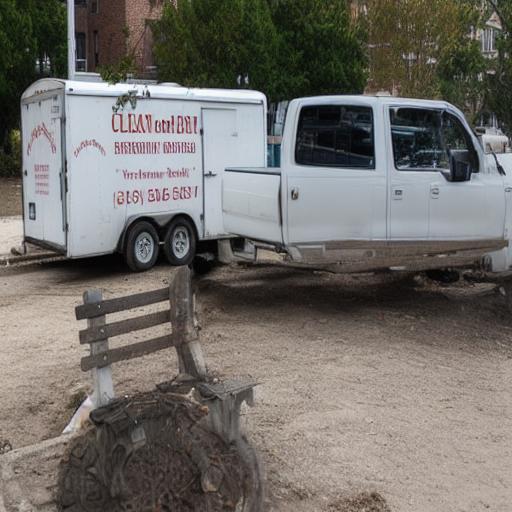} &
        \includegraphics[width=0.105\textwidth]{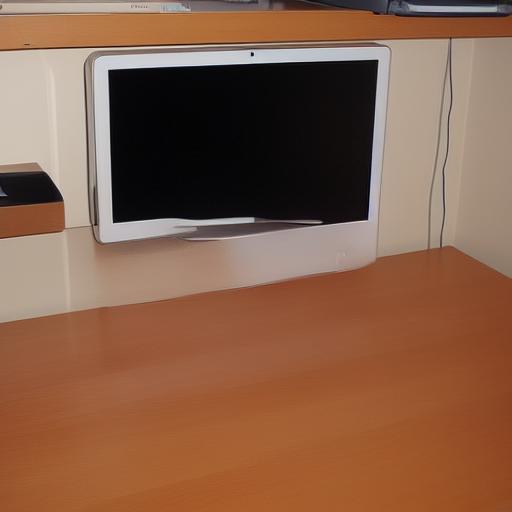} &\\
        
        {\raisebox{0.37in}{\multirow{1}{*}{\begin{tabular}{c}DragonDiffusion \\ (50 steps, 23s)\end{tabular}}}} &
        \includegraphics[width=0.105\textwidth]{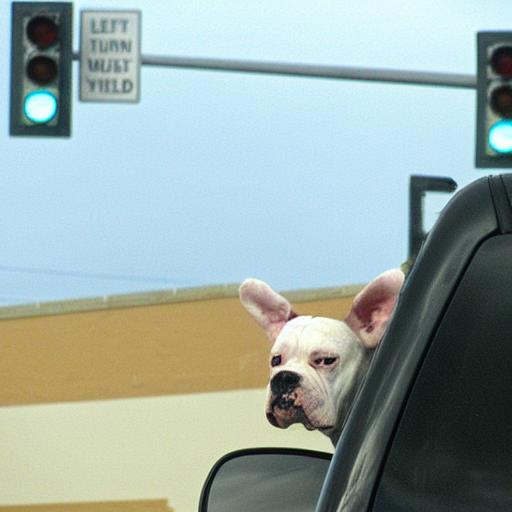} &
        \includegraphics[width=0.105\textwidth]{images/comparison/COCOEE/000000061097_GT_dd50.jpg} &
        \includegraphics[width=0.105\textwidth]{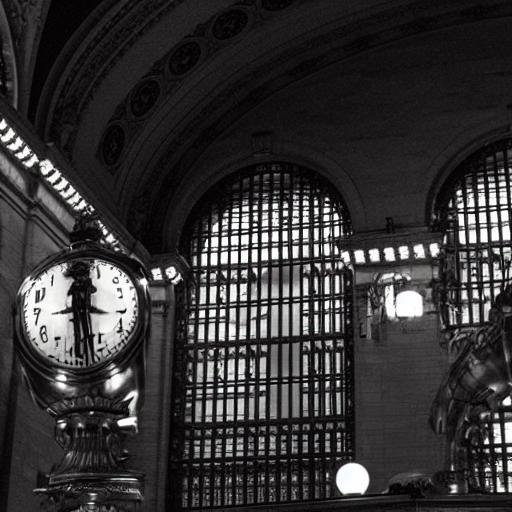} &
        \includegraphics[width=0.105\textwidth]{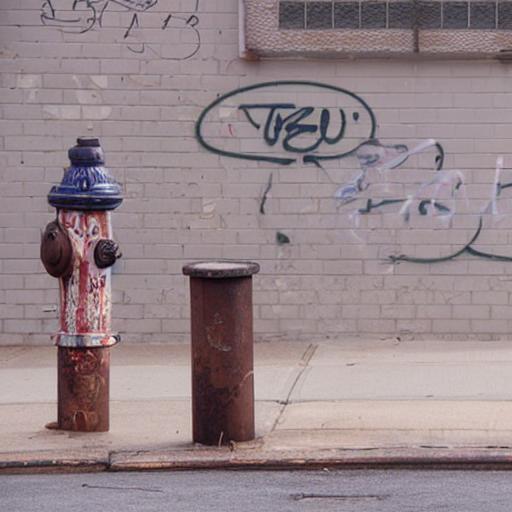} &
        \includegraphics[width=0.105\textwidth]{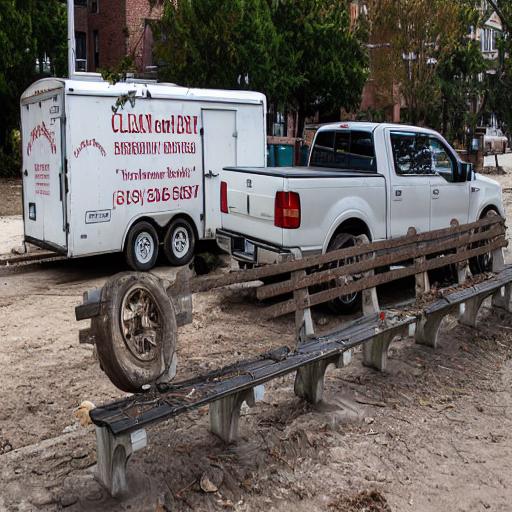} &
        \includegraphics[width=0.105\textwidth]{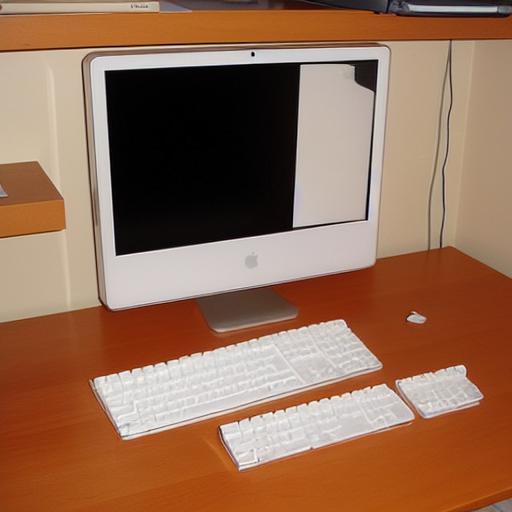} &\\

        {\raisebox{0.37in}{\multirow{1}{*}{\begin{tabular}{c}DragonDiffusion \\ (16 steps, 9s)\end{tabular}}}} &
        \includegraphics[width=0.105\textwidth]{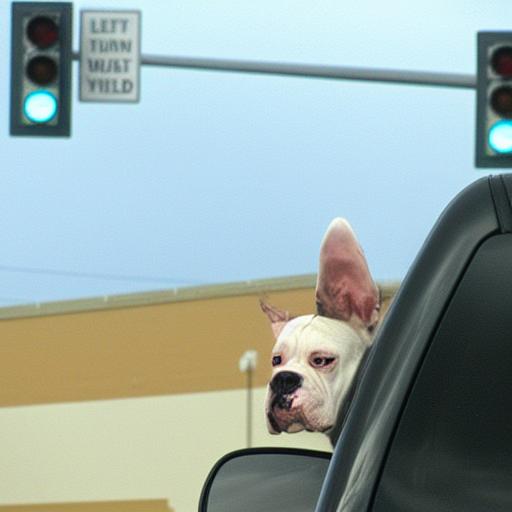} &
        \includegraphics[width=0.105\textwidth]{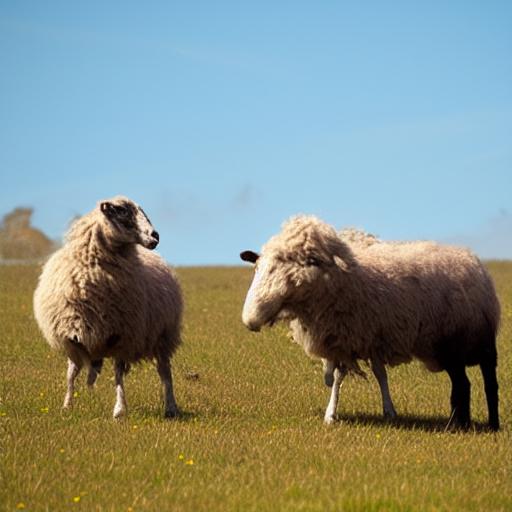} &
        \includegraphics[width=0.105\textwidth]{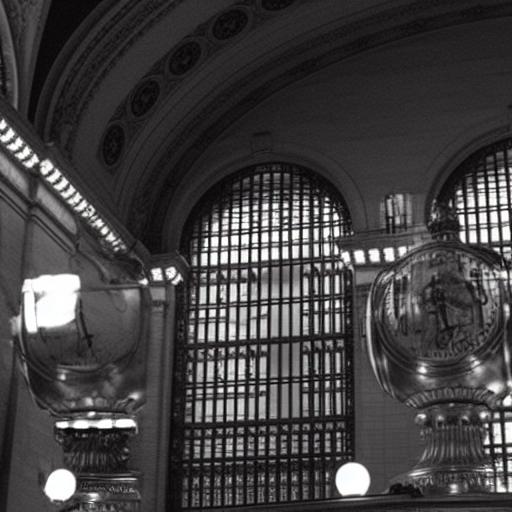} &
        \includegraphics[width=0.105\textwidth]{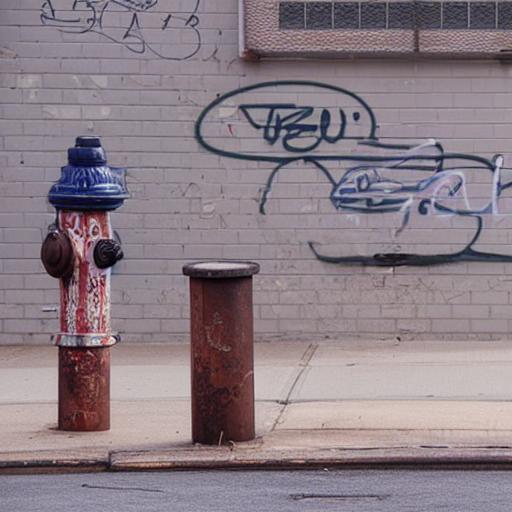} &
        \includegraphics[width=0.105\textwidth]{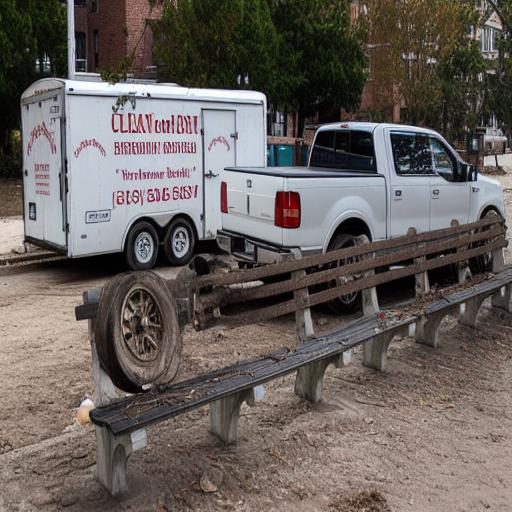} &
        \includegraphics[width=0.105\textwidth]{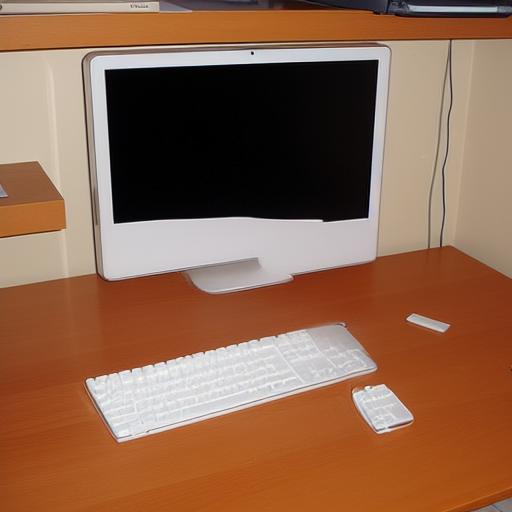} &\\

        {\raisebox{0.37in}{\multirow{1}{*}{\begin{tabular}{c}DiffEditor \\ (50 steps, 24s)\end{tabular}}}} &
        \includegraphics[width=0.105\textwidth]{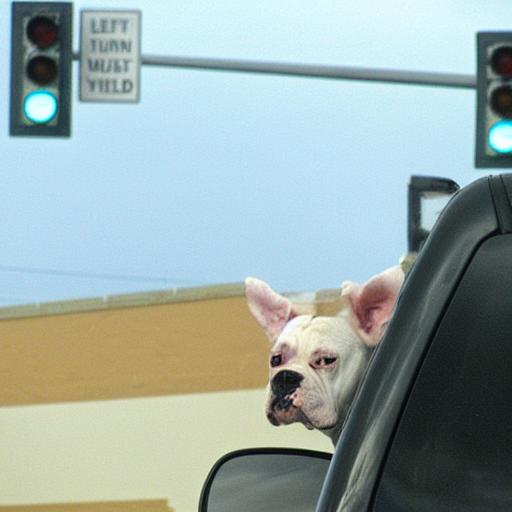} &
        \includegraphics[width=0.105\textwidth]{images/comparison/COCOEE/000000061097_GT_de50.jpg} &
        \includegraphics[width=0.105\textwidth]{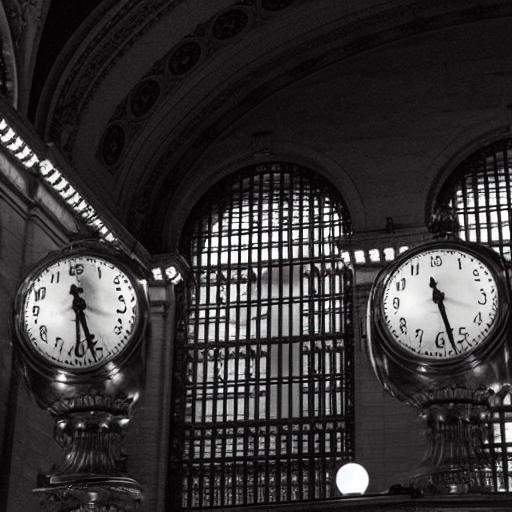} &
        \includegraphics[width=0.105\textwidth]{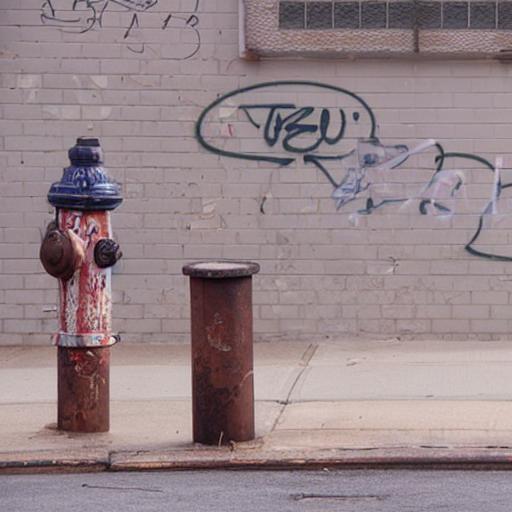} &
        \includegraphics[width=0.105\textwidth]{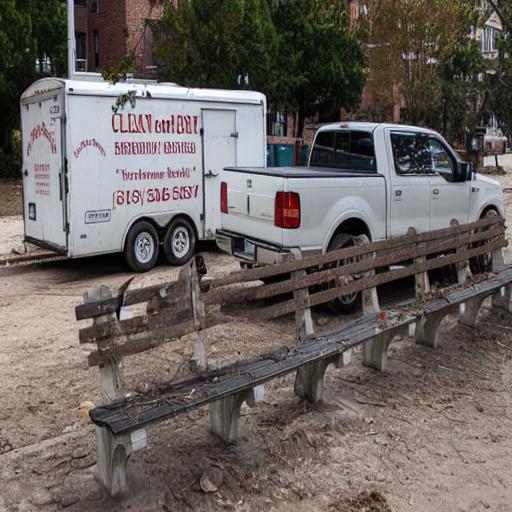} &
        \includegraphics[width=0.105\textwidth]{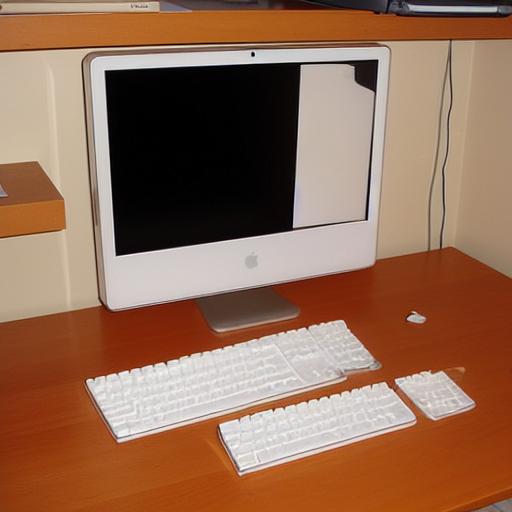} &\\

        {\raisebox{0.37in}{\multirow{1}{*}{\begin{tabular}{c}DiffEditor \\ (16 steps, 9s)\end{tabular}}}} &
        \includegraphics[width=0.105\textwidth]{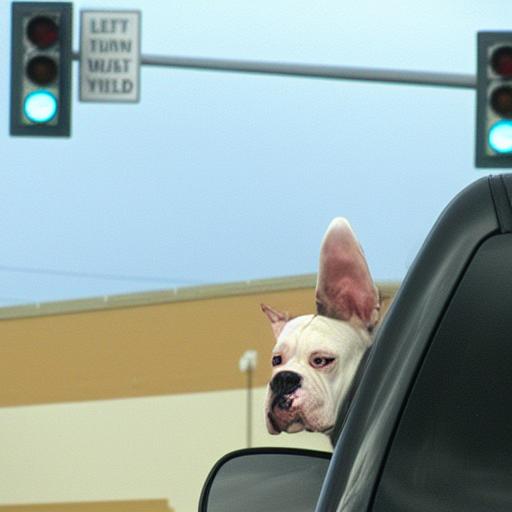} &
        \includegraphics[width=0.105\textwidth]{images/comparison/COCOEE/000000061097_GT_de16.jpg} &
        \includegraphics[width=0.105\textwidth]{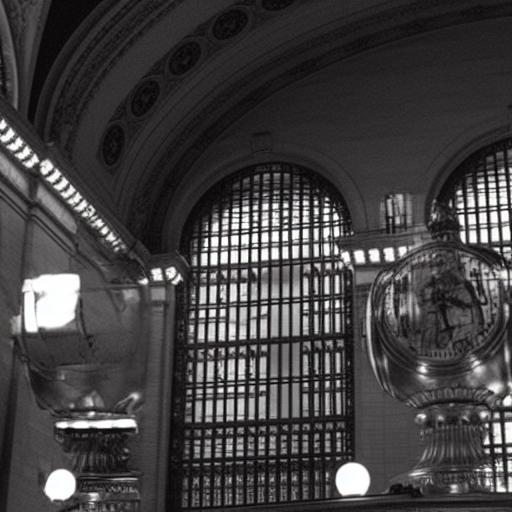} &
        \includegraphics[width=0.105\textwidth]{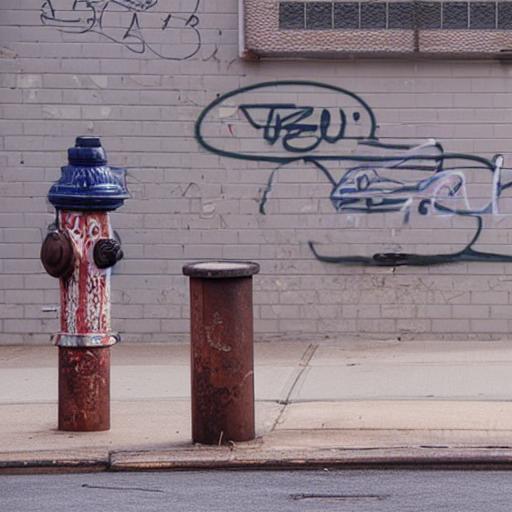} &
        \includegraphics[width=0.105\textwidth]{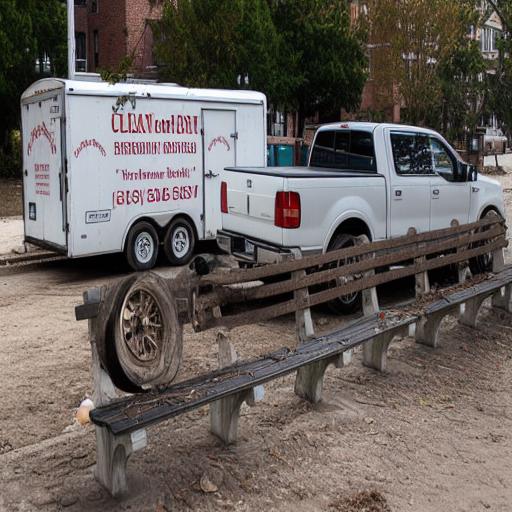} &
        \includegraphics[width=0.105\textwidth]{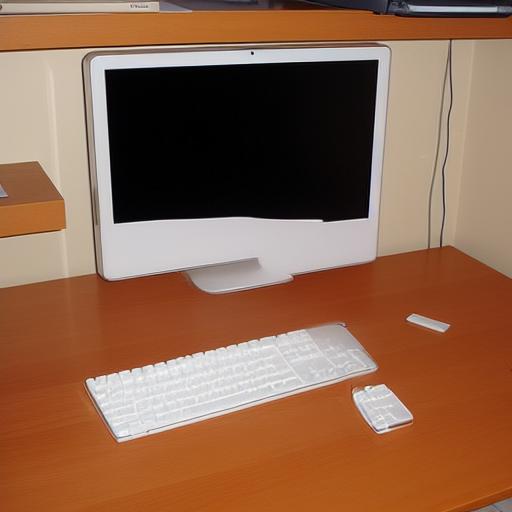} &\\

        {\raisebox{0.37in}{\multirow{1}{*}{\begin{tabular}{c}\textbf{PixelMan} \\ (50 steps, 27s)\end{tabular}}}} &
        \includegraphics[width=0.105\textwidth]{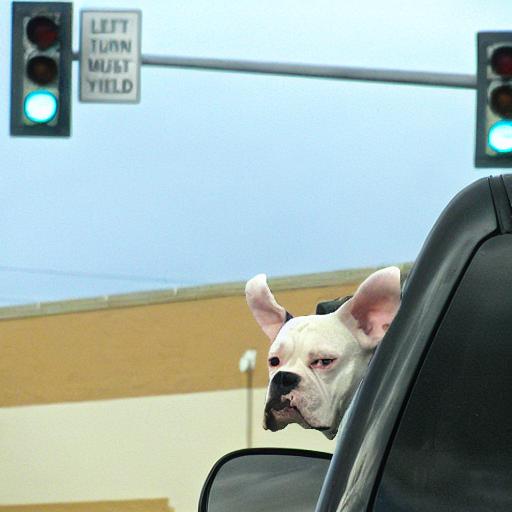} &
        \includegraphics[width=0.105\textwidth]{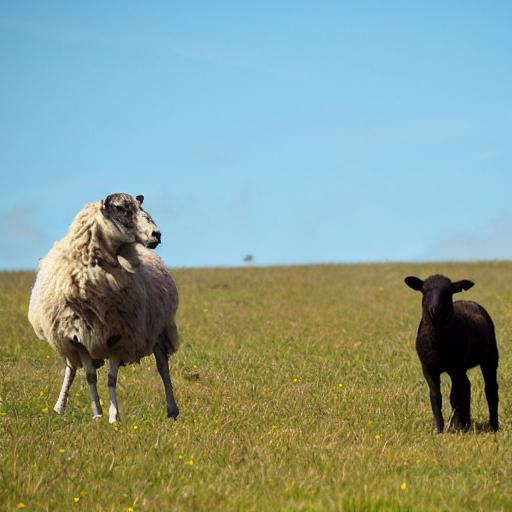} &
        \includegraphics[width=0.105\textwidth]{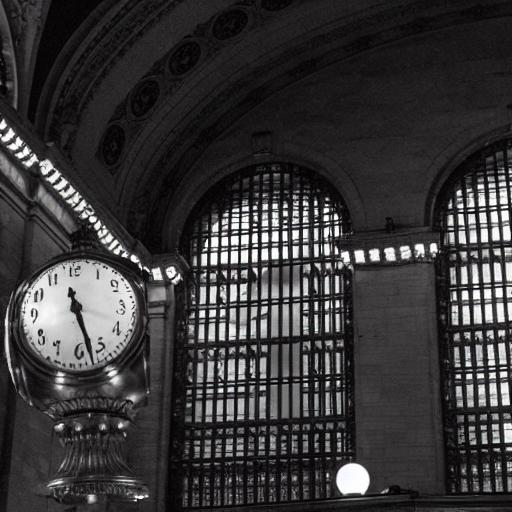} &
        \includegraphics[width=0.105\textwidth]{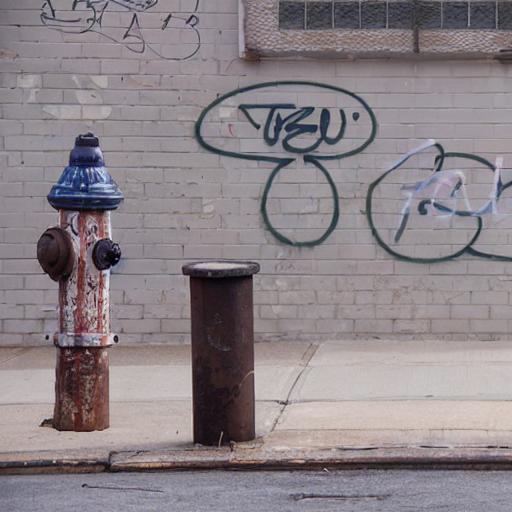} &
        \includegraphics[width=0.105\textwidth]{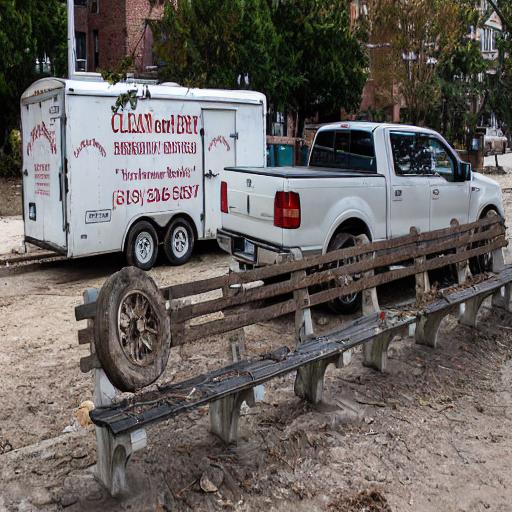} &
        \includegraphics[width=0.105\textwidth]{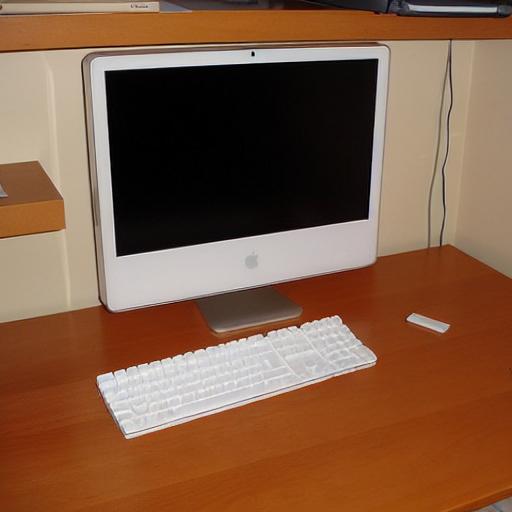} &\\
        
        {\raisebox{0.37in}{\multirow{1}{*}{\begin{tabular}{c}\textbf{PixelMan} \\ (16 steps, 9s)\end{tabular}}}} &
        \includegraphics[width=0.105\textwidth]{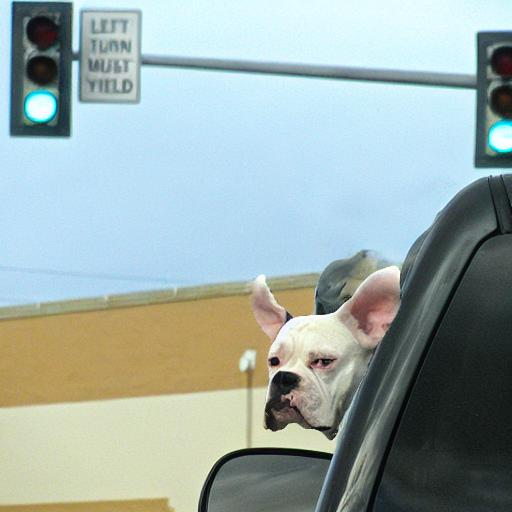} &
        \includegraphics[width=0.105\textwidth]{images/comparison/COCOEE/000000061097_GT_ours16.jpg} &
        \includegraphics[width=0.105\textwidth]{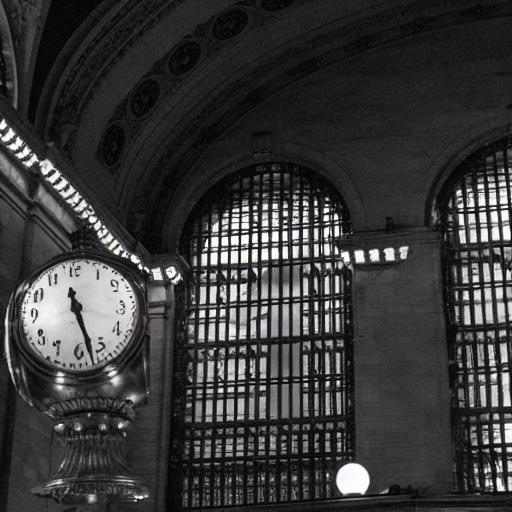} &
        \includegraphics[width=0.105\textwidth]{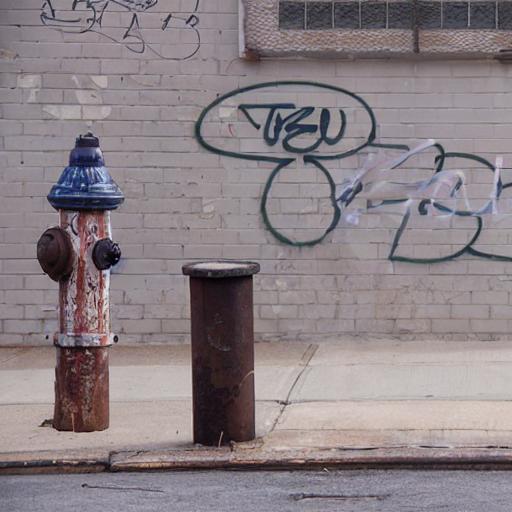} &
        \includegraphics[width=0.105\textwidth]{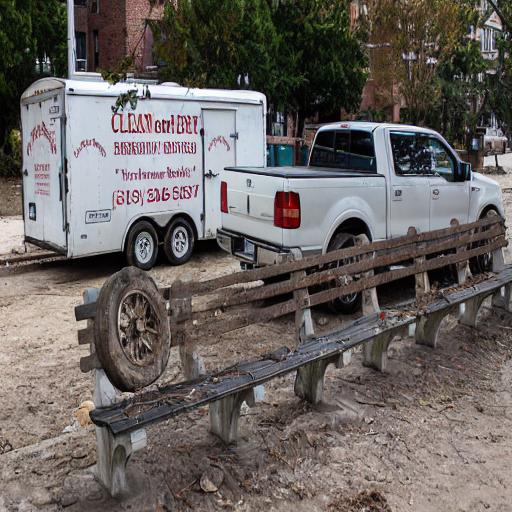} &
        \includegraphics[width=0.105\textwidth]{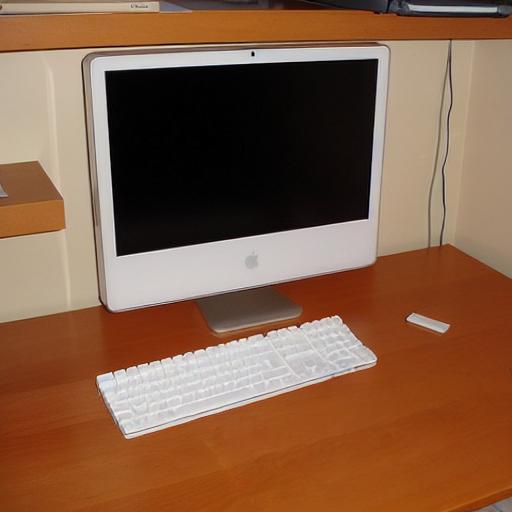} &\\

    \end{tabular}
    }
    \caption{
        \textbf{Additional qualitative comparison} on the COCOEE dataset at both 16 and 50 steps. 
    }
    \label{fig:examples_full_1}
\end{figure*}
\begin{figure*}[hbt!]
    \centering
    \setlength{\tabcolsep}{0.4pt}
    \renewcommand{\arraystretch}{0.4}
    {\footnotesize
    \begin{tabular}{c c c c c c c c}
        &
        \multicolumn{1}{c}{(a)} &
        \multicolumn{1}{c}{(b)} &
        \multicolumn{1}{c}{(c)} &
        \multicolumn{1}{c}{(d)} &
        \multicolumn{1}{c}{(e)} &
        \multicolumn{1}{c}{(f)} \\

        {\raisebox{0.34in}{
        \multirow{1}{*}{\rotatebox{0}{Input}}}} &
        \includegraphics[width=0.105\textwidth]{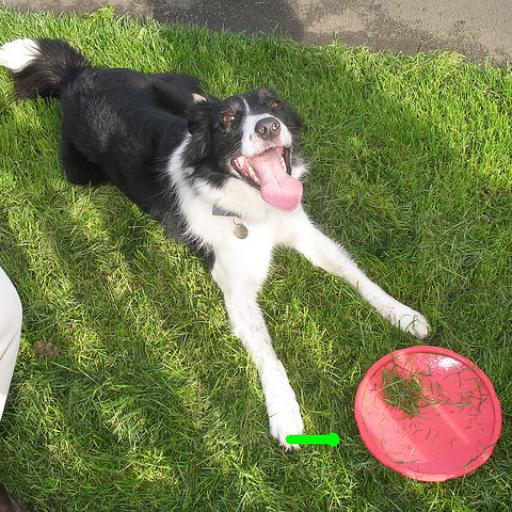} &
        \includegraphics[width=0.105\textwidth]{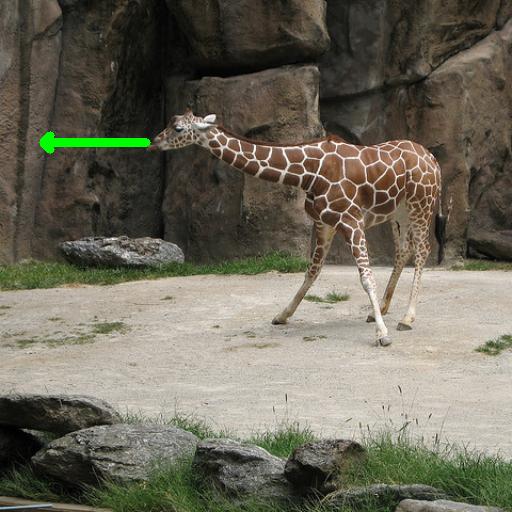} &
        \includegraphics[width=0.105\textwidth]{images/comparison/COCOEE/000001816267_GT_source.jpg} &
        \includegraphics[width=0.105\textwidth]{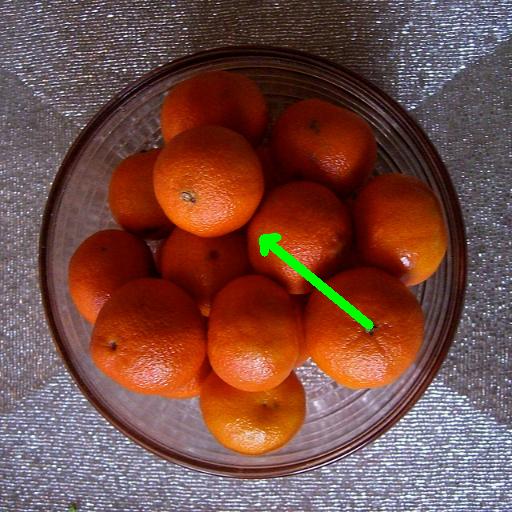} &
        \includegraphics[width=0.105\textwidth]{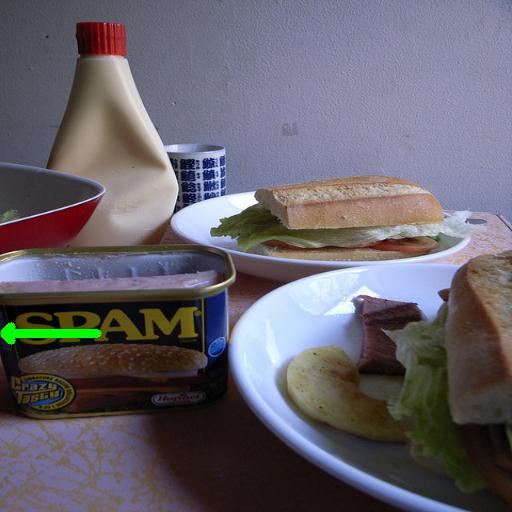} &
        \includegraphics[width=0.105\textwidth]{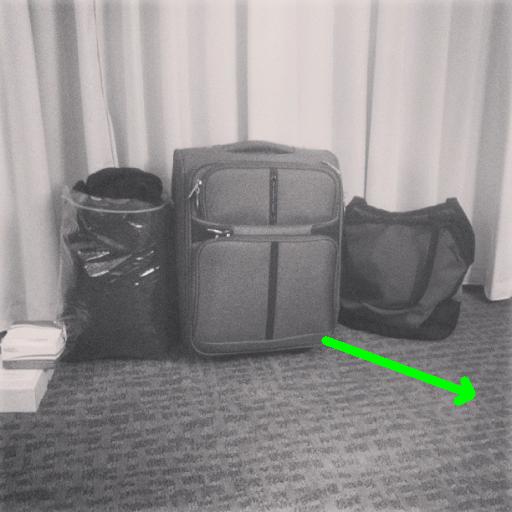} &\\

        {\raisebox{0.47in}{\multirow{1}{*}{\begin{tabular}{c}SDv2-Inpainting\\+AnyDoor \\ (50 steps, 15s)\end{tabular}}}}
        &
        \includegraphics[width=0.105\textwidth]{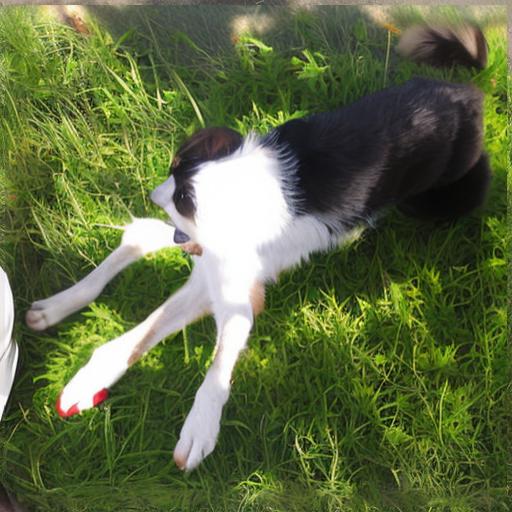} &
        \includegraphics[width=0.105\textwidth]{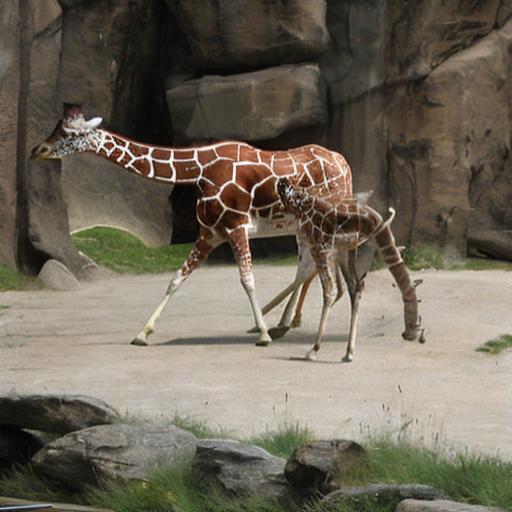} &
        \includegraphics[width=0.105\textwidth]{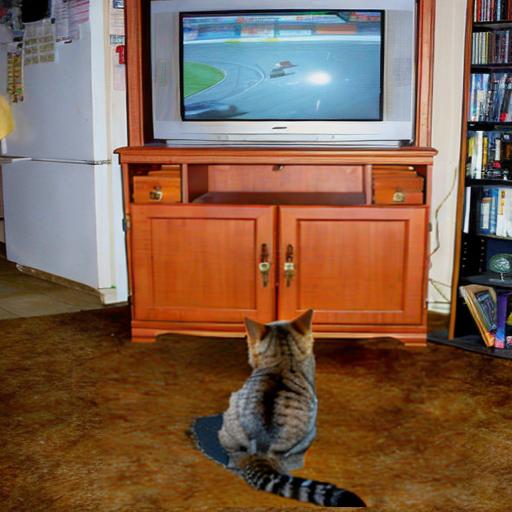} &
        \includegraphics[width=0.105\textwidth]{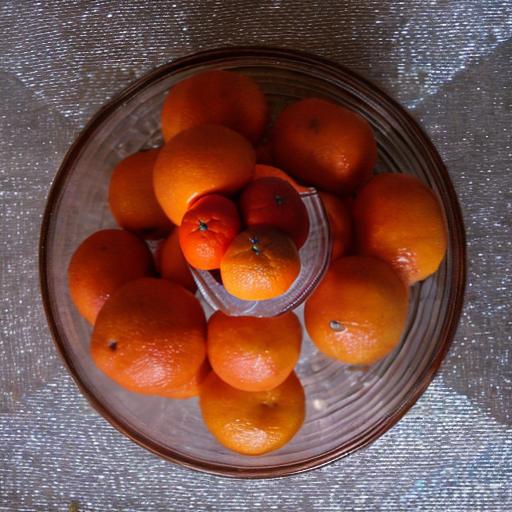} &
        \includegraphics[width=0.105\textwidth]{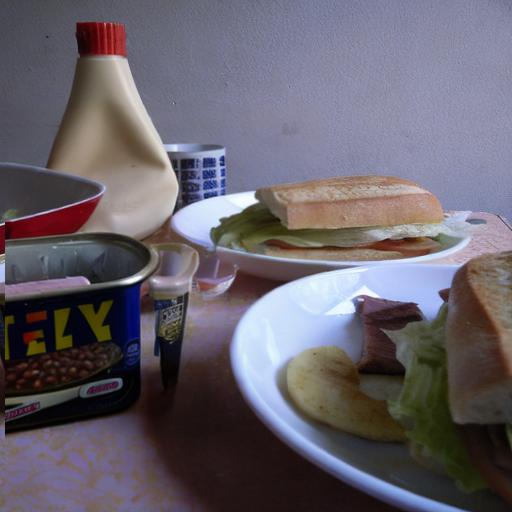} &
        \includegraphics[width=0.105\textwidth]{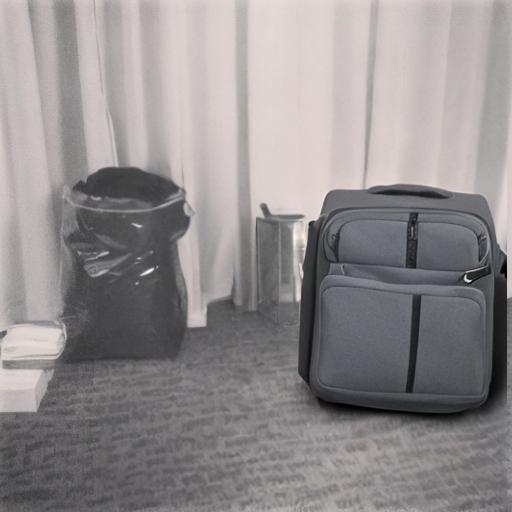} &\\

        {\raisebox{0.47in}{\multirow{1}{*}{\begin{tabular}{c}SDv2-Inpainting\\+AnyDoor \\ (16 steps, 5s)\end{tabular}}}}
        &
        \includegraphics[width=0.105\textwidth]{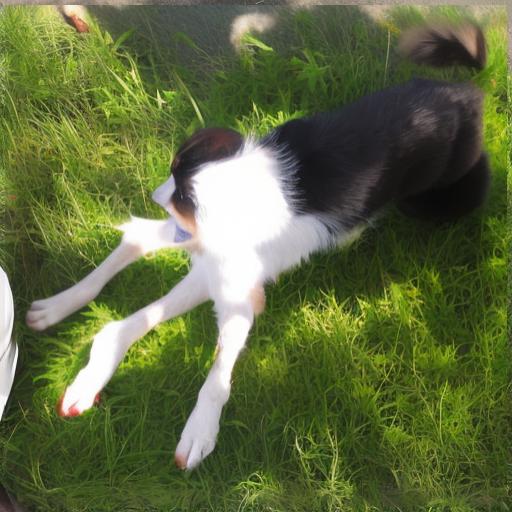} &
        \includegraphics[width=0.105\textwidth]{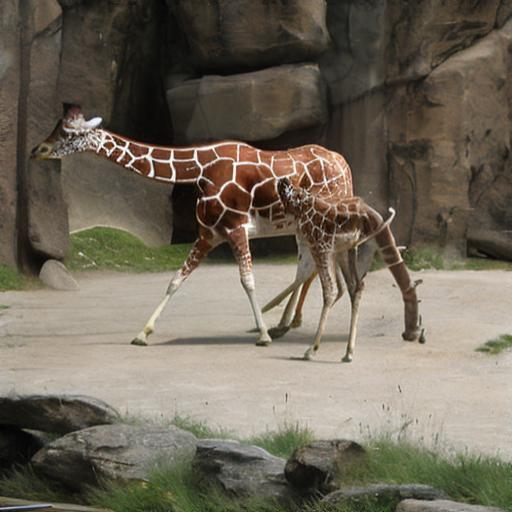} &
        \includegraphics[width=0.105\textwidth]{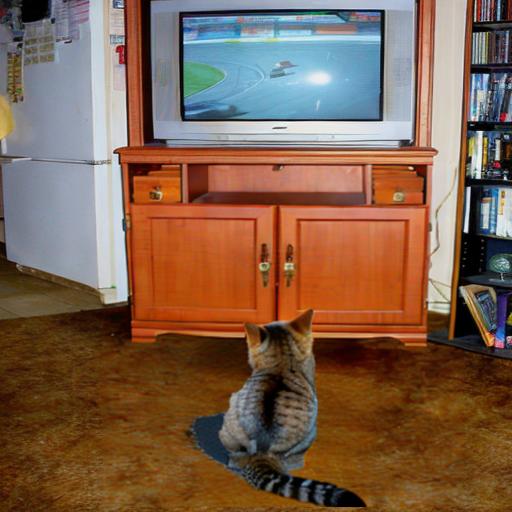} &
        \includegraphics[width=0.105\textwidth]{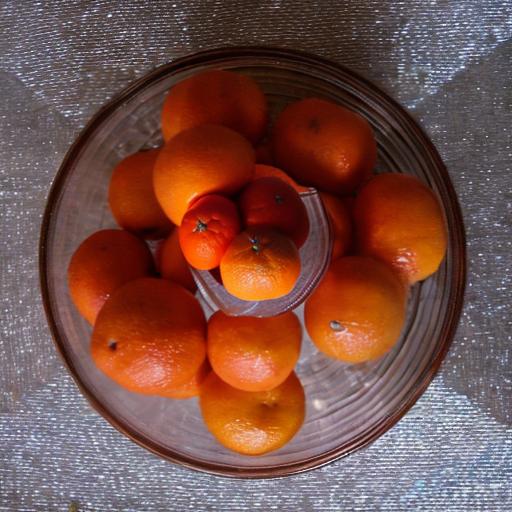} &
        \includegraphics[width=0.105\textwidth]{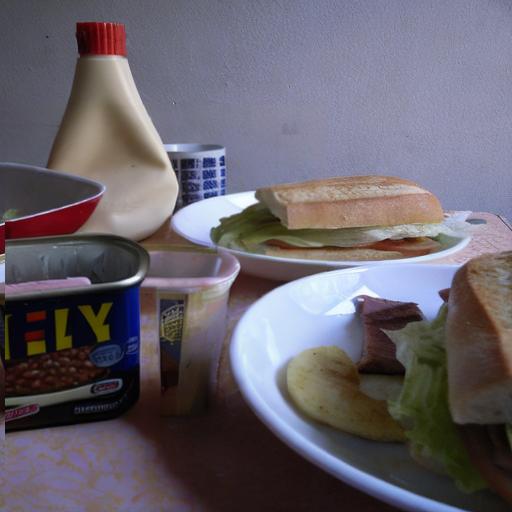} &
        \includegraphics[width=0.105\textwidth]{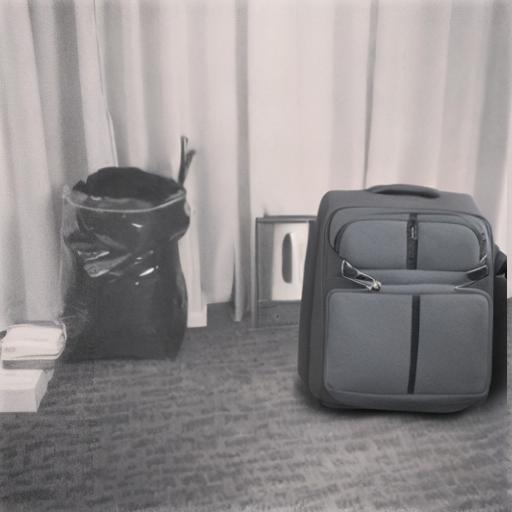} &\\
        
        {\raisebox{0.37in}{\multirow{1}{*}{\begin{tabular}{c}SelfGuidance \\ (50 steps, 11s)\end{tabular}}}} &
        \includegraphics[width=0.105\textwidth]{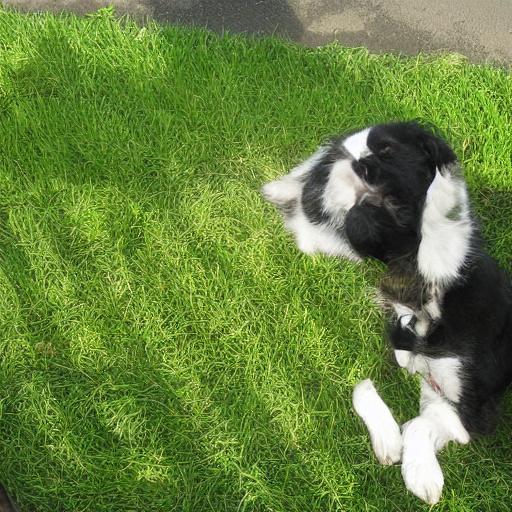} &
        \includegraphics[width=0.105\textwidth]{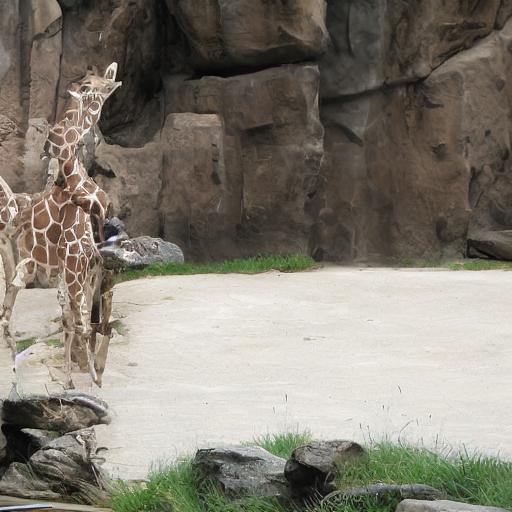} &
        \includegraphics[width=0.105\textwidth]{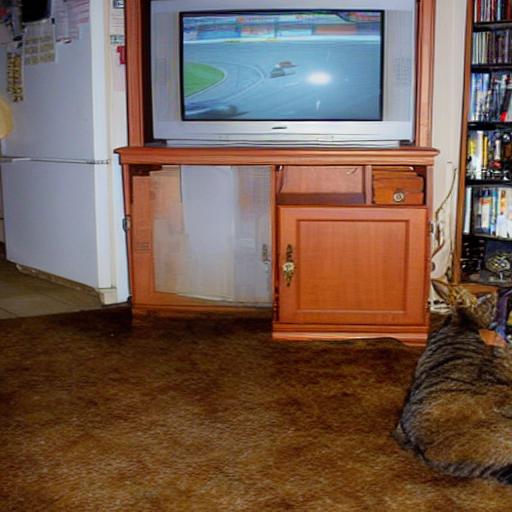} &
        \includegraphics[width=0.105\textwidth]{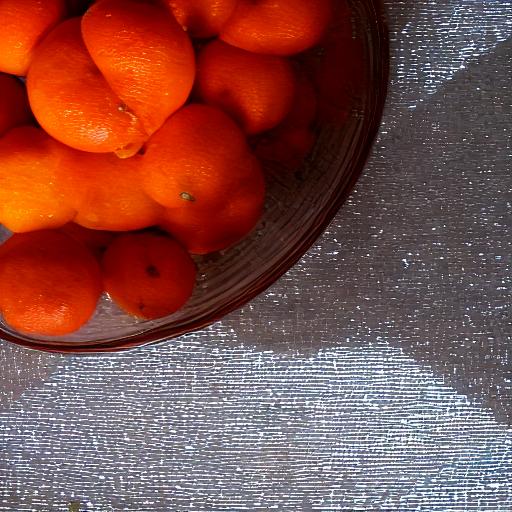} &
        \includegraphics[width=0.105\textwidth]{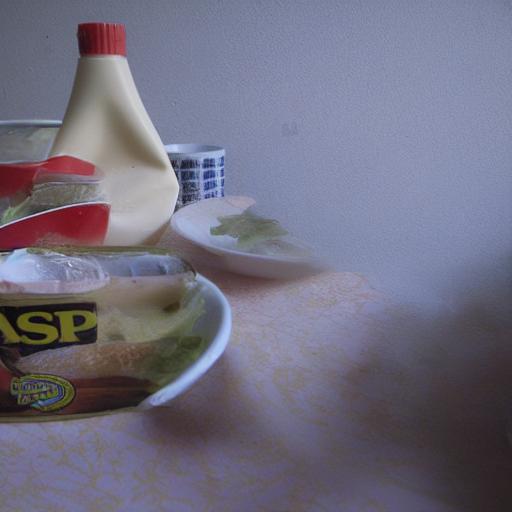} &
        \includegraphics[width=0.105\textwidth]{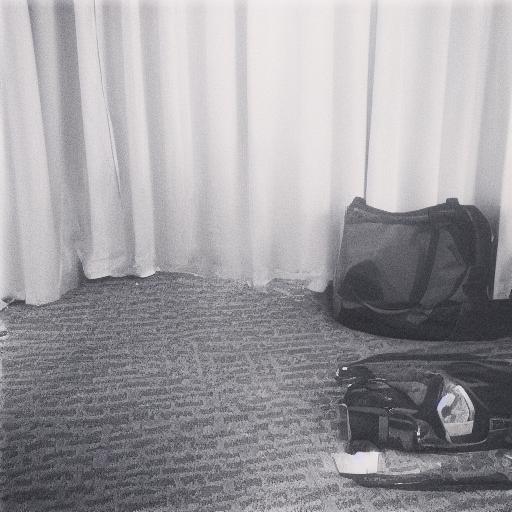} &\\
        
        {\raisebox{0.37in}{\multirow{1}{*}{\begin{tabular}{c}SelfGuidance \\ (16 steps, 4s)\end{tabular}}}} &
        \includegraphics[width=0.105\textwidth]{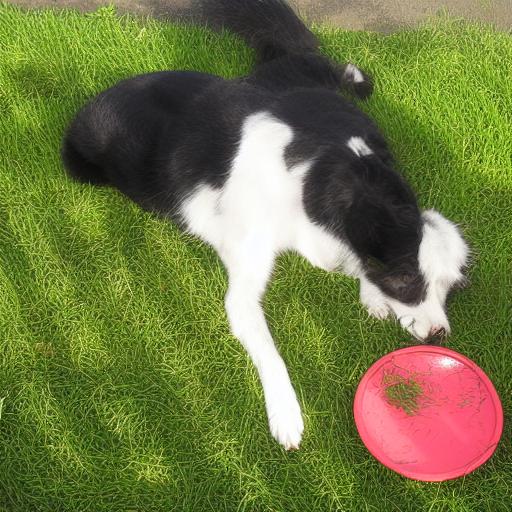} &
        \includegraphics[width=0.105\textwidth]{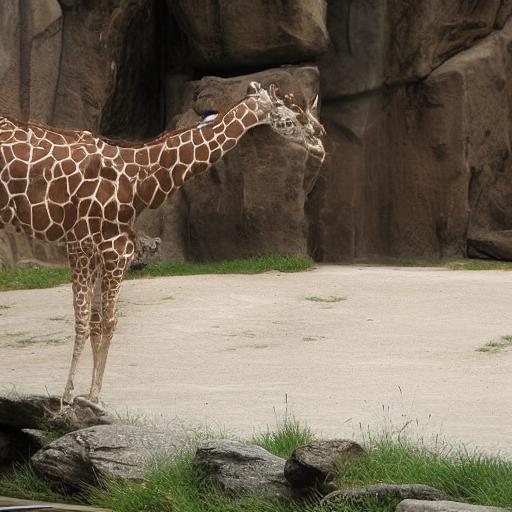} &
        \includegraphics[width=0.105\textwidth]{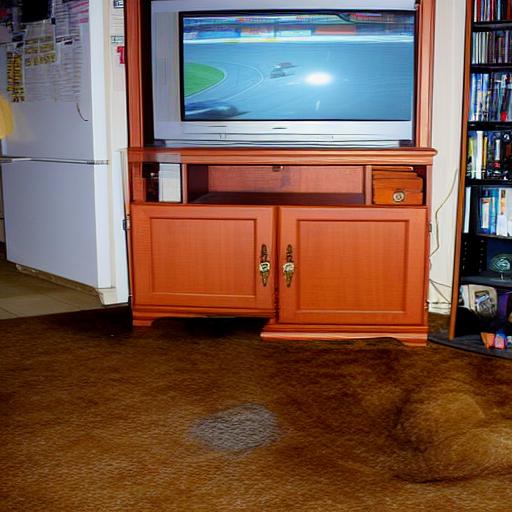} &
        \includegraphics[width=0.105\textwidth]{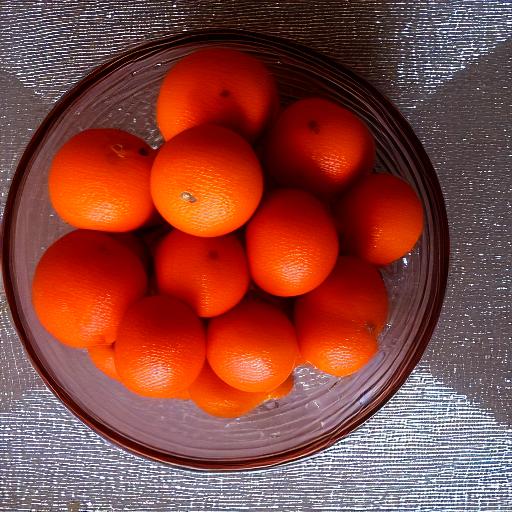} &
        \includegraphics[width=0.105\textwidth]{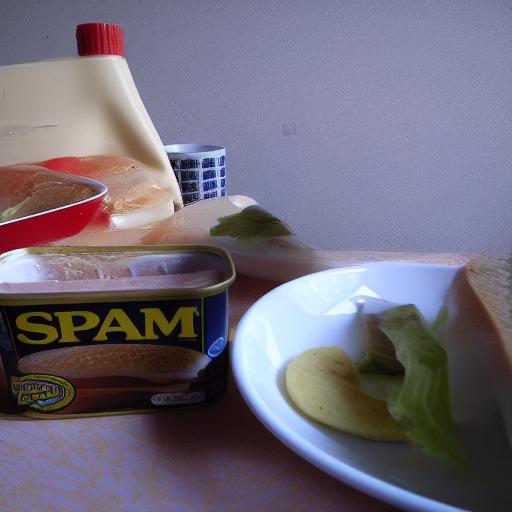} &
        \includegraphics[width=0.105\textwidth]{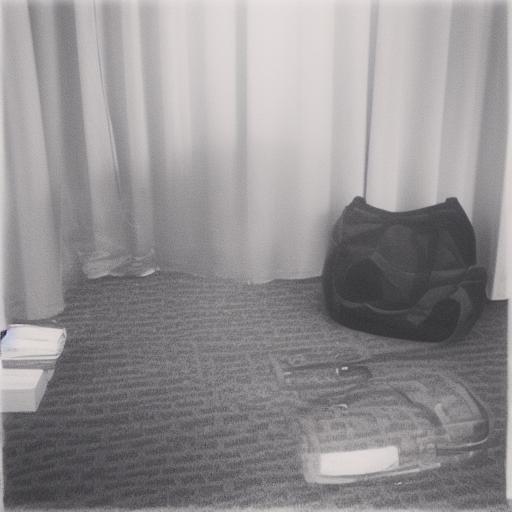} &\\
        
        {\raisebox{0.37in}{\multirow{1}{*}{\begin{tabular}{c}DragonDiffusion \\ (50 steps, 23s)\end{tabular}}}} &
        \includegraphics[width=0.105\textwidth]{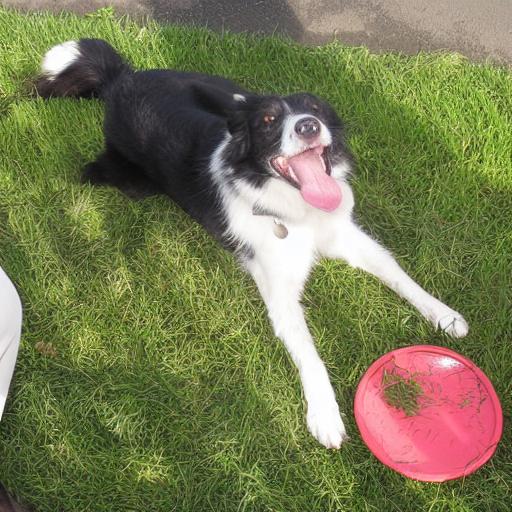} &
        \includegraphics[width=0.105\textwidth]{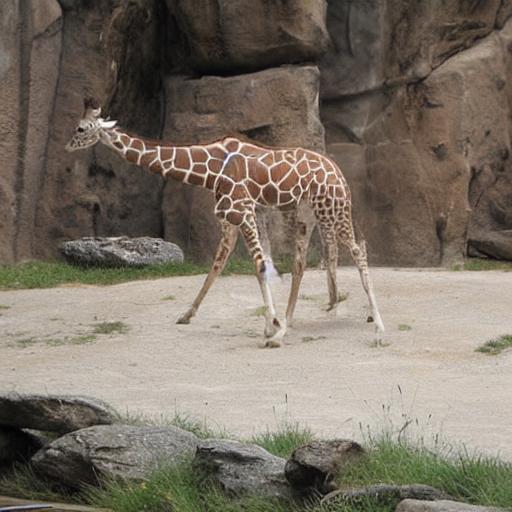} &
        \includegraphics[width=0.105\textwidth]{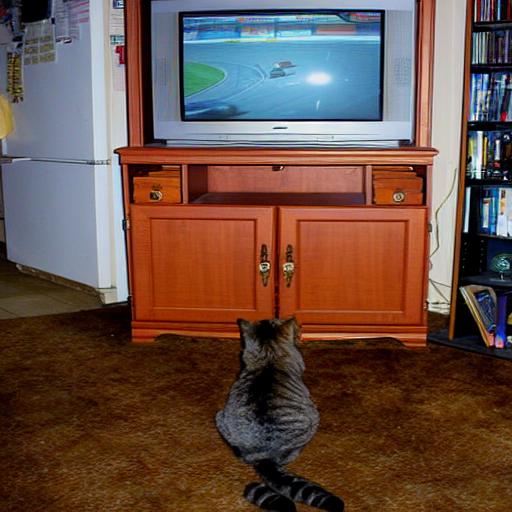} &
        \includegraphics[width=0.105\textwidth]{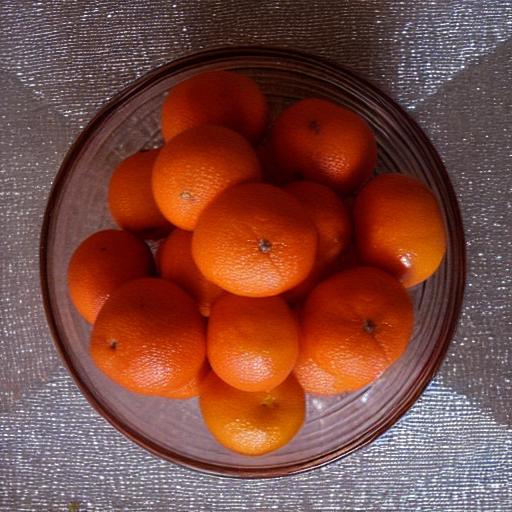} &
        \includegraphics[width=0.105\textwidth]{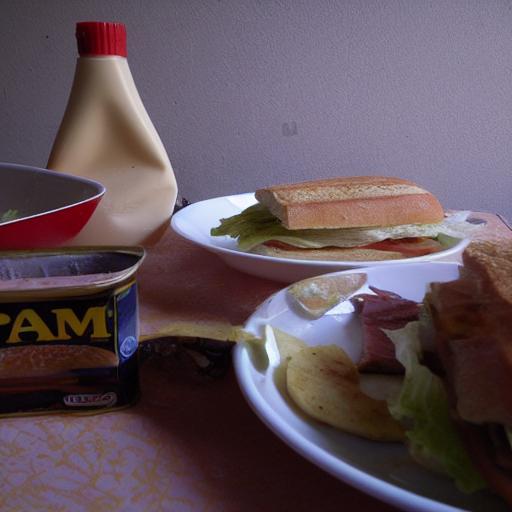} &
        \includegraphics[width=0.105\textwidth]{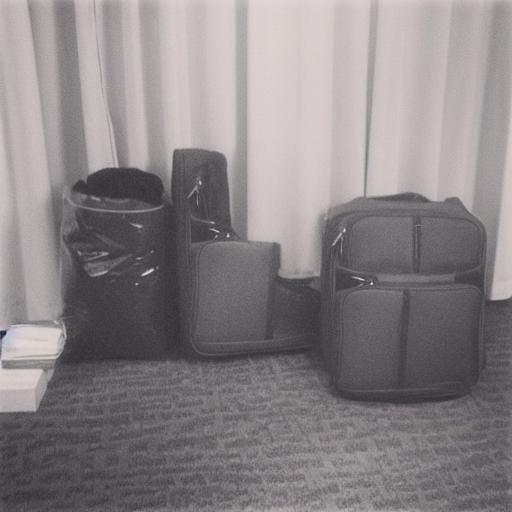} &\\

        {\raisebox{0.37in}{\multirow{1}{*}{\begin{tabular}{c}DragonDiffusion \\ (16 steps, 9s)\end{tabular}}}} &
        \includegraphics[width=0.105\textwidth]{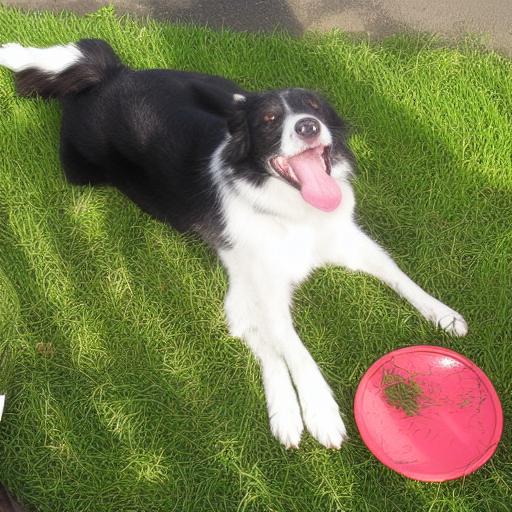} &
        \includegraphics[width=0.105\textwidth]{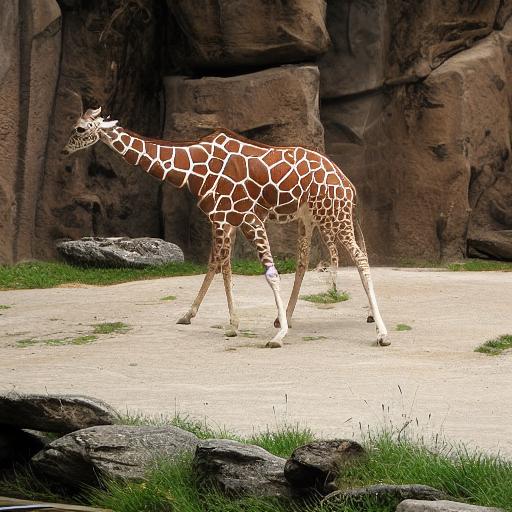} &
        \includegraphics[width=0.105\textwidth]{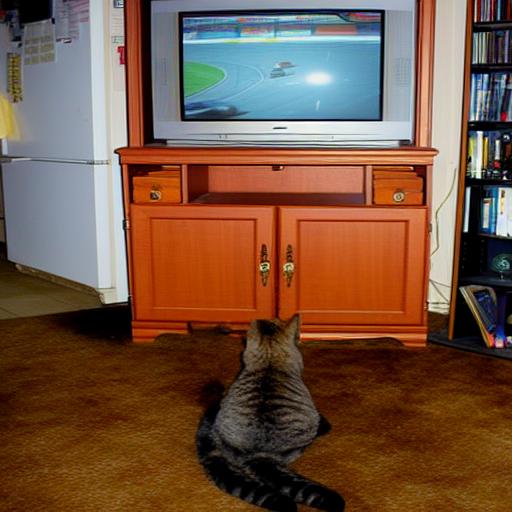} &
        \includegraphics[width=0.105\textwidth]{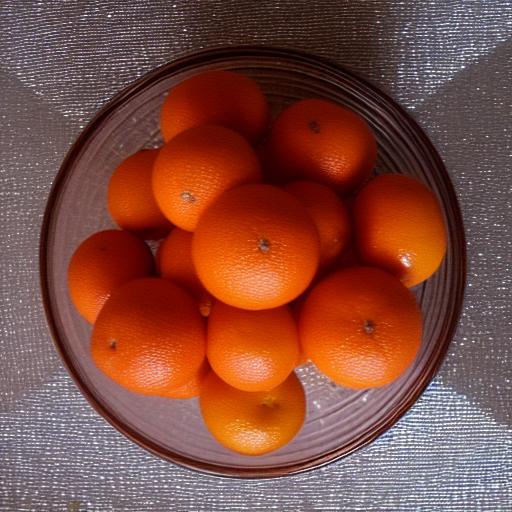} &
        \includegraphics[width=0.105\textwidth]{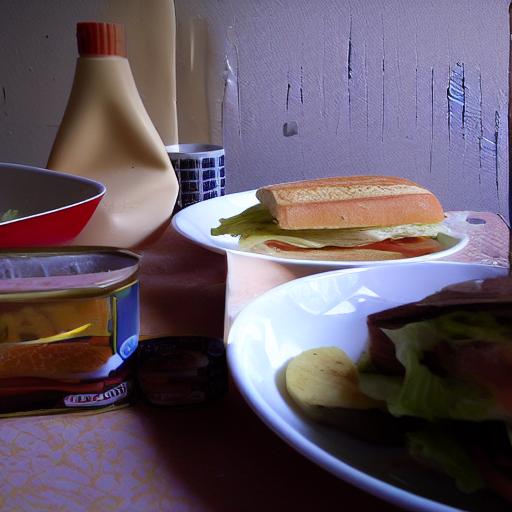} &
        \includegraphics[width=0.105\textwidth]{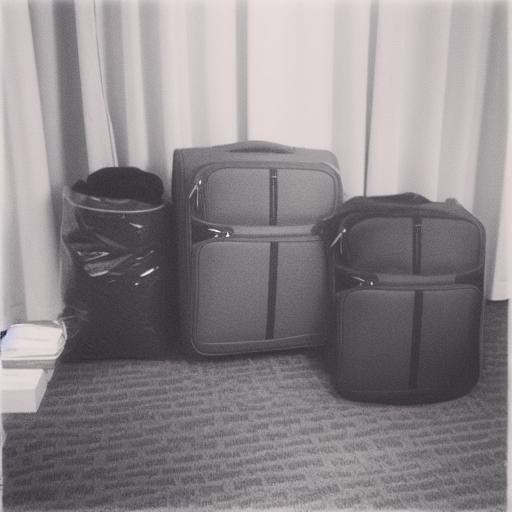} &\\

        {\raisebox{0.37in}{\multirow{1}{*}{\begin{tabular}{c}DiffEditor \\ (50 steps, 24s)\end{tabular}}}} &
        \includegraphics[width=0.105\textwidth]{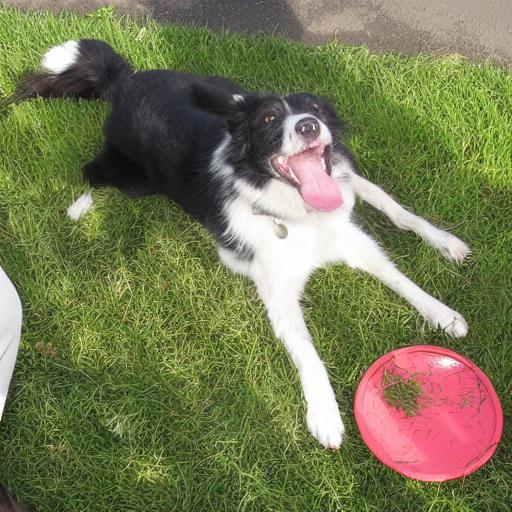} &
        \includegraphics[width=0.105\textwidth]{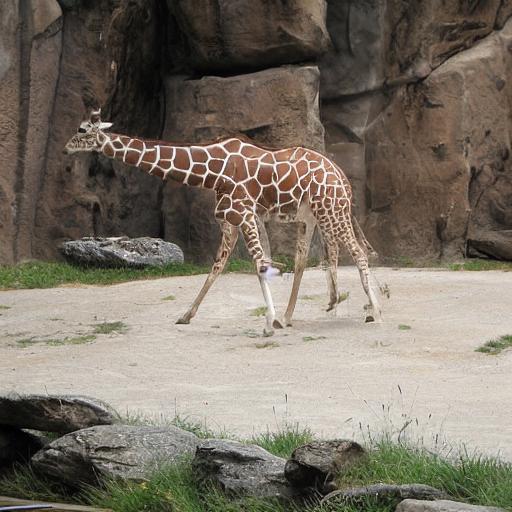} &
        \includegraphics[width=0.105\textwidth]{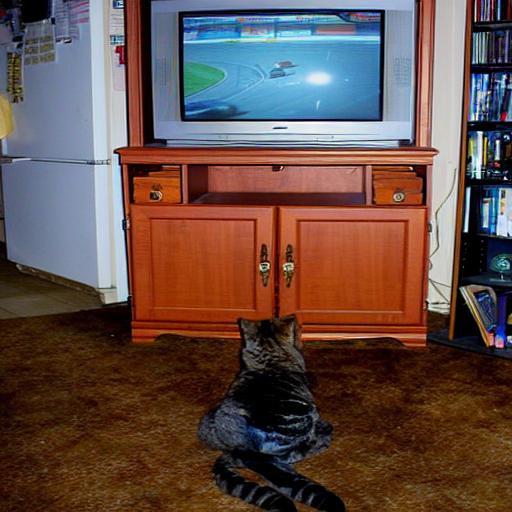} &
        \includegraphics[width=0.105\textwidth]{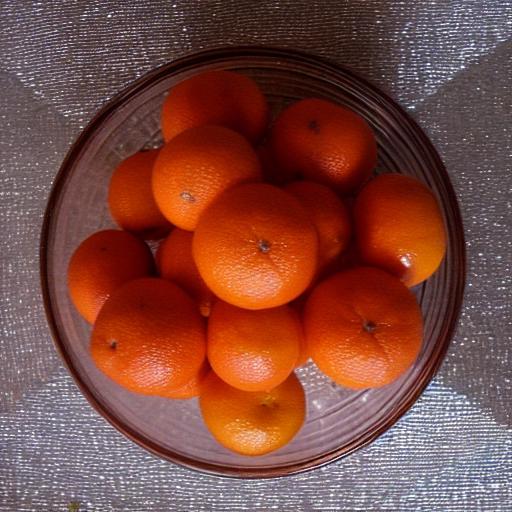} &
        \includegraphics[width=0.105\textwidth]{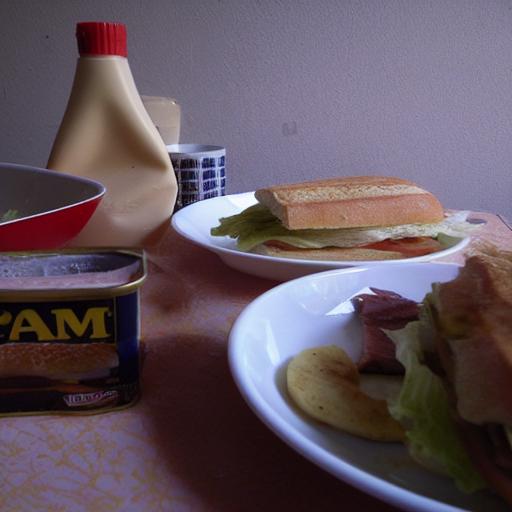} &
        \includegraphics[width=0.105\textwidth]{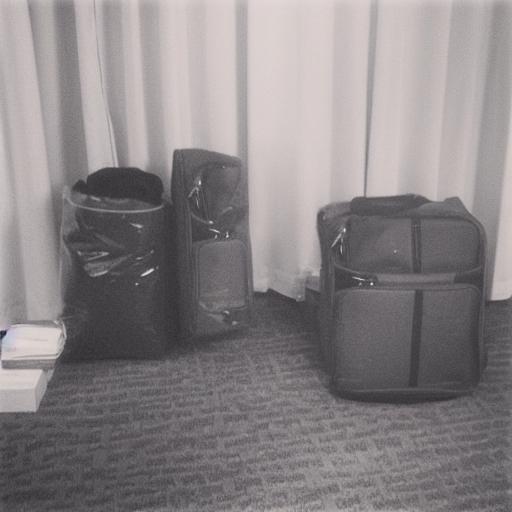} &\\

        {\raisebox{0.37in}{\multirow{1}{*}{\begin{tabular}{c}DiffEditor \\ (16 steps, 9s)\end{tabular}}}} &
        \includegraphics[width=0.105\textwidth]{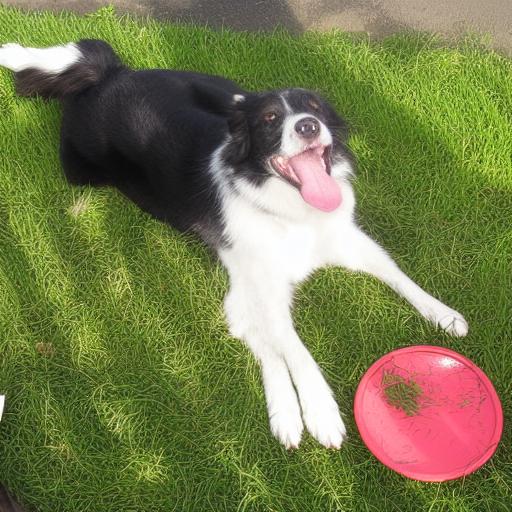} &
        \includegraphics[width=0.105\textwidth]{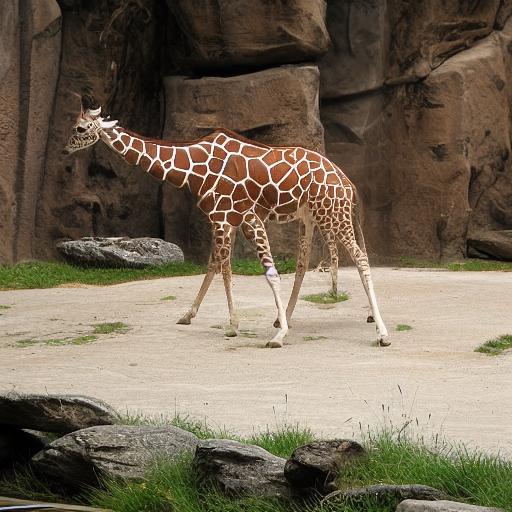} &
        \includegraphics[width=0.105\textwidth]{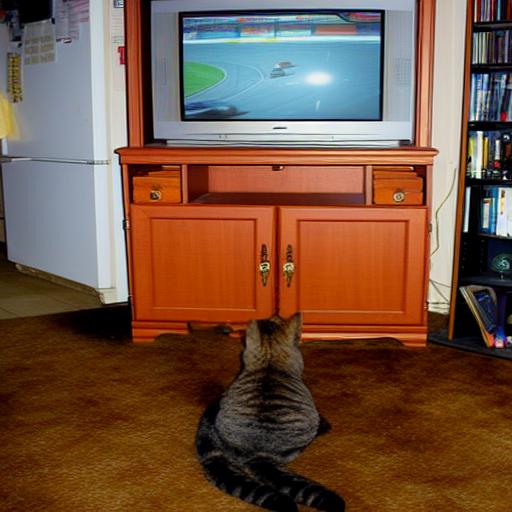} &
        \includegraphics[width=0.105\textwidth]{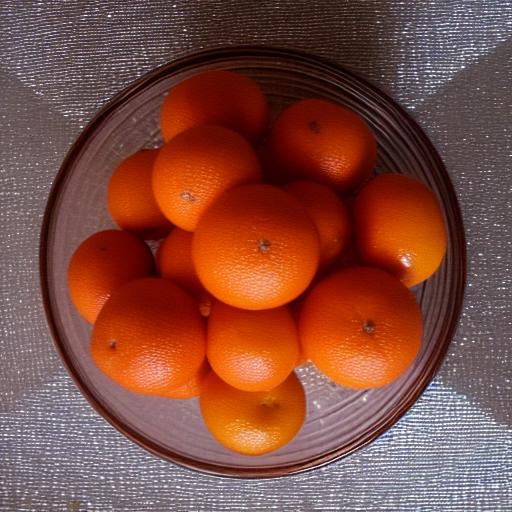} &
        \includegraphics[width=0.105\textwidth]{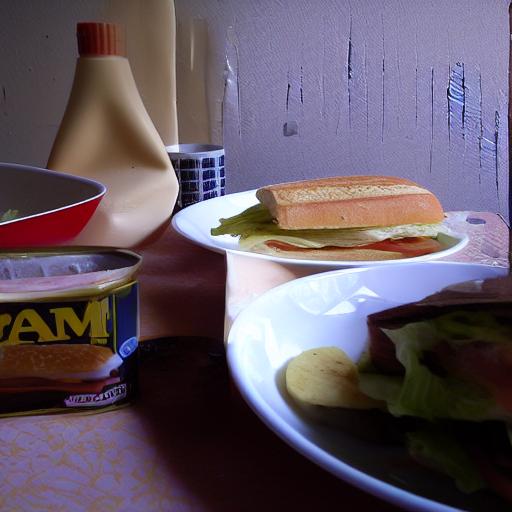} &
        \includegraphics[width=0.105\textwidth]{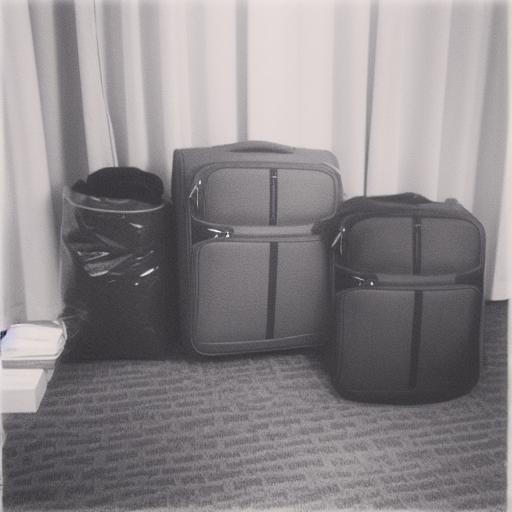} &\\

        {\raisebox{0.37in}{\multirow{1}{*}{\begin{tabular}{c}\textbf{PixelMan} \\ (50 steps, 27s)\end{tabular}}}} &
        \includegraphics[width=0.105\textwidth]{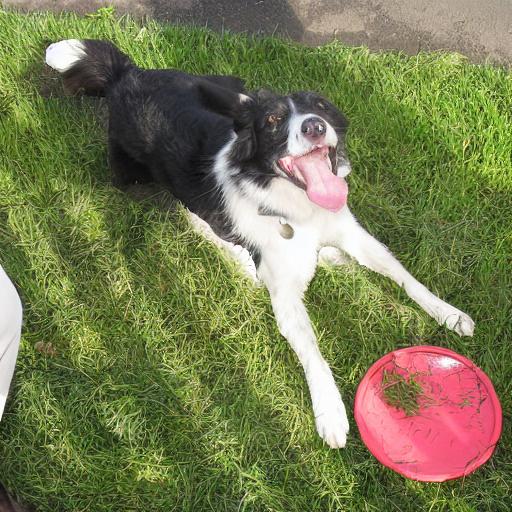} &
        \includegraphics[width=0.105\textwidth]{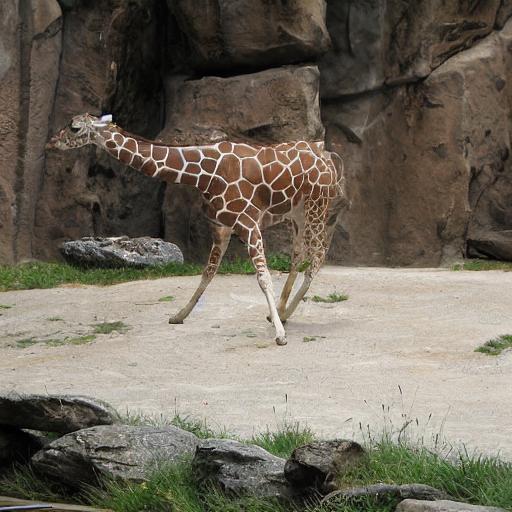} &
        \includegraphics[width=0.105\textwidth]{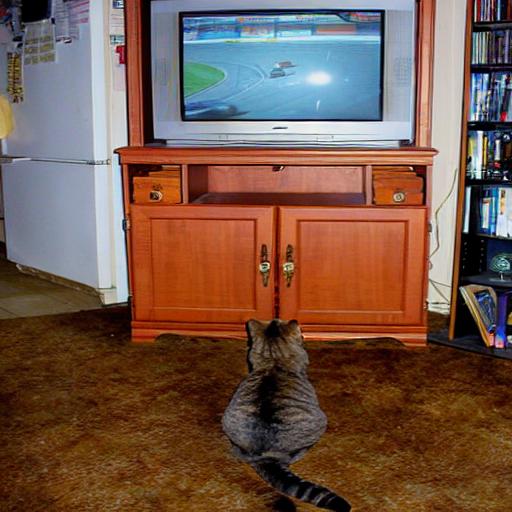} &
        \includegraphics[width=0.105\textwidth]{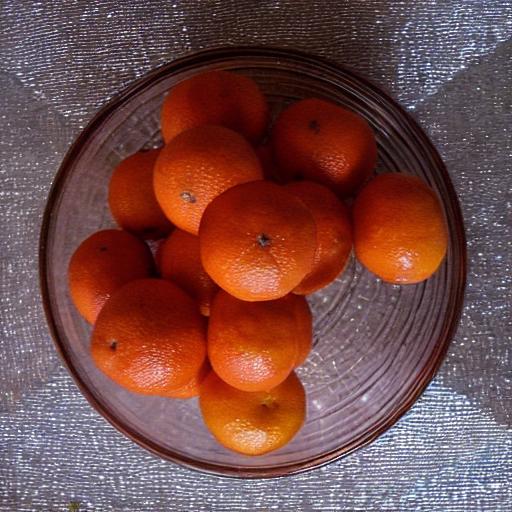} &
        \includegraphics[width=0.105\textwidth]{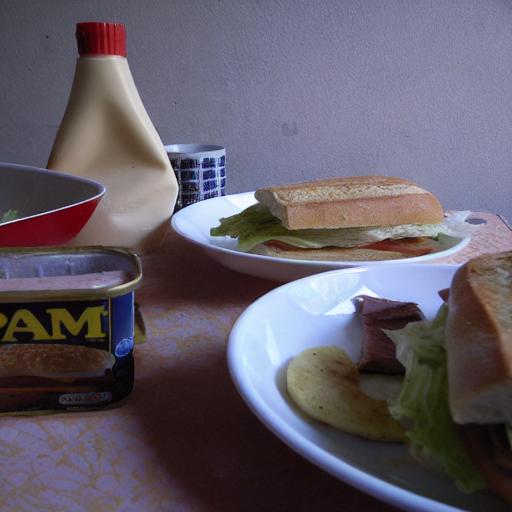} &
        \includegraphics[width=0.105\textwidth]{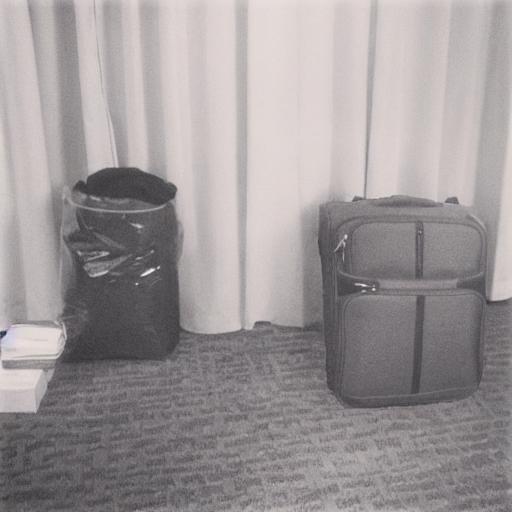} &\\
        
        {\raisebox{0.37in}{\multirow{1}{*}{\begin{tabular}{c}\textbf{PixelMan} \\ (16 steps, 9s)\end{tabular}}}} &
        \includegraphics[width=0.105\textwidth]{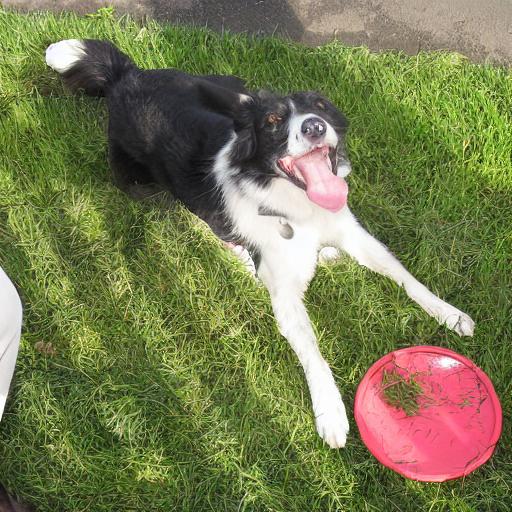} &
        \includegraphics[width=0.105\textwidth]{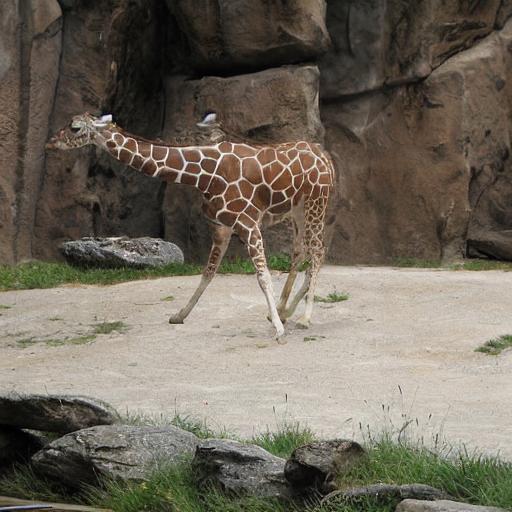} &
        \includegraphics[width=0.105\textwidth]{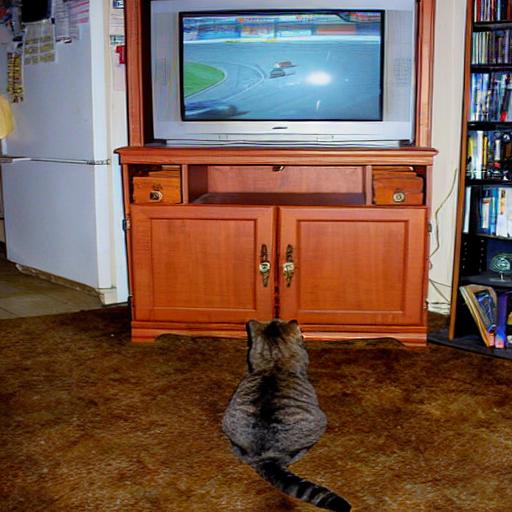} &
        \includegraphics[width=0.105\textwidth]{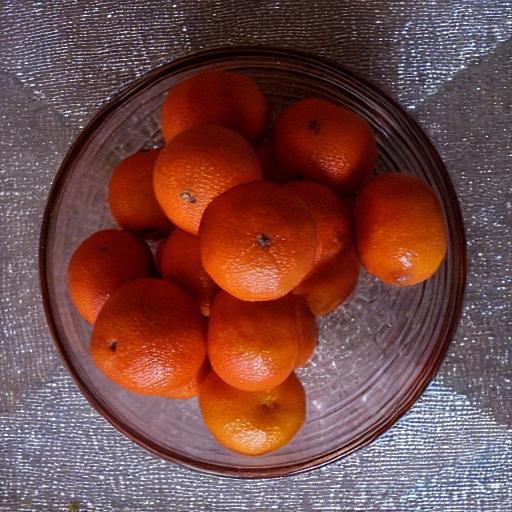} &
        \includegraphics[width=0.105\textwidth]{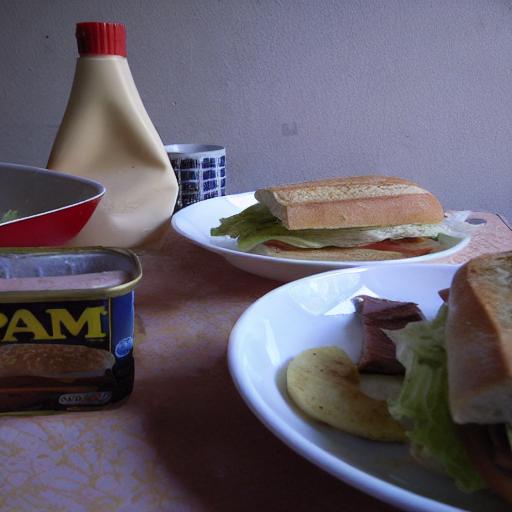} &
        \includegraphics[width=0.105\textwidth]{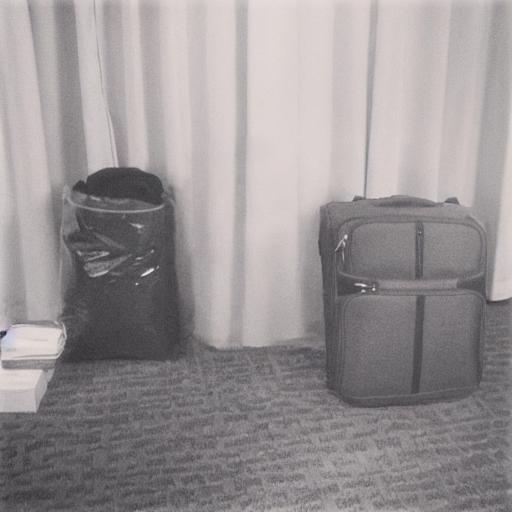} &\\

    \end{tabular}
    }
    \caption{
        \textbf{Additional qualitative comparison} on the COCOEE dataset at both 16 and 50 steps. 
    }
    \label{fig:examples_full_2}
\end{figure*}
\begin{figure*}[hbt!]
    \centering
    \setlength{\tabcolsep}{0.4pt}
    \renewcommand{\arraystretch}{0.4}
    {\footnotesize
    \begin{tabular}{c c c c c c c c}
        &
        \multicolumn{1}{c}{(a)} &
        \multicolumn{1}{c}{(b)} &
        \multicolumn{1}{c}{(c)} &
        \multicolumn{1}{c}{(d)} &
        \multicolumn{1}{c}{(e)} &
        \multicolumn{1}{c}{(f)} \\

        {\raisebox{0.34in}{
        \multirow{1}{*}{\rotatebox{0}{Input}}}} &
        \includegraphics[width=0.138\textwidth]{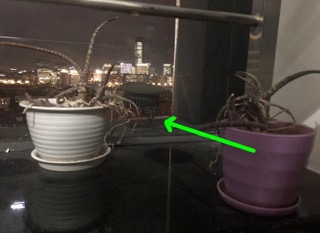} &
        \includegraphics[width=0.138\textwidth]{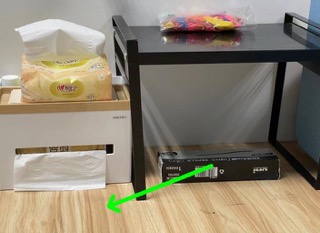} &
        \includegraphics[width=0.138\textwidth]{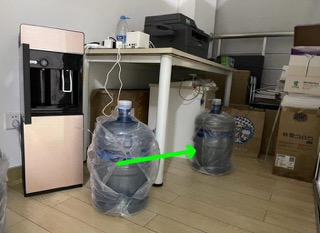} &
        \includegraphics[width=0.138\textwidth]{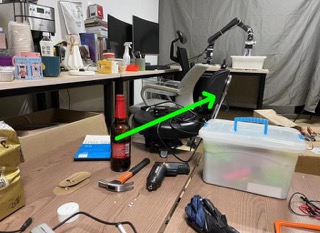} &
        \includegraphics[width=0.138\textwidth]{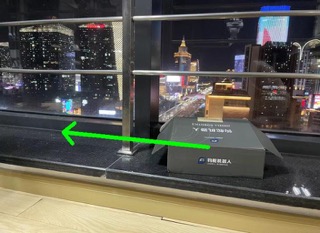} &
        \includegraphics[width=0.138\textwidth]{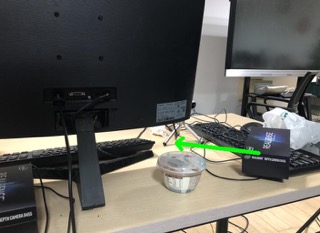} &\\

        {\raisebox{0.47in}{\multirow{1}{*}{\begin{tabular}{c}SDv2-Inpainting\\+AnyDoor \\ (50 steps, 16s)\end{tabular}}}}
        &
        \includegraphics[width=0.138\textwidth]{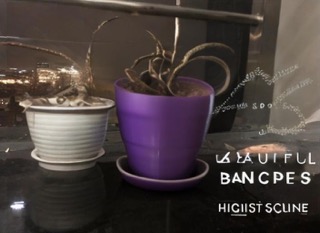} &
        \includegraphics[width=0.138\textwidth]{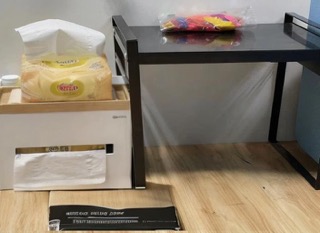} &
        \includegraphics[width=0.138\textwidth]{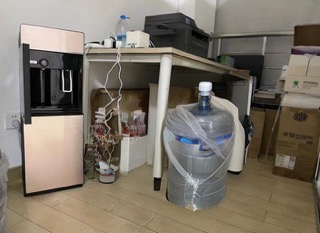} &
        \includegraphics[width=0.138\textwidth]{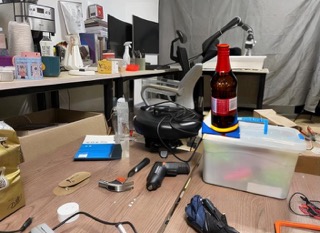} &
        \includegraphics[width=0.138\textwidth]{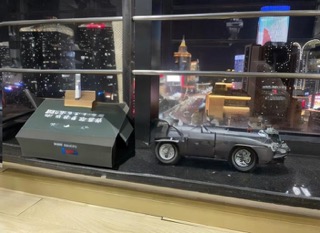} &
        \includegraphics[width=0.138\textwidth]{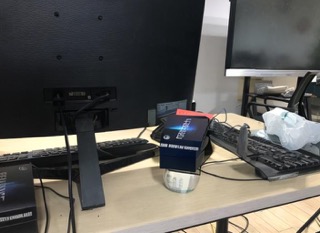} &\\

        {\raisebox{0.47in}{\multirow{1}{*}{\begin{tabular}{c}SDv2-Inpainting\\+AnyDoor \\ (16 steps, 6s)\end{tabular}}}}
        &
        \includegraphics[width=0.138\textwidth]{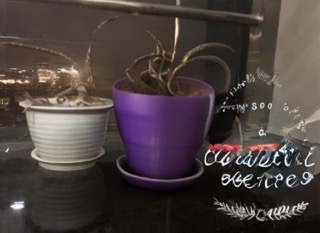} &
        \includegraphics[width=0.138\textwidth]{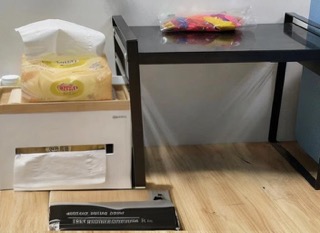} &
        \includegraphics[width=0.138\textwidth]{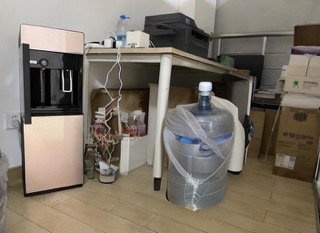} &
        \includegraphics[width=0.138\textwidth]{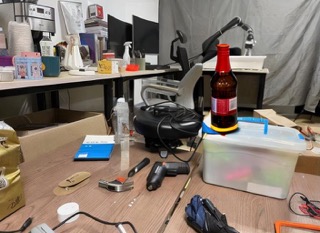} &
        \includegraphics[width=0.138\textwidth]{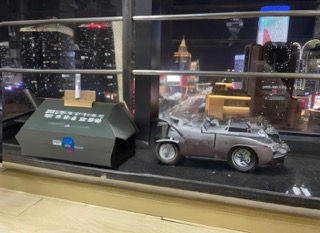} &
        \includegraphics[width=0.138\textwidth]{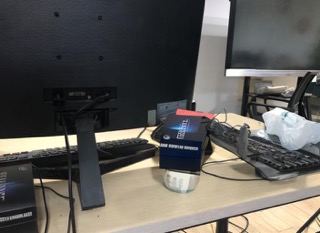} &\\
        
        {\raisebox{0.37in}{\multirow{1}{*}{\begin{tabular}{c}SelfGuidance \\ (50 steps, 14s)\end{tabular}}}} &
        \includegraphics[width=0.138\textwidth]{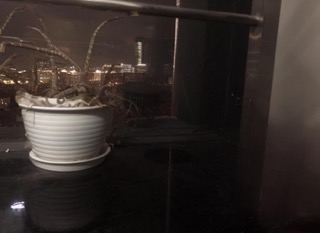} &
        \includegraphics[width=0.138\textwidth]{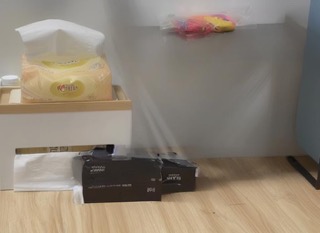} &
        \includegraphics[width=0.138\textwidth]{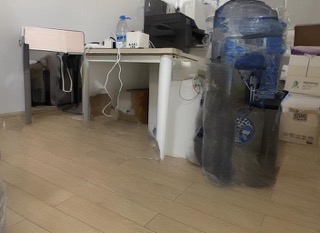} &
        \includegraphics[width=0.138\textwidth]{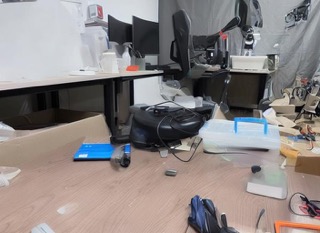} &
        \includegraphics[width=0.138\textwidth]{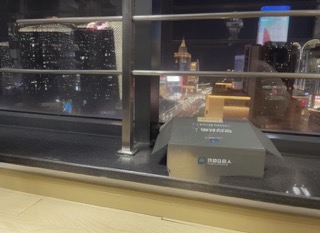} &
        \includegraphics[width=0.138\textwidth]{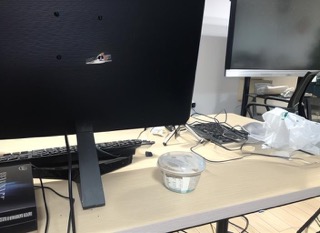} &\\
        
        {\raisebox{0.37in}{\multirow{1}{*}{\begin{tabular}{c}SelfGuidance \\ (16 steps, 6s)\end{tabular}}}} &
        \includegraphics[width=0.138\textwidth]{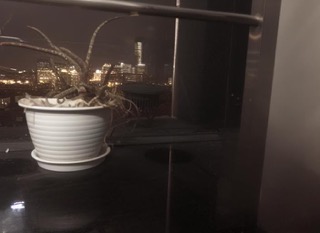} &
        \includegraphics[width=0.138\textwidth]{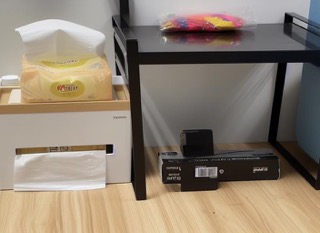} &
        \includegraphics[width=0.138\textwidth]{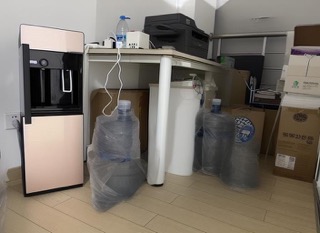} &
        \includegraphics[width=0.138\textwidth]{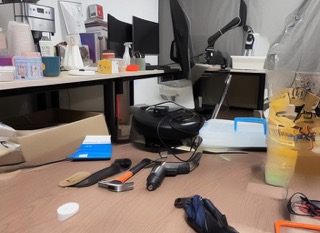} &
        \includegraphics[width=0.138\textwidth]{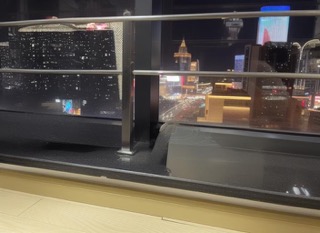} &
        \includegraphics[width=0.138\textwidth]{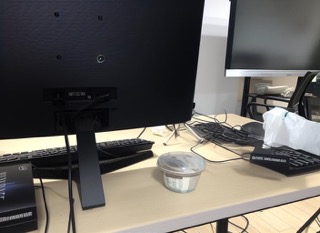} &\\
        
        {\raisebox{0.37in}{\multirow{1}{*}{\begin{tabular}{c}DragonDiffusion \\ (50 steps, 30s)\end{tabular}}}} &
        \includegraphics[width=0.138\textwidth]{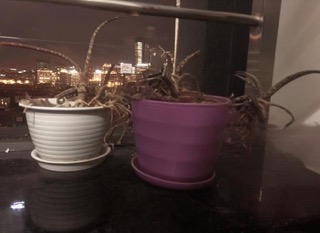} &
        \includegraphics[width=0.138\textwidth]{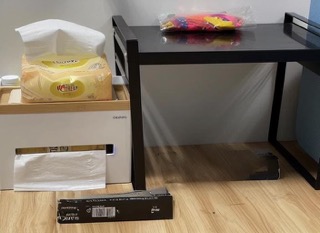} &
        \includegraphics[width=0.138\textwidth]{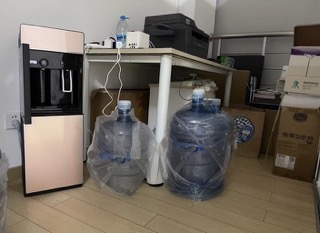} &
        \includegraphics[width=0.138\textwidth]{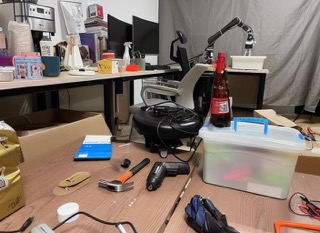} &
        \includegraphics[width=0.138\textwidth]{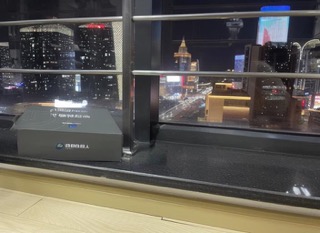} &
        \includegraphics[width=0.138\textwidth]{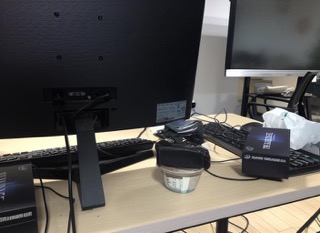} &\\

        {\raisebox{0.37in}{\multirow{1}{*}{\begin{tabular}{c}DragonDiffusion \\ (16 steps, 12s)\end{tabular}}}} &
        \includegraphics[width=0.138\textwidth]{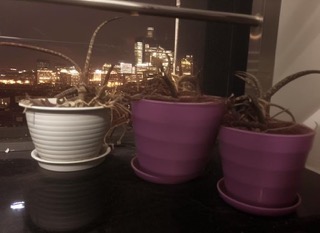} &
        \includegraphics[width=0.138\textwidth]{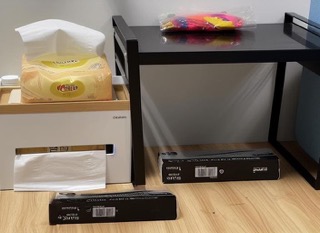} &
        \includegraphics[width=0.138\textwidth]{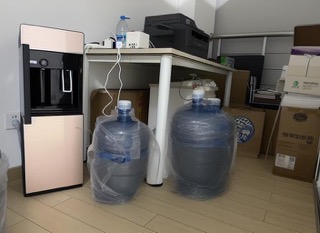} &
        \includegraphics[width=0.138\textwidth]{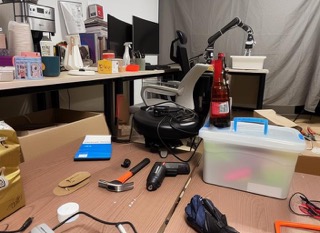} &
        \includegraphics[width=0.138\textwidth]{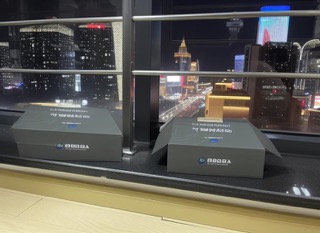} &
        \includegraphics[width=0.138\textwidth]{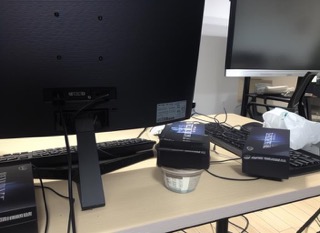} &\\

        {\raisebox{0.37in}{\multirow{1}{*}{\begin{tabular}{c}DiffEditor \\ (50 steps, 32s)\end{tabular}}}} &
        \includegraphics[width=0.138\textwidth]{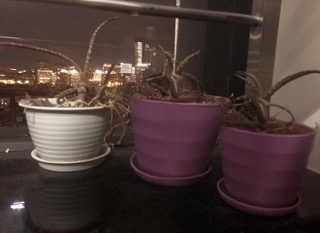} &
        \includegraphics[width=0.138\textwidth]{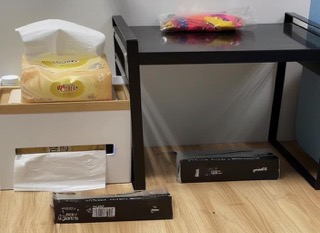} &
        \includegraphics[width=0.138\textwidth]{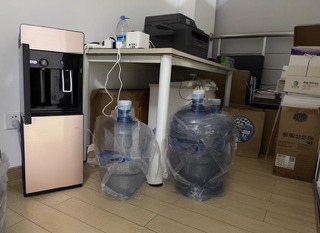} &
        \includegraphics[width=0.138\textwidth]{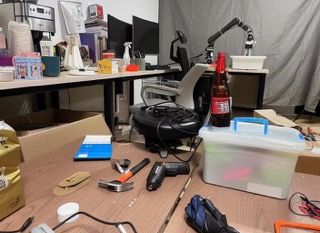} &
        \includegraphics[width=0.138\textwidth]{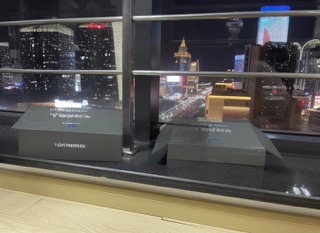} &
        \includegraphics[width=0.138\textwidth]{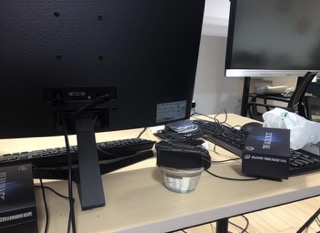} &\\

        {\raisebox{0.37in}{\multirow{1}{*}{\begin{tabular}{c}DiffEditor \\ (16 steps, 11s)\end{tabular}}}} &
        \includegraphics[width=0.138\textwidth]{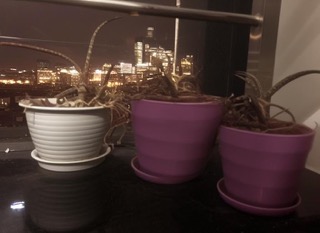} &
        \includegraphics[width=0.138\textwidth]{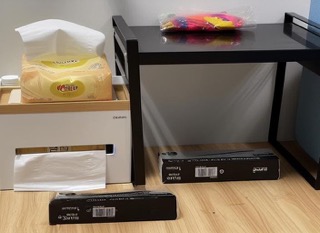} &
        \includegraphics[width=0.138\textwidth]{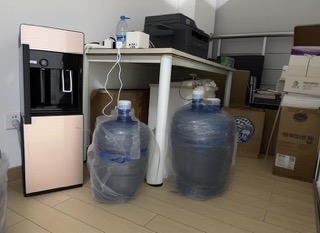} &
        \includegraphics[width=0.138\textwidth]{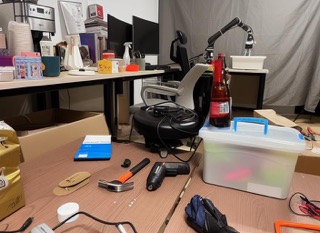} &
        \includegraphics[width=0.138\textwidth]{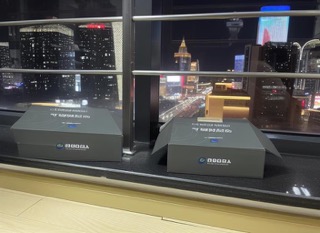} &
        \includegraphics[width=0.138\textwidth]{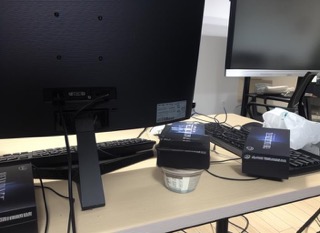} &\\

        {\raisebox{0.37in}{\multirow{1}{*}{\begin{tabular}{c}\textbf{PixelMan} \\ (50 steps, 34s)\end{tabular}}}} &
        \includegraphics[width=0.138\textwidth]{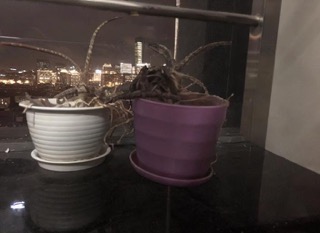} &
        \includegraphics[width=0.138\textwidth]{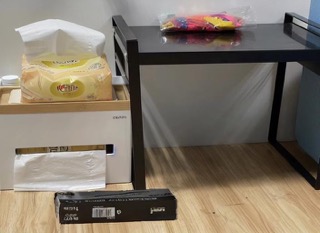} &
        \includegraphics[width=0.138\textwidth]{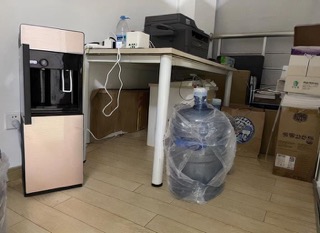} &
        \includegraphics[width=0.138\textwidth]{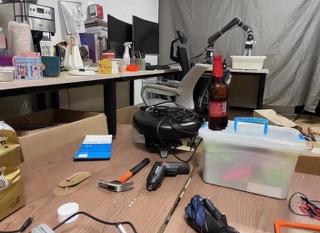} &
        \includegraphics[width=0.138\textwidth]{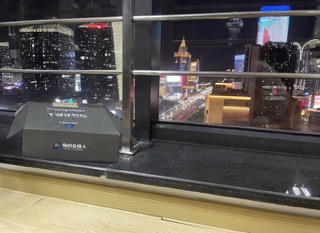} &
        \includegraphics[width=0.138\textwidth]{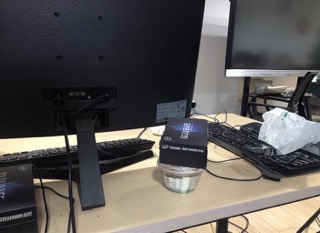} &\\
        
        {\raisebox{0.37in}{\multirow{1}{*}{\begin{tabular}{c}\textbf{PixelMan} \\ (16 steps, 11s)\end{tabular}}}} &
        \includegraphics[width=0.138\textwidth]{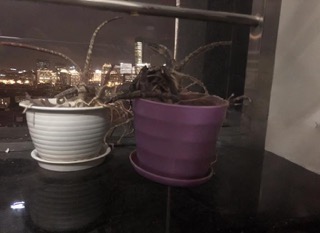} &
        \includegraphics[width=0.138\textwidth]{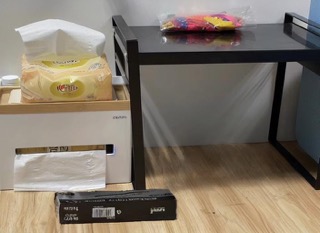} &
        \includegraphics[width=0.138\textwidth]{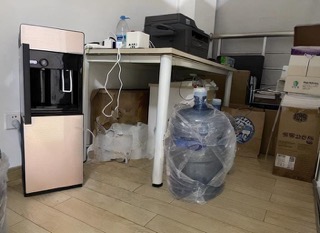} &
        \includegraphics[width=0.138\textwidth]{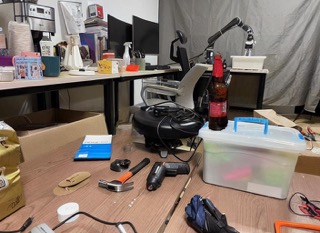} &
        \includegraphics[width=0.138\textwidth]{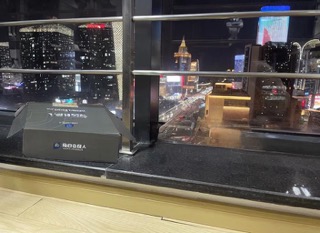} &
        \includegraphics[width=0.138\textwidth]{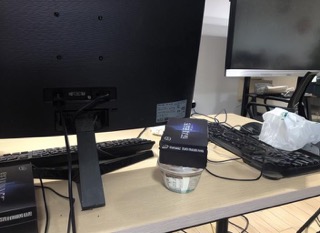} &\\

    \end{tabular}
    }
    \caption{
        \textbf{Additional qualitative comparison} on the ReS dataset at both 16 and 50 steps. 
    }
    \label{fig:examples_res_1}
\end{figure*}

\begin{figure*}[hbt!]
    \centering
    \setlength{\tabcolsep}{0.4pt}
    \renewcommand{\arraystretch}{0.4}
    {\footnotesize
    \begin{tabular}{c c c c c c c c}
        &
        \multicolumn{1}{c}{(a)} &
        \multicolumn{1}{c}{(b)} &
        \multicolumn{1}{c}{(c)} &
        \multicolumn{1}{c}{(d)} &
        \multicolumn{1}{c}{(e)} &
        \multicolumn{1}{c}{(f)} \\

        {\raisebox{0.34in}{
        \multirow{1}{*}{\rotatebox{0}{Input}}}} &
        \includegraphics[width=0.138\textwidth]{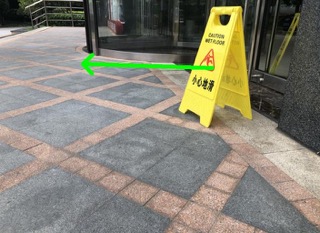} &
        \includegraphics[width=0.138\textwidth]{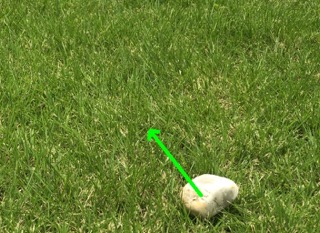} &
        \includegraphics[width=0.138\textwidth]{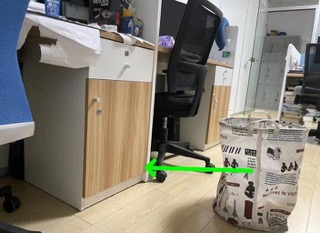} &
        \includegraphics[width=0.138\textwidth]{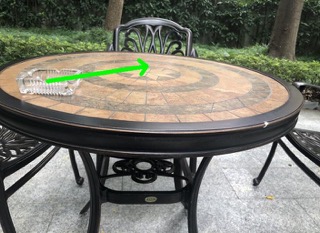} &
        \includegraphics[width=0.138\textwidth]{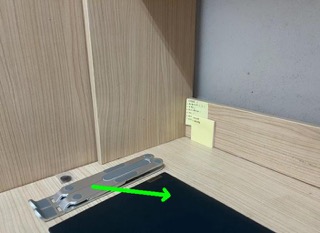} &
        \includegraphics[width=0.138\textwidth]{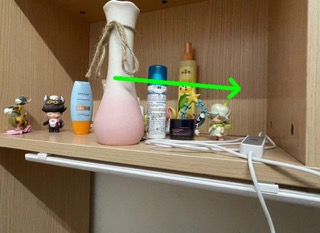} &\\

        {\raisebox{0.47in}{\multirow{1}{*}{\begin{tabular}{c}SDv2-Inpainting\\+AnyDoor \\ (50 steps, 16s)\end{tabular}}}}
        &
        \includegraphics[width=0.138\textwidth]{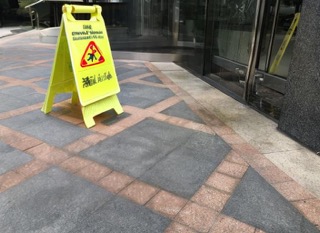} &
        \includegraphics[width=0.138\textwidth]{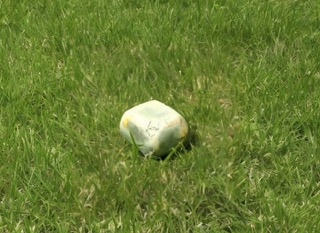} &
        \includegraphics[width=0.138\textwidth]{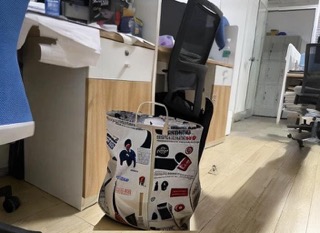} &
        \includegraphics[width=0.138\textwidth]{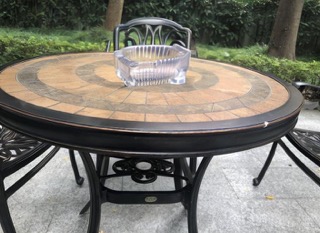} &
        \includegraphics[width=0.138\textwidth]{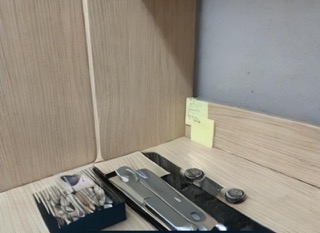} &
        \includegraphics[width=0.138\textwidth]{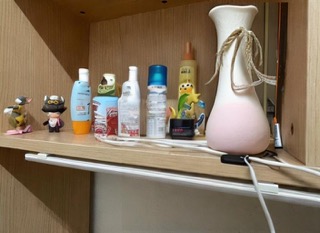} &\\

        {\raisebox{0.47in}{\multirow{1}{*}{\begin{tabular}{c}SDv2-Inpainting\\+AnyDoor \\ (16 steps, 6s)\end{tabular}}}}
        &
        \includegraphics[width=0.138\textwidth]{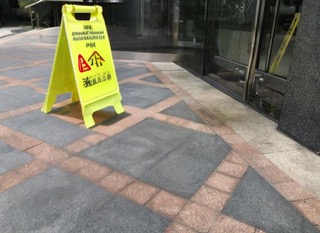} &
        \includegraphics[width=0.138\textwidth]{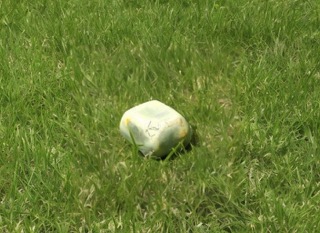} &
        \includegraphics[width=0.138\textwidth]{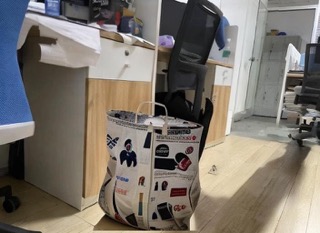} &
        \includegraphics[width=0.138\textwidth]{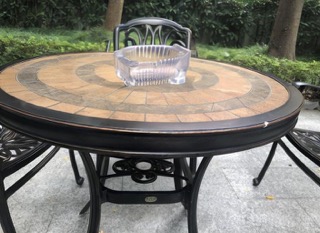} &
        \includegraphics[width=0.138\textwidth]{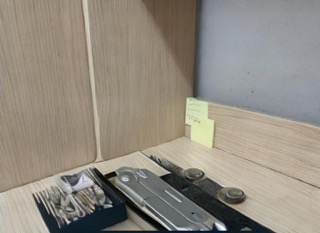} &
        \includegraphics[width=0.138\textwidth]{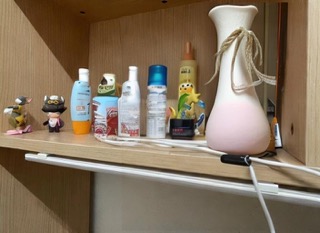} &\\
        
        {\raisebox{0.37in}{\multirow{1}{*}{\begin{tabular}{c}SelfGuidance \\ (50 steps, 14s)\end{tabular}}}} &
        \includegraphics[width=0.138\textwidth]{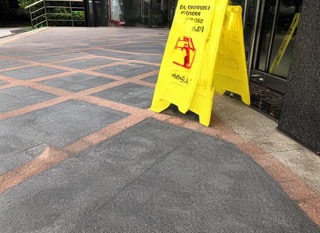} &
        \includegraphics[width=0.138\textwidth]{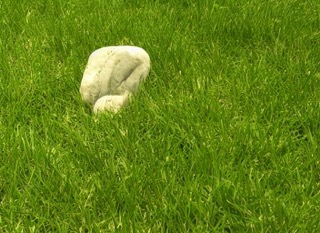} &
        \includegraphics[width=0.138\textwidth]{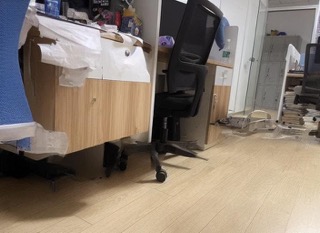} &
        \includegraphics[width=0.138\textwidth]{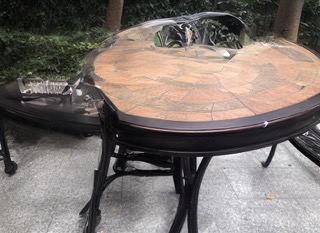} &
        \includegraphics[width=0.138\textwidth]{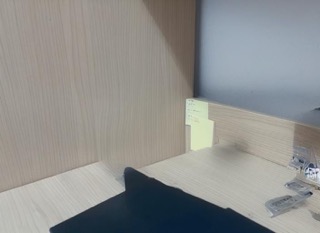} &
        \includegraphics[width=0.138\textwidth]{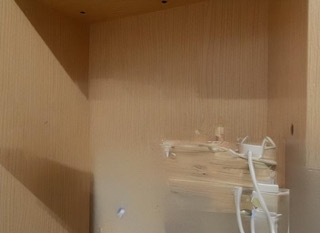} &\\
        
        {\raisebox{0.37in}{\multirow{1}{*}{\begin{tabular}{c}SelfGuidance \\ (16 steps, 6s)\end{tabular}}}} &
        \includegraphics[width=0.138\textwidth]{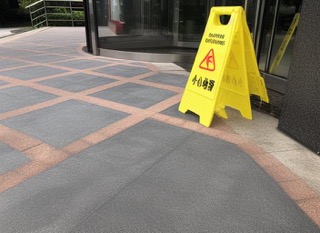} &
        \includegraphics[width=0.138\textwidth]{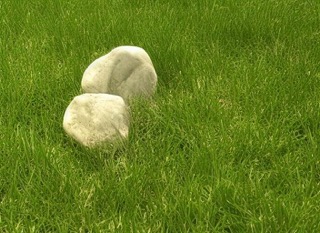} &
        \includegraphics[width=0.138\textwidth]{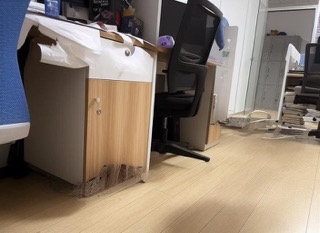} &
        \includegraphics[width=0.138\textwidth]{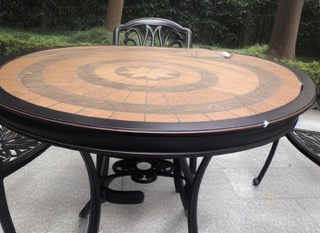} &
        \includegraphics[width=0.138\textwidth]{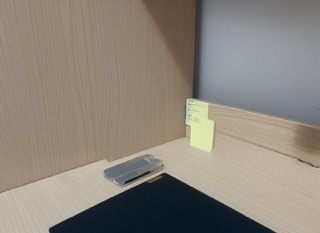} &
        \includegraphics[width=0.138\textwidth]{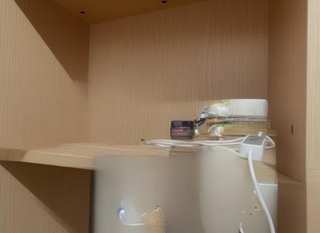} &\\
        
        {\raisebox{0.37in}{\multirow{1}{*}{\begin{tabular}{c}DragonDiffusion \\ (50 steps, 30s)\end{tabular}}}} &
        \includegraphics[width=0.138\textwidth]{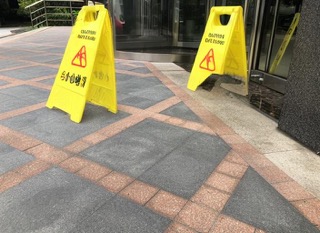} &
        \includegraphics[width=0.138\textwidth]{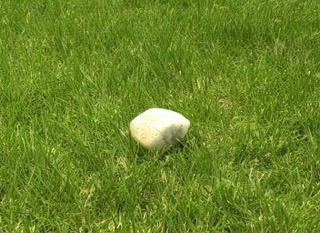} &
        \includegraphics[width=0.138\textwidth]{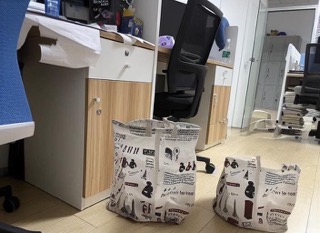} &
        \includegraphics[width=0.138\textwidth]{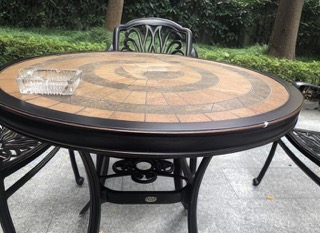} &
        \includegraphics[width=0.138\textwidth]{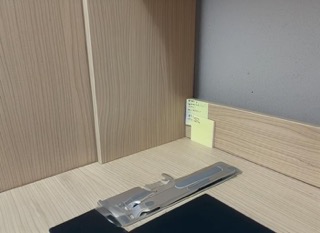} &
        \includegraphics[width=0.138\textwidth]{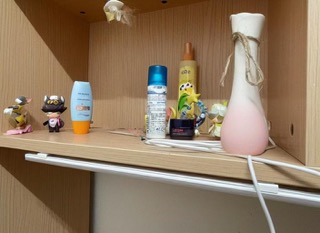} &\\

        {\raisebox{0.37in}{\multirow{1}{*}{\begin{tabular}{c}DragonDiffusion \\ (16 steps, 12s)\end{tabular}}}} &
        \includegraphics[width=0.138\textwidth]{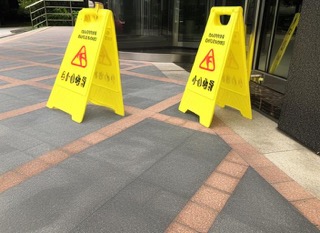} &
        \includegraphics[width=0.138\textwidth]{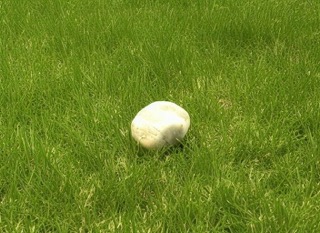} &
        \includegraphics[width=0.138\textwidth]{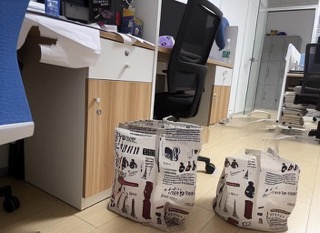} &
        \includegraphics[width=0.138\textwidth]{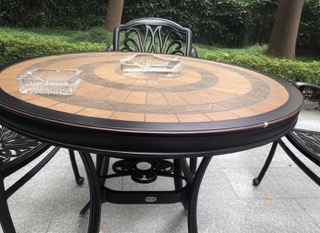} &
        \includegraphics[width=0.138\textwidth]{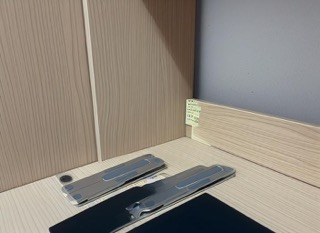} &
        \includegraphics[width=0.138\textwidth]{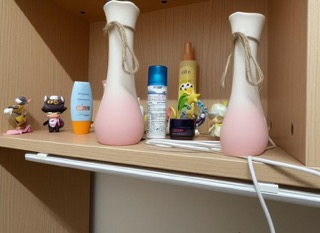} &\\

        {\raisebox{0.37in}{\multirow{1}{*}{\begin{tabular}{c}DiffEditor \\ (50 steps, 32s)\end{tabular}}}} &
        \includegraphics[width=0.138\textwidth]{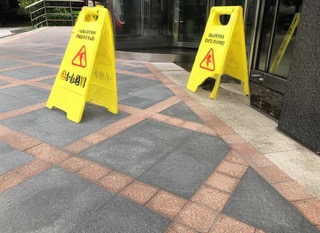} &
        \includegraphics[width=0.138\textwidth]{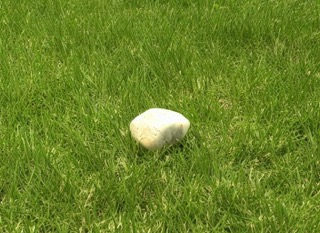} &
        \includegraphics[width=0.138\textwidth]{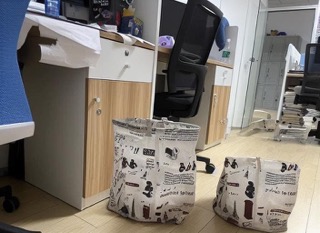} &
        \includegraphics[width=0.138\textwidth]{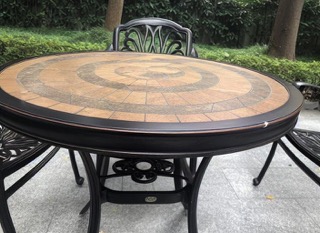} &
        \includegraphics[width=0.138\textwidth]{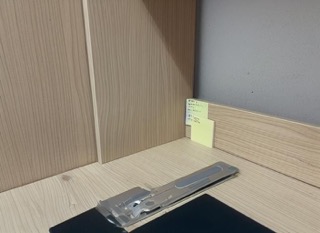} &
        \includegraphics[width=0.138\textwidth]{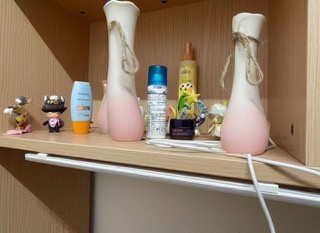} &\\

        {\raisebox{0.37in}{\multirow{1}{*}{\begin{tabular}{c}DiffEditor \\ (16 steps, 11s)\end{tabular}}}} &
        \includegraphics[width=0.138\textwidth]{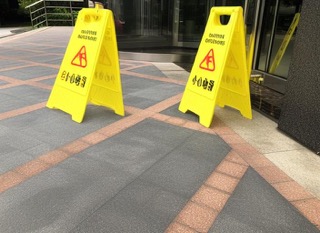} &
        \includegraphics[width=0.138\textwidth]{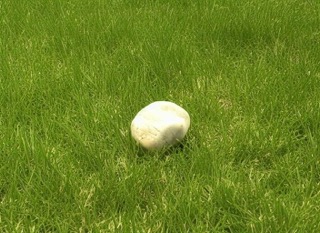} &
        \includegraphics[width=0.138\textwidth]{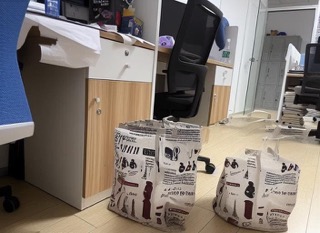} &
        \includegraphics[width=0.138\textwidth]{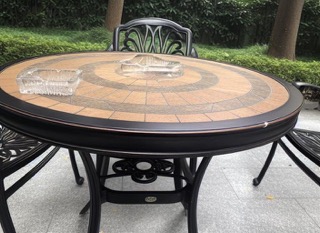} &
        \includegraphics[width=0.138\textwidth]{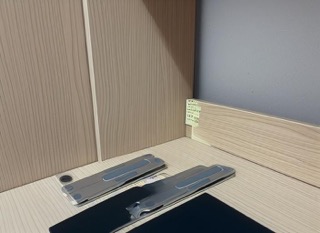} &
        \includegraphics[width=0.138\textwidth]{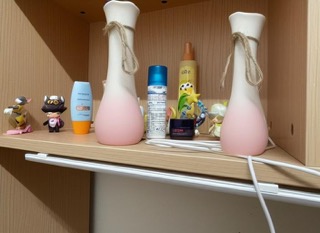} &\\

        {\raisebox{0.37in}{\multirow{1}{*}{\begin{tabular}{c}\textbf{PixelMan} \\ (50 steps, 34s)\end{tabular}}}} &
        \includegraphics[width=0.138\textwidth]{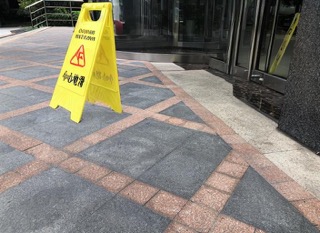} &
        \includegraphics[width=0.138\textwidth]{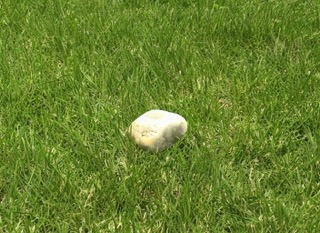} &
        \includegraphics[width=0.138\textwidth]{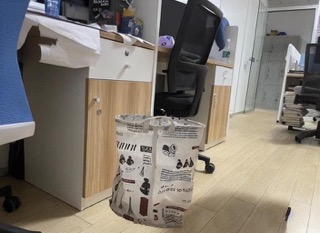} &
        \includegraphics[width=0.138\textwidth]{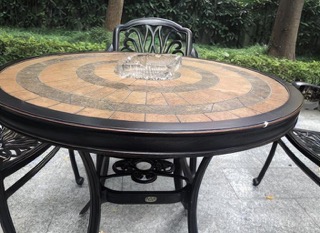} &
        \includegraphics[width=0.138\textwidth]{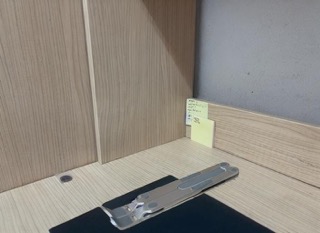} &
        \includegraphics[width=0.138\textwidth]{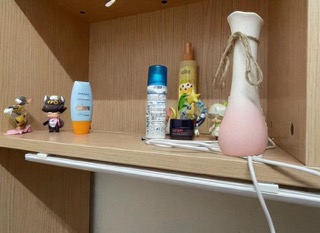} &\\
        
        {\raisebox{0.37in}{\multirow{1}{*}{\begin{tabular}{c}\textbf{PixelMan} \\ (16 steps, 11s)\end{tabular}}}} &
        \includegraphics[width=0.138\textwidth]{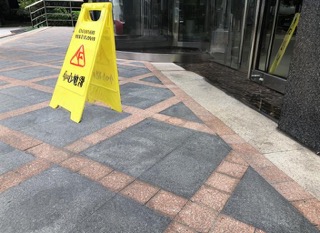} &
        \includegraphics[width=0.138\textwidth]{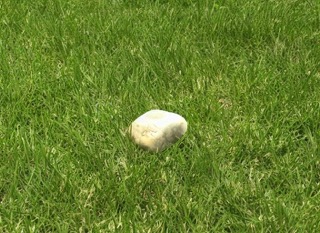} &
        \includegraphics[width=0.138\textwidth]{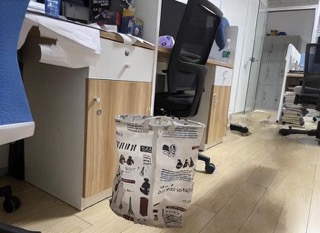} &
        \includegraphics[width=0.138\textwidth]{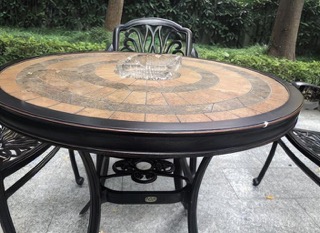} &
        \includegraphics[width=0.138\textwidth]{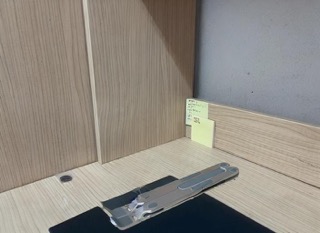} &
        \includegraphics[width=0.138\textwidth]{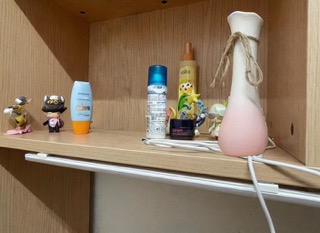} &\\

    \end{tabular}
    }
    \caption{
        \textbf{Additional qualitative comparison} on the ReS dataset at both 16 and 50 steps. 
    }
    \label{fig:examples_res_2}
\end{figure*}

\begin{figure*}[hbt!]
    \centering
    \setlength{\tabcolsep}{0.4pt}
    \renewcommand{\arraystretch}{0.4}
    {\footnotesize
    \begin{tabular}{c c c c c c c c}
        &
        \multicolumn{1}{c}{(a)} &
        \multicolumn{1}{c}{(b)} &
        \multicolumn{1}{c}{(c)} &
        \multicolumn{1}{c}{(d)} &
        \multicolumn{1}{c}{(e)} &
        \multicolumn{1}{c}{(f)} \\

        {\raisebox{0.44in}{
        \multirow{1}{*}{\rotatebox{0}{Input}}}} &
        \includegraphics[width=0.138\textwidth]{images/comparison/COCOEE/000000337109_GT_source.jpg} &
        \includegraphics[width=0.138\textwidth]{images/comparison/COCOEE/000000061097_GT_source.jpg} &
        \includegraphics[width=0.138\textwidth]{images/comparison/COCOEE/000000485981_GT_source.jpg} &
        \includegraphics[width=0.138\textwidth]{images/comparison/COCOEE/000001557820_GT_source.jpg} &
        \includegraphics[width=0.138\textwidth]{images/comparison/COCOEE/000000036603_GT_source.jpg} &
        \includegraphics[width=0.138\textwidth]{images/comparison/COCOEE/000001114763_GT_source.jpg} &\\

        {\raisebox{0.57in}{\multirow{1}{*}{\begin{tabular}{c}SDv2-Inpainting\\+AnyDoor \\ (8 steps, 3s)\end{tabular}}}}
        &
        \includegraphics[width=0.138\textwidth]{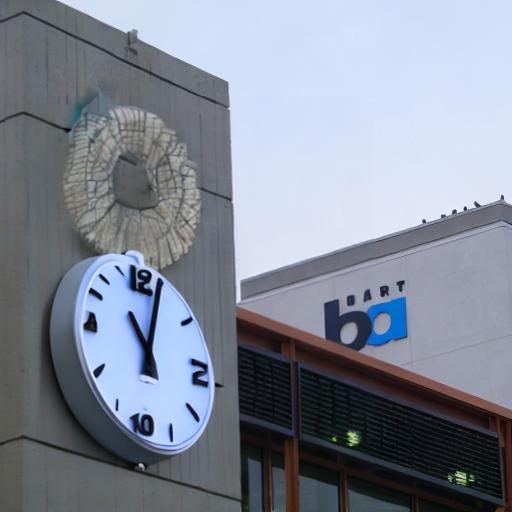} &
        \includegraphics[width=0.138\textwidth]{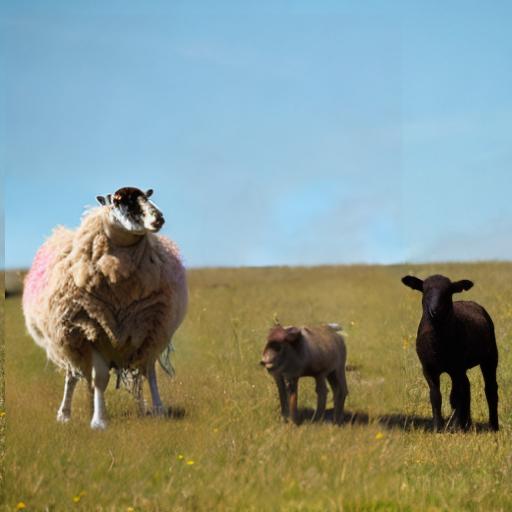} &
        \includegraphics[width=0.138\textwidth]{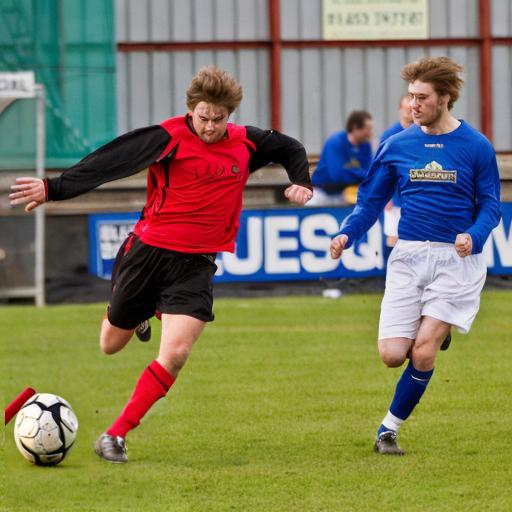} &
        \includegraphics[width=0.138\textwidth]{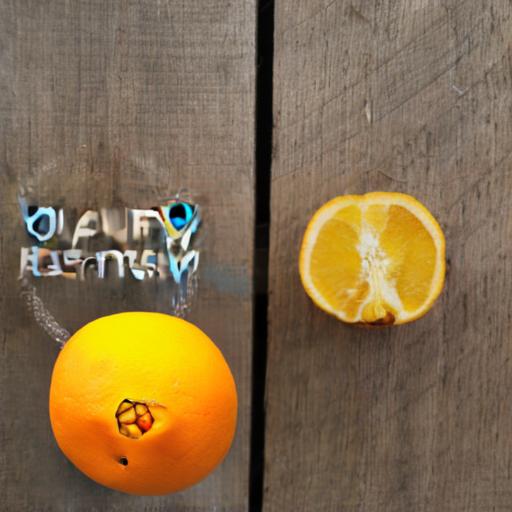} &
        \includegraphics[width=0.138\textwidth]{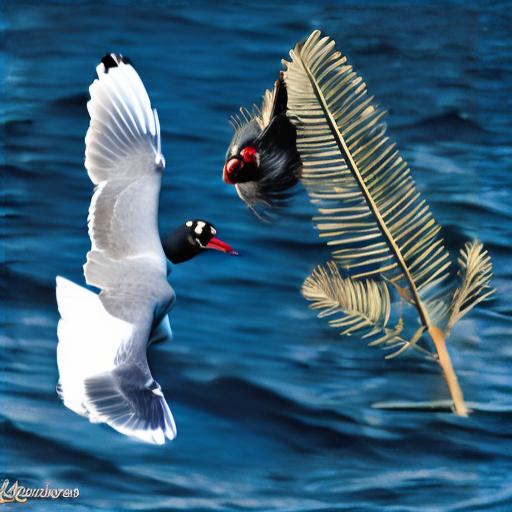} &
        \includegraphics[width=0.138\textwidth]{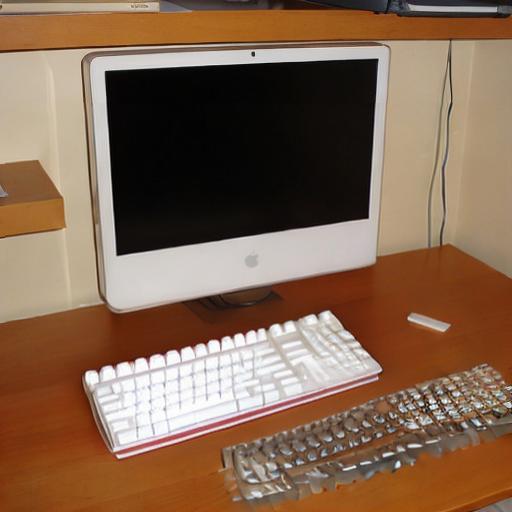} &\\
        
        {\raisebox{0.47in}{\multirow{1}{*}{\begin{tabular}{c}SelfGuidance \\ (8 steps, 2s)\end{tabular}}}} &
        \includegraphics[width=0.138\textwidth]{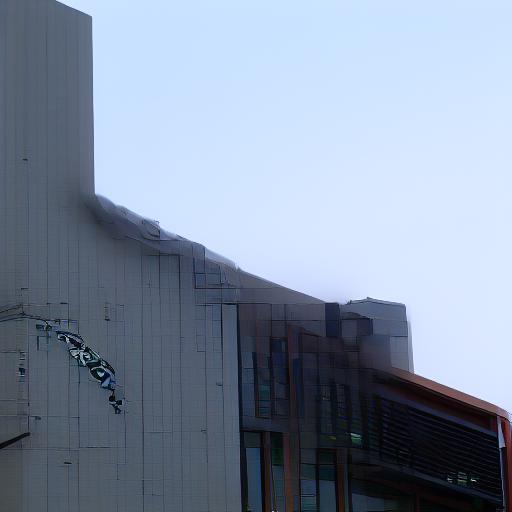} &
        \includegraphics[width=0.138\textwidth]{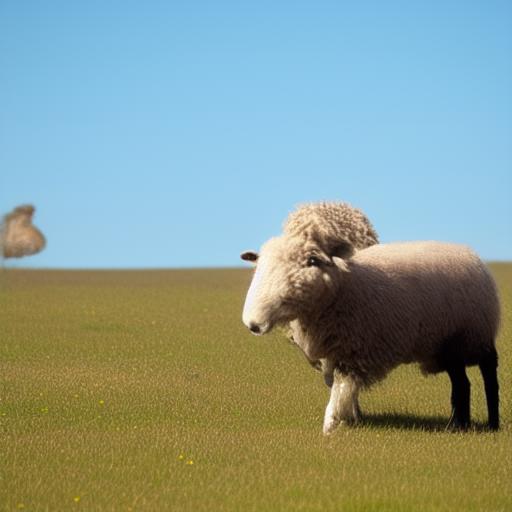} &
        \includegraphics[width=0.138\textwidth]{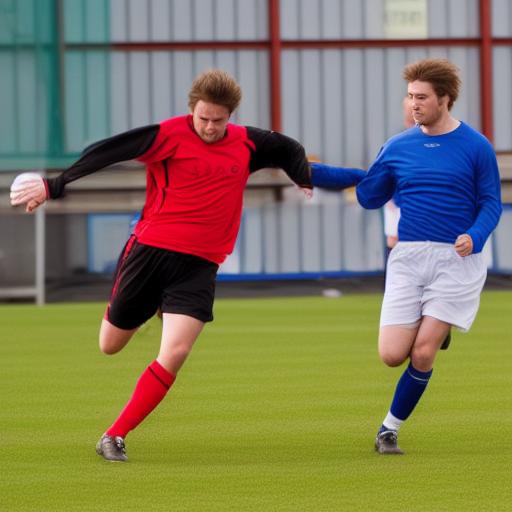} &
        \includegraphics[width=0.138\textwidth]{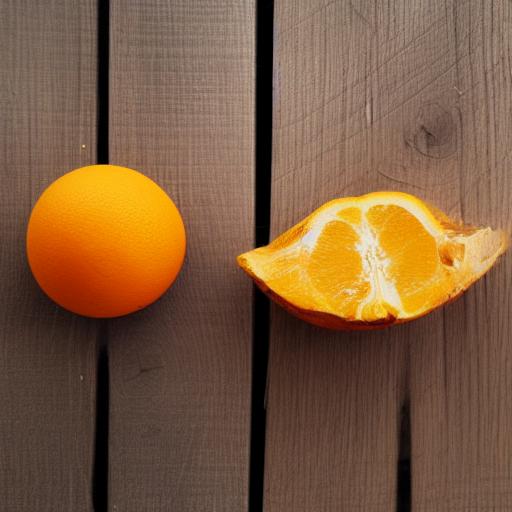} &
        \includegraphics[width=0.138\textwidth]{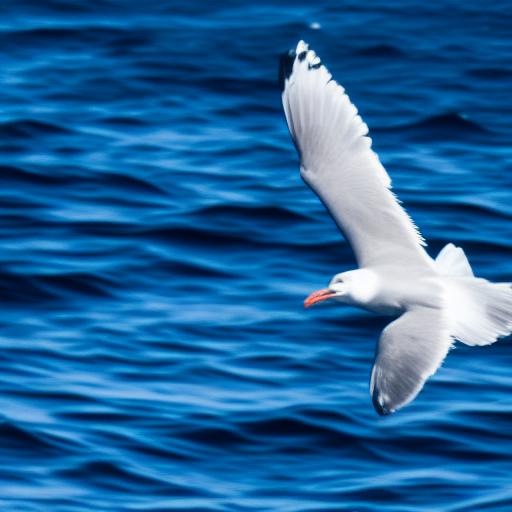} &
        \includegraphics[width=0.138\textwidth]{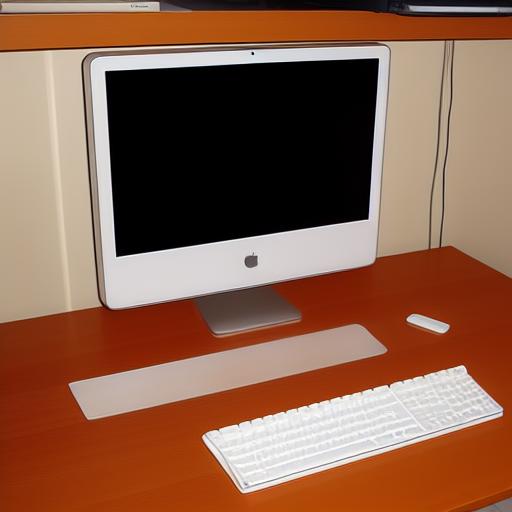} &\\

        {\raisebox{0.47in}{\multirow{1}{*}{\begin{tabular}{c}DragonDiffusion \\ (8 steps, 5s)\end{tabular}}}} &
        \includegraphics[width=0.138\textwidth]{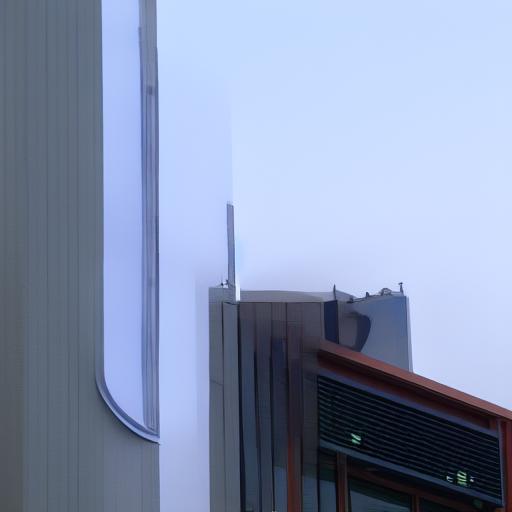} &
        \includegraphics[width=0.138\textwidth]{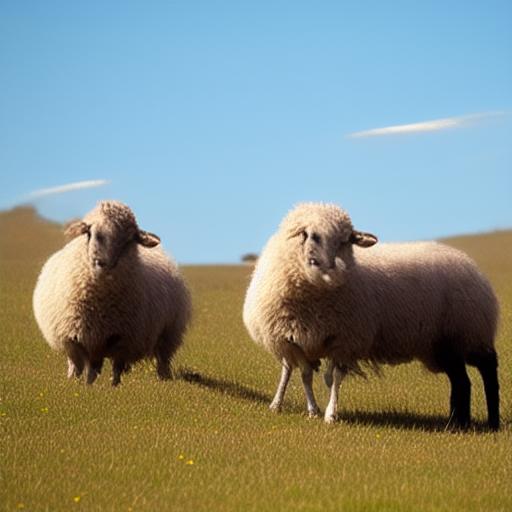} &
        \includegraphics[width=0.138\textwidth]{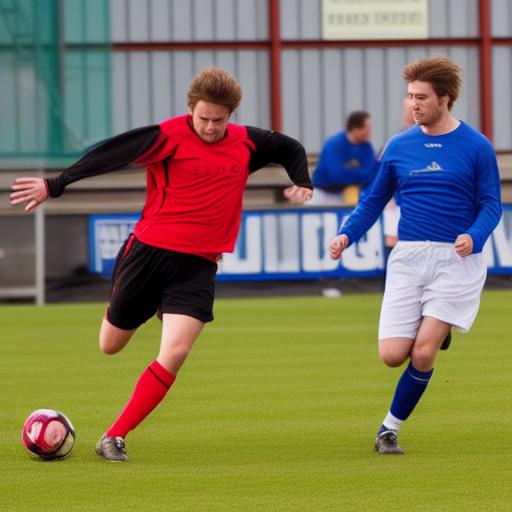} &
        \includegraphics[width=0.138\textwidth]{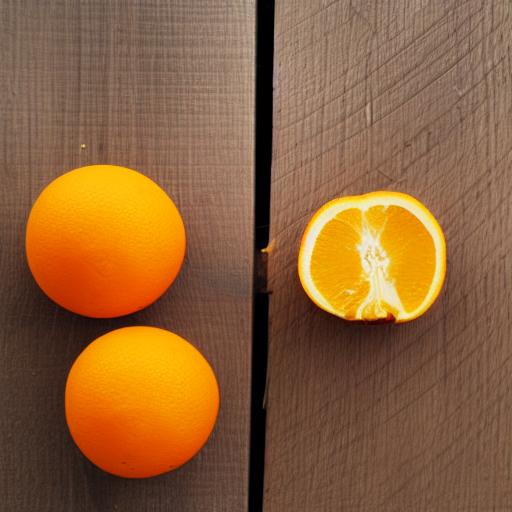} &
        \includegraphics[width=0.138\textwidth]{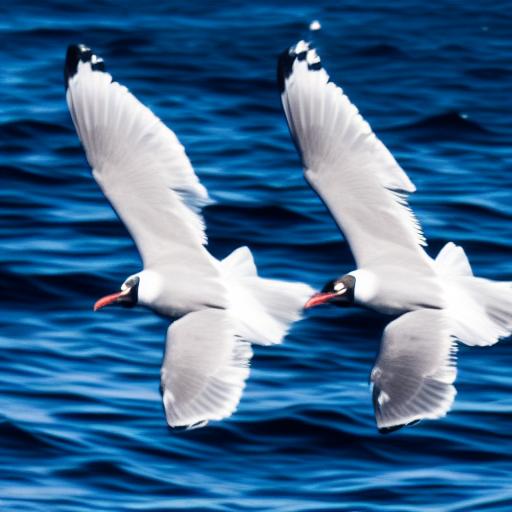} &
        \includegraphics[width=0.138\textwidth]{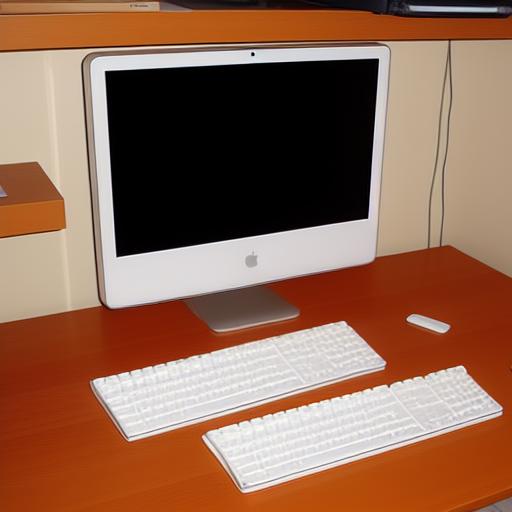} &\\

        {\raisebox{0.47in}{\multirow{1}{*}{\begin{tabular}{c}DiffEditor \\ (8 steps, 5s)\end{tabular}}}} &
        \includegraphics[width=0.138\textwidth]{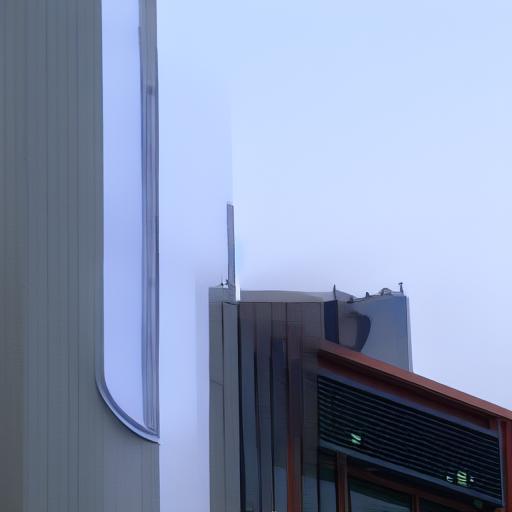} &
        \includegraphics[width=0.138\textwidth]{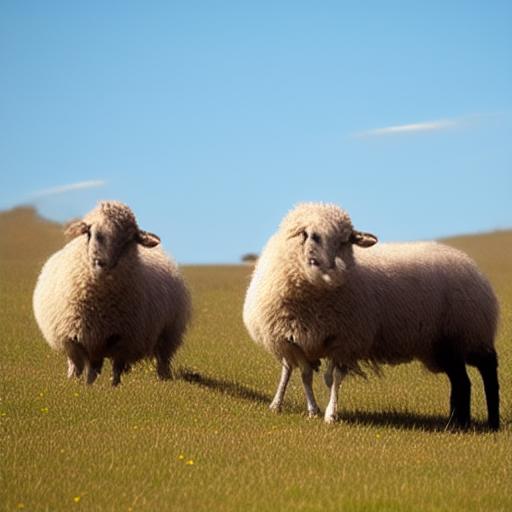} &
        \includegraphics[width=0.138\textwidth]{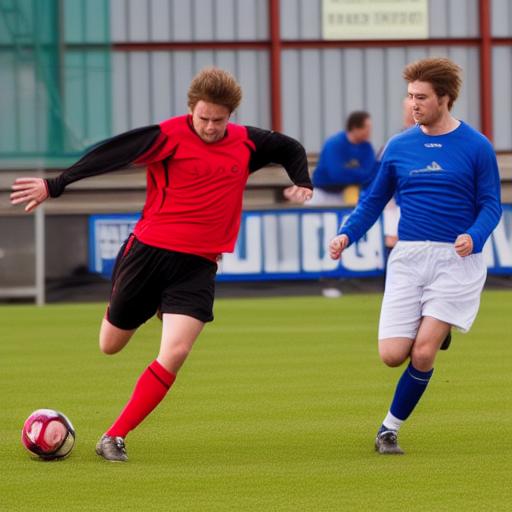} &
        \includegraphics[width=0.138\textwidth]{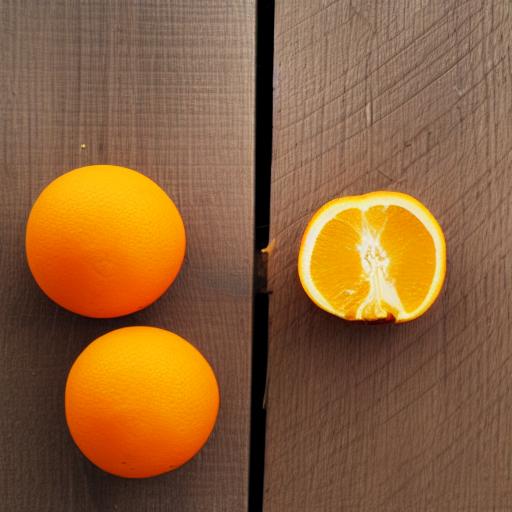} &
        \includegraphics[width=0.138\textwidth]{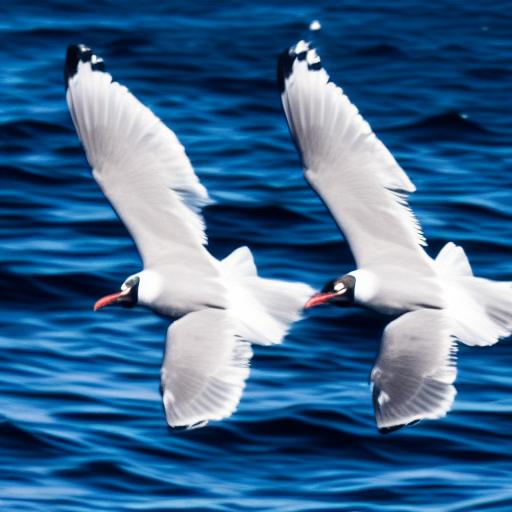} &
        \includegraphics[width=0.138\textwidth]{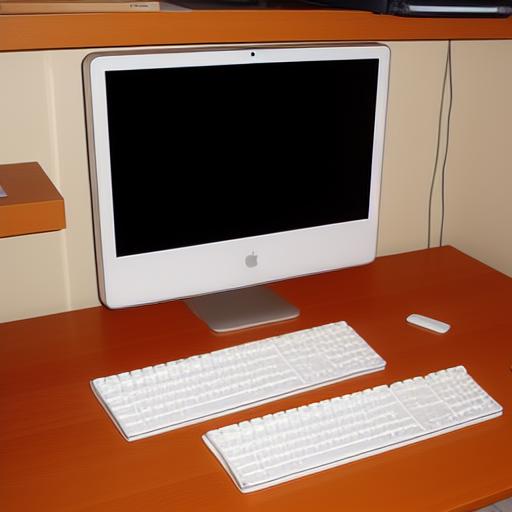} &\\
        
        {\raisebox{0.47in}{\multirow{1}{*}{\begin{tabular}{c}\textbf{PixelMan} \\ (8 steps, 4s)\end{tabular}}}} &
        \includegraphics[width=0.138\textwidth]{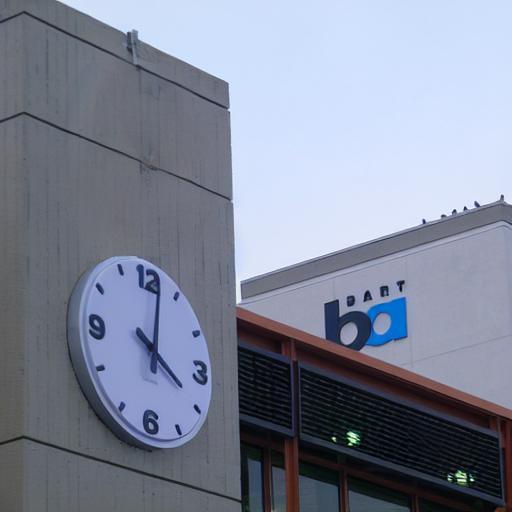} &
        \includegraphics[width=0.138\textwidth]{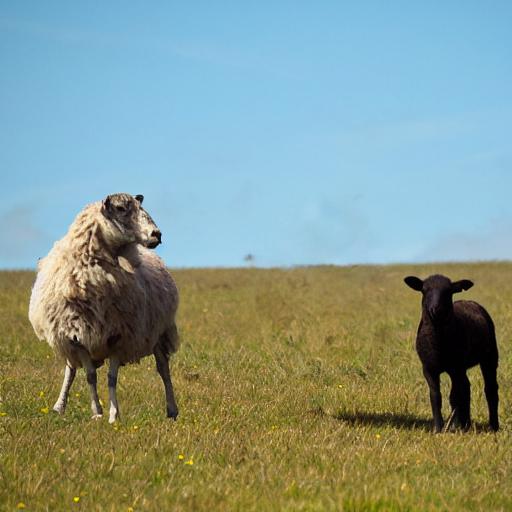} &
        \includegraphics[width=0.138\textwidth]{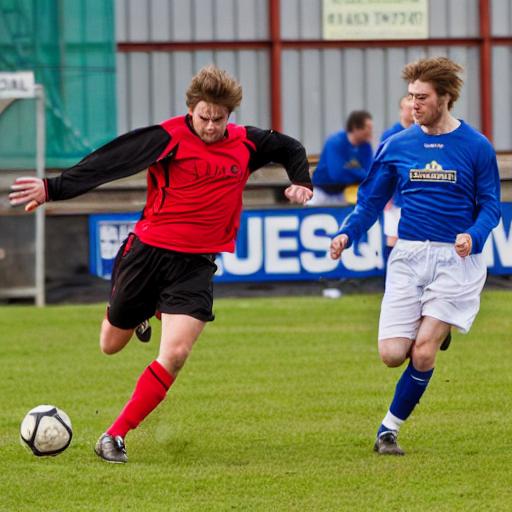} &
        \includegraphics[width=0.138\textwidth]{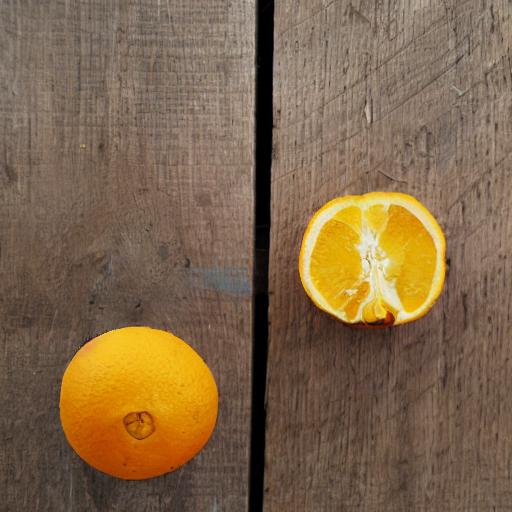} &
        \includegraphics[width=0.138\textwidth]{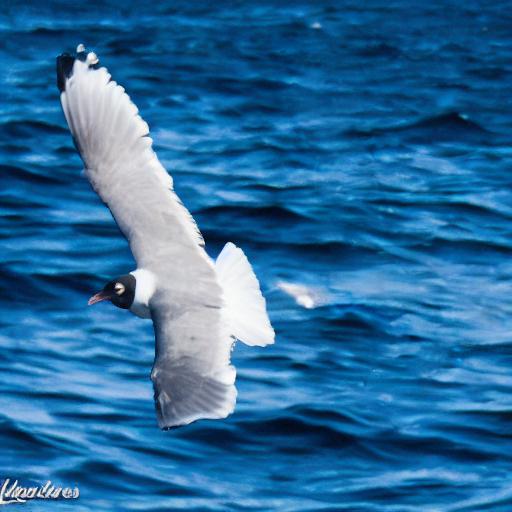} &
        \includegraphics[width=0.138\textwidth]{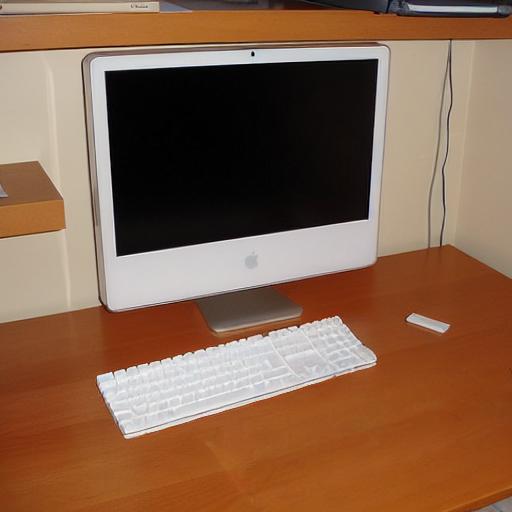} &\\

    \end{tabular}
    }
    \caption{
        \textbf{Additional qualitative comparison} on the COCOEE dataset at 8 steps. 
    }
    \label{fig:examples_8step_cocoee}
\end{figure*}
\begin{figure*}[hbt!]
    \centering
    \setlength{\tabcolsep}{0.4pt}
    \renewcommand{\arraystretch}{0.4}
    {\footnotesize
    \begin{tabular}{c c c c c c c c}
        &
        \multicolumn{1}{c}{(a)} &
        \multicolumn{1}{c}{(b)} &
        \multicolumn{1}{c}{(c)} &
        \multicolumn{1}{c}{(d)} &
        \multicolumn{1}{c}{(e)} &
        \multicolumn{1}{c}{(f)} \\

        {\raisebox{0.34in}{
        \multirow{1}{*}{\rotatebox{0}{Input}}}} &
        \includegraphics[width=0.138\textwidth]{images/comparison/ReS/p100_2_source.jpg} &
        \includegraphics[width=0.138\textwidth]{images/comparison/ReS/p16_1_source.jpg} &
        \includegraphics[width=0.138\textwidth]{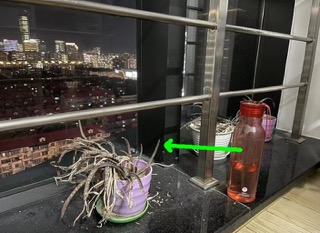} &
        \includegraphics[width=0.138\textwidth]{images/comparison/ReS/p57_2_source.jpg} &
        \includegraphics[width=0.138\textwidth]{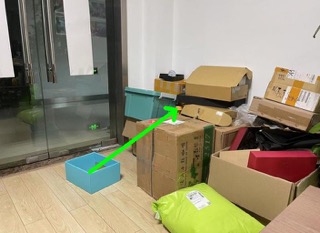} &
        \includegraphics[width=0.138\textwidth]{images/comparison/ReS/p58_2_source.jpg} &\\

        {\raisebox{0.47in}{\multirow{1}{*}{\begin{tabular}{c}SDv2-Inpainting\\+AnyDoor \\ (8 steps, 3s)\end{tabular}}}}
        &
        \includegraphics[width=0.138\textwidth]{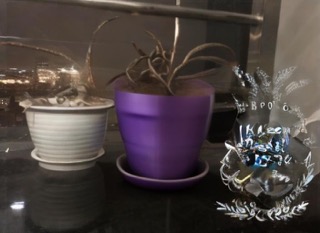} &
        \includegraphics[width=0.138\textwidth]{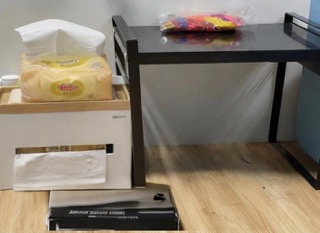} &
        \includegraphics[width=0.138\textwidth]{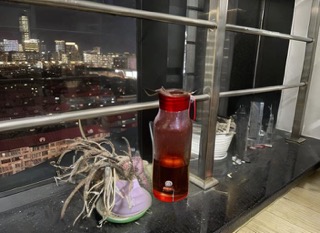} &
        \includegraphics[width=0.138\textwidth]{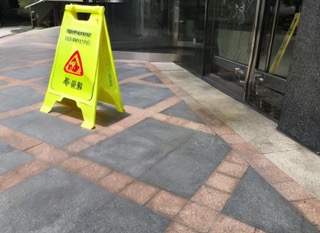} &
        \includegraphics[width=0.138\textwidth]{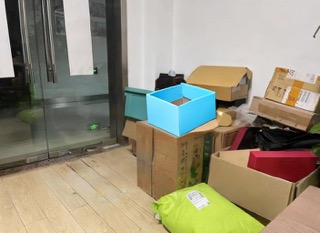} &
        \includegraphics[width=0.138\textwidth]{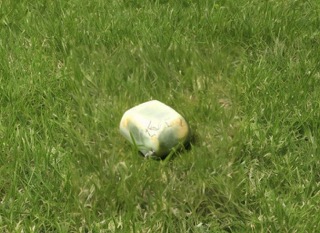} &\\
                
        {\raisebox{0.37in}{\multirow{1}{*}{\begin{tabular}{c}SelfGuidance \\ (8 steps, 3s)\end{tabular}}}} &
        \includegraphics[width=0.138\textwidth]{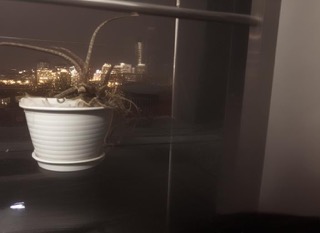} &
        \includegraphics[width=0.138\textwidth]{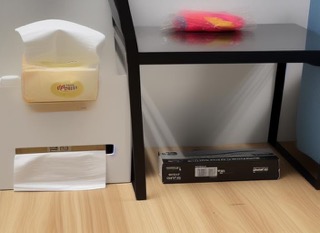} &
        \includegraphics[width=0.138\textwidth]{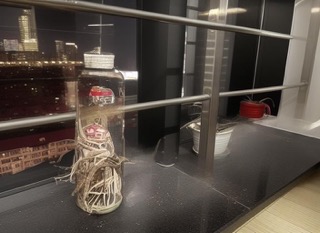} &
        \includegraphics[width=0.138\textwidth]{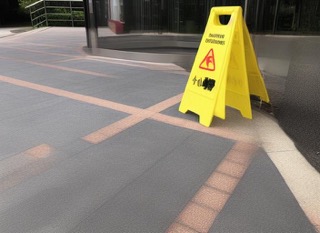} &
        \includegraphics[width=0.138\textwidth]{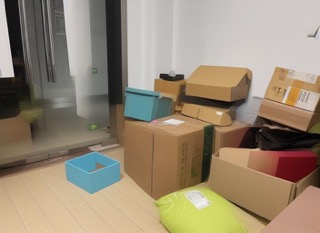} &
        \includegraphics[width=0.138\textwidth]{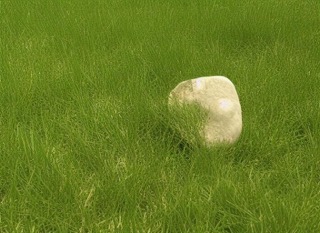} &\\
        
        {\raisebox{0.37in}{\multirow{1}{*}{\begin{tabular}{c}DragonDiffusion \\ (8 steps, 6s)\end{tabular}}}} &
        \includegraphics[width=0.138\textwidth]{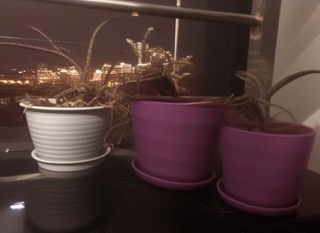} &
        \includegraphics[width=0.138\textwidth]{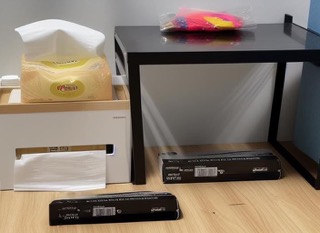} &
        \includegraphics[width=0.138\textwidth]{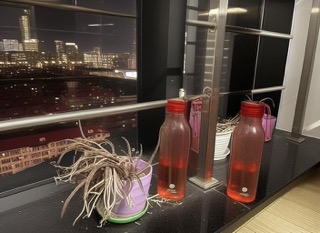} &
        \includegraphics[width=0.138\textwidth]{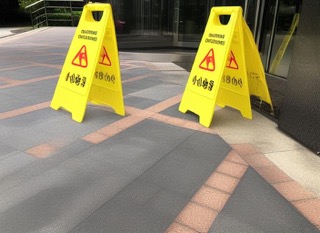} &
        \includegraphics[width=0.138\textwidth]{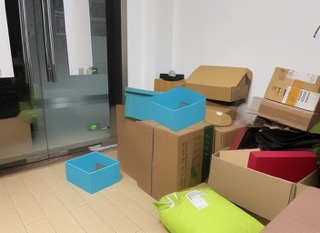} &
        \includegraphics[width=0.138\textwidth]{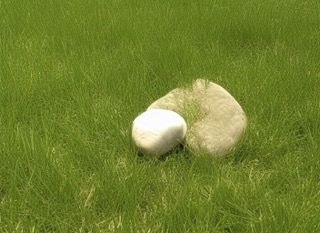} &\\

        {\raisebox{0.37in}{\multirow{1}{*}{\begin{tabular}{c}DiffEditor \\ (8 steps, 6s)\end{tabular}}}} &
        \includegraphics[width=0.138\textwidth]{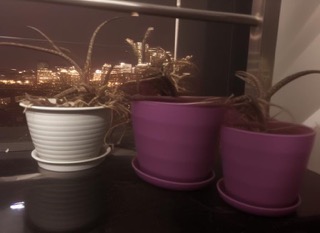} &
        \includegraphics[width=0.138\textwidth]{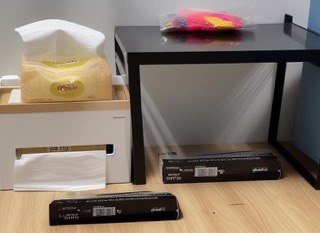} &
        \includegraphics[width=0.138\textwidth]{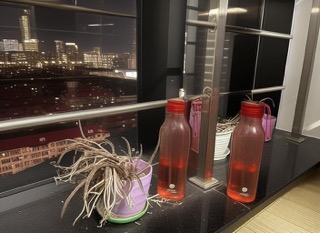} &
        \includegraphics[width=0.138\textwidth]{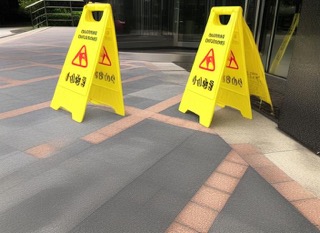} &
        \includegraphics[width=0.138\textwidth]{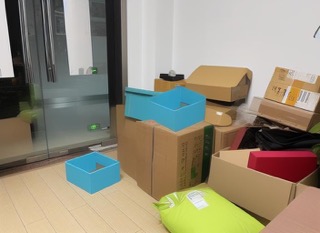} &
        \includegraphics[width=0.138\textwidth]{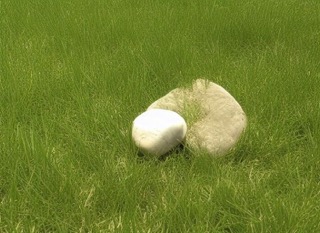} &\\

        {\raisebox{0.37in}{\multirow{1}{*}{\begin{tabular}{c}\textbf{PixelMan} \\ (8 steps, 5s)\end{tabular}}}} &
        \includegraphics[width=0.138\textwidth]{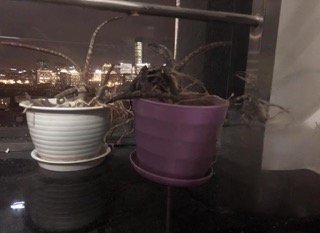} &
        \includegraphics[width=0.138\textwidth]{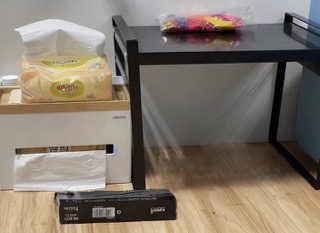} &
        \includegraphics[width=0.138\textwidth]{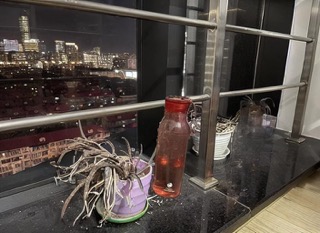} &
        \includegraphics[width=0.138\textwidth]{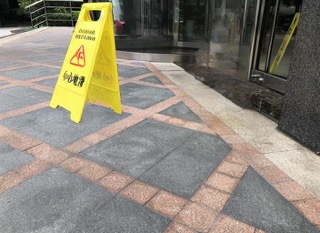} &
        \includegraphics[width=0.138\textwidth]{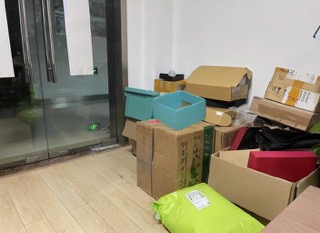} &
        \includegraphics[width=0.138\textwidth]{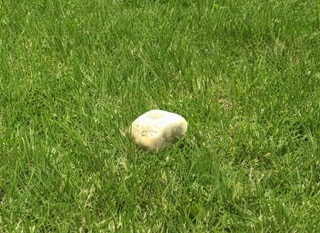} &\\

    \end{tabular}
    }
    \caption{
        \textbf{Additional qualitative comparison} on the ReS dataset at 8 steps. 
    }
    \label{fig:examples_8step_res}
\end{figure*}

\begin{figure*}[hbt!]
    \centering
    \setlength{\tabcolsep}{0.4pt}
    \renewcommand{\arraystretch}{0.4}
    {\footnotesize
    \begin{tabular}{c c c c c c c c}
        &
        \multicolumn{1}{c}{\begin{tabular}{c}(a)\\Object Enlarging\\\end{tabular}} &
        \multicolumn{1}{c}{\begin{tabular}{c}(b)\\Object Enlarging\\\end{tabular}} &
        \multicolumn{1}{c}{\begin{tabular}{c}(c)\\Object Shrinking\\(With Moving)\end{tabular}} &
        \multicolumn{1}{c}{\begin{tabular}{c}(d)\\Object Enlarging\\(With Moving)\end{tabular}} &
        \multicolumn{1}{c}{\begin{tabular}{c}(e)\\Object Pasting\\(With Shrinking)\end{tabular}} &
        \multicolumn{1}{c}{\begin{tabular}{c}(f)\\Object Pasting\\(With Enlarging)\end{tabular}} \\

        {\raisebox{0.48in}{
        \multirow{1}{*}{\begin{tabular}{c}Reference\\ Image\end{tabular}}}} &
        \includegraphics[width=0.138\textwidth]{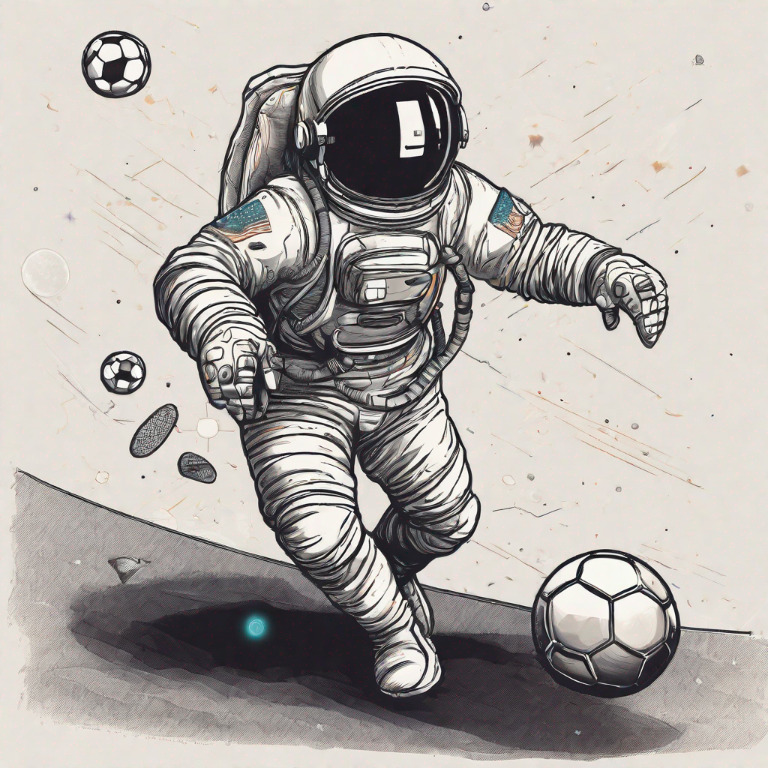} &
        \includegraphics[width=0.138\textwidth]{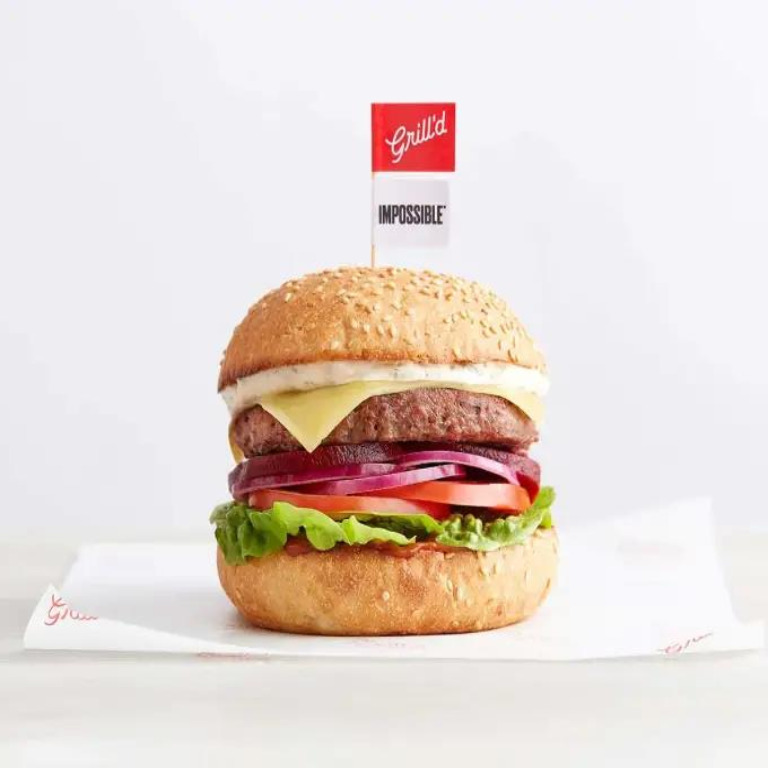} &
        \includegraphics[width=0.138\textwidth]
        {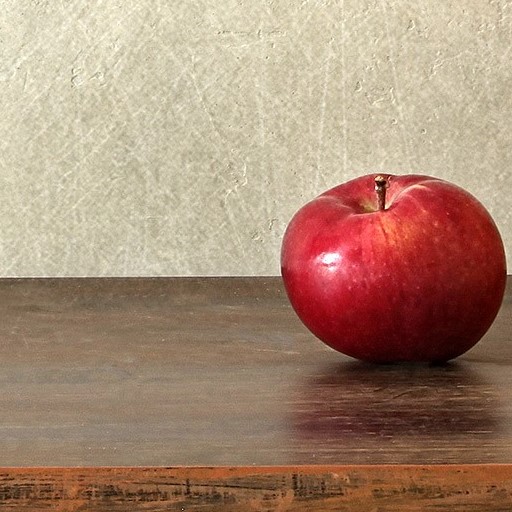} &
        \includegraphics[width=0.138\textwidth]{images/resize_and_paste/input/input_astronaut.jpg} &
        \includegraphics[width=0.138\textwidth]{images/resize_and_paste/input/reference_burger.jpg} &
        \includegraphics[width=0.138\textwidth]{images/resize_and_paste/input/reference_burger.jpg} &\\
        
        {\raisebox{0.44in}{
        \multirow{1}{*}{\begin{tabular}{c}Input\end{tabular}}}} &
        \includegraphics[width=0.138\textwidth]{images/resize_and_paste/input/input_astronaut.jpg} &
        \includegraphics[width=0.138\textwidth]{images/resize_and_paste/input/reference_burger.jpg} &
        \includegraphics[width=0.138\textwidth]{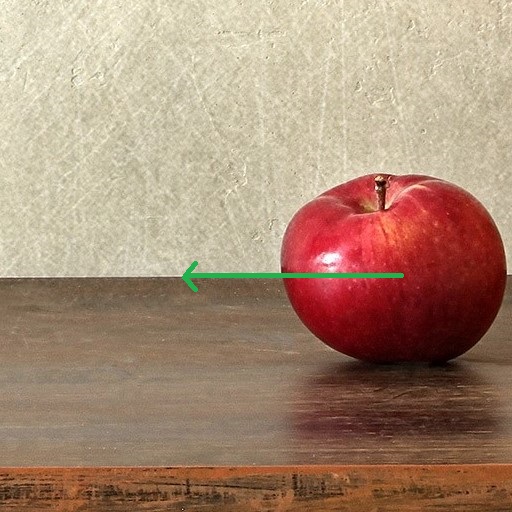} &
        \includegraphics[width=0.138\textwidth]{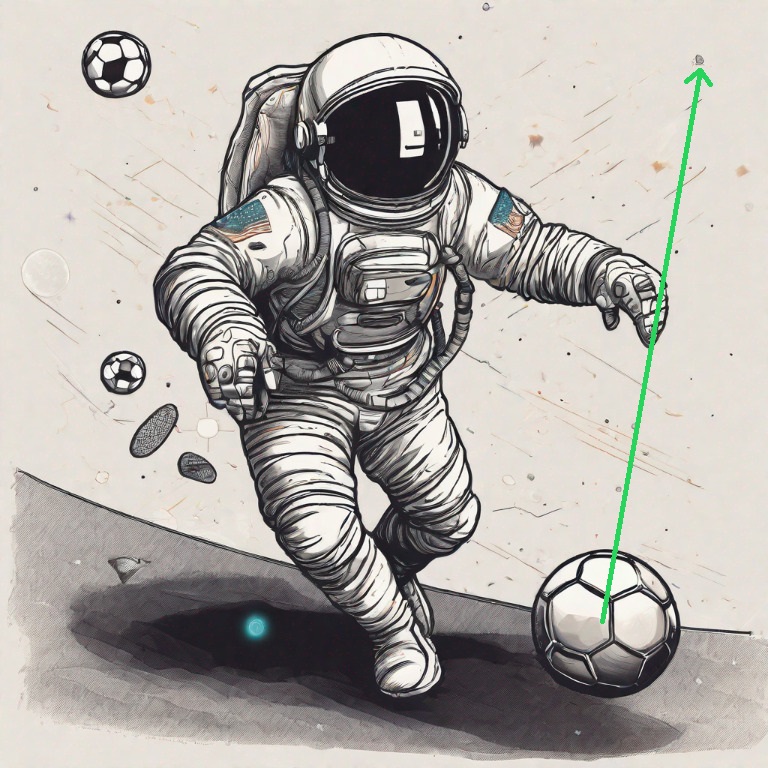} &
        \includegraphics[width=0.138\textwidth]{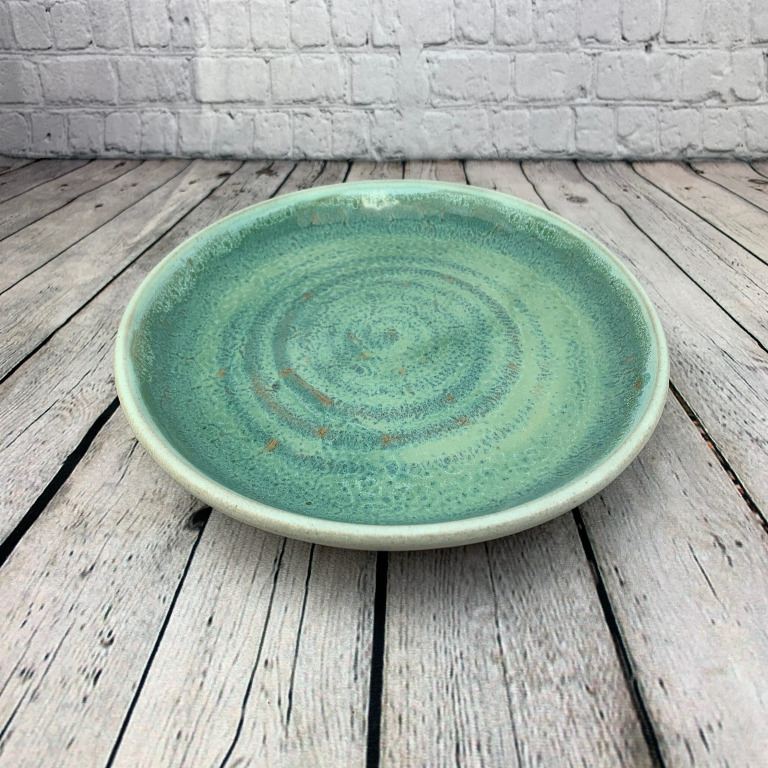} &
        \includegraphics[width=0.138\textwidth]{images/resize_and_paste/input/input_burger.jpg} &\\

        {\raisebox{0.56in}{\multirow{1}{*}{\begin{tabular}{c}DragonDiffusion\\ (160 NFEs)  \\ (50 steps, 23s)\end{tabular}}}} &
        \includegraphics[width=0.138\textwidth]{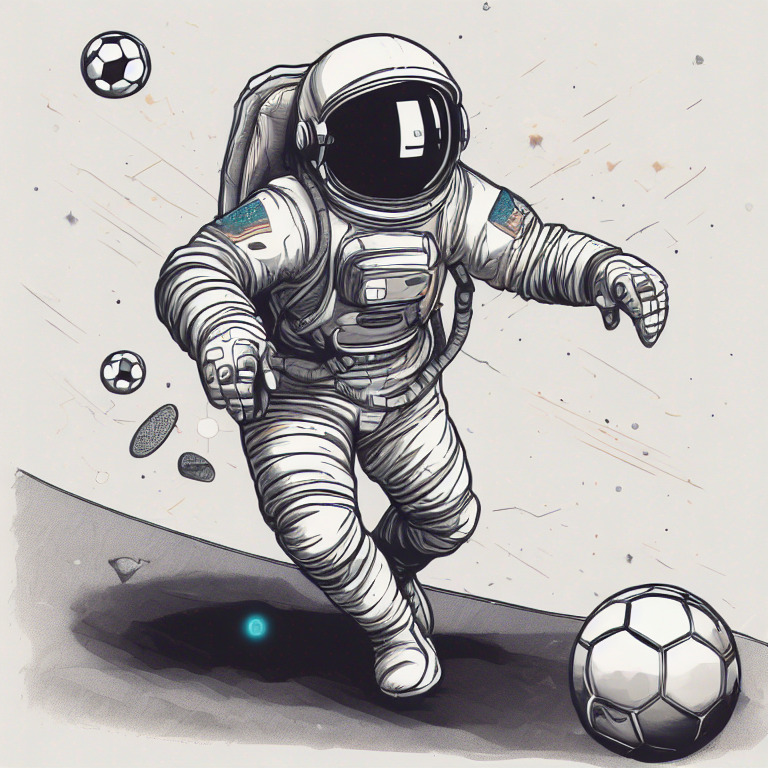} &
        \includegraphics[width=0.138\textwidth]{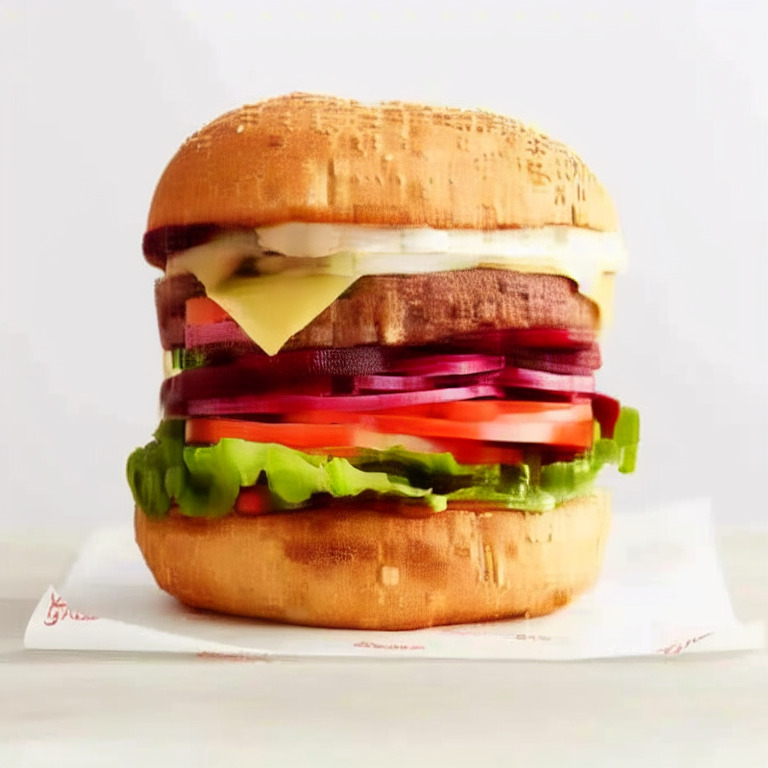} &
        \includegraphics[width=0.138\textwidth]{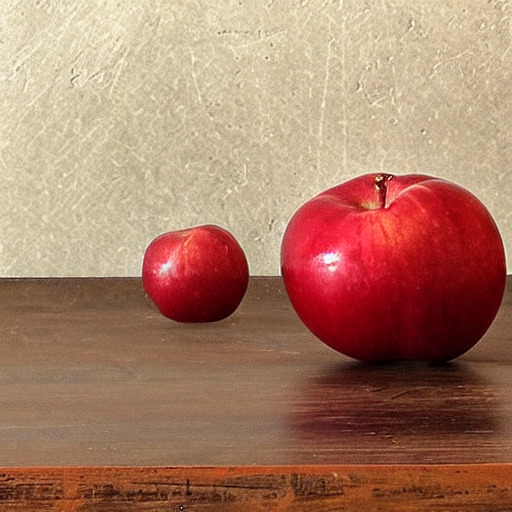} &
        \includegraphics[width=0.138\textwidth]{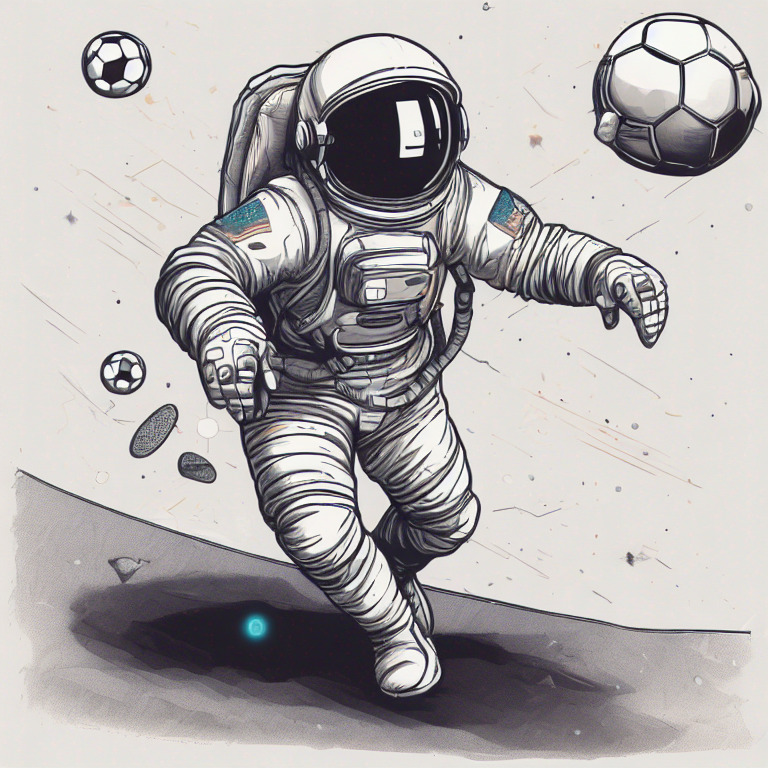} &
        \includegraphics[width=0.138\textwidth]{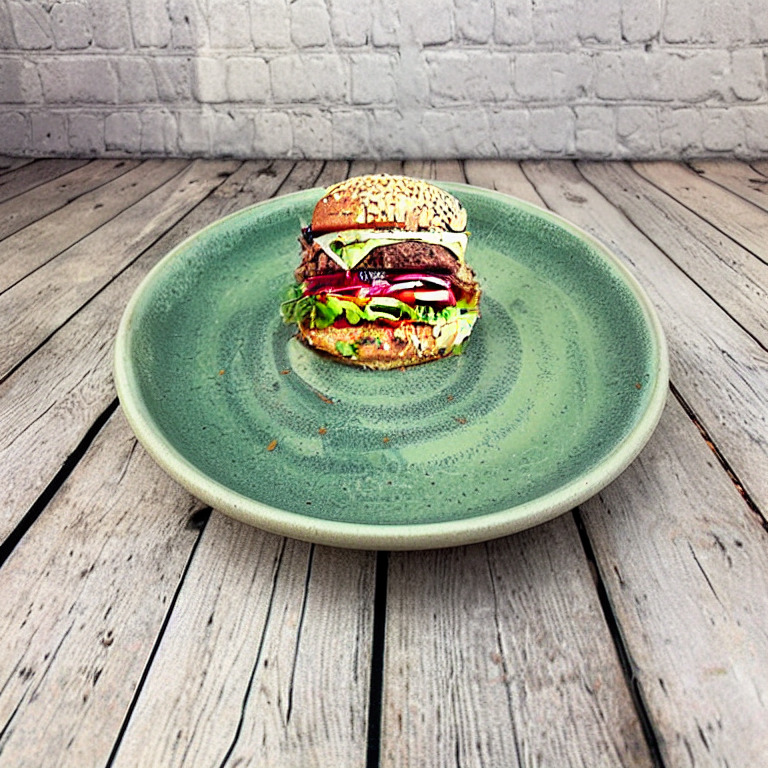} &
        \includegraphics[width=0.138\textwidth]{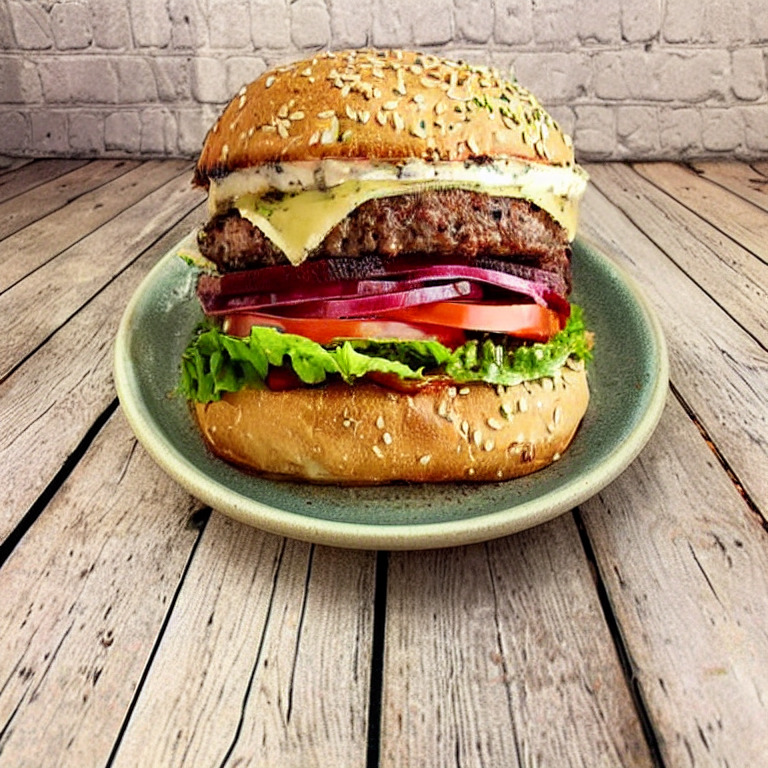} &\\

        {\raisebox{0.56in}{\multirow{1}{*}{\begin{tabular}{c}DiffEditor\\ (176 NFEs) \\ (50 steps, 24s)\end{tabular}}}} &
        \includegraphics[width=0.138\textwidth]{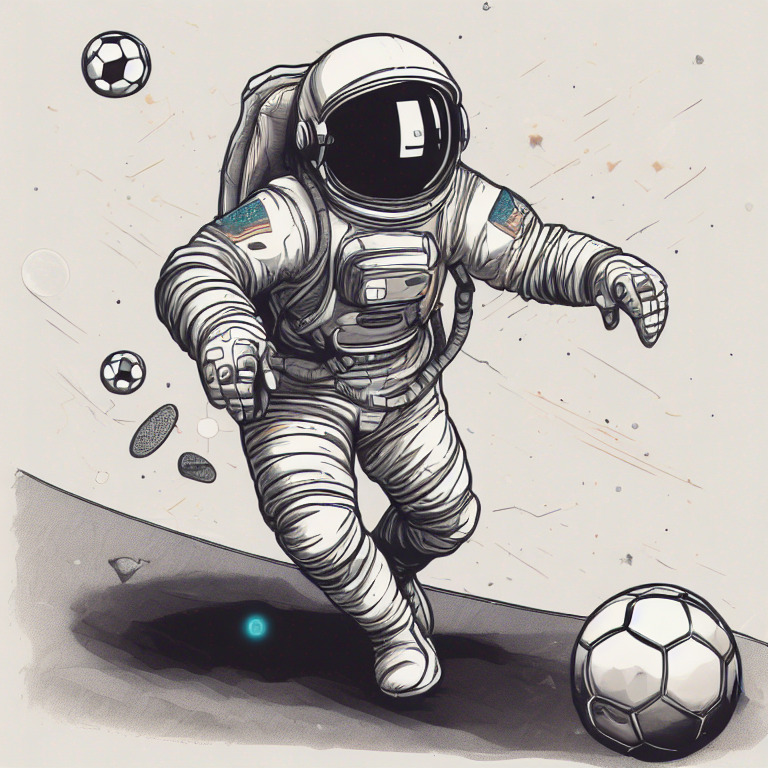} &
        \includegraphics[width=0.138\textwidth]{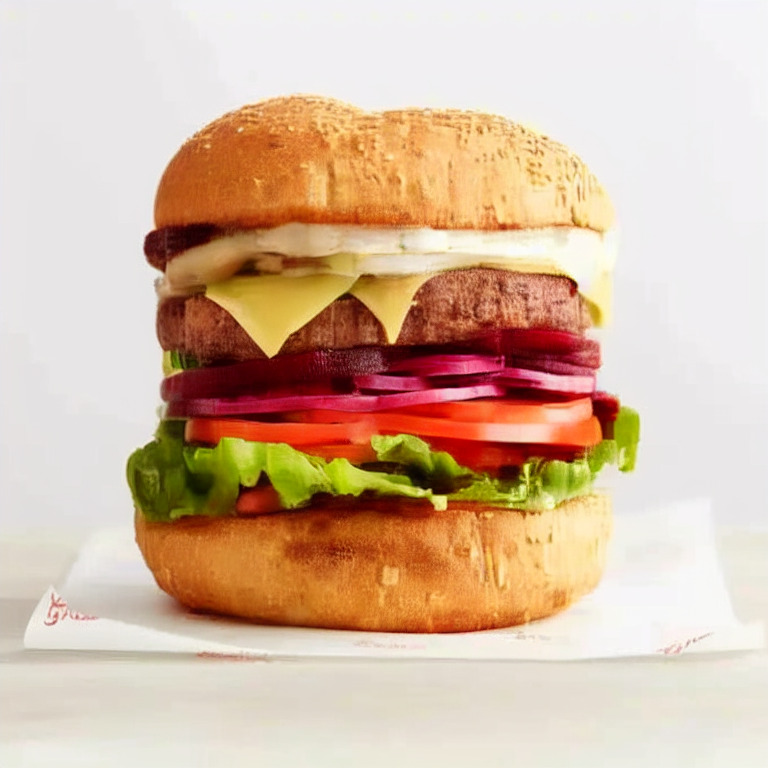} &
        \includegraphics[width=0.138\textwidth]{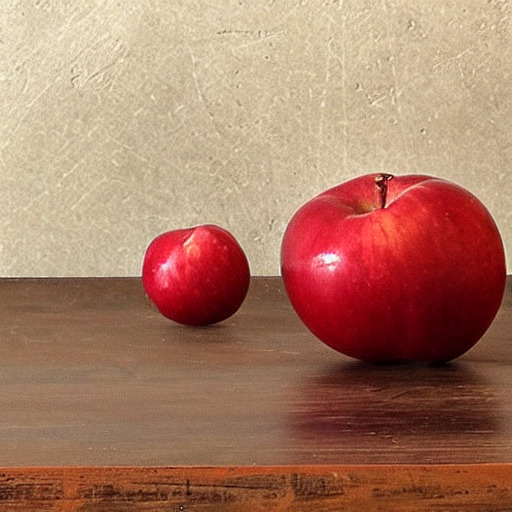} &
        \includegraphics[width=0.138\textwidth]{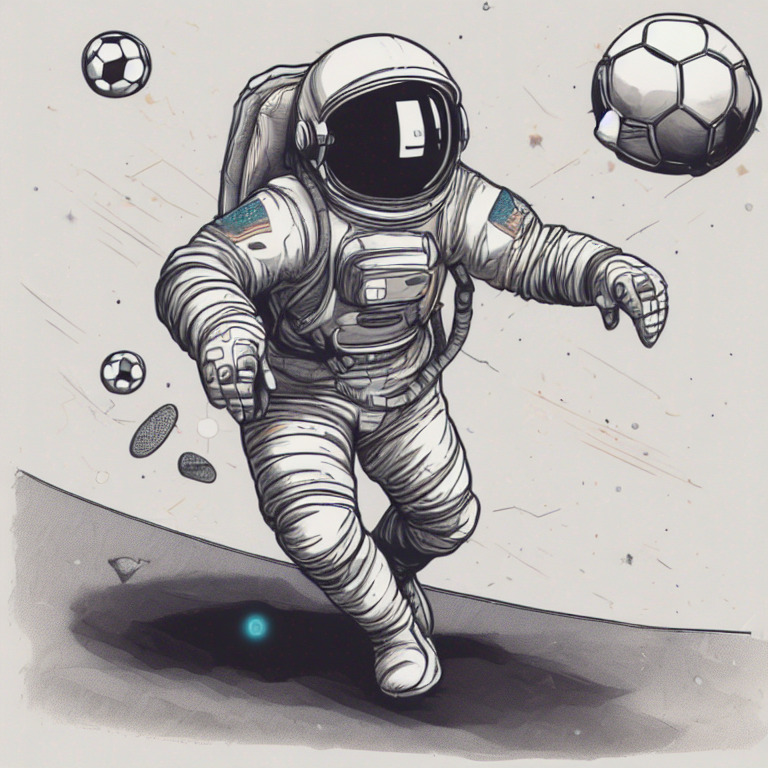} &
        \includegraphics[width=0.138\textwidth]{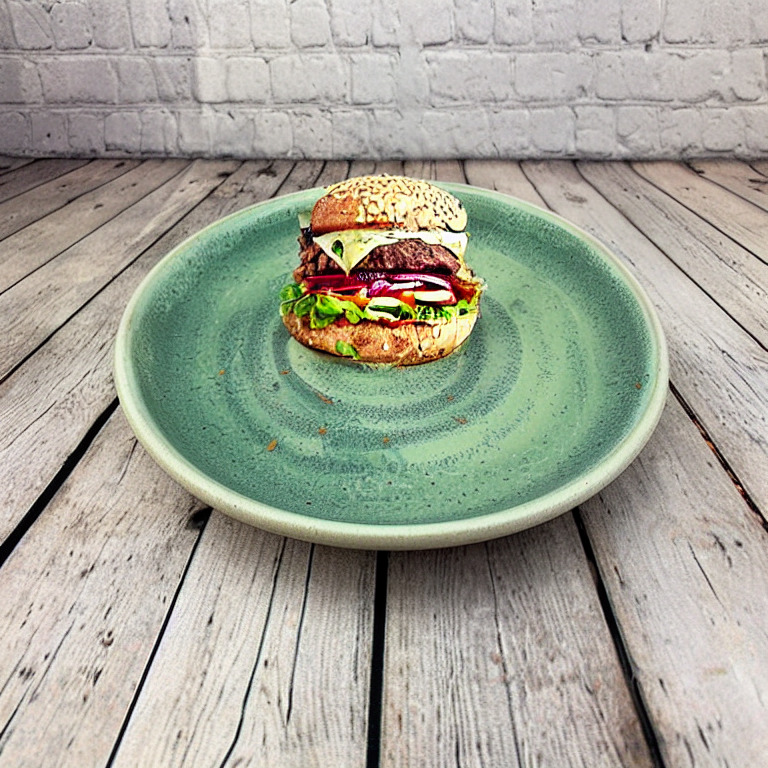} &
        \includegraphics[width=0.138\textwidth]{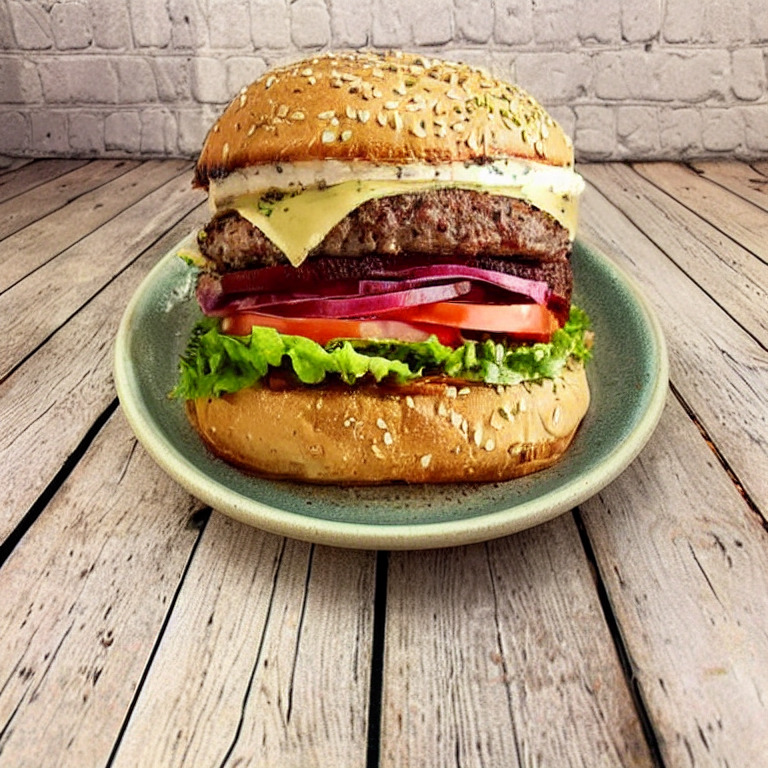} &\\
        
        {\raisebox{0.56in}{\multirow{1}{*}{\begin{tabular}{c}\textbf{PixelMan}\\ (64 NFEs) \\ (16 steps, 9s)\end{tabular}}}} &
        \includegraphics[width=0.138\textwidth]{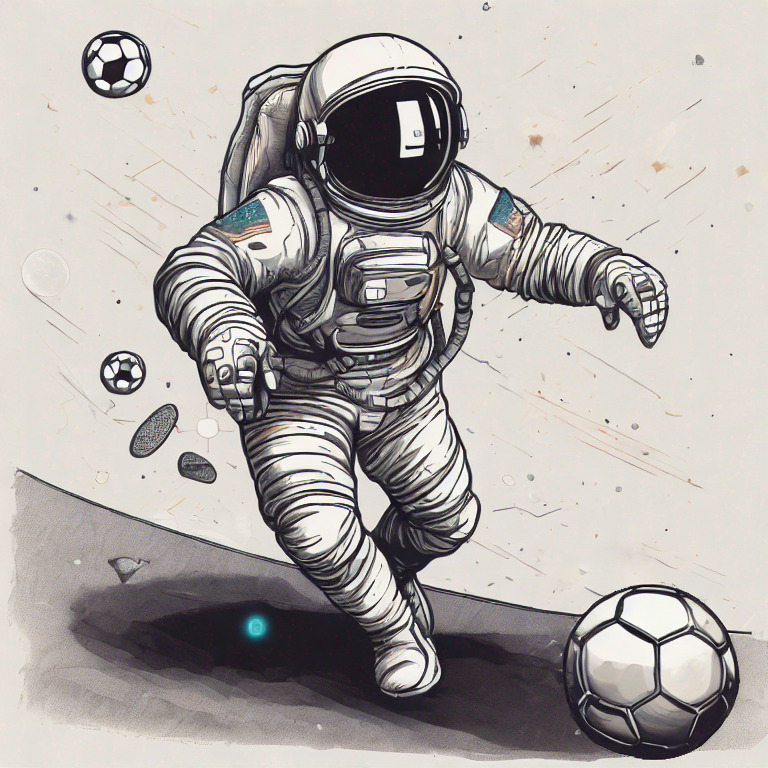} &
        \includegraphics[width=0.138\textwidth]{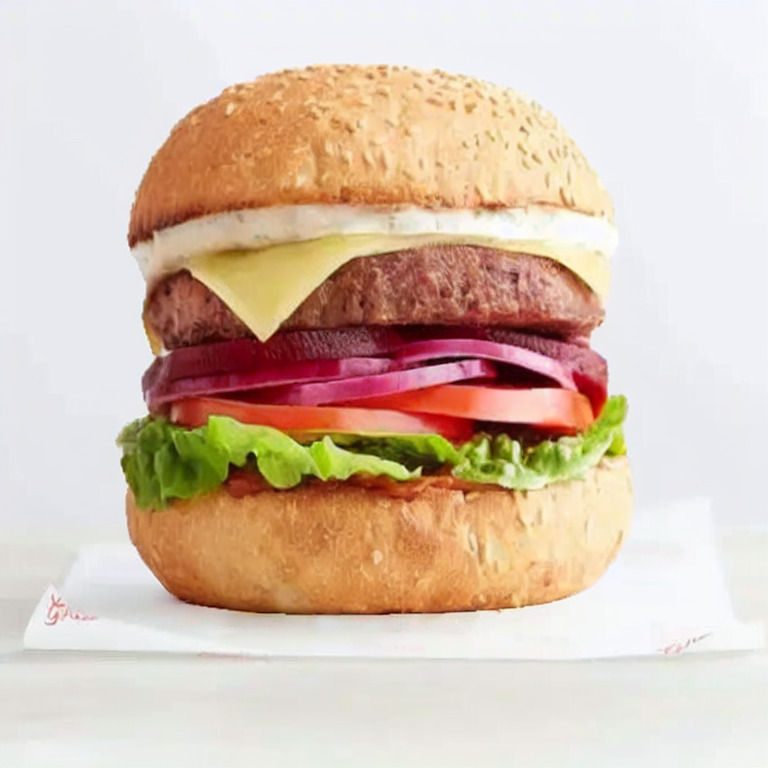} &
        \includegraphics[width=0.138\textwidth]{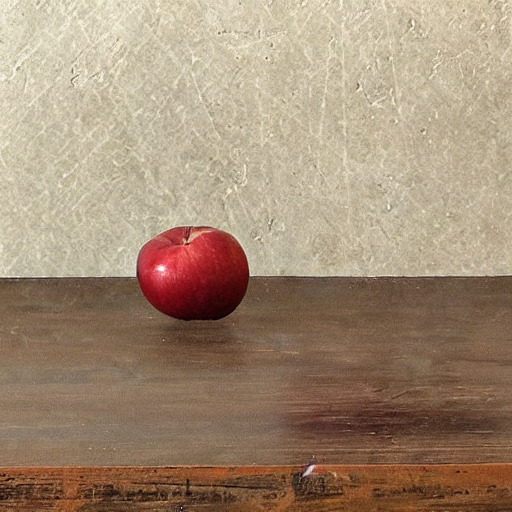} &
        \includegraphics[width=0.138\textwidth]{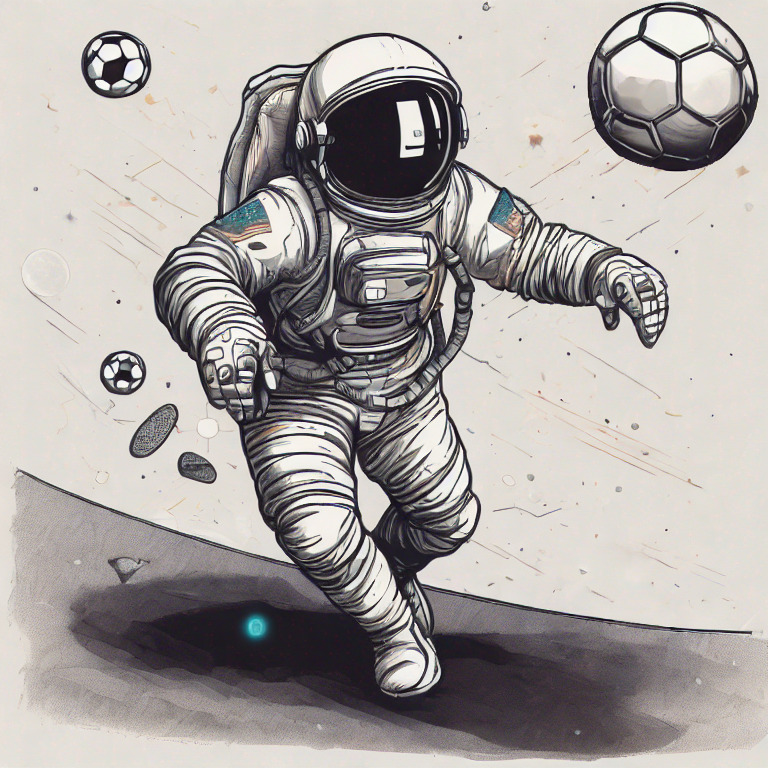} &
        \includegraphics[width=0.138\textwidth]{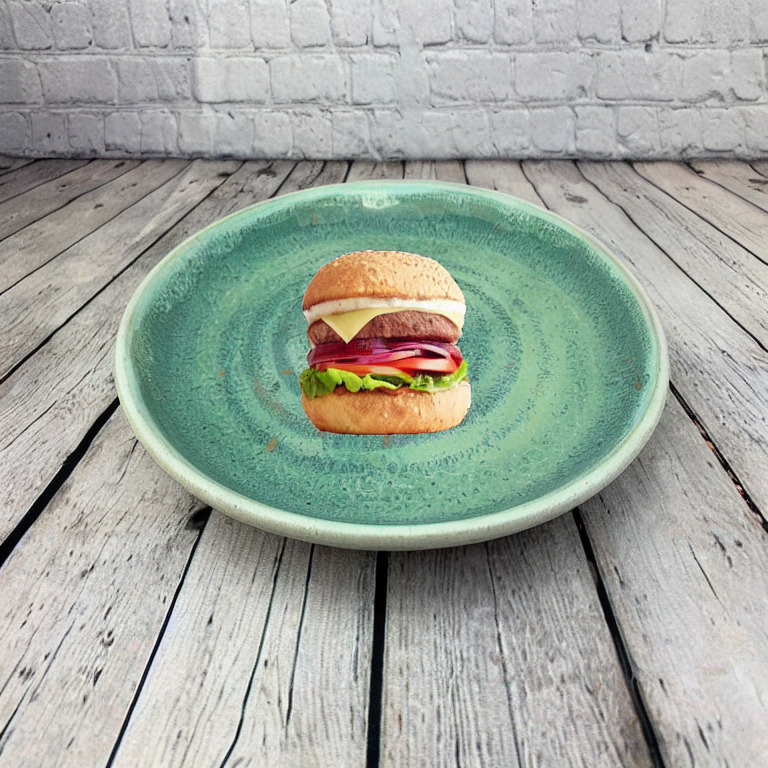} &
        \includegraphics[width=0.138\textwidth]{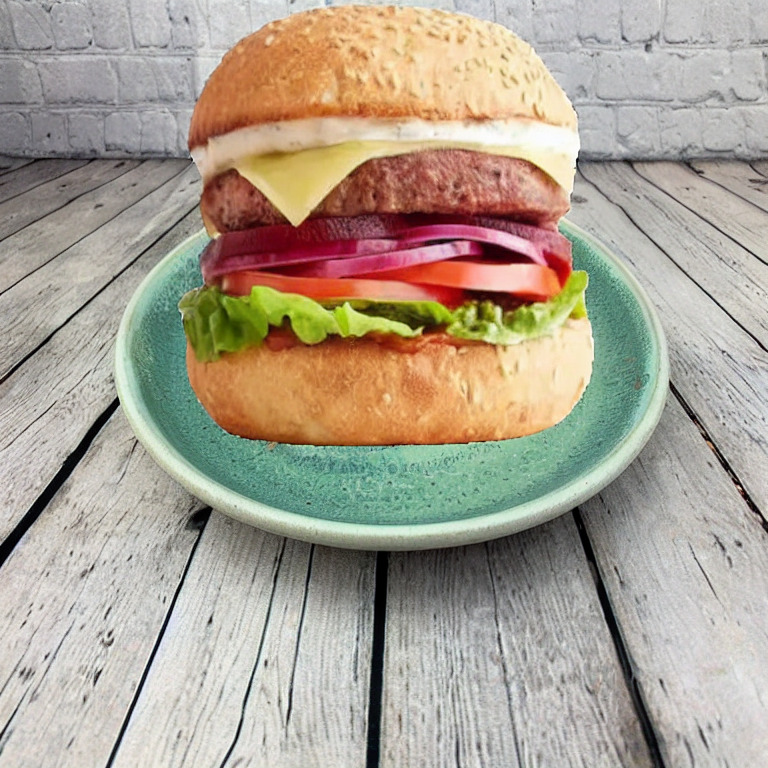} &\\

    \end{tabular}
    }
    \caption{
        \textbf{Qualitative examples on other consistent object editing tasks} including object resizing, and object pasting.
    }
    \label{fig:examples_other_tasks}
\end{figure*}

\begin{figure*}[hbt!]
    \centering
    \setlength{\tabcolsep}{0.4pt}
    \renewcommand{\arraystretch}{0.4}
    {\footnotesize
    \begin{tabular}{c c c c c c c c}
        &
        \multicolumn{1}{c}{(a)} &
        \multicolumn{1}{c}{(b)} &
        \multicolumn{1}{c}{(c)} &
        \multicolumn{1}{c}{(d)} &
        \multicolumn{1}{c}{(e)} &
        \multicolumn{1}{c}{(f)} \\

        {\raisebox{0.34in}{
        \multirow{1}{*}{\rotatebox{0}{Input}}}} &
        \includegraphics[width=0.105\textwidth]{images/comparison/COCOEE/000000485981_GT_source.jpg} &
        \includegraphics[width=0.105\textwidth]{images/comparison/COCOEE/000000595445_GT_source.jpg} &
        \includegraphics[width=0.105\textwidth]{images/comparison/COCOEE/000000111930_GT_source.jpg} &
        \includegraphics[width=0.105\textwidth]{images/comparison/COCOEE/000000327431_GT_source.jpg} &
        \includegraphics[width=0.105\textwidth]{images/comparison/COCOEE/000000309203_GT_source.jpg} &
        \includegraphics[width=0.105\textwidth]{images/comparison/COCOEE/000001557820_GT_source.jpg} &\\

        {\raisebox{0.37in}{\multirow{1}{*}{\begin{tabular}{c}PAIR Diffusion \\ (50 steps, 32s)\end{tabular}}}} &
        \includegraphics[width=0.105\textwidth]{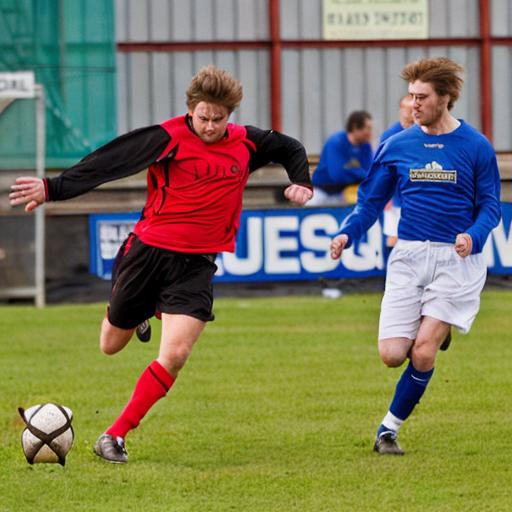} &
        \includegraphics[width=0.105\textwidth]{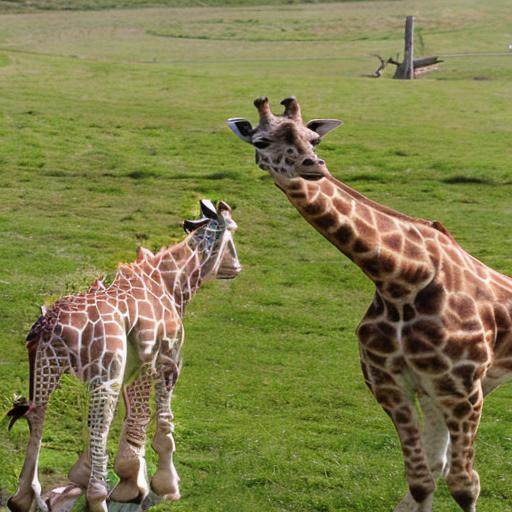} &
        \includegraphics[width=0.105\textwidth]{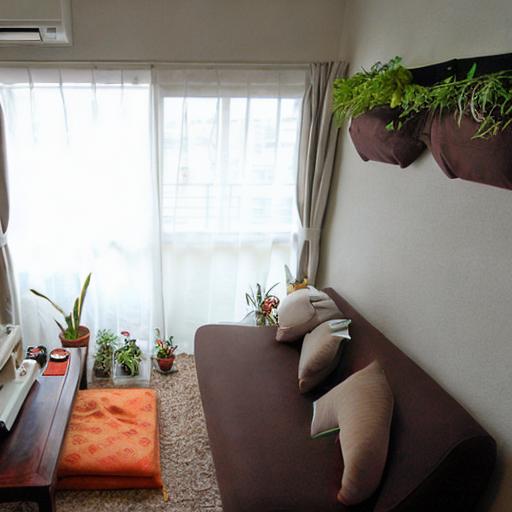} &
        \includegraphics[width=0.105\textwidth]{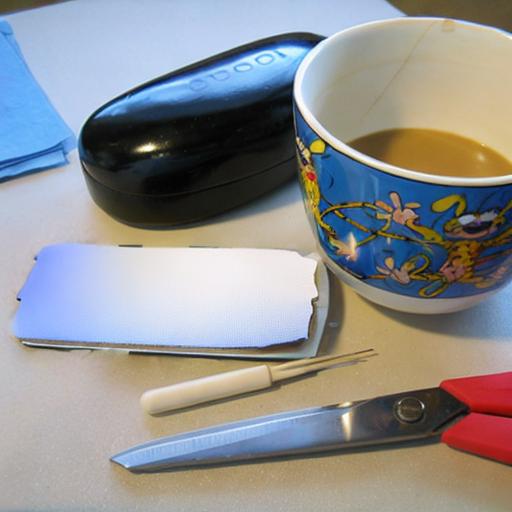} &
        \includegraphics[width=0.105\textwidth]{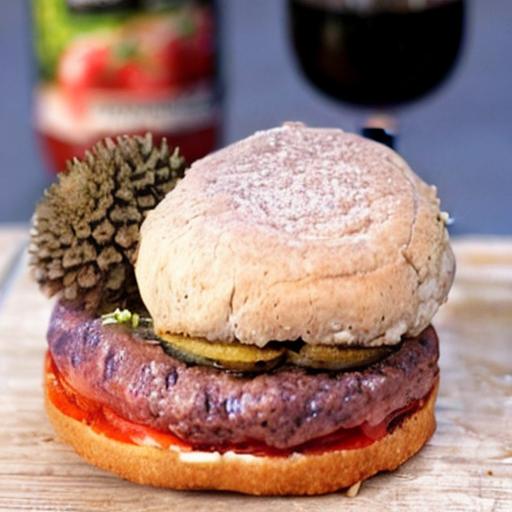} &
        \includegraphics[width=0.105\textwidth]{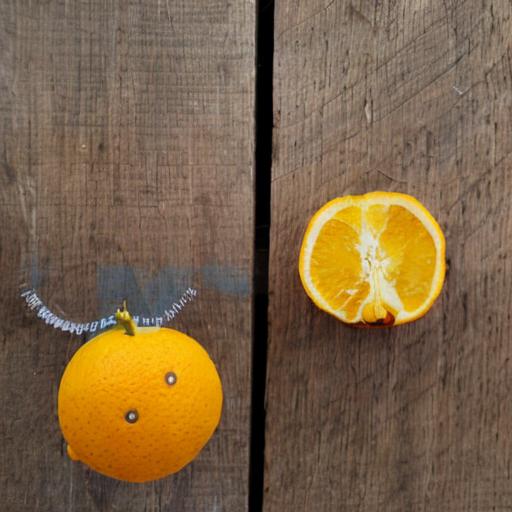} &\\

        {\raisebox{0.37in}{\multirow{1}{*}{\begin{tabular}{c}InfEdit \\ (50 steps, 35s)\end{tabular}}}} &
        \includegraphics[width=0.105\textwidth]{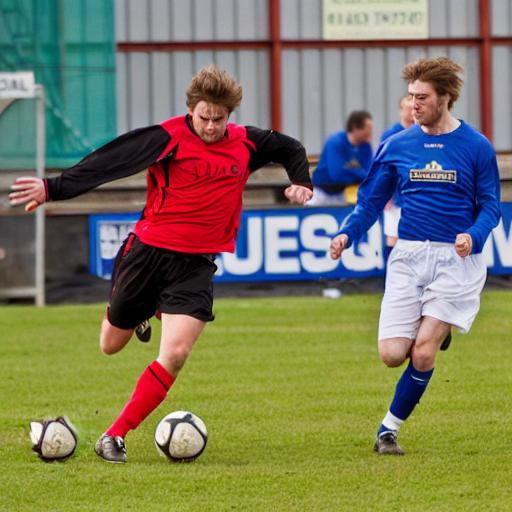} &
        \includegraphics[width=0.105\textwidth]{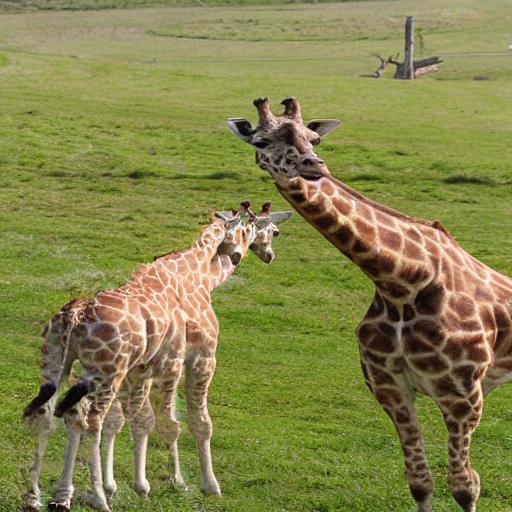} &
        \includegraphics[width=0.105\textwidth]{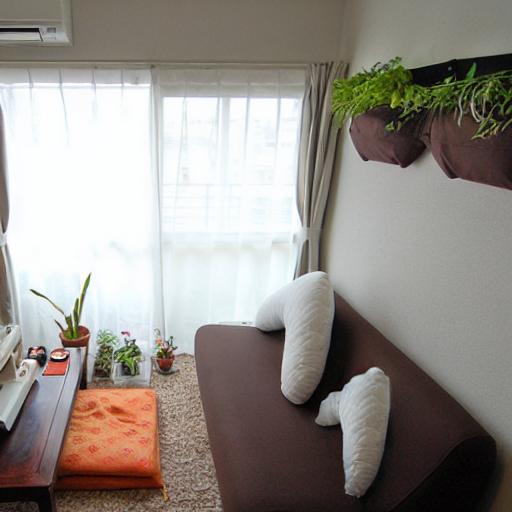} &
        \includegraphics[width=0.105\textwidth]{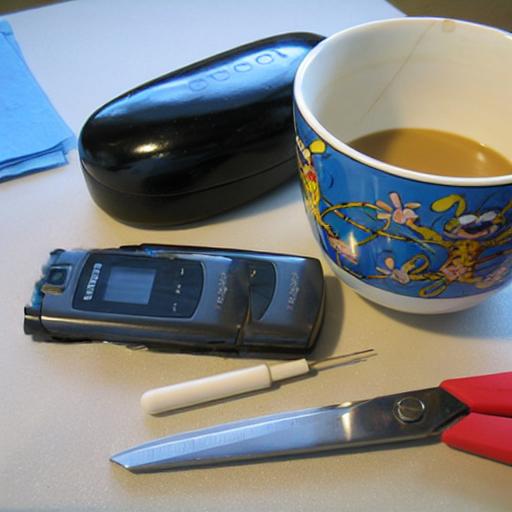} &
        \includegraphics[width=0.105\textwidth]{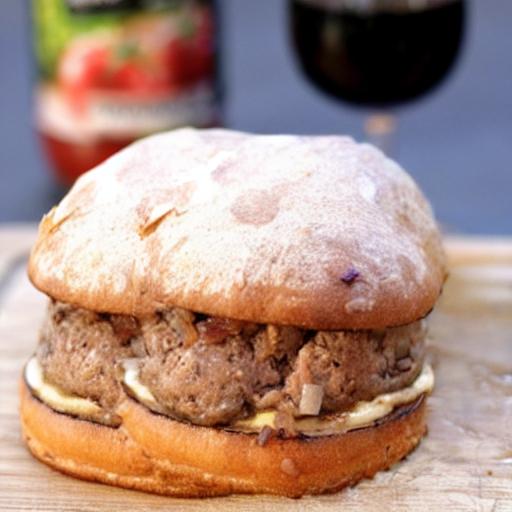} &
        \includegraphics[width=0.105\textwidth]{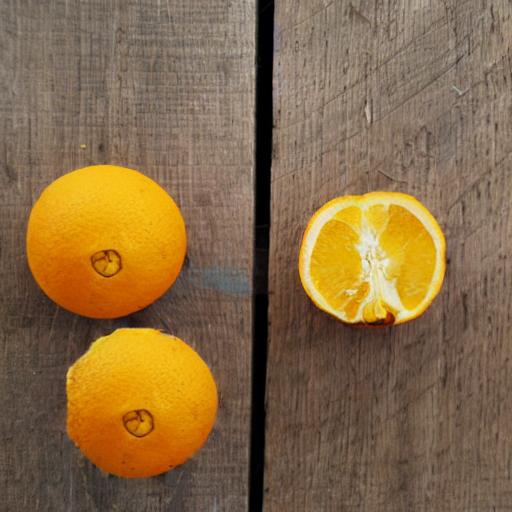} &\\

        {\raisebox{0.37in}{\multirow{1}{*}{\begin{tabular}{c}\textbf{PixelMan} \\ (50 steps, 27s)\end{tabular}}}} &
        \includegraphics[width=0.105\textwidth]{images/comparison/COCOEE/000000485981_GT_ours50.jpg} &
        \includegraphics[width=0.105\textwidth]{images/comparison/COCOEE/000000595445_GT_ours50.jpg} &
        \includegraphics[width=0.105\textwidth]{images/comparison/COCOEE/000000111930_GT_ours50.jpg} &
        \includegraphics[width=0.105\textwidth]{images/comparison/COCOEE/000000327431_GT_ours50.jpg} &
        \includegraphics[width=0.105\textwidth]{images/comparison/COCOEE/000000309203_GT_ours50.jpg} &
        \includegraphics[width=0.105\textwidth]{images/comparison/COCOEE/000001557820_GT_ours50.jpg} &\\
        
        {\raisebox{0.37in}{\multirow{1}{*}{\begin{tabular}{c}PAIR Diffusion \\ (16 steps, 12s)\end{tabular}}}} &
        \includegraphics[width=0.105\textwidth]{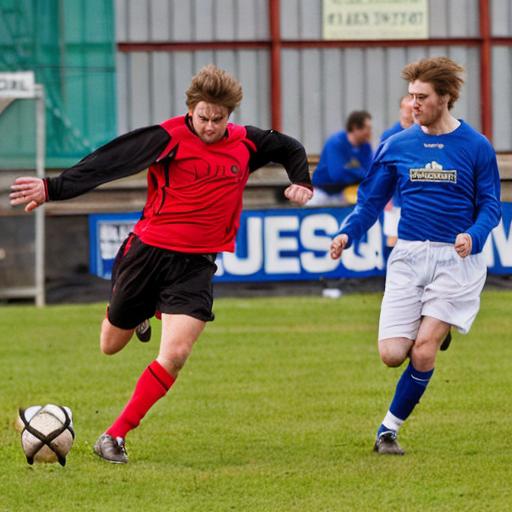} &
        \includegraphics[width=0.105\textwidth]{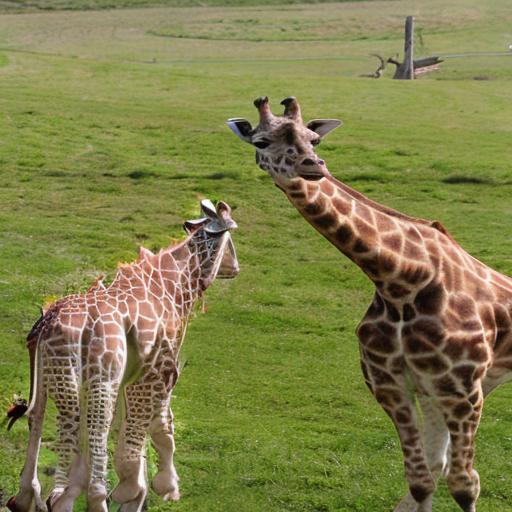} &
        \includegraphics[width=0.105\textwidth]{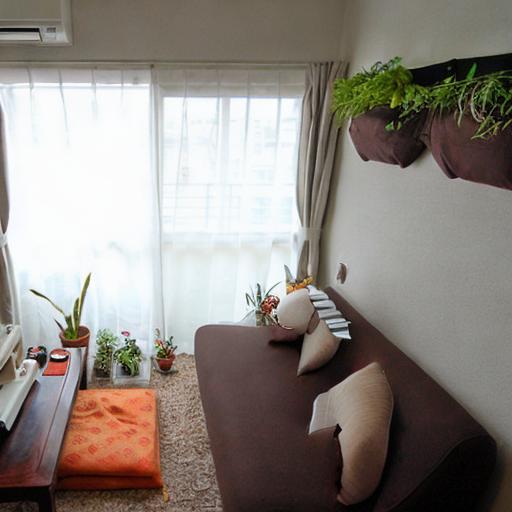} &
        \includegraphics[width=0.105\textwidth]{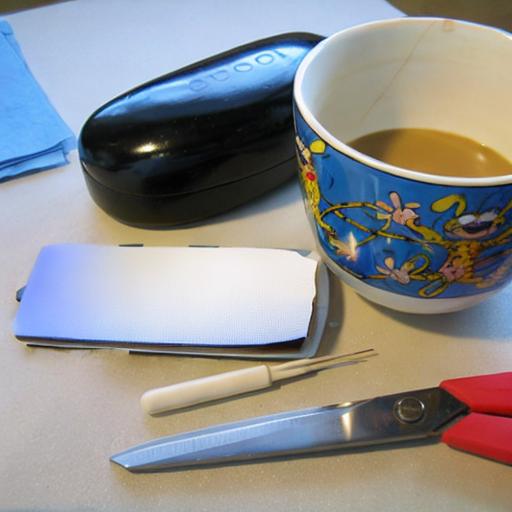} &
        \includegraphics[width=0.105\textwidth]{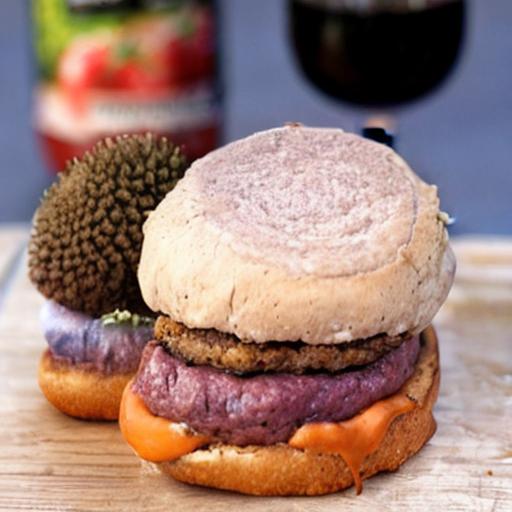} &
        \includegraphics[width=0.105\textwidth]{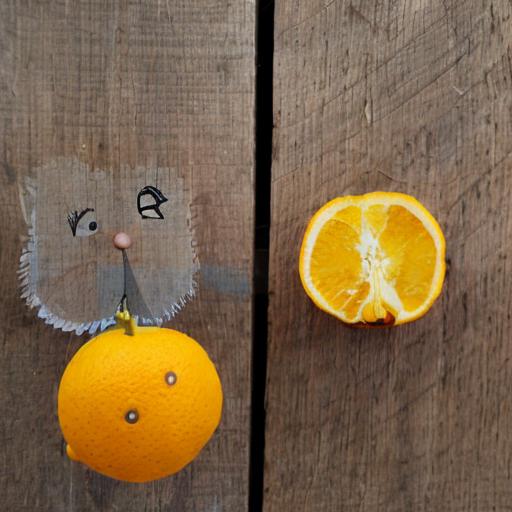} &\\

        {\raisebox{0.37in}{\multirow{1}{*}{\begin{tabular}{c}InfEdit \\ (16 steps, 1s)\end{tabular}}}} &
        \includegraphics[width=0.105\textwidth]{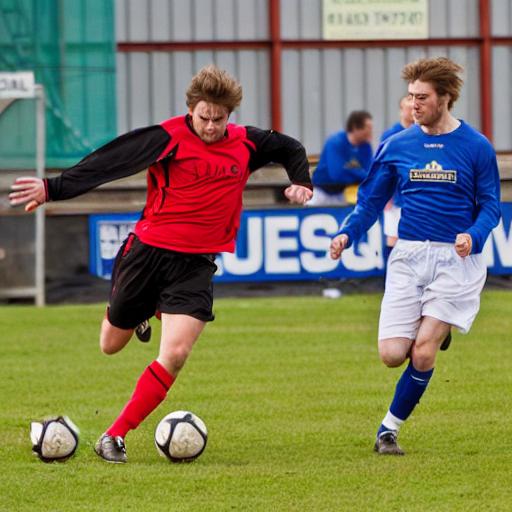} &
        \includegraphics[width=0.105\textwidth]{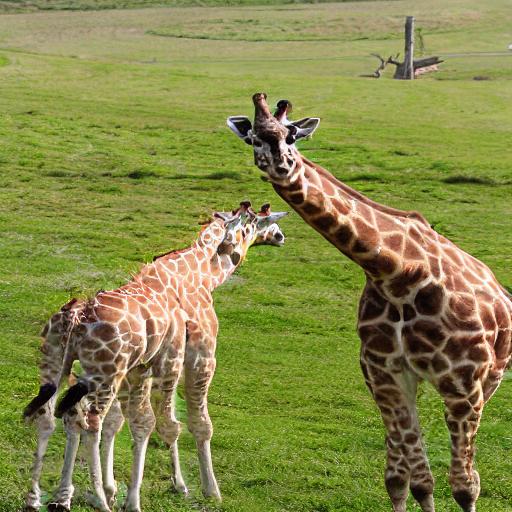} &
        \includegraphics[width=0.105\textwidth]{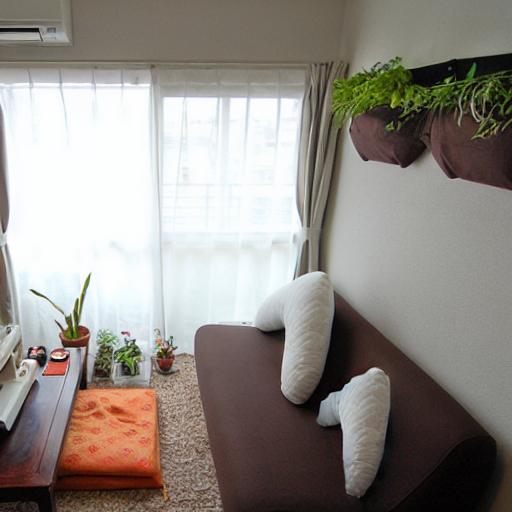} &
        \includegraphics[width=0.105\textwidth]{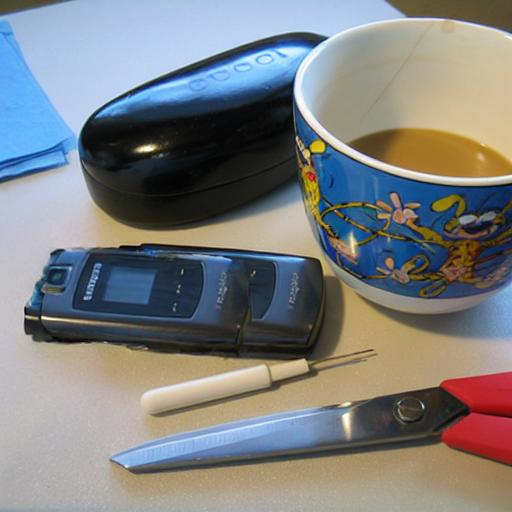} &
        \includegraphics[width=0.105\textwidth]{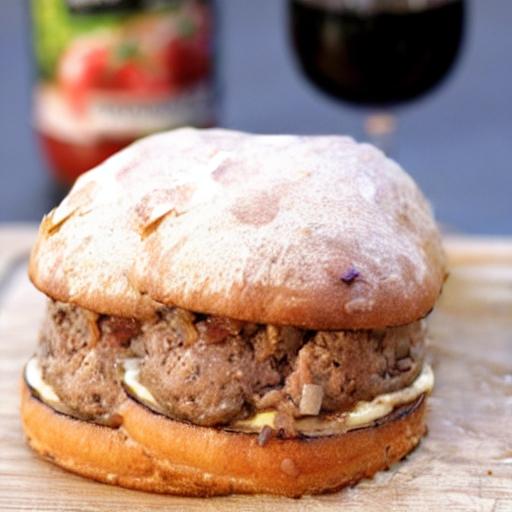} &
        \includegraphics[width=0.105\textwidth]{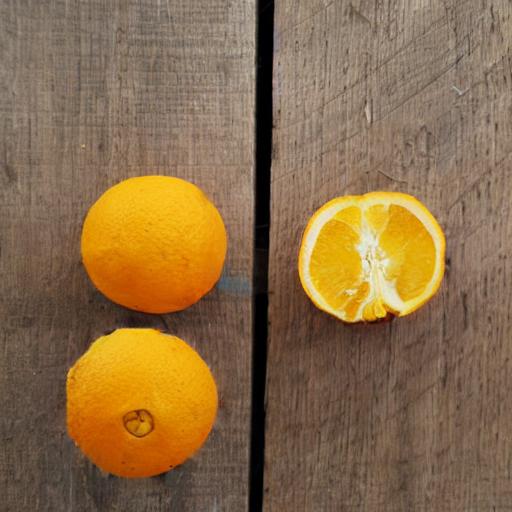} &\\

        {\raisebox{0.37in}{\multirow{1}{*}{\begin{tabular}{c}\textbf{PixelMan} \\ (16 steps, 9s)\end{tabular}}}} &
        \includegraphics[width=0.105\textwidth]{images/comparison/COCOEE/000000485981_GT_ours16.jpg} &
        \includegraphics[width=0.105\textwidth]{images/comparison/COCOEE/000000595445_GT_ours16.jpg} &
        \includegraphics[width=0.105\textwidth]{images/comparison/COCOEE/000000111930_GT_ours16.jpg} &
        \includegraphics[width=0.105\textwidth]{images/comparison/COCOEE/000000327431_GT_ours16.jpg} &
        \includegraphics[width=0.105\textwidth]{images/comparison/COCOEE/000000309203_GT_ours16.jpg} &
        \includegraphics[width=0.105\textwidth]{images/comparison/COCOEE/000001557820_GT_ours16.jpg} &\\
        
        {\raisebox{0.37in}{\multirow{1}{*}{\begin{tabular}{c}PAIR Diffusion \\ (8 steps, 7s)\end{tabular}}}} &
        \includegraphics[width=0.105\textwidth]{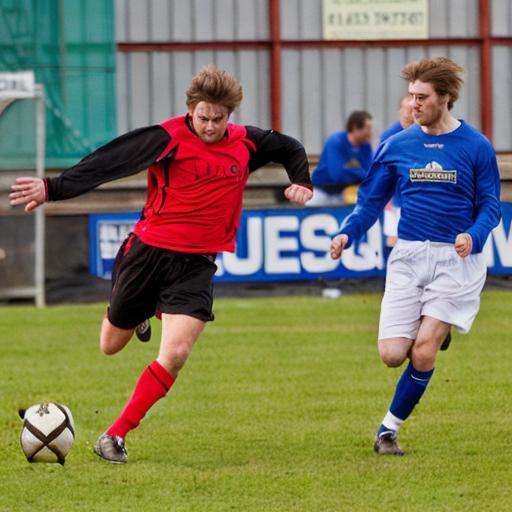} &
        \includegraphics[width=0.105\textwidth]{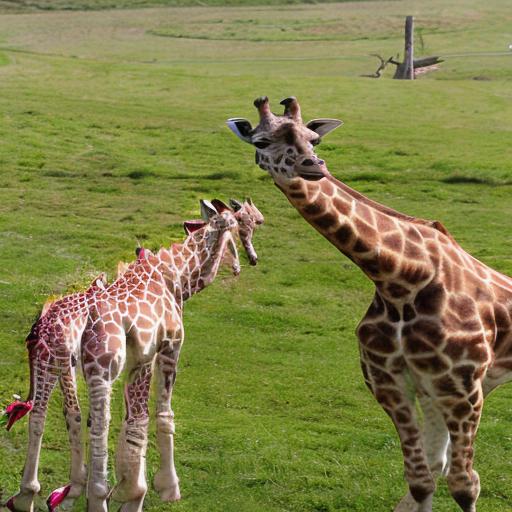} &
        \includegraphics[width=0.105\textwidth]{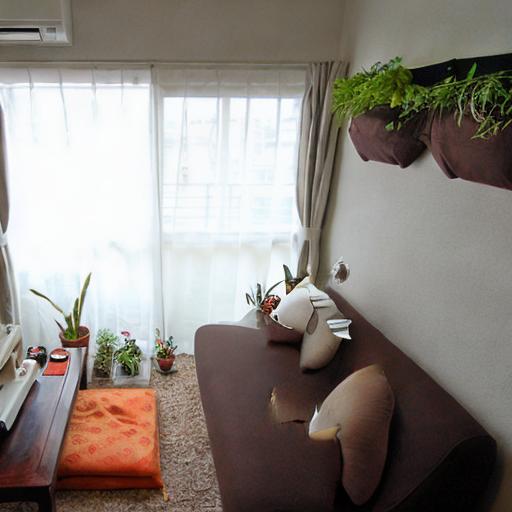} &
        \includegraphics[width=0.105\textwidth]{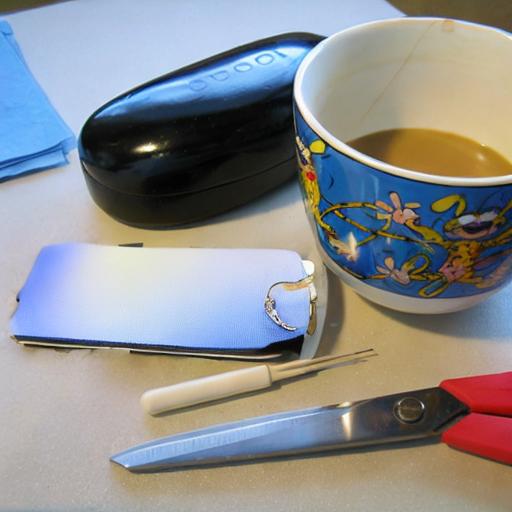} &
        \includegraphics[width=0.105\textwidth]{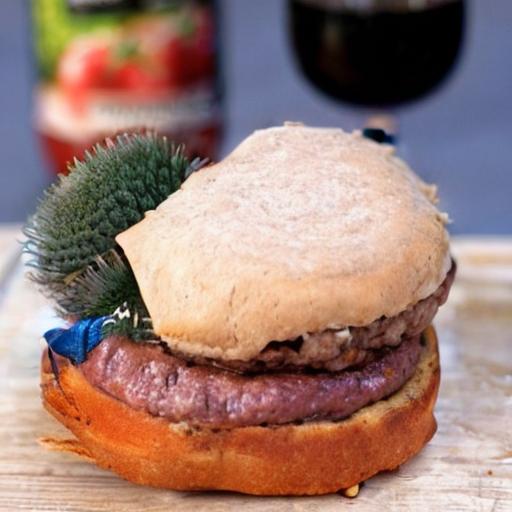} &
        \includegraphics[width=0.105\textwidth]{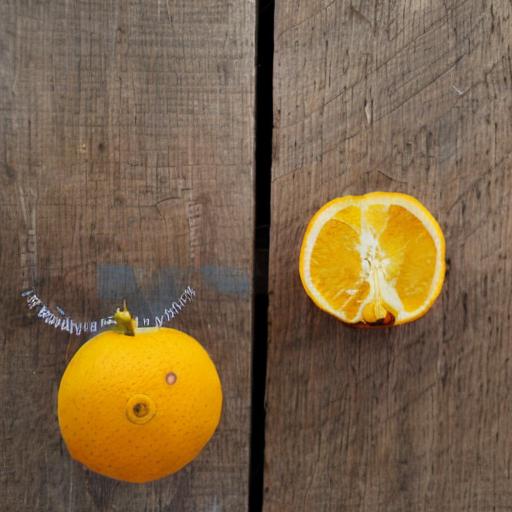} &\\
        
        {\raisebox{0.37in}{\multirow{1}{*}{\begin{tabular}{c}InfEdit \\ (8 steps, 5s)\end{tabular}}}} &
        \includegraphics[width=0.105\textwidth]{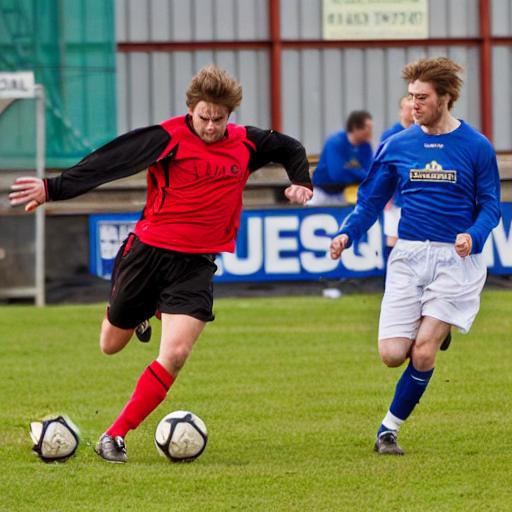} &
        \includegraphics[width=0.105\textwidth]{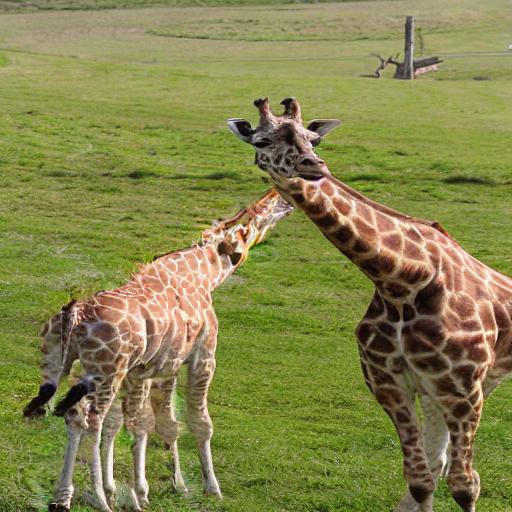} &
        \includegraphics[width=0.105\textwidth]{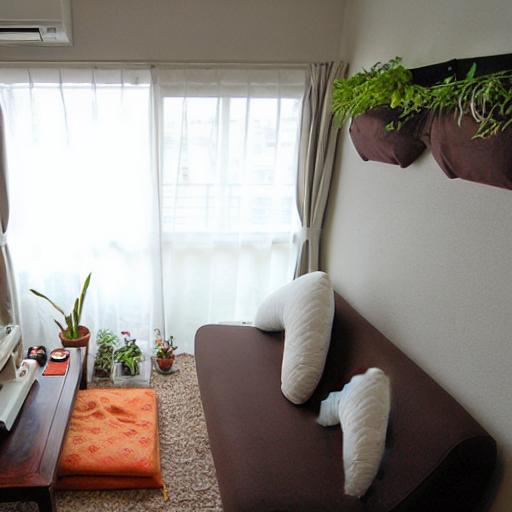} &
        \includegraphics[width=0.105\textwidth]{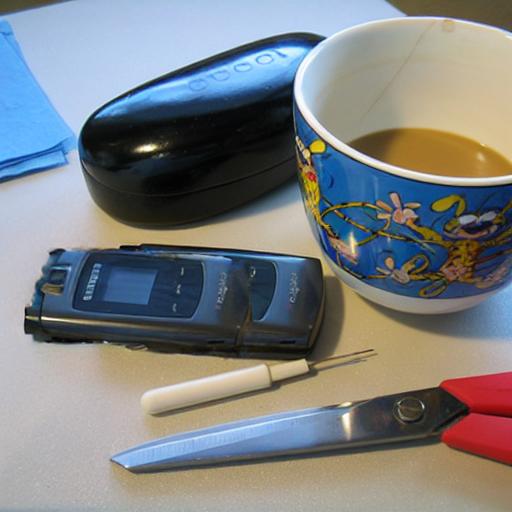} &
        \includegraphics[width=0.105\textwidth]{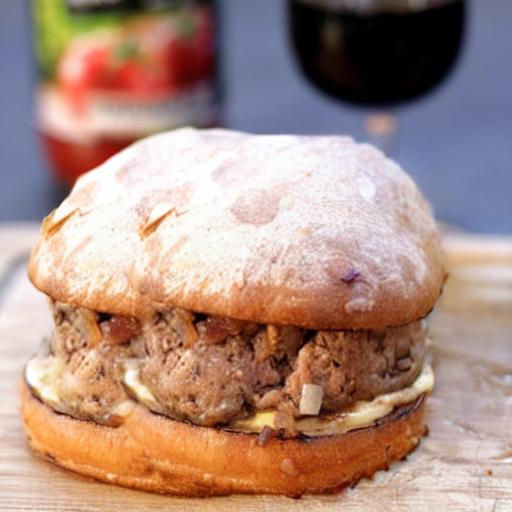} &
        \includegraphics[width=0.105\textwidth]{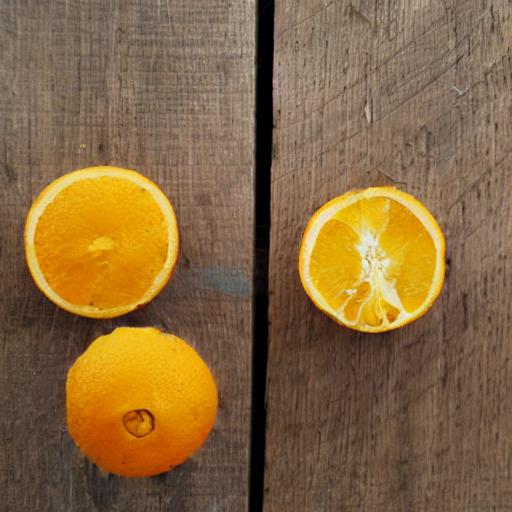} &\\
        
        {\raisebox{0.37in}{\multirow{1}{*}{\begin{tabular}{c}\textbf{PixelMan} \\ (8 steps, 4s)\end{tabular}}}} &
        \includegraphics[width=0.105\textwidth]{images/comparison/COCOEE/000000485981_GT_ours8.jpg} &
        \includegraphics[width=0.105\textwidth]{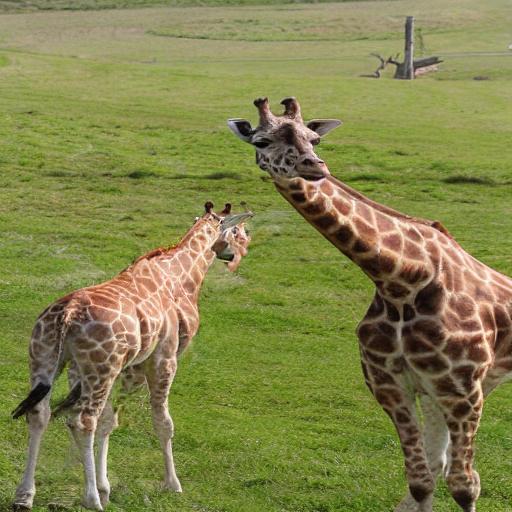} &
        \includegraphics[width=0.105\textwidth]{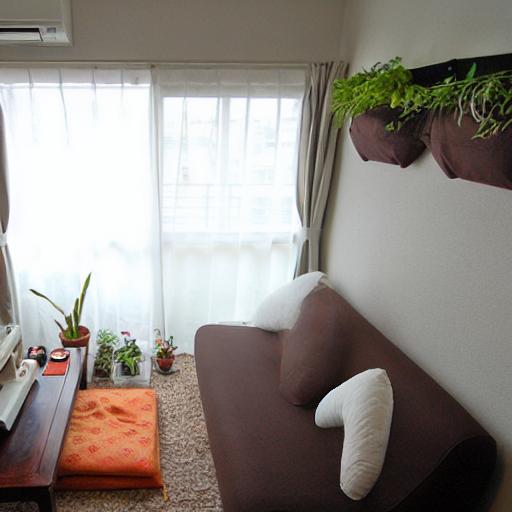} &
        \includegraphics[width=0.105\textwidth]{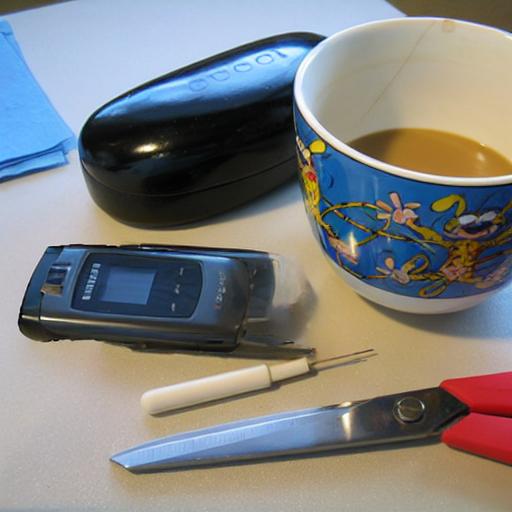} &
        \includegraphics[width=0.105\textwidth]{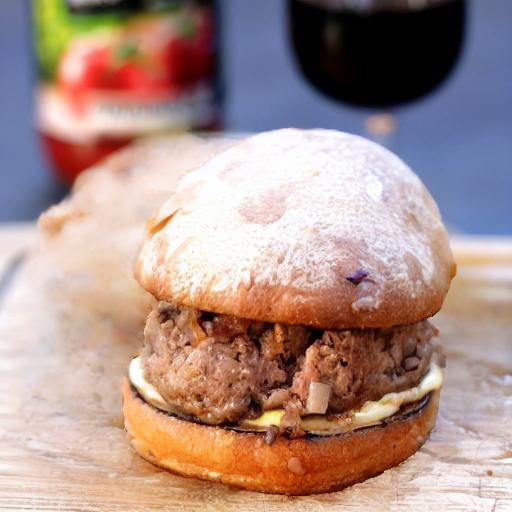} &
        \includegraphics[width=0.105\textwidth]{images/comparison/COCOEE/000001557820_GT_ours8.jpg} &\\

    \end{tabular}
    }
    \caption{
        \textbf{Additional qualitative comparisons} to PAIR Diffusion~\cite{goel2023pair} and InfEdit~\cite{xu2023infedit} on the COCOEE dataset at 50, 16 and 8 steps. 
    }
    \label{fig:examples_additional_cocoee}
\end{figure*}

\begin{figure*}[hbt!]
    \centering
    \setlength{\tabcolsep}{0.4pt}
    \renewcommand{\arraystretch}{0.4}
    {\footnotesize
    \begin{tabular}{c c c c c c c c}
        &
        \multicolumn{1}{c}{(a)} &
        \multicolumn{1}{c}{(b)} &
        \multicolumn{1}{c}{(c)} &
        \multicolumn{1}{c}{(d)} &
        \multicolumn{1}{c}{(e)} &
        \multicolumn{1}{c}{(f)} \\

        {\raisebox{0.34in}{
        \multirow{1}{*}{\rotatebox{0}{Input}}}} &
        \includegraphics[width=0.138\textwidth]{images/comparison/ReS/p100_2_source.jpg} &
        \includegraphics[width=0.138\textwidth]{images/comparison/ReS/p16_1_source.jpg} &
        \includegraphics[width=0.138\textwidth]{images/comparison/ReS/p17_2_source.jpg} &
        \includegraphics[width=0.138\textwidth]{images/comparison/ReS/p25_1_source.jpg} &
        \includegraphics[width=0.138\textwidth]{images/comparison/ReS/p31_2_source.jpg} &
        \includegraphics[width=0.138\textwidth]{images/comparison/ReS/p52_2_source.jpg} &\\

        {\raisebox{0.47in}{\multirow{1}{*}{\begin{tabular}{c}PAIR Diffusion \\ (50 steps, 49s)\end{tabular}}}}
        &
        \includegraphics[width=0.138\textwidth]{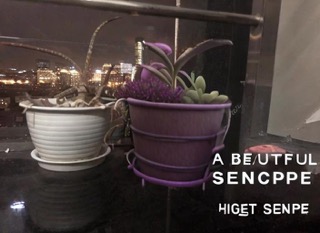} &
        \includegraphics[width=0.138\textwidth]{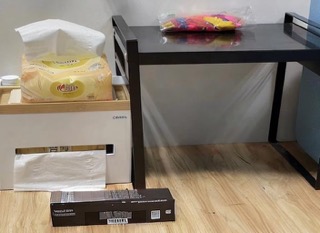} &
        \includegraphics[width=0.138\textwidth]{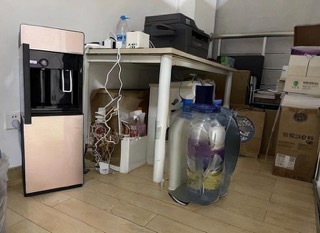} &
        \includegraphics[width=0.138\textwidth]{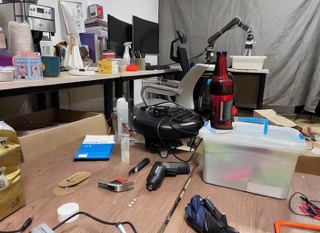} &
        \includegraphics[width=0.138\textwidth]{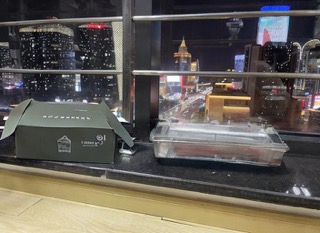} &
        \includegraphics[width=0.138\textwidth]{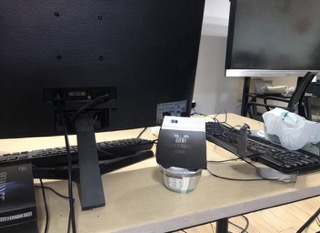} &\\

        {\raisebox{0.47in}{\multirow{1}{*}{\begin{tabular}{c}InfEdit \\ (50 steps, 47s)\end{tabular}}}}
        &
        \includegraphics[width=0.138\textwidth]{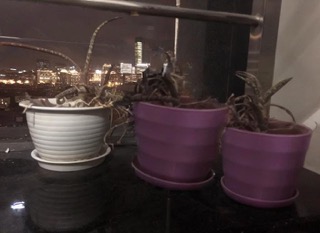} &
        \includegraphics[width=0.138\textwidth]{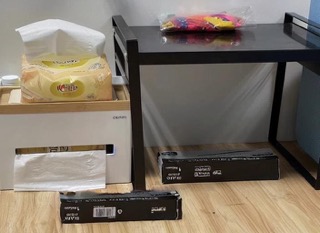} &
        \includegraphics[width=0.138\textwidth]{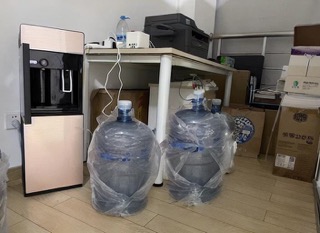} &
        \includegraphics[width=0.138\textwidth]{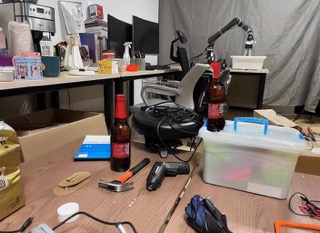} &
        \includegraphics[width=0.138\textwidth]{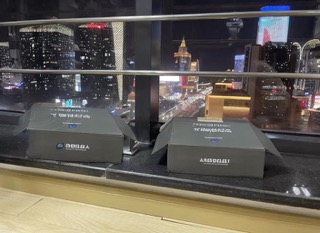} &
        \includegraphics[width=0.138\textwidth]{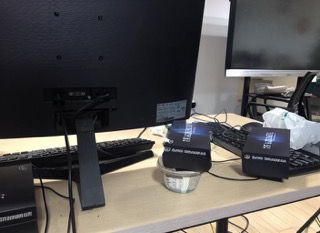} &\\
        
        {\raisebox{0.37in}{\multirow{1}{*}{\begin{tabular}{c}\textbf{PixelMan} \\ (50 steps, 34s)\end{tabular}}}} &
        \includegraphics[width=0.138\textwidth]{images/comparison/ReS/p100_2_ours50.jpg} &
        \includegraphics[width=0.138\textwidth]{images/comparison/ReS/p16_1_ours50.jpg} &
        \includegraphics[width=0.138\textwidth]{images/comparison/ReS/p17_2_ours50.jpg} &
        \includegraphics[width=0.138\textwidth]{images/comparison/ReS/p25_1_ours50.jpg} &
        \includegraphics[width=0.138\textwidth]{images/comparison/ReS/p31_2_ours50.jpg} &
        \includegraphics[width=0.138\textwidth]{images/comparison/ReS/p52_2_ours50.jpg} &\\
        
        {\raisebox{0.37in}{\multirow{1}{*}{\begin{tabular}{c}PAIR Diffusion \\ (16 steps, 18s)\end{tabular}}}} &
        \includegraphics[width=0.138\textwidth]{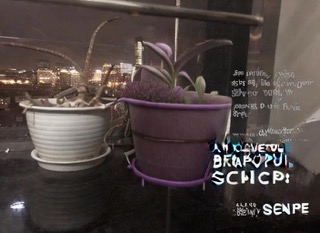} &
        \includegraphics[width=0.138\textwidth]{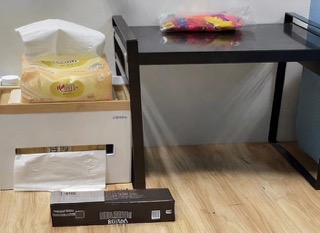} &
        \includegraphics[width=0.138\textwidth]{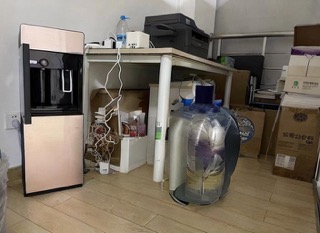} &
        \includegraphics[width=0.138\textwidth]{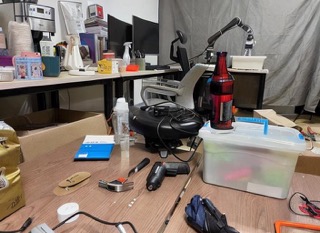} &
        \includegraphics[width=0.138\textwidth]{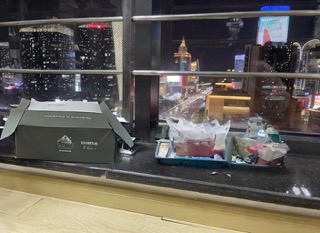} &
        \includegraphics[width=0.138\textwidth]{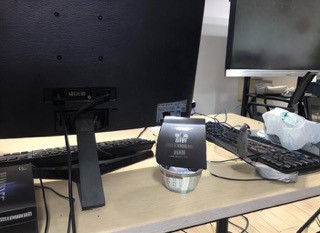} &\\
        
        {\raisebox{0.37in}{\multirow{1}{*}{\begin{tabular}{c}InfEdit \\ (16 steps, 15s)\end{tabular}}}} &
        \includegraphics[width=0.138\textwidth]{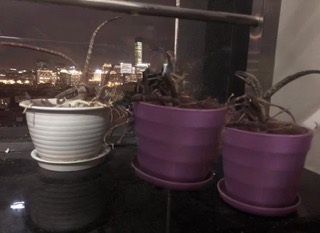} &
        \includegraphics[width=0.138\textwidth]{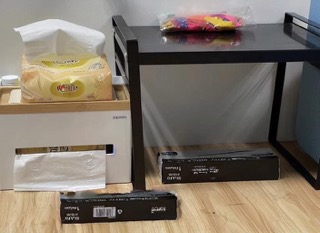} &
        \includegraphics[width=0.138\textwidth]{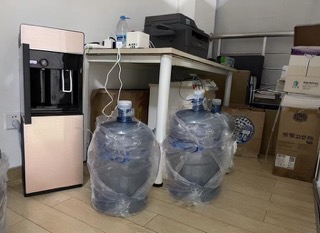} &
        \includegraphics[width=0.138\textwidth]{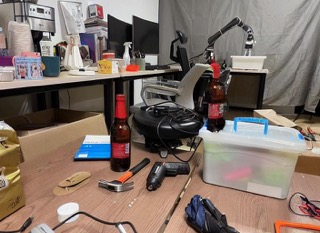} &
        \includegraphics[width=0.138\textwidth]{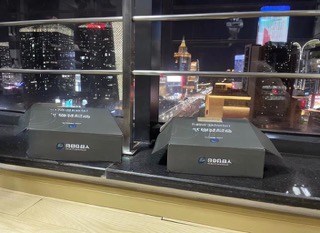} &
        \includegraphics[width=0.138\textwidth]{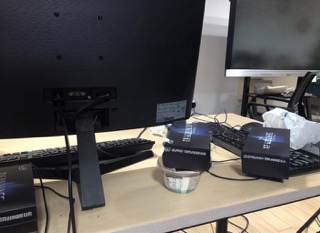} &\\

        {\raisebox{0.37in}{\multirow{1}{*}{\begin{tabular}{c}\textbf{PixelMan} \\ (16 steps, 11s)\end{tabular}}}} &
        \includegraphics[width=0.138\textwidth]{images/comparison/ReS/p100_2_ours16.jpg} &
        \includegraphics[width=0.138\textwidth]{images/comparison/ReS/p16_1_ours16.jpg} &
        \includegraphics[width=0.138\textwidth]{images/comparison/ReS/p17_2_ours16.jpg} &
        \includegraphics[width=0.138\textwidth]{images/comparison/ReS/p25_1_ours16.jpg} &
        \includegraphics[width=0.138\textwidth]{images/comparison/ReS/p31_2_ours16.jpg} &
        \includegraphics[width=0.138\textwidth]{images/comparison/ReS/p52_2_ours16.jpg} &\\

        {\raisebox{0.37in}{\multirow{1}{*}{\begin{tabular}{c}PAIR Diffusion \\ (8 steps, 11s)\end{tabular}}}} &
        \includegraphics[width=0.138\textwidth]{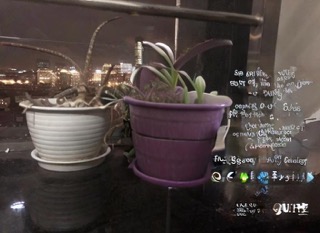} &
        \includegraphics[width=0.138\textwidth]{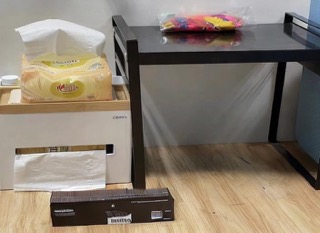} &
        \includegraphics[width=0.138\textwidth]{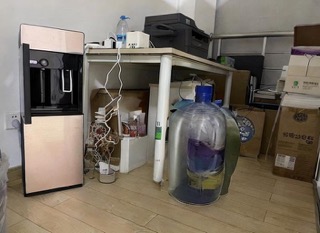} &
        \includegraphics[width=0.138\textwidth]{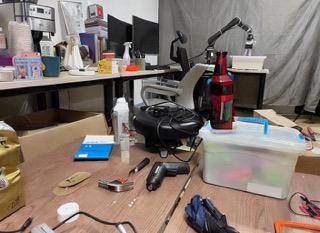} &
        \includegraphics[width=0.138\textwidth]{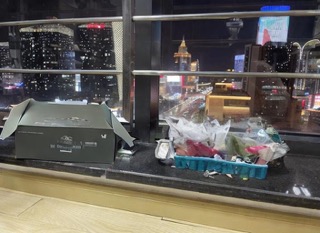} &
        \includegraphics[width=0.138\textwidth]{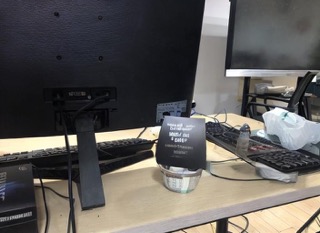} &\\

        {\raisebox{0.37in}{\multirow{1}{*}{\begin{tabular}{c}InfEdit \\ (8 steps, 7s)\end{tabular}}}} &
        \includegraphics[width=0.138\textwidth]{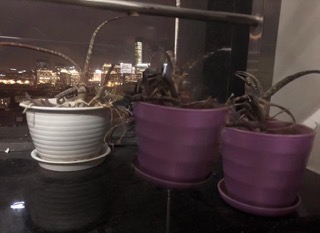} &
        \includegraphics[width=0.138\textwidth]{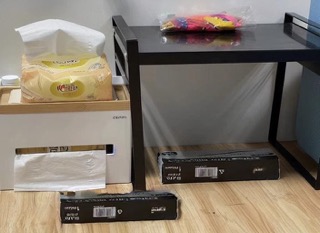} &
        \includegraphics[width=0.138\textwidth]{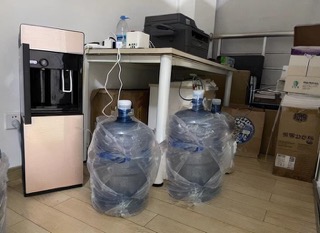} &
        \includegraphics[width=0.138\textwidth]{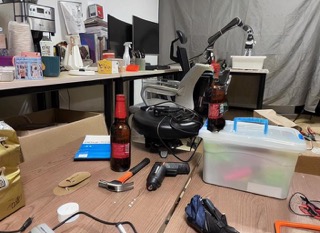} &
        \includegraphics[width=0.138\textwidth]{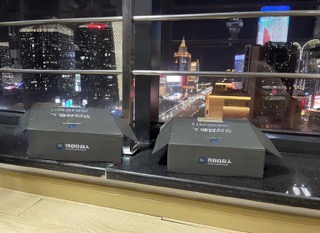} &
        \includegraphics[width=0.138\textwidth]{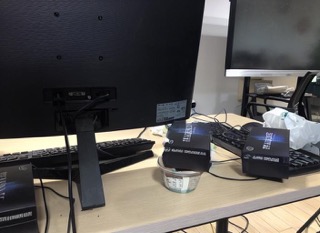} &\\

        {\raisebox{0.37in}{\multirow{1}{*}{\begin{tabular}{c}\textbf{PixelMan} \\ (8 steps, 5s)\end{tabular}}}} &
        \includegraphics[width=0.138\textwidth]{images/comparison/ReS/p100_2_ours8.jpg} &
        \includegraphics[width=0.138\textwidth]{images/comparison/ReS/p16_1_ours8.jpg} &
        \includegraphics[width=0.138\textwidth]{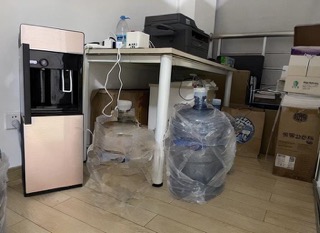} &
        \includegraphics[width=0.138\textwidth]{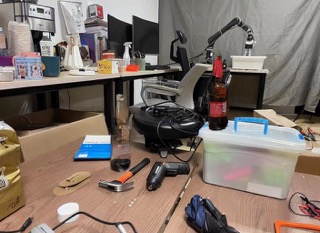} &
        \includegraphics[width=0.138\textwidth]{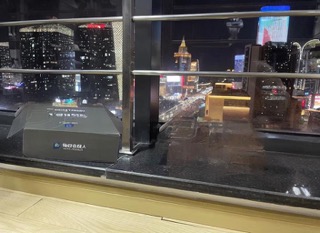} &
        \includegraphics[width=0.138\textwidth]{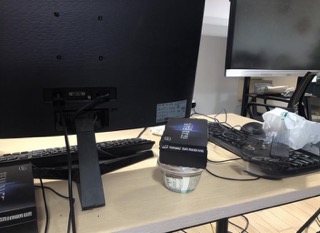} &\\

    \end{tabular}
    }
    \caption{
        \textbf{Additional qualitative comparisons} to PAIR Diffusion~\cite{goel2023pair} and InfEdit~\cite{xu2023infedit} on the ReS dataset at 50, 16 and 8 steps. 
    }
    \label{fig:examples_additional_res}
\end{figure*}

\end{document}